\pdfoutput=1

\documentclass[11pt]{article}

\usepackage[final]{acl}

\usepackage{times}
\usepackage{latexsym}

\usepackage[T1]{fontenc}

\usepackage[utf8]{inputenc}

\usepackage{microtype}

\usepackage{inconsolata}

\usepackage{graphicx}
\usepackage{hyperref}
\usepackage{url}
\usepackage{subcaption}
\usepackage{booktabs}
\usepackage{multirow}
\usepackage{bbding}
\usepackage[most]{tcolorbox}
\usepackage{comment}
\usepackage{amsmath}
\usepackage{amssymb}
\usepackage[normalem]{ulem}
\title{
Enhancing Chain-of-Thought
Reasoning 
with \\Critical Representation Fine-tuning 
}

\author{
 \textbf{Chenxi Huang\textsuperscript{1,2}\thanks{Work done during an internship at Alibaba Cloud Computing}},
\textbf{Shaotian Yan\textsuperscript{2}},
\textbf{Liang Xie\textsuperscript{2,3}},
\textbf{Binbin Lin\textsuperscript{4}\thanks{Correspondence: \href{binbinlin@zju.edu.cn}{Binbin Lin\textless binbinlin@zju.edu.cn\textgreater}, \href{jason.sc@alibaba-inc.com}{Chen Shen\textless jason.sc@alibaba-inc.com\textgreater}}},
\\
 \textbf{Sinan Fan\textsuperscript{2}},
 \textbf{Yue Xin\textsuperscript{2}},
 \textbf{Deng Cai\textsuperscript{1}},
  \textbf{Chen Shen\textsuperscript{2$\dag$}\thanks{Project Lead.}},
 \textbf{Jieping Ye\textsuperscript{2}}
\\
\\
 \textsuperscript{1}the State Key Laboratory of CAD\&CG, Zhejiang University
 \\
 \textsuperscript{2}Alibaba Cloud Computing
\\
 \textsuperscript{3}College of Computer Science and Technology, Zhejiang University of Technology
 \\
  \textsuperscript{4}College of Software, Zhejiang University
}

\begin{document}
\maketitle

\begin{abstract}

Representation Fine-tuning (ReFT), a recently proposed Parameter-Efficient Fine-Tuning (PEFT) method, has attracted widespread attention for significantly improving parameter efficiency by editing representation space alone. In this work, we investigate applying ReFT to complex reasoning tasks. However, directly using the native ReFT method, which modifies fixed representations at the beginning and end of each layer, yields suboptimal performance, as these fixed-position representations have uncertain impact on the outputs. 
We observe that, in complex reasoning tasks, there often exist certain critical representations. These representations either integrate significant information from preceding layers or regulate subsequent layer representations. Through layer-by-layer propagation, they exert a substantial influence on the final output. Naturally, fine-tuning these critical representations has the potential to greatly enhance reasoning performance.
Building upon these insights, we propose \textbf{C}ritical \textbf{R}epresentation \textbf{F}ine-\textbf{T}uning (CRFT), a novel method that identifies and optimizes these critical representations through information flow analysis. CRFT operates within a supervised learning framework, dynamically optimizing critical representations in a low-rank linear subspace while freezing the base model. 
The effectiveness and efficiency of our method are validated across eight benchmarks for arithmetic and commonsense reasoning, using LLaMA and Mistral model families. 
Notably, our method improves the accuracy of LLaMA-2-7B and ReFT by $18.2\%$ and $3.8\%$, respectively, on GSM8K, while using only $0.016\%$ of the model parameters, significantly less than other PEFT methods.
Furthermore, our method also adapts effectively to few-shot settings, boosting one-shot accuracy by $16.4\%$.
Our work highlights the untapped potential of representation-level optimization for CoT reasoning, offering a lightweight yet powerful alternative to traditional PEFT methods.\looseness=-1

\end{abstract}

\begin{figure}[tbp]
\centering
\includegraphics[width=0.98\linewidth]{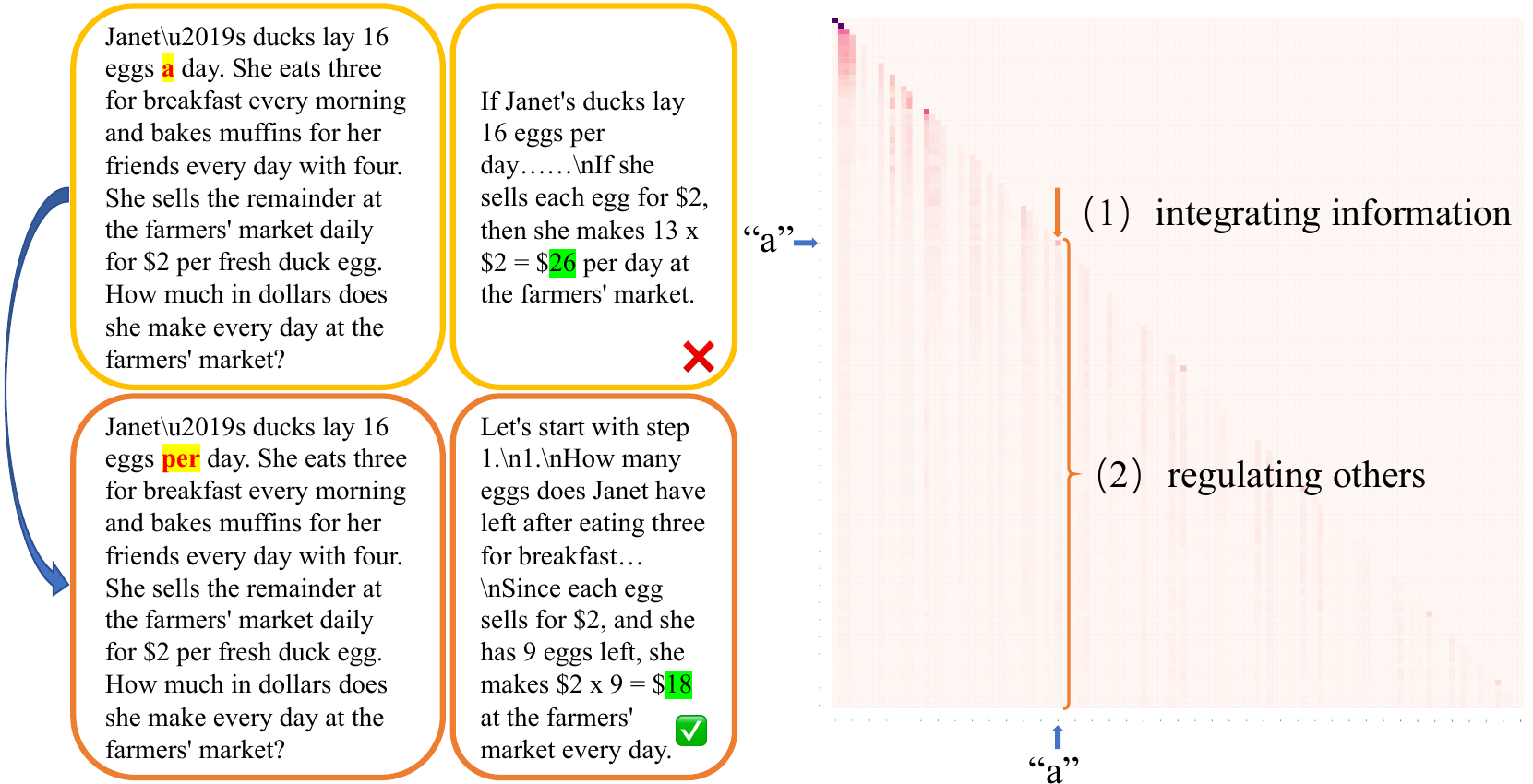}
\caption{\textbf{Examples of modifying a critical representation in the first layer (an input token).} This example, conducted on LLaMA-2-13B, illustrates (1) two strategies of identifying critical representations and (2) the impact of modifying these representations on the output.\looseness=-1}
\label{Fig: Example_short}
\end{figure}

\section{Introduction}
Large language models (LLMs) have made significant advances in the treatment of complex reasoning tasks~\citep{chu2023survey, yao2024tree, besta2024graph}, which demand intricate logical reasoning and comprehensive explanations. These tasks differ from simpler in-context tasks that mainly involve straightforward information retrieval or classification. A pivotal element in these advancements is the Chain-of-Thought (CoT)~\citep{wei2022chain}, decomposing the reasoning process into several intermediary steps, particularly used in the domains of arithmetic~\citep{lu2022dynamic, imani2023mathprompter, lightman2023let} and commonsense~\citep{trinh2018simple, ling2017program, patel2021nlp}.

Representation Fine-Tuning (ReFT)~\citep{wu2024reft} has emerged as a promising approach, offering parameter efficiency by operating at the representation level. Representations are considered fundamental as they reveal the inner reasoning processes of large language models (LLMs).
However, ReFT yields suboptimal performance in complex reasoning tasks, due to its reliance on altering fixed representations at the beginning and end of each layer, coupled with the unpredictable effects these changes have on the output. 
Through empirical analysis, we observe that in complex reasoning tasks, certain critical representations exist within each layer, as illustrated in Figure~\ref{Fig: Example_short}. These representations either aggregate significant information from the previous layer or modulate other representations in the subsequent layer.
Through layer-by-layer propagation, they exert a substantial influence on the final reasoning output. 
To further validate their importance, introducing random perturbations ($0.01$ Gaussian noise) to a random representation in each layer of LLaMA-2-7B on GSM8K resulted in a $1.4\%$ accuracy drop, underscoring the sensitivity of model performance to these representations. 
Naturally, fine-tuning these critical representations holds significant potential to enhance reasoning performance.
Building upon these insights, we propose a novel PEFT method termed \textbf{C}ritical \textbf{R}epresentation \textbf{F}ine-\textbf{T}uning (CRFT).

We employ information flow analytics~\citep{wang2023label}, utilizing attention and saliency scores~\citep{simonyan2013deep} as explicit indicators to identify critical representations. Specifically, for representations that aggregate significant information from the preceding layer, we prioritize those with predominant self-information flow, as they effectively consolidate gathered information. For representations that modulate subsequent layers, we focus on those with substantial outgoing information flow, reflecting their significant regulatory influence.
However, optimizing critical representations poses a significant challenge due to their context-dependent nature.  While some representations positively contribute to outputs and require no optimization, others adversely affect performance, with necessary adjustments varying across contexts.
To address this, we introduce adaptive learning within a supervised framework. Building on recent advances in parameter-efficient fine-tuning (PEFT) at the representation level~\cite{wu2024reft, wu2024advancing}, we freeze the base model and optimize critical representations by learning updated directions in a low-rank linear subspace.\looseness=-1

We conducted comprehensive experiments on eight reasoning datasets in two scenarios: arithmetic and commonsense~\citep{talmor2018commonsenseqa}, using four base models covering the LLaMA and Mistral families.
The experimental results demonstrate the effectiveness of our intervention. Specifically, our method achieves improvements of $18.2\%$ over LLaMA-2-7B on the GSM8K dataset with only the $0.016\%$ parameters of the model.
Furthermore, our method can be easily extended to few-shot learning, achieving increases of $16.4\%$ and $9.8\%$ in one-shot and two-shot learning, respectively.
Our work highlights the untapped potential of representation-level optimization for CoT reasoning, offering a lightweight yet powerful alternative to traditional prompt-centric and weight-centric methods.

\section{Method}
Our method, CRFT, consists of identifying and optimizing critical representations. We begin by introducing the problem formulation in Section~\ref{sec: problem}. Next, we propose two strategies for identifying critical representations by analyzing the information flow, as presented in Section~\ref{sec: identify}. Finally, we describe the way of optimizing critical representations in Section~\ref{sec:update}.

\subsection{Problem Formulation}
\label{sec: problem}
Given a sequence of $n$ input tokens $\boldsymbol{x}=(x_1, \dots, x_n)$, the language model commences by embedding these tokens into a list of representations $\boldsymbol{h}^{(0)} = (\boldsymbol{h}_1^{(0)}, \dots, \boldsymbol{h}_n^{(0)})$. 
Since the vast majority of state-of-the-art language models are currently constructed based on the transformer~\citep{vaswani2017attention} architecture, we focus solely on this architecture, which consists of $L$ layers of transformer blocks.
Subsequently, the $L$ layers successively compute the $l$-th list of hidden representations $\boldsymbol{h}^{(l)}$ as a function of the previous list of hidden representations $\boldsymbol{h}^{(l-1)}$. 
Each hidden representation is a vector $\boldsymbol{h} \in \mathbb{R}^d$. Finally, the model leverages the last layer of hidden representations $\boldsymbol{h}^{(L)}$ to produce its predictions.
Specifically, as a reasoning task, the model incrementally produces $k$ tokens following the probability expression $p(x_{n+k}|x_1, \ldots, x_n, x_{n+1}, \ldots, x_{n+k-1})$.
Our method aims to improve accuracy by identifying and optimizing critical representations $\boldsymbol{M}(\boldsymbol{h})$.

\begin{figure}[tbp]
\centering
\begin{subfigure}[b]{0.47\linewidth}
    \centering
    \includegraphics[width=\linewidth]{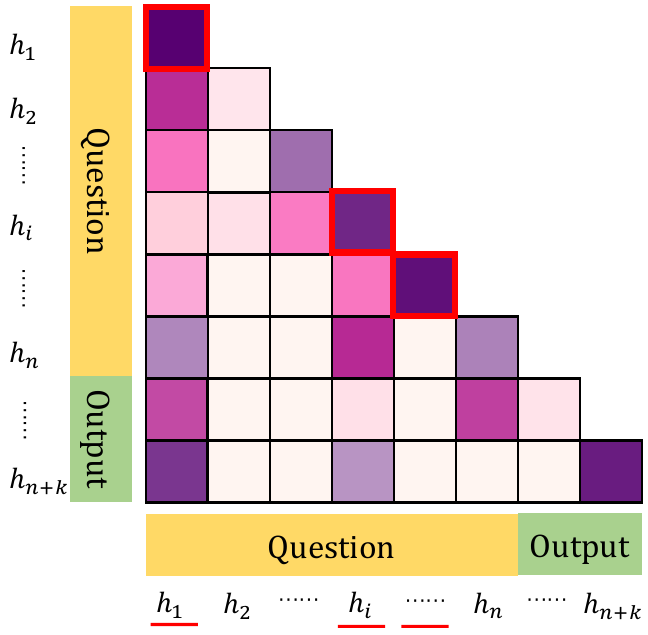}
    \caption{Self-referential filtering.}
    \label{fig:SRF}
\end{subfigure}
\hspace{0.02\linewidth}
\begin{subfigure}[b]{0.47\linewidth}
    \centering
    \includegraphics[width=\linewidth]{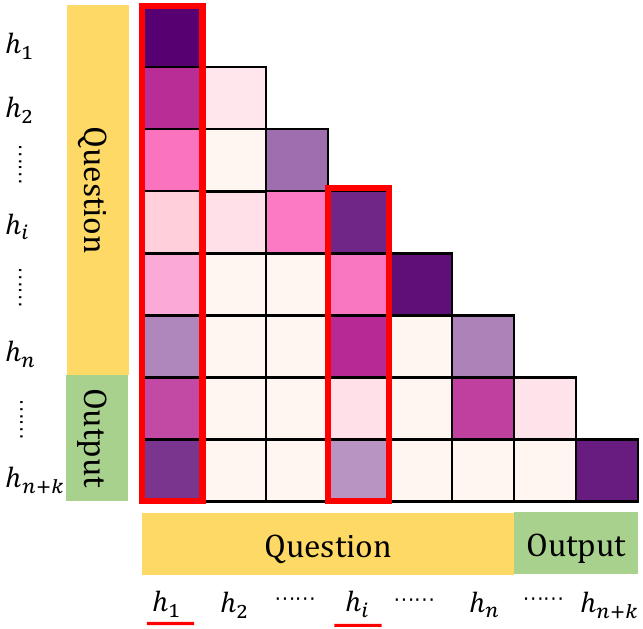}
    \caption{Multi-referential filtering.}
    \label{fig:MRF}
\end{subfigure}
\caption{\textbf{The illustration of self-referential filtering and multi-referential filtering.} We use \textcolor[RGB]{255,0,0}{red boxes} to highlight the diagonal cells in Figure~\ref{fig:SRF} and the column averages in Figure~\ref{fig:MRF} that exceed the threshold $\alpha$. The corresponding representations are marked with \textcolor[RGB]{255,0,0}{red lines} and are referred to as critical representations.}
\label{fig:self-identity}
\end{figure}

\subsection{Identify Critical Representations}
\label{sec: identify}
Previous representation editing works also involve modification of representations but depend on empirical observations or general knowledge to locate representations for editing, which limits their adaptability and performance.
For example, ReFT~\citep{wu2024reft} requires training and testing on other datasets to determine the optimal number of continuous representations to edit, specified as the first $x$ and last $y$ representations.
This selection process is not only cumbersome, but also lacks interpretability.
Our work identifies critical representations $\boldsymbol{M(h)}$, which significantly influence reasoning abilities and output correctness.
\begin{align}
    \boldsymbol{M}(\boldsymbol h) = \{ \boldsymbol{h}_i\mid & \text{Is correct}(\text{model}( \boldsymbol{h}_i+\epsilon)) \notag\\
    &\not=\text{Is correct}(\text{model}( \boldsymbol{h}_i))\},
\label{eq:anchor}
\end{align}
\noindent where $\epsilon$ is a small perturbation in a vector space. 
For simplicity, we use the abbreviation $\boldsymbol{M}^{(l)}$ to represent $\boldsymbol{M}(\boldsymbol{h}^{(l)})$ in the following text.
When all critical representations contribute positively to the output, accuracy is largely ensured.

As in the examples in Figure~\ref{Fig: Example_short}, whether a representation is an critical representation cannot be determined by itself but rather by its relationship with other representations.
So, we utilize the information flow~\citep{wang2023label}, leveraging attention and saliency scores as indicators. 
As shown in Figure~\ref{fig:self-identity}, we use a grid to visualize the information interaction between representations, where cell $(i, j)$ indicates the information interaction between representation $j$ and representation $i$.
The value of the cell $(i, j)$ is indicated by attention scores or saliency scores, with darker colors signifying richer information interactions.
The critical representations can be categorized into two functional roles: (1) integrating significant information from the preceding layer and (2) regulating the subsequent layer representations.
Specifically, for the former, we focus on representations that consistently receive information flow from itself, indicating effective information accumulation. 
For the latter, we target representations that disseminate information to multiple others, indicating its rich information interaction.
Consequently, we design two strategies to filter critical representations: \textbf{\textit{self-referential filtering}} and \textbf{\textit{multi-referential filtering}}, respectively.\looseness=-1

\subsubsection{Self-Referential Filtering}
If information from representation $i$ mainly flows back to itself in the subsequent layer, it means that representation $i$ contains important information or has effectively accumulated significant information.
Consequently, we use $\operatorname{Info}(i, i)$ as a critical metric to assess this retention. 
If $\operatorname{Info}(i, i)$ is large, then $\operatorname{Info}(i, j), j\not= i$ will be small since the values in a row are normalized through the softmax function. This situation suggests that the information flow from the representation $i$ is predominantly directed toward itself, confirming that the representation $i$ is indeed crucial.\looseness=-1
\begin{equation}
\label{eq:DF}
    \begin{aligned}
        \boldsymbol M_{\text{diag}}^{(l)} = \{ \boldsymbol{h}^{(l)}_i\mid\operatorname{Info}^{(l-1)}(i, i) > \alpha\},
        i \in \{1, \dots, n\}.
    \end{aligned}
\end{equation}

To quantify information interactions, we employ attention scores and saliency scores as indicators, thus proposing two distinct ways: Self-Referential Attention Filtering (SAF) and Self-Referential Saliency Filtering (SSF), separately. 

\noindent\textbf{Self-Referential Attention Filtering (SAF).}
We utilize normal attention scores $A^{(l)}_i$, described in Eq.~\ref{eq:attenion}, as an explicit indicator to filter critical representations, since they quantify the relevance and degree of emphasis assigned to various representations within a sequence. 
This mechanism enables the model to dynamically concentrate on interactions and enhance its understanding capabilities.
\begin{equation}
\label{eq:attenion}
    \operatorname{Info}^{(l)}_{\mathrm{SAF}}(i, i)=A^{(l)}_i = \text{softmax}(\boldsymbol{h}_{i}^{(l)}(\boldsymbol{h}^{(l)})^{\mathrm{T}}/\sqrt{d}),
\end{equation}
\noindent\textbf{Self-Referential Saliency Filtering (SSF).}
We also leverage saliency scores to filter critical representations.
As saliency score is a widely accepted interpretation tool~\citep{simonyan2013deep}, comprehensively considers attention scores and gradient values, highlighting interactions from critical representations to the model output, as shown in Eq.~\ref{eq:saliency},
\begin{equation}
\label{eq:saliency}
    \operatorname{Info}^{(l)}_{\mathrm{SSF}}(i,i) = A^{(l)}_i \odot \frac{\partial \mathcal L(x)}{\partial A^{(l)}_i}, \end{equation}
where $\odot$ denotes the element-wise multiplication, and $\mathcal L(\cdot)$ represents the cross entropy loss function of the predicted probability distribution and the predicted class indices.

\subsubsection{Multi-Referential Filtering}
If information from representation $j$ significantly affects multiple other representations, including producing representations, then representation $j$ is crucial. Specifically, we calculate the average of cells in the column $j$ as a critical metric to represent the influence of $j$ on other representations. If the average of $\ \operatorname{Info}(\cdot, j)$ is large, then representation $j$ has a substantial influence on others and plays a crucial role.
As shown in Eq.~\ref{eq:CF}, we use the threshold $\beta$ to filter the critical representations,
\begin{equation}
\label{eq:CF}
    \boldsymbol M_{\text{col}}^{(l)} = \left\{ \boldsymbol{h}^{(l)}_j \; \left\vert \; \frac{\sum_{i=j}^{n+k} \operatorname{Info}^{(l)}(i, j)}{n+k-j+1}> \right. \beta\right\},
\end{equation}
where $k$ is the number of output representations. 

We also use the attention score and the saliency score to quantify the influence of representation $j$ on representation $i$, which is termed Multi-Referential Attention Filtering (MAF) and Multi-Referential Saliency Filtering (MSF), respectively.

\begin{figure}[tbp]
\centering
\includegraphics[width=0.98\linewidth]{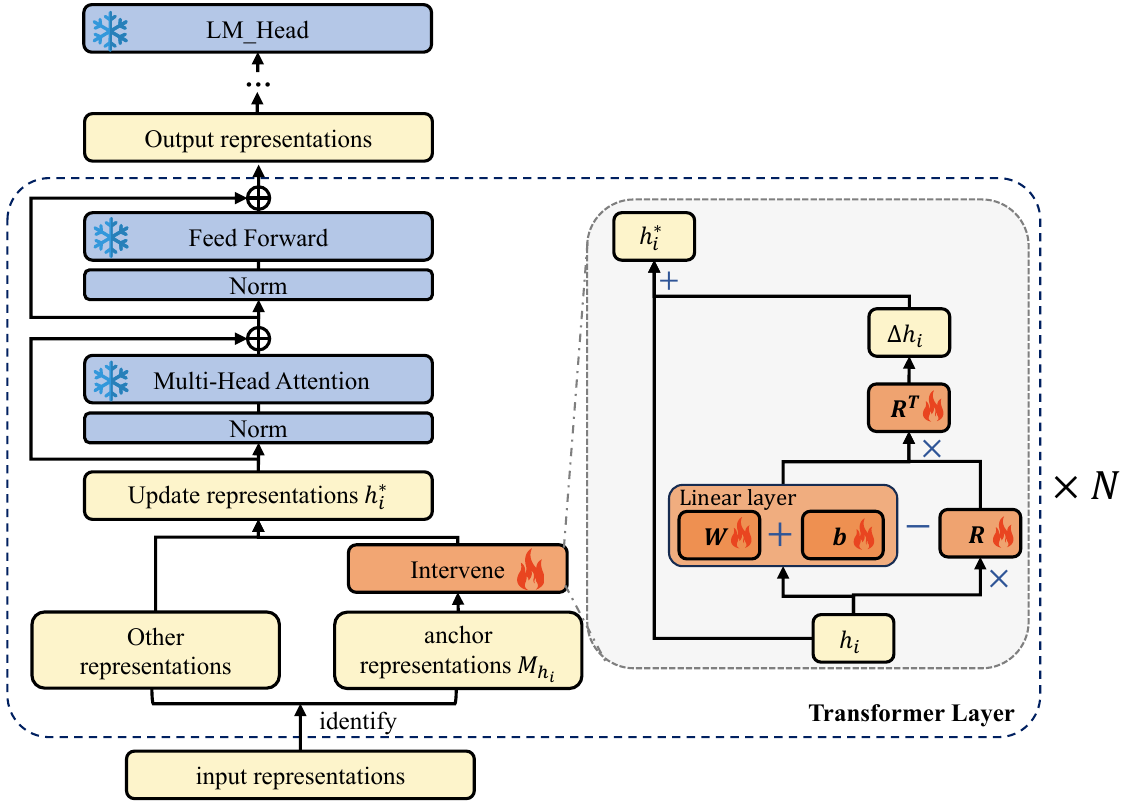}
\caption{\textbf{The pipeline of optimizing critical representations.} \textcolor[RGB]{255,128,0}{Orange} highlights the parameters to be learned, while \textcolor[RGB]{0,0,255}{blue} indicates the parameters that remain frozen.}
\label{fig:training}
\end{figure}

\subsection{Optimize Critical Representations}
\label{sec:update}
Upon identifying critical representations, it becomes imperative to optimize them to ensure that their influence on reasoning tasks is accurately aligned.
However, the direction of this modification remains uncertain and may not be unique. Consequently, we model the adjustment as a learnable vector $\Delta \boldsymbol{h}$, which is learned during the training process to rectify the critical representations adaptively. 
Following~\citep{wu2024reft, huang2024ravel}, we restrict our optimized vectors to a low-rank linear subspace employing a projection matrix with orthonormal rows $\boldsymbol{R}\in \mathbb{R}^{r\times d}$, where $r$ indicates the dimensionality of the subspace we are intervening in.
We learn a projected source through a linear layer $\text{Linear}(\boldsymbol h)=\boldsymbol W\boldsymbol h + \boldsymbol b$. 
Consequently, we modify the representation within the $r$-dimensional subspace spanned by the rows of $\boldsymbol{R}$ to adopt the values derived from our linear projection source, $\text{Linear}(\boldsymbol h)$.
The overall optimization mechanism is depicted in Eq.~\ref{eq:update}, 
\begin{equation}
\label{eq:update}
\small 
\Phi(\boldsymbol h) = \begin{cases}\boldsymbol h+\boldsymbol R^T(\boldsymbol W\boldsymbol h + \boldsymbol b - \boldsymbol R\boldsymbol h), & \text{if }\boldsymbol h \in \boldsymbol M(\boldsymbol h)\\
\boldsymbol h, & \text{otherwise.}\end{cases}
\end{equation}

\begin{table}[tb]
\centering
\resizebox{\linewidth}{!}{
\begin{tabular}{@{}lccc@{}}
\toprule
\textbf{PEFT} & \textbf{Param (\%)} & \textbf{Identify} & \textbf{Accuracy ($\uparrow$)}\\
\midrule
None & - & - & 14.6\\
\hline
LoRA (r=64) & 0.826\% & - & 38.5\\
LoRA (r=8) & 0.103\% & - & 36.7\\
RoSA ($r=48$) & 0.819\% &  - & 30.5\\
RoSA ($r=32$) & 0.816\% &  - & 32.2\\
RoSA ($r=16$) & 0.812\% &  - & 32.8\\
SpA & 0.809\% & - & 29.6\\
\hline
ReFT ($r=8$) & 0.031\% & $p7+s7$ & 29.0\\
\hline
\multirow{6}{*}{CRFT (ours)} & \multirow{6}{*}{0.016\%} & SAF & 30.4 | 29.6\\
& & MAF & 32.0 | \underline{32.1}\\ 
& & Union(attn) & 31.2 | \textbf{32.8}\\ 
\cmidrule{3-4}
& & SSF & 31.4 | 30.4\\
& & MSF & 31.4 | 30.3\\
& & Union(sal) & \textbf{32.8} | 31.5\\
\bottomrule
\end{tabular}}
\caption{\textbf{Quantitative comparison of PEFT methods on GSM8K with LLaMA-2-7B.} The best performance is highlighted in \textbf{bold}, while the second-best is \underline{underlined}.\looseness=-1}
\label{Tab: compare}
\end{table}

\begin{table*}[tb]
\centering
\resizebox{0.98\textwidth}{!}{
\begin{tabular}{@{}llcccc|cccc@{}}
\toprule
\multirow{2}{*}{\textbf{Model}} & \multirow{2}{*}{\textbf{PEFT}} & \multirow{2}{*}{\textbf{Identify}} & \multicolumn{7}{c}{\textbf{Accuracy ($\uparrow$)}}\\
\cmidrule{4-10}
& & & \textbf{AQuA} & \textbf{MAWPS} & \textbf{SVAMP} & \textbf{BoolQ} & \textbf{SocialIQA} & \textbf{WinoGrande} & \textbf{OpenBookQA}\\
\midrule
\multirow{5}{*}{LLaMA-2-7B} 
& ReFT & p7+s7 & 21.7 & 80.7 & 52.2 & 50.7 & 61.2 & 51.7 & 58.6 \\
\cmidrule{2-10}
& \multirow{4}{*}{CRFT (ours)} & SAF & 25.6 | 26.0 & 78.6 | \textbf{84.5} & \textbf{53.4} | 52.6 & 60.0 | 53.7 & 62.5 | \textbf{67.4} & \underline{60.6} | 55.3 & 57.0 | \underline{62.2}\\
& & MAF & \textbf{27.6} | 24.8 & \underline{81.1} | 80.7 & 52.4 | \textbf{53.4} & 60.5 | \underline{61.8} & 52.8 | 64.9 & \textbf{68.4} | 51.8 & 50.6 | \textbf{66.4}\\
\cmidrule{3-10}
& & SSF & 26.0 | 26.8 & 80.7 | 79.8 & 52.5 | \underline{53.3} & \textbf{62.0} | 54.3 & \underline{67.1} | 64.4 & 60.2 | 60.1 & 58.4 | 58.6\\
& & MSF & \underline{27.2} | 22.8 & 79.4 | 80.7 & 52.3 | 52.5 & 60.0 | 59.7 & 65.8 | 63.4 & 54.5 | 54.2 & 59.0 | 56.4\\ \midrule
\multirow{5}{*}{LLaMA-3-8B} 
& ReFT & p7+s7 & 46.9 & 87.0 & 74.2 & 62.1 & 60.2 & 56.0 & 66.0\\
\cmidrule{2-10}
& \multirow{4}{*}{CRFT (ours)} & SAF & 47.2 | 47.2 & \underline{89.9} | 88.2 & 75.5 | 76.1 & 63.0 | 66.4 & 68.2 | 67.1 & 62.6 | 56.3 & 71.0 | 73.6\\
& & MAF & 48.4 | \underline{50.4} & \textbf{90.8} | \textbf{90.8} & 77.1 | 77.9 & 62.4 | 66.2 & 66.5 | 62.7 & \textbf{67.2} | \underline{62.9} & 73.8 | 72.6\\
\cmidrule{3-10}
& & SSF & 50.0 | 49.2 & 86.6 | 86.6 & \underline{78.0} | \textbf{78.1} & 64.0 | 66.6 & \textbf{74.7} | \underline{74.2} & 60.3 | 62.0 & \underline{75.6} | \textbf{77.0}\\
& & MSF & 48.0 | \textbf{51.6} & 87.0 | 87.4 & 75.2 | 74.8 & \underline{67.0} | \textbf{67.9} & 67.4 | 69.7 & 62.3 | 62.8 & 70.0 | 68.6\\
\midrule 
\multirow{5}{*}{Mistral-7B} 
& ReFT & p7+s7 & 32.3 & 84.9 & 67.4 & 62.5  & 64.6 & 58.5 & 63.8\\
\cmidrule{2-10}
& \multirow{4}{*}{CRFT (ours)} & SAF & 36.2 | 38.6 & \underline{87.0} | 85.7 & 65.9 | 66.2 & 63.0 | \textbf{66.5} & 66.7 | \textbf{75.6} & 61.5 | 62.9 & \underline{72.6} | \underline{72.6}\\
& & MAF & \underline{39.0} | 38.2 & 84.9 | 85.3 & 66.3 | 65.3 & 62.1 | 60.8  & 66.9 | 71.5 & 61.2 | \underline{63.7} & 64.2 | 69.6\\
\cmidrule{3-10}
& & SSF & 37.4 | 33.5 & 85.3 | 84.5 & \underline{70.3} | \textbf{70.6} & 62.3 | 64.8 & 64.9 | 62.9 & \textbf{64.3} | 61.4 & 61.6 | 66.4\\
& & MSF & \textbf{41.3} | 37.8 & \textbf{87.4} | 85.3 & 66.0 | 66.9 & 62.5 | \underline{65.0} & 69.3 | \underline{71.8} & 62.3 | 59.5 & \textbf{72.8} | 68.6\\
\bottomrule
\end{tabular}}
\caption{\textbf{Quantitative comparison on arithmetic and commonsense reasoning datasets with three base models: LLaMA-2-7B, LLaMA-3-8B, and Mistral-7B.} We train on Math10k and report results on AQuA, MAWPS, and SVAMP for arithmetic reasoning datasets; and we train on our combined commonsense datasets Commonsense60k and report results on four datasets: BoolQ, SocialIQA, WinoGrande, and OpenBookQA. 
}
\label{Tab: math}
\end{table*}

\section{Experiments}
To validate the effectiveness of our method, CRFT, we performed experiments in two scenarios covering eight datasets: GSM8K~\citep{cobbe2021gsm8k}, AQuA~\citep{ling2017program}, MAWPS~\citep{koncel-kedziorski-etal-2016-mawps}, SVAMP~\citep{patel2021nlp}, BoolQ~\citep{clark2019boolq}, SocialIQA~\citep{sap2019socialiqa}, WinoGrande~\citep{sakaguchi2021winogrande}, and OpenBookQA~\citep{mihaylov2018can}. 
In particular, for the Commonsense task, previous work used the Commonsense170K dataset, which only provides the answers and lacks a reasoning process. We synthesized a Commonsense60K dataset with reasoning steps based on six commonly used commonsense datasets: CommonsenseQA~\citep{talmor2018commonsenseqa}, CoS-e~\citep{rajani2019explain}, OpenBookQA~\citep{mihaylov2018can}, SocialIQA~\citep{sap2019socialiqa}, StrategyQA~\citep{geva2021strategyqa}, WorldTree~\citep{jansen2018worldtree}. 
All experiments were conducted on the Pyvene~\citep{wu-etal-2024-pyvene} codebase using a single GPU, either an NVIDIA A100 (80GB) or an L20 (40GB).
And our method requires $4$ hours for training on GSM8K with LLaMA-2-7B.
We set the scoring method to the ``order'', with $\alpha$ and $\beta$ both set to $0.05$.
The ablation studies of these hyperparameters are discussed in Section~\ref{sec:hyper}. We adopt SAF strategy in Section~\ref{sec:few-shot} and Section~\ref{sec:hyper}.
The details of all datasets and other implementations are reported in Appendix~\ref{sec:implement_details}.
Our evaluation focused exclusively on the accuracy of the final numerical or multiple-choice answers. 
Generation examples are reported in Appendix~\ref{sec:case}.\looseness=-1

\subsection{Quantitative Results}
Table~\ref{Tab: compare} summarizes the comparison of our method, CRFT, with other PEFT methods on GSM8k with LLaMA-2-7B.
For each strategy, we report two accuracy values: the first value involves selecting critical representations by further filtering those identified as critical representations from the previous layer, while the second value focuses on identifying critical representations by filtering only within the current layer.
Given that the optimal strategy may differ by context, we recommend a combined approach of self-referential and multi-referential filtering. Since the scoring systems of these two strategies are not directly comparable, the union of the filtered sets is employed.
To ensure a fair comparison, the same number of critical representations is maintained, which may lead to the omission of some highly important ones. Consequently, the combined method may exhibit slightly lower performance in certain situations. Alternatively, adjusting the threshold $\alpha$ and $\beta$ provides a solution: lowering $\alpha$ ($\beta$) increases interventions for improved performance, while raising $\alpha$ ($\beta$) decreases interventions for enhanced efficiency.

Without bells and whistles, our method is comparable with other PEFT methods with fewer learnable parameters. 
For example, one of our strategies, union with attention scores, outperforms LLaMA-2-7B and ReFT by $18.2\%$ and $3.8\%$, respectively.
Furthermore, the percentage of trainable parameters, calculated by dividing the trainable parameters by the total parameters of the model, highlights their substantial efficiency.
Our method requires only $1/6$ of the learnable parameters used by LoRA and $1/2$ of those used by ReFT with the same rank.

Furthermore, our method, CRFT, consistently exhibits better performance on different models in arithmetic and commonsense scenarios.
We report the results on different model sizes and model families on GSM8K: LLaMA-2-7B, LLaMA-2-13B, LLaMA-3-8B, and Mistral-7B, as shown in Appendix~\ref{sec:moremodels}.
In addition, we present additional experimental results in arithmetic and commonsense scenarios, as shown in Table~\ref{Tab: math}. We use the official public code of ReFT to report performance, as it only reports the results on LLaMA-1. And following the experimental conclusion of ReFT, we adopt the best intervention parameters $p7+s7$, indicating the intervention in the first and the last seven representations.
The consistent improvements observed in different reasoning tasks and different models underscore the robustness and versatility of our approach.\looseness=-1

\begin{table}[tb]
\centering
\resizebox{\linewidth}{!}{
\begin{tabular}{@{}l|ccc@{}}
\toprule
\textbf{Few-shot} & \textbf{zero-shot} & \textbf{one-shot} & \textbf{two-shot}\\
\midrule
None & 14.6 & 16.2 & 20.5\\
CRFT & 29.6 & 28.7 | 32.6 & 29.0 | 30.3\\
\hline
\textbf{Improvement} & \textbf{+15.0} & \textbf{+12.5} | \textbf{+16.4} & \textbf{+8.5} | \textbf{+9.8}\\
\bottomrule
\end{tabular}}
\caption{\textbf{Expand our method to few-shot learning on GSM8K using Llama-2-7B with the SAF strategy.}}
\label{Tab: few-shot}
\end{table}

\subsection{Expand to Few-shot Learning}
\label{sec:few-shot}
Our method can easily be extended to few-shot learning. 
Intuitively, demonstrations should not directly affect the output; they are usually used to gain a higher-level semantic understanding, which then affects the output. However, representations in the question, such as numbers, can indeed have a direct impact.
Consequently, we present experiments in Table~\ref{Tab: few-shot} to examine whether the demonstration and the question should be learned independently. 
The first value suggests that the demonstration and the question are interdependent, leading to a single update vector for the critical representations. Conversely, the second value implies that the demonstration and the question are independent, resulting in distinct update vectors.
These results 
prove the necessity of differentiating update directions between demonstrations and the question.
Due to memory limitation, we only experimented with one-shot and two-shot.

\begin{table}[t]
\centering
\resizebox{0.75\linewidth}{!}{
\begin{tabular}{@{}l|cccc@{}}
\toprule
\textbf{Threshold} & 1.0 & 0.25 & 0.05 & 0.01\\
\hline
\textbf{Accuracy ($\uparrow$)} & 24.7 & 30.0 & 29.6 & \textbf{33.2}\\
\bottomrule
\end{tabular}}
\caption{\textbf{Ablation study on threshold $\alpha$ ($\beta$) on GSM8K using Llama-2-7B with the SAF strategy.}}
\label{Tab: hyper-parameters-threshold}
\end{table}

\begin{table}[t]
\centering
\resizebox{0.75\linewidth}{!}{
\begin{tabular}{@{}l|cccc@{}}
\toprule
\textbf{Number} & 0 & 14 & 20 & 30\\
\hline
\textbf{Accuracy ($\uparrow$)} & 14.6 & 29.6 & \textbf{30.3} & 27.7\\
\bottomrule
\end{tabular}}
\caption{\textbf{Ablation study on the number of intervention representations on GSM8K using Llama-2-7B with the SAF strategy.}}
\label{Tab: hyper-parameters-length}
\end{table}

\subsection{Hyperparameter Configurations}
\label{sec:hyper}
We conducted extensive ablation studies on GSM8K using Llama-2-7B with the SAF strategy to systematically investigate hyperparameters, including the threshold $\alpha$ and $\beta$, the number of intervention representations and selection criteria.

The threshold $\alpha$ and $\beta$ determines the degree to which the critical representations are. 
Given that the threshold values for $\alpha$ and $\beta$ lie on the same dimension, we apply a unified threshold for both self-referential filtering and multi-referential filtering. We investigated four values, as shown in Table~\ref{Tab: hyper-parameters-threshold}, and found that a threshold of $0.01$ yields the best results. 
As the threshold decreases, the number of selected representations increases, but altering these selected representations can be more difficult. However, a lower threshold also carries the risk of excluding significant representations.

In the implementation, the number of intervention representations for each layer is fixed. If the number of critical representations obtained through the SAF strategy exceeds the number of intervention representations, we sample them using specific selection criteria. Conversely, if the number of critical representations is fewer than needed, we use a placeholder value of $-1$ to pad the length. For a fair comparison with the ReFT method, we set the default number of intervention representations to $14$. An ablation study on the number of representations, shown in Table~\ref{Tab: hyper-parameters-length}, revealed that the results were optimal when set to $20$. When the number of intervention representations becomes too large, it hinders the learning of the update direction, leading to suboptimal results compared to using fewer representations.\looseness=-1

For selection criteria, we designed three approaches to sample critical representations: positional order, score ranking, and random selection. The results, shown in Table~\ref{Tab: hyper-parameters-select}, indicate that positional order selection is superior, while random selection yields significantly lower results compared to the other two criteria.

The ablation study presented above suggests that careful selection of hyperparameters can further enhance the performance of our CRFT method.

\begin{table}[tb]
\centering
\resizebox{0.75\linewidth}{!}{
\begin{tabular}{@{}l|ccc@{}}
\toprule
\textbf{Criteria} & order & score & random\\
\hline
\textbf{Accuracy ($\uparrow$)} & \textbf{29.6} & 28.7 & 23.1\\
\bottomrule
\end{tabular}}
\caption{\textbf{Ablation study on selection criteria on GSM8K using Llama-2-7B with the SAF strategy.}}
\label{Tab: hyper-parameters-select}
\end{table}

\begin{table}[tb]
\centering
\resizebox{\linewidth}{!}{
\begin{tabular}{@{}l|cccccc@{}}
\toprule
\textbf{Layer} & None & 0 & 31 & 0-15 & 16-31 & all\\
\hline
\textbf{Acc. ($\uparrow$)} & 14.6 & 24.9 & 22.7 & 30.5 & 24.6 & 29.6\\
\bottomrule
\end{tabular}
}
\caption{\textbf{The validation of effectiveness in each layer on GSM8K using Llama-2-7B with the SAF strategy.}}
\label{Tab: lora}
\end{table}

\begin{table*}[tb]
\centering
\resizebox{\linewidth}{!}{
\begin{tabular}{@{}l|c|c|c|ccccccccccc@{}}
\toprule
\multirow{2}{*}{\textbf{Location}} & \multirow{2}{*}{\textbf{None}} & \textbf{ReFT} & \textbf{Our CRFT} & \multicolumn{11}{c}{\textbf{Uniform Random}}\\
\cmidrule{5-15}
 & & \textbf{($p7+s7$)} & \textbf{(MAF)} & \textbf{37} & \textbf{38} & \textbf{39} & \textbf{40} & \textbf{41} & \textbf{42} & \textbf{43} & \textbf{44} & \textbf{45} & \textbf{46} & \textbf{47}\\
\midrule
\textbf{Accuracy ($\uparrow$)} & 14.6 & \underline{29.0} & \textbf{32.1} & 26.6 & 26.6 & 28.1 & 27.3 & 25.5 & 24.5 & 27.8 & 27.5 & 28.1 & 26.2 & 26.4\\
\bottomrule
\end{tabular}}
\caption{\textbf{The necessity of identifying critical representations.} We present the results with LLaMA-2-7B on GSM8K. The best way to identify critical representations is highlighted in \textbf{bold}, while the second-best is \underline{underlined}.\looseness=-1}
\label{Tab: randomseed}
\end{table*}

\begin{figure}[t]
\centering
\includegraphics[width=0.95\linewidth]{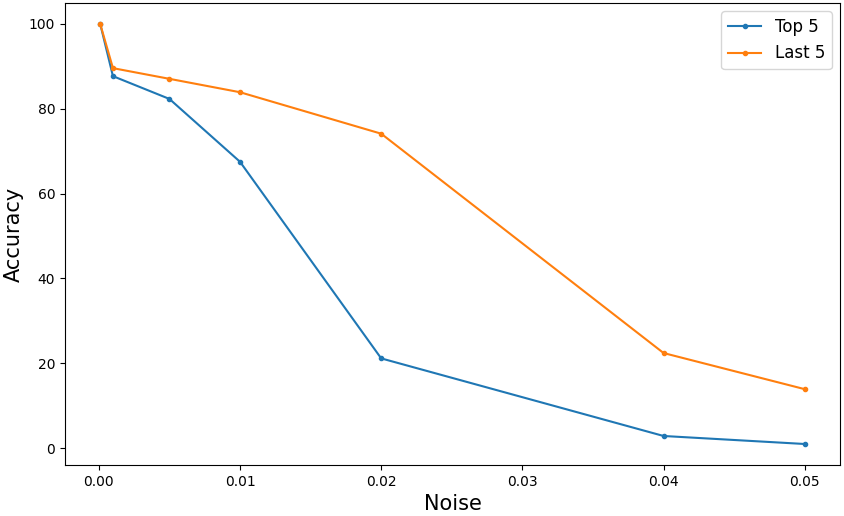}
\caption{\textbf{The validation of critical representations identification.}  Accuracy of originally correct examples under noise in the top $5$ and last $5$ representations.}
\label{Fig: change_anchor}
\end{figure}

\subsection{Are critical representations instrumental?} 
We validate that the selected representations are critical representations with a significant impact on the output. Using SAF as an identification strategy, we selected the top $5$ and last $5$ representations based on their scores for each layer on GSM8K with LLaMA-2-7B. The effects of adding noise to these representations are presented in Figure~\ref{Fig: change_anchor}, where the x-axis represents the magnitude of the noise, and the y-axis shows the proportion of originally correct examples remaining correct. 
We observe that the accuracy of the top $5$ representations decreases rapidly with increasing noise. When the noise level is $0.02$, the accuracy of the top $5$ representations drops to $21.1\%$, whereas the last $5$ representations maintain an accuracy of $74.1\%$.
This result demonstrates the significant impact of critical representations on output performance.

In addition, we investigated the necessity of identifying critical representations.
We tested random intervention locations using seed values ranging from $37$ to $47$. 
As shown in Table~\ref{Tab: randomseed}, the interventions of random representations during training can surpass the original LLaMA-2-7B, as the update direction is learnable. 
However, it remains inferior to the results achieved through our careful identification of critical representations, highlighting the necessity of this process.

Furthermore, we verified that intervention is necessary at each layer. As shown in Table~\ref{Tab: lora}, we intervene in the first layer (Layer $0$), the final layer (Layer $31$), the first half of the layers $\left[0-15\right]$, the last half of the layers $\left[16-31\right]$, and all layers. 
We found that each intervention improved accuracy and that interventions in the earlier layers have a greater impact on the results. However, the best performance was achieved by intervening in the first half of the layers, as earlier feature representations are more closely aligned with the task objectives and can propagate throughout the model.

\begin{figure*}[tbhp]
\centering
\begin{subfigure}[b]{0.47\textwidth}
    \centering
    \includegraphics[width=\textwidth]{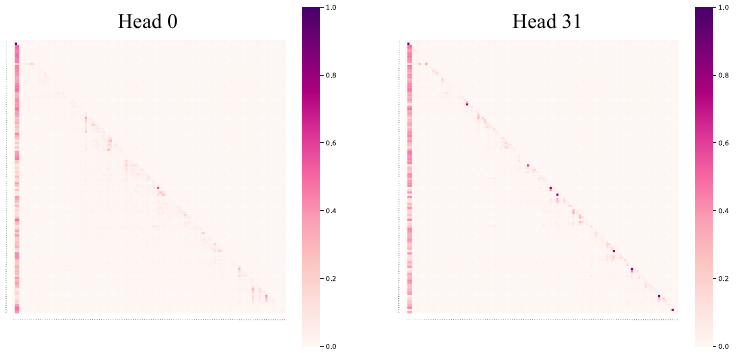}
    \caption{Original LLaMA-2-7B.}
    \label{fig:ori_attention_score}
\end{subfigure}
\hspace{0.04\textwidth}
\begin{subfigure}[b]{0.47\textwidth}
    \centering
    \includegraphics[width=\textwidth]{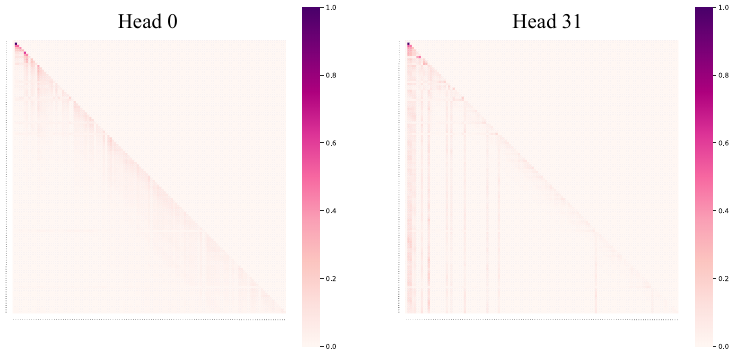}
    \caption{Our method CRFT (SAF).}
    \label{fig:dde_attention_score}
\end{subfigure}
\caption{\textbf{Visualization of attention scores for the first and last heads in the last layer.}}
\label{Fig: attentionscore_LLaMA-2-7B_layer31}
\end{figure*}

\subsection{How do critical representations impact information flow?}
We visualize attention maps to capture the variations in the information flow. The first and last heads in the final layer (Layer $32$) of both LLaMA-2-7B and our proposed method, CRFT (SAF), are illustrated in Figure~\ref{Fig: attentionscore_LLaMA-2-7B_layer31}. 
A comprehensive comparison of all heads is provided in Appendix~\ref{sec:vis},
and the phenomenon is consistent.
We have identified three observations, as follows:
\begin{itemize}
    \item \textbf{Excessive information interaction in the representation $\boldsymbol{h}_0$ is reduced.} In column $0$, the absence of prominent color indicates a diminished influence of representation $\boldsymbol{h}_0$ on other representations. The initial representation $\boldsymbol{h}_0$ in LLaMA-2-7B lacks semantic information, but attracts a high level of attention.
    Previous works~\citep{xiao2023efficient, yu2024unveiling} have referred to this phenomenon as the ``attention sink''. By applying our method, the representation $\boldsymbol{h}_0$ receives less undue attention, leading to a more balanced distribution of attention.\looseness=-1
    \item \textbf{Increased information interaction between representations.} The increase in the number of vertical lines signifies a heightened interaction among the representations.
    \item \textbf{Broader information flow.} The presence of high attention scores along the diagonal has shifted from a few isolated peaks to multiple cells. This denotes a broader information flow from various representations.
\end{itemize}
Based on the visualized results above, our method alters the direction of information flow, guiding it towards a more optimal path, and enriching the overall information interaction.

\section{Related Work}
\noindent\textbf{Intervention in LLMs.}
Intervention strategies encompass various techniques designed to influence the behavior of large-scale models during the inference phase. Common strategies include activation editing~\citep{li2024focus}, weight editing~\citep{dai2022can}, and the use of guidance vectors~\citep{zou2023representation}, as well as altering the output distribution through comparative analysis~\citep{li2022contrastive, chuang2023dola}.
As representations encode rich information, some methods~\citep{geiger2021causal, wu2024reft, alkhamissi2024llm} change the output by editing representations. Although representation interventions can serve as powerful tools for model control, previous methods intervene in representations based on empirical observations~\citep{wu2024advancing, wu2024reft} or general knowledge~\citep{zhang2023tell}. 
The above approaches are not general and time-consuming, which limits their adaptability and performance. In contrast, our method precisely identifies the representations to intervene.\looseness=-1

\noindent\textbf{Information Flow Analysis.}
Recent studies~\citep{yuaninstance,yandon,fanimproving} have utilized attention mechanisms to analyze their impact on model performance. For example, StreamLLM~\citep{xiao2023efficient} discovered that the initial token of an input text often receives an excessive amount of attention, despite frequently lacking semantic significance. It suggests that we should preserve these tokens when processing long input sequences to prevent forgetting. 
Furthermore, ACT~\citep{yu2024unveiling} found that attention sinks can occur not only at the initial token but also throughout the entire sequence. Moreover, it was discovered that these attention sinks are not always beneficial to model performance. ACT optimizes attention distributions during inference, but not all heads can benefit from the calibration. 
Similarly, PASTA~\citep{zhang2023tell} demonstrates that increasing the attention score of defined tokens in specific heads can improve the ability of LLM to follow instructions. However, tokens need to be manually defined.
Our method addresses these challenges by adaptively learning the updated direction of critical representations during training, leading to better overall performance.\looseness=-1

\section{Conclusion}
We propose a novel Chain-of-Thought (CoT) reasoning method, termed Critical Representation Fine-Tuning (CRFT), which focuses exclusively on critical representations to influence model outputs.
CRFT first identifies critical representations by analyzing the information flow through attention and saliency scores, and subsequently optimizes them via supervised fine-tuning within a low-rank subspace.
Comprehensive experiments conducted across various models and datasets validate the effectiveness and efficiency, providing a new perspective on CoT reasoning tasks, particularly in long CoTs. Furthermore, CRFT exhibits sufficient flexibility to be readily adapted to a few-shot learning scenarios, underscoring its potential to enhance reasoning capabilities within models.

\section*{Limitation}
For identification, we currently focus on searching for representations that significantly impact the model output. However, it is important to note that representations with minor impacts may still have an influence, even if their effects are often negligible.
A more effective strategy could involve prioritizing the correction of representations with negative impacts, although identifying such representations remains a challenge. 
Furthermore, while our optimizations are currently restricted to linear spaces, there is potential to explore alternative optimization methods that could enhance our framework.\looseness=-1

\section*{Acknowledgements}
This work was supported in part by the National Nature Science Foundation of China (Grant Nos: 62273302, 62273303, 62036009), in part by the Yongjiang Talent Introduction Programme (Grant Nos.: 2022A-240-G), in part by Ningbo Key R\&D Program (Nos.: 2023Z231, 2023Z229), in part by the Alibaba Research Intern Program. We sincerely thank Lei Li for his suggestions in writing.

\clearpage

\bibliography{custom}

\newpage
\pagebreak

\appendix
\section{Implement Details}
\label{sec:implement_details}
\subsection{Datasets}
The test datasets that we use across two scenarios covering eight datasets: GSM8K~\citep{cobbe2021gsm8k}, AQuA~\citep{ling2017program}, MAWPS~\citep{koncel-kedziorski-etal-2016-mawps}, SVAMP~\citep{patel2021nlp}, BoolQ~\citep{clark2019boolq}, SocialIQA~\citep{sap2019socialiqa}, WinoGrande~\citep{sakaguchi2021winogrande}, and OpenBookQA~\citep{mihaylov2018can}.

\noindent\textbf{GSM8K.} GSM8K, which comprises grade-school math word problems requiring multi-step reasoning, usually takes between $2$ and $8$ steps to solve problems using basic arithmetic operations $+, -, \times, \div$. We used the last $300$ samples in the training set as the validation set and reported the results on its test set.

\noindent\textbf{Arithmetic Reasoning Scenarios.} Following the experimental setup established in~\citet{hu2023llm}, we fine-tune a combined dataset of seven arithmetic reasoning tasks, called Math10K, utilizing LM-generated chain-of-thought steps. We report performance metrics in three test sets: AQuA, MAWPS, and SVAMP.\looseness=-1

\noindent\textbf{Commonsense Reasoning Scenarios.} For commonsense reasoning scenarios, we opted not to use Commonsense170K from~\citet{hu2023llm}, as it does not incorporate CoT steps. We create a suitable Commonsense60k training set, combining six commonsense reasoning tasks: CommonsenseQA~\citep{talmor2018commonsenseqa}, CoS-e~\citep{rajani2019explain}, OpenBookQA~\citep{mihaylov2018can}, SocialIQA~\citep{sap2019socialiqa}, StrategyQA~\citep{geva2021strategyqa}, WorldTree~\citep{jansen2018worldtree}.
We report performance metrics in four test sets: BoolQ, SocialIQA, WinoGrande, and OpenBookQA.\looseness=-1

\subsection{Base Models}
We finetune our models on LLaMA-2-7B, LLaMA-2-13B, LLaMA-3-8B and Mistral-7B.
We use the ``chat'' version of LLaMA-2, and the ``instruct'' version of LLaMA-3 and Mistral-7B.

\subsection{Hyperparameters}
For a fair comparison with ReFT ($p7+s7$), we selected $14$ intervention representations and maintained a rank of $8$, consistent with the parameters used in ReFT. We set the hyperparameters of $\alpha$ to $0.05$. And we used the ``order'' selection criteria by default.
To ensure a fair comparison, we maintain the same training principle, details are shown in Table~\ref{Tab: hyperparameters}.
For all tasks, model outputs are generated with greedy search.

\begin{table}[tb]
\centering
\resizebox{\linewidth}{!}{
\begin{tabular}{@{}l|c@{}}
\toprule
\textbf{HyperParameters} & \textbf{Values}\\
\midrule
Rank & $8$\\
Number of representations & $14$\\
threshold $\alpha$ & $0.05$\\
selection criteria & Order(default) \\
\hline
\multirow{2}{*}{Number of Epochs} & $12$ for arithmetic,\\
& $6$ for commonsense\\
Batch Size & $2$\\
Gradient accumulation steps & $16$\\
seed & $42$\\
Optimizer & AdamW\\
Learning Rate Schedule & Linear\\
Learning  Rate & $9e-4$\\
dropout & $0.05$ for GSM8K, $0$ for others\\
Weight Decay & $0.06$ for GSM8K, $0$ for others\\
Warmup ratio & $0$ for GSM8K, $0.1$ for others\\
\bottomrule
\end{tabular}}
\caption{\textbf{The values of hyperparameters.}}
\label{Tab: hyperparameters}
\end{table}

\begin{table*}[tbhp]
\centering
\begin{tabular}{@{}lccccc@{}}
\toprule
\multirow{2}{*}{\textbf{PEFT}} & \multirow{2}{*}{\textbf{Identify}} & \multicolumn{4}{c}{\textbf{Accuracy ($\uparrow$)}}\\
\cmidrule{3-6}
& & LLaMA-2-7B & LLaMA-2-13B & LLaMA-3-8B & Mistral-7B\\
\midrule
None & - & 14.6 & 30.9 & 64.5 & 38.4\\
ReFT & p7+s7 & 29.0 & 37.9 & 64.7 & 46.5\\
\hline
\multirow{6}{*}{CRFT (ours)} & SAF & 30.4 | 29.6 & 38.7 | 39.6 & \underline{70.8} | 70.6 & 46.4 | 46.9\\
& MAF & 32.0 | \underline{32.1} & 38.3 | 38.0 & 67.5 | 64.8 & 48.0 | 47.3\\
& Union(attn) & 31.2 | \textbf{32.8} & \textbf{40.3} | 39.4 & 64.4 | \textbf{71.0} & \underline{48.1} | 47.7\\
\cmidrule{2-6}
& SSF & 31.4 | 30.4 & \underline{40.1} | 38.4 & 64.6 | 64.5 & 46.4 | 46.5\\
& MSF & 31.4 | 30.3 & 38.3 | 38.3 & 64.5 | 65.1 & 46.9 | 47.7\\
& Union(sal) & \textbf{32.8} | 31.5  & 38.3 | 38.3 & 63.8 | 64.0 & 48.0 | \textbf{48.2}\\
\bottomrule
\end{tabular}
\caption{\textbf{Quantitative comparison on GSM8K with four base models: LLaMA-2-7B, LLaMA-2-13B, LLaMA-3-8B, and Mistral-7B.} 
The best performance is highlighted in \textbf{bold}, while the second-best is \underline{underlined}. }
\label{Tab: gsm8k}
\end{table*}

\subsection{Prompt}
We use a prompt for each task.
\begin{tcolorbox}[colback=white, colframe=black, title=GSM8K]
[question]

Answer the above question. First, think step by step and then answer the final number.
\end{tcolorbox}

\begin{tcolorbox}[colback=white, colframe=black, title=Other Arithmetic Scenario]
Below are instructions for a task. Write a response that appropriately completes the request.

\#\#\# Instruction:

[Question]

\#\#\# Response:
\end{tcolorbox}
\begin{tcolorbox}[colback=white, colframe=black, title=Commonsense Scenario]
[Question]

the correct answer is 
\end{tcolorbox}

\section{Further Ablation Studies}
To further demonstrate the effectiveness of our proposed method CRFT, we provide additional experimental results. 

\subsection{Performance under Different Base Models}\label{sec:moremodels}
We verify our method, CRFT, in different models, as shown in Table~\ref{Tab: gsm8k}. 
We tested four basic models (LLaMA-2-7B, LLaMA-2-13B, LLaMA-3-8B, Mistral-7B) on the GSM8K dataset, covering different sizes and families.
The consistent improvement of our experimental results demonstrates the effectiveness of our method.

\begin{table}[tbhp]
\centering
\begin{tabular}{@{}lcc@{}}
\toprule
\textbf{PEFT} & \textbf{Identify} & \textbf{Accuracy ($\uparrow$)}\\
\midrule
ReFT & - & 29.0\\
\midrule
\multirow{6}{*}{CRFT(ours)} & SAF & 31.1 | \textbf{33.1}\\
& MAF & 32.2 | 30.3\\
& Union(attn) & \underline{32.9} | \underline{32.9}\\
\cmidrule{2-3}
& SSF & 32.0 | 32.2\\
& MSF & 31.8 | 30.0\\
& Union(attn) & 30.6 | 31.8\\
\bottomrule
\end{tabular}
\caption{\textbf{Quantitative performance on GSM8K using LLaMA-2-7B with $\alpha=0.01$.} 
}
\label{Tab: compare2}
\end{table}

\subsection{Performance under Optimized Parameter Configuration}
We present the performance of our approach on the GSM8K dataset using the LLaMA-2-7B base model with an optimized threshold of $0.01$ in Table~\ref{Tab: compare2}. These results indicate that setting the threshold to $0.01$ indeed leads to improved performance. 

\subsection{Efficacy of the Union Strategy}
We present supplementary experiments to validate the efficacy of the Union strategy, in Table~\ref{Tab: union}. Specifically, we evaluate our method on the AQuA (mathematics) and BoolQ (commonsense reasoning) datasets, employing both LLaMA-2-7B and LLaMA-3-8B as base models. The results demonstrate that the Union strategy consistently achieves gains without requiring manual selection.

\begin{table*}[tb]
\centering
\begin{tabular}{@{}lccc|cc@{}}
\toprule
\multirow{3}{*}{\textbf{PEFT}} & \multirow{3}{*}{\textbf{Identify}} & \multicolumn{4}{c}{\textbf{Accuracy ($\uparrow$)}}\\
\cmidrule{3-6}
& & \multicolumn{2}{c}{\textbf{AQuA}} & \multicolumn{2}{c}{\textbf{BoolQ}}\\
& & LLaMA-2-7B & LLaMA-3-8B & LLaMA-2-7B & LLaMA-3-8B\\
\midrule
ReFT & - & 21.7 & 46.9 & 50.7 & 62.1\\
\midrule
\multirow{6}{*}{CRFT(ours)} & SAF & 25.6 | 26.0 & 47.2 | 47.2 & 60.0 | 53.7 & 63.0 | 64.4\\
& MAF & 27.6 | 24.8 & 48.4 | 50.4 & 60.5 | \underline{61.8} & 62.4 | 66.2\\
& Union(attn) & \underline{28.4} | \textbf{30.3} & 49.6 | \underline{53.2} & 57.4 | 57.5 & \textbf{69.4} | 66.4\\
\cmidrule{2-6}
& SSF & 26.0 | 26.8 & 50.0 | 49.2 & \textbf{62.0} | 54.3 & 64.0 | 66.6\\
& MSF & 27.2 | 22.8 & 48.0 | 51.6 & 60.0 | 59.7 & 67.0 | 67.9\\
& Union(sal) & 26.8 | 27.6 & \textbf{53.5} | \underline{53.2} & 59.7 | 61.7 & \underline{68.8} | 67.8\\
\bottomrule
\end{tabular}
\caption{\textbf{The efficacy of the union strategy.} We present the results with LLaMA-2-7B and LLaMA-3-8B on AQuA and BoolQ.
The best performance is highlighted in \textbf{bold}, while the second-best is \underline{underlined}.}
\label{Tab: union}
\end{table*}

\section{Generation Examples}\label{sec:case}
We give a generation example for each test dataset. 

\begin{figure*}[h]
\centering
\includegraphics[width=0.8\textwidth]{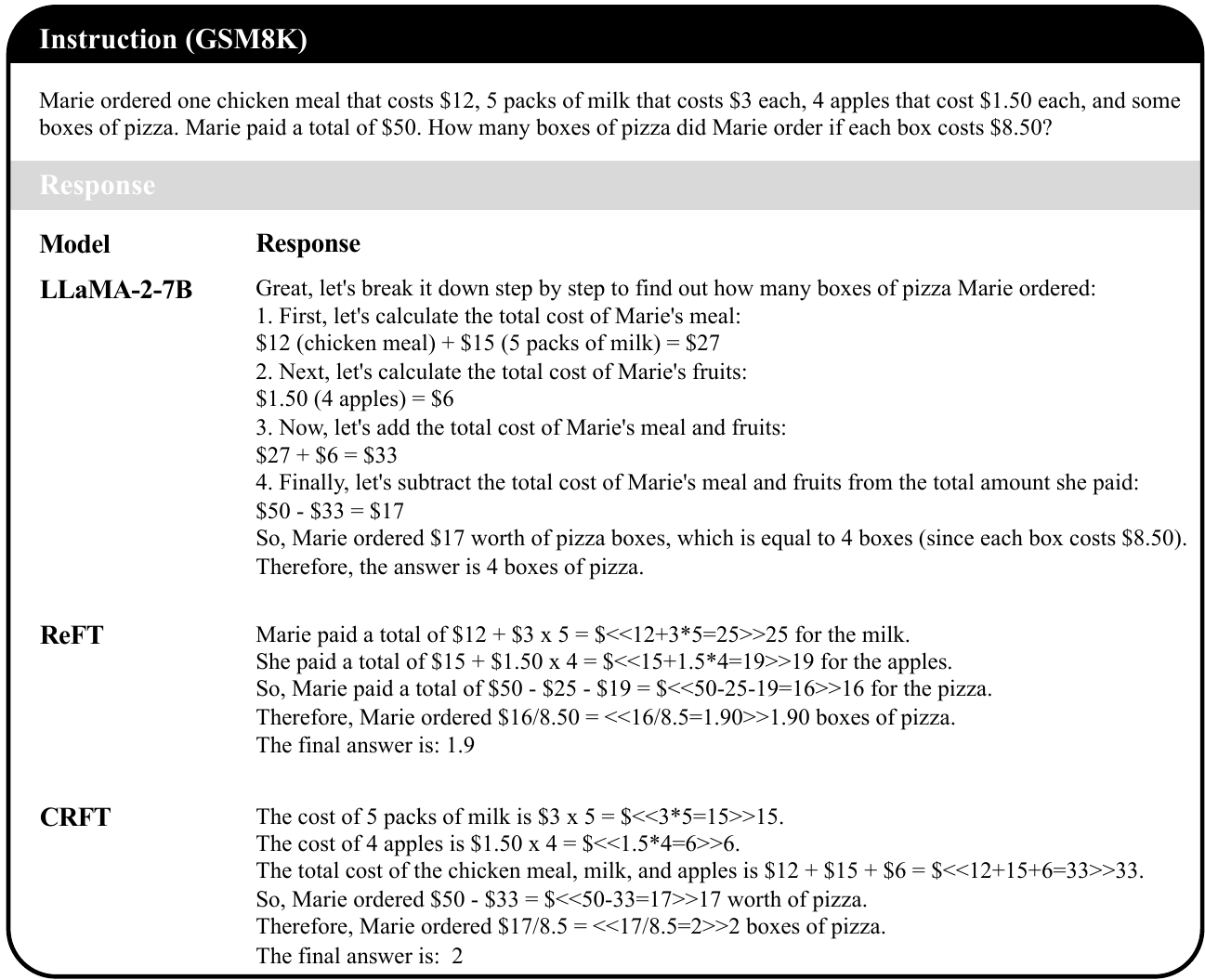}
\label{Fig: GSM8K}
\end{figure*}

\begin{figure*}[h]
\centering
\includegraphics[width=0.8\textwidth]{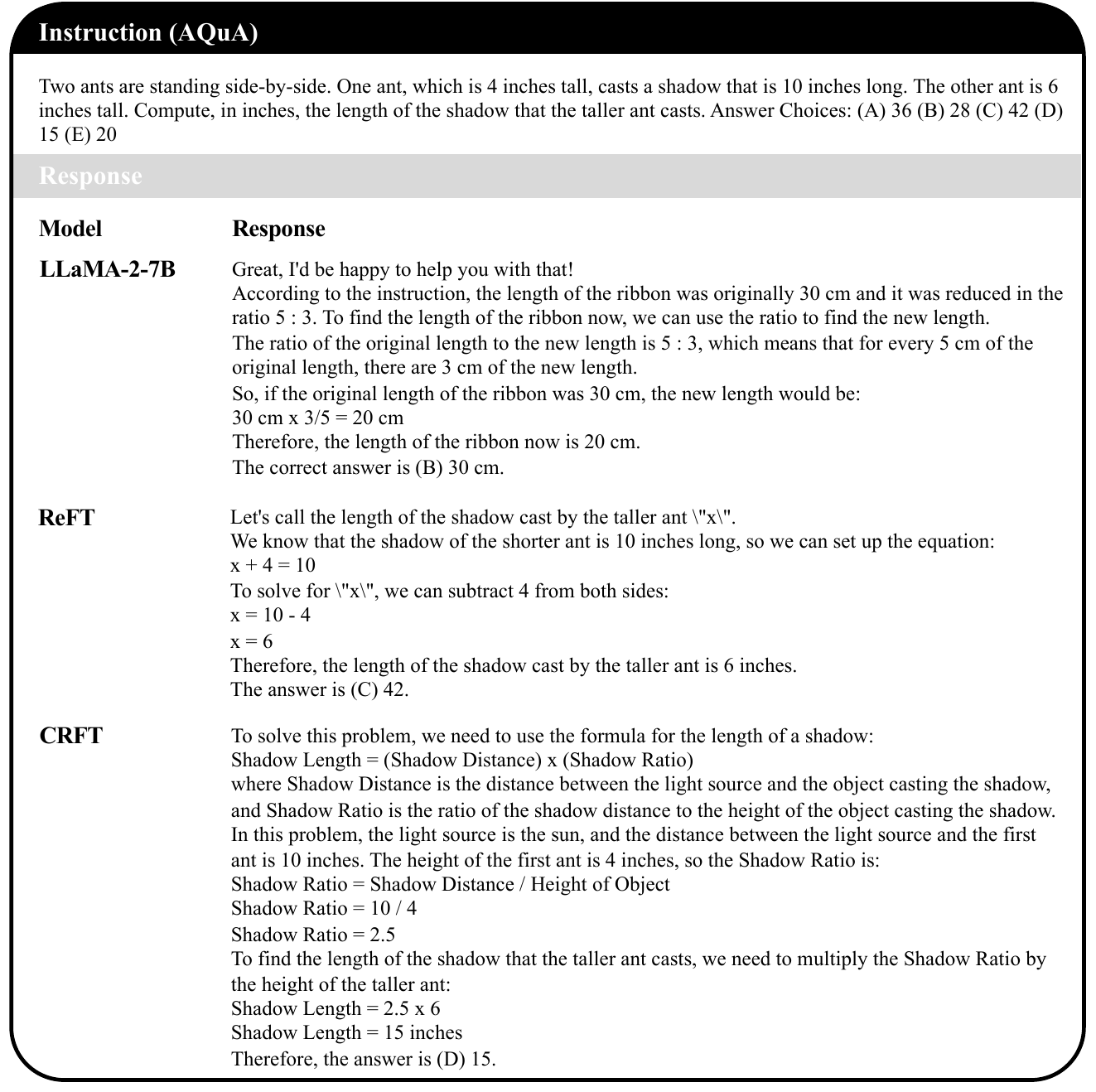}
\label{Fig: AQuA}
\end{figure*}

\begin{figure*}[h]
\centering
\includegraphics[width=0.8\textwidth]{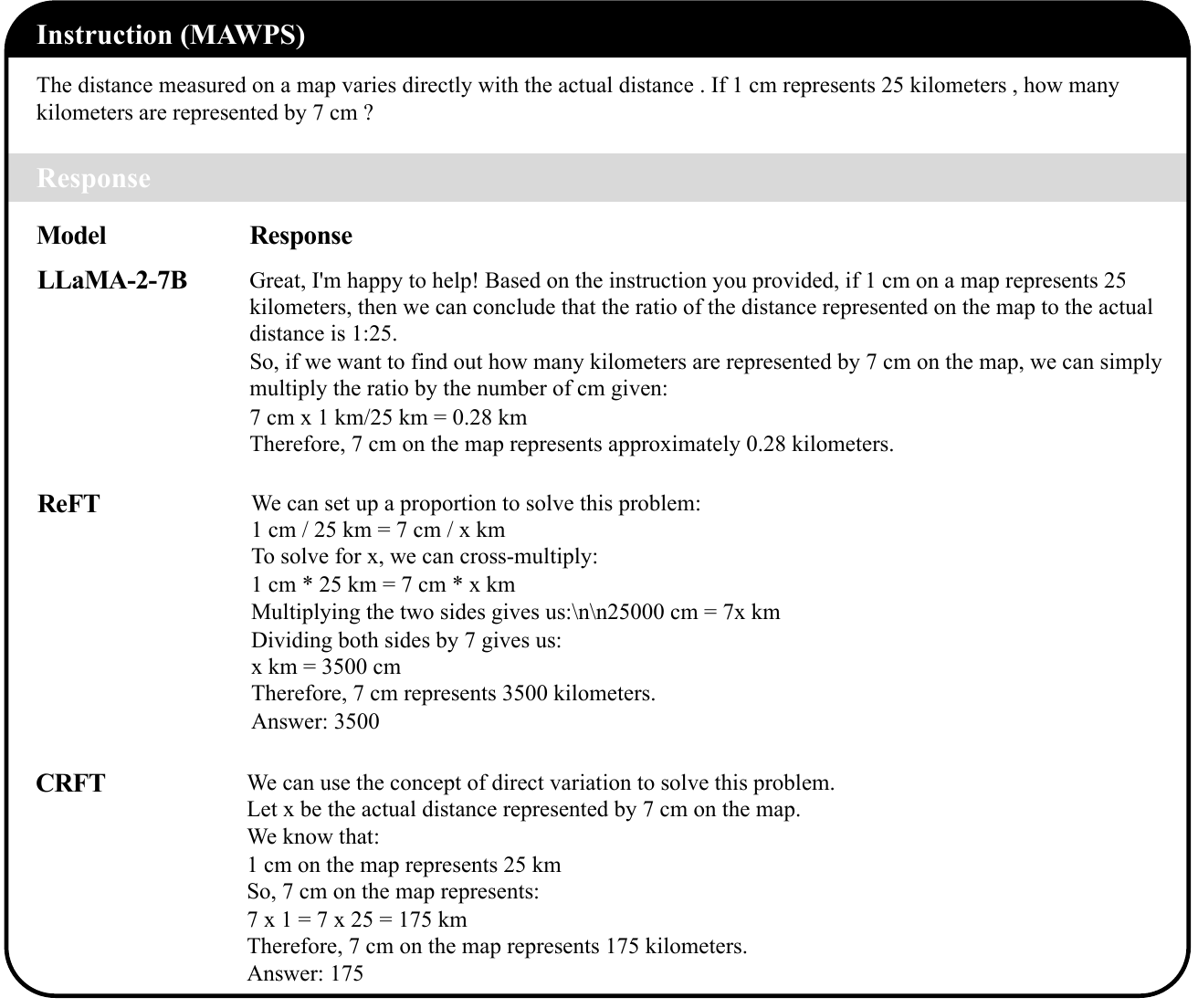}
\label{Fig: MAWPS}
\end{figure*}

\begin{figure*}[h]
\centering
\includegraphics[width=0.8\textwidth]{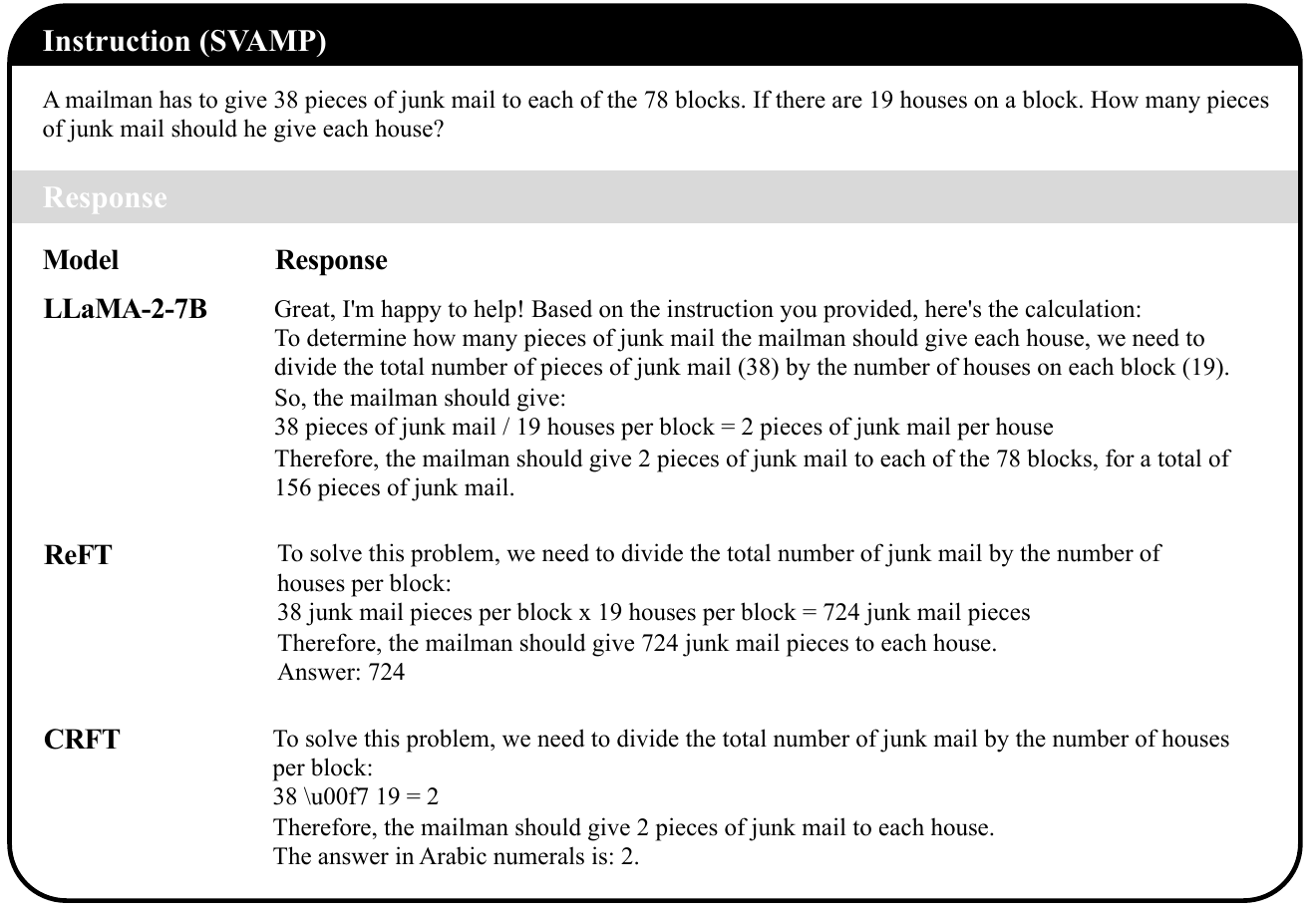}
\label{Fig: SVAMP}
\end{figure*}

\begin{figure*}[h]
\centering
\includegraphics[width=0.8\textwidth]{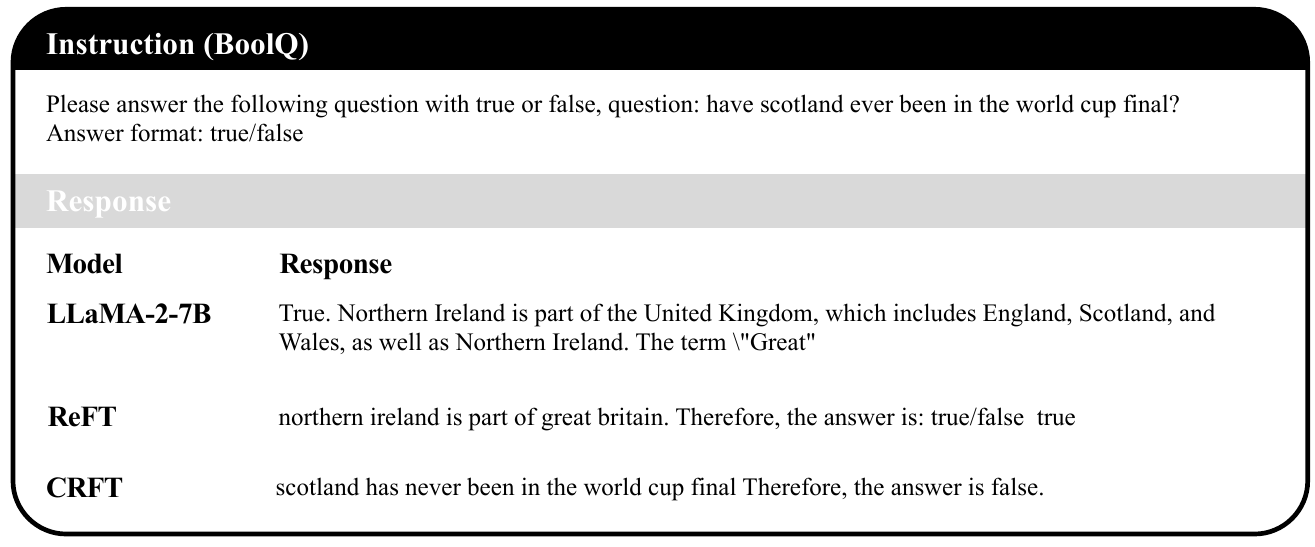}
\label{Fig: BoolQ}
\end{figure*}

\begin{figure*}[htbp]
\centering
\includegraphics[width=0.8\textwidth]{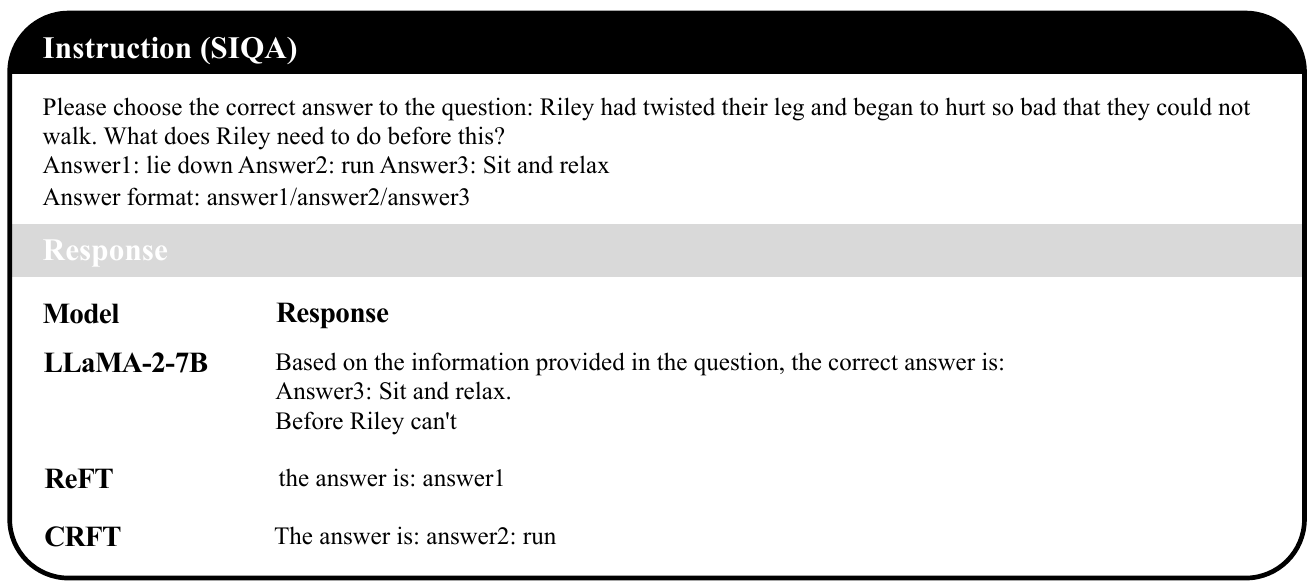}
\label{Fig: SIQA}
\end{figure*}

\begin{figure*}[htbp]
\centering
\includegraphics[width=0.8\textwidth]{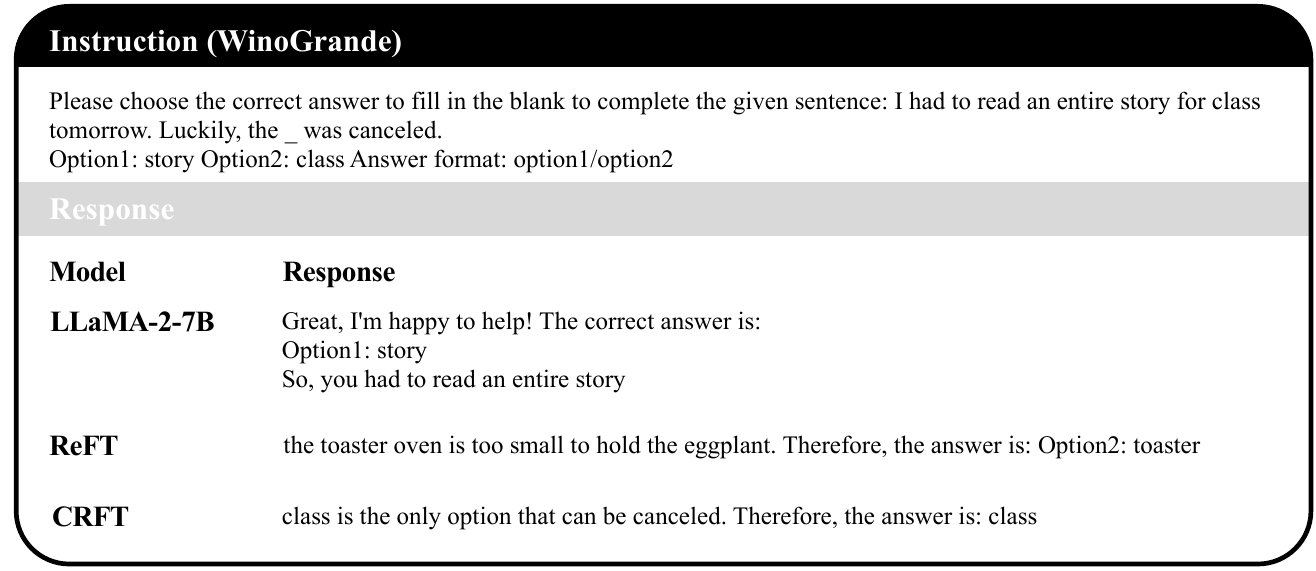}
\label{Fig: WinoG}
\end{figure*}

\begin{figure*}[htbp]
\centering
\includegraphics[width=0.8\textwidth]{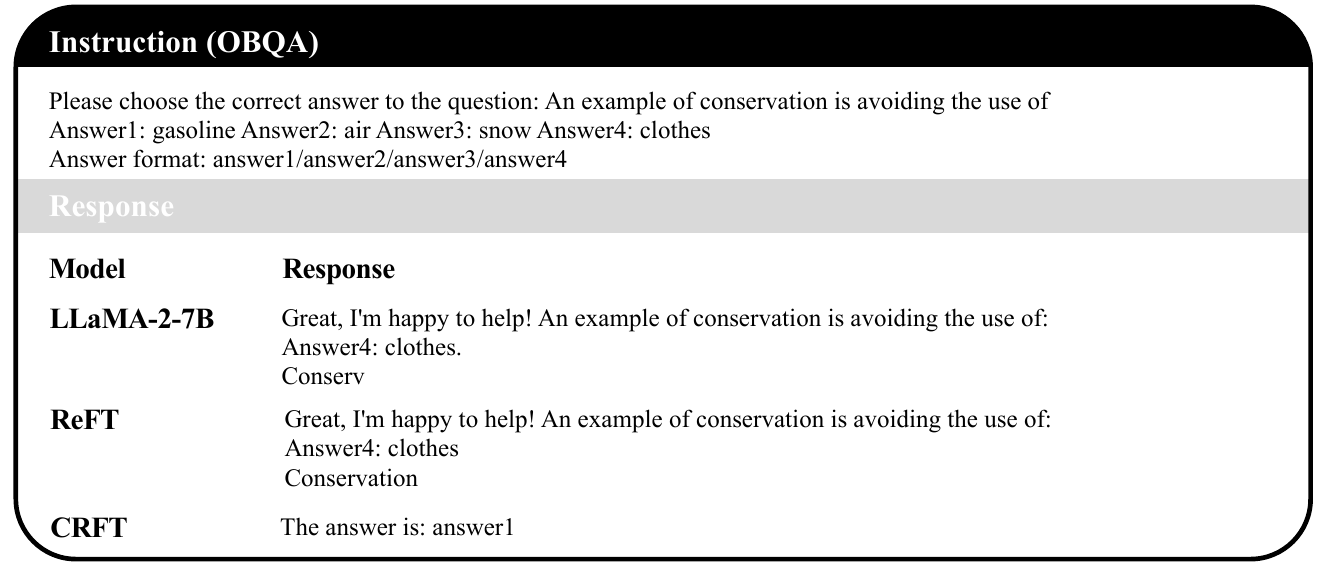}
\label{Fig: OBQA}
\end{figure*}

\section{Attention Analysis}\label{sec:vis}
We visualize the attention score of all $32$ heads in the final layer and the last head in all layers, which illustrates that our method indeed enriches information interactions.

\begin{figure*}[hp]
\centering
    \begin{minipage}{0.23\textwidth}
        \includegraphics[width=\linewidth]{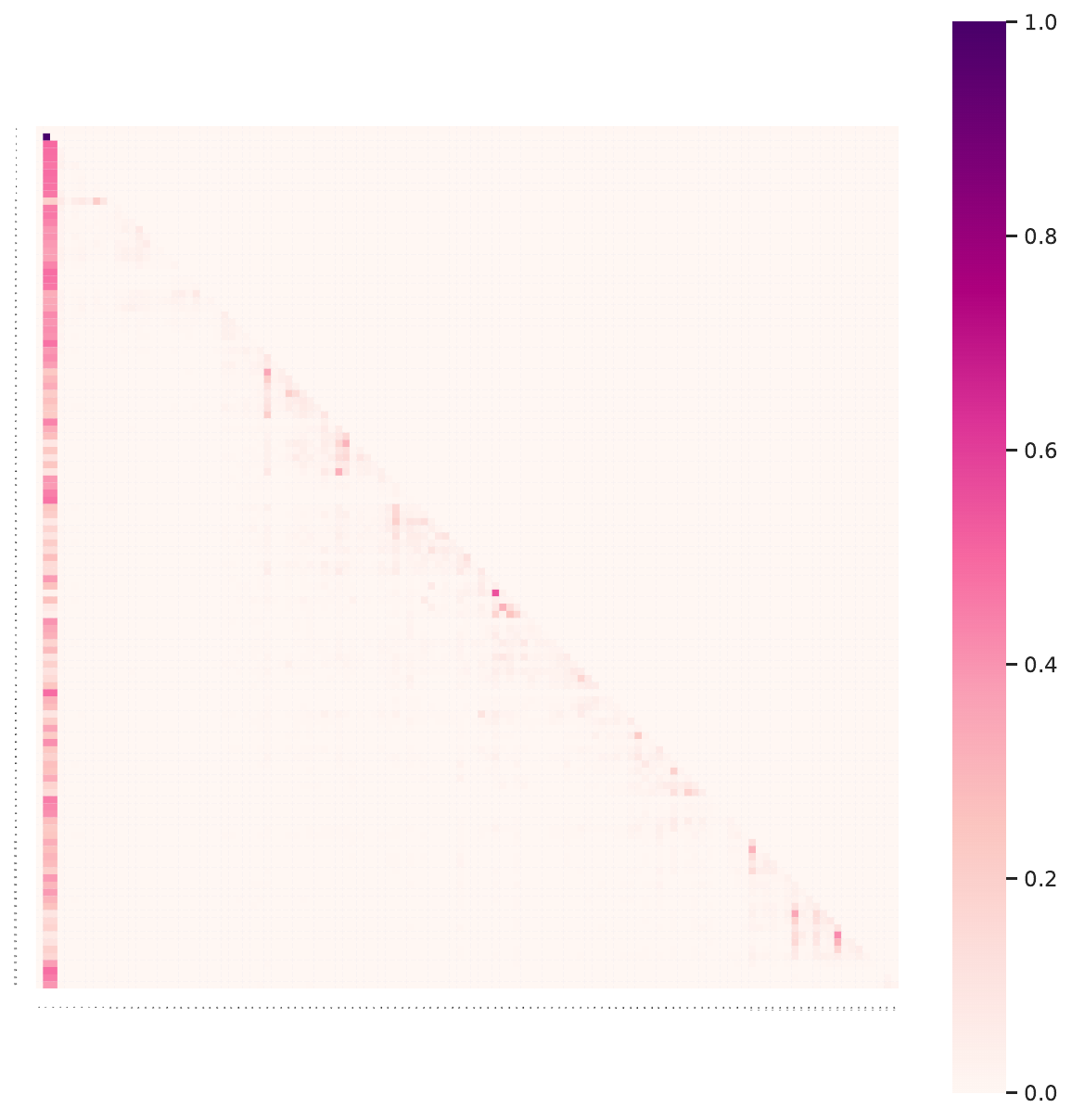}
        \caption*{Head 0}
    \end{minipage}
    \begin{minipage}{0.23\textwidth}
        \includegraphics[width=\linewidth]{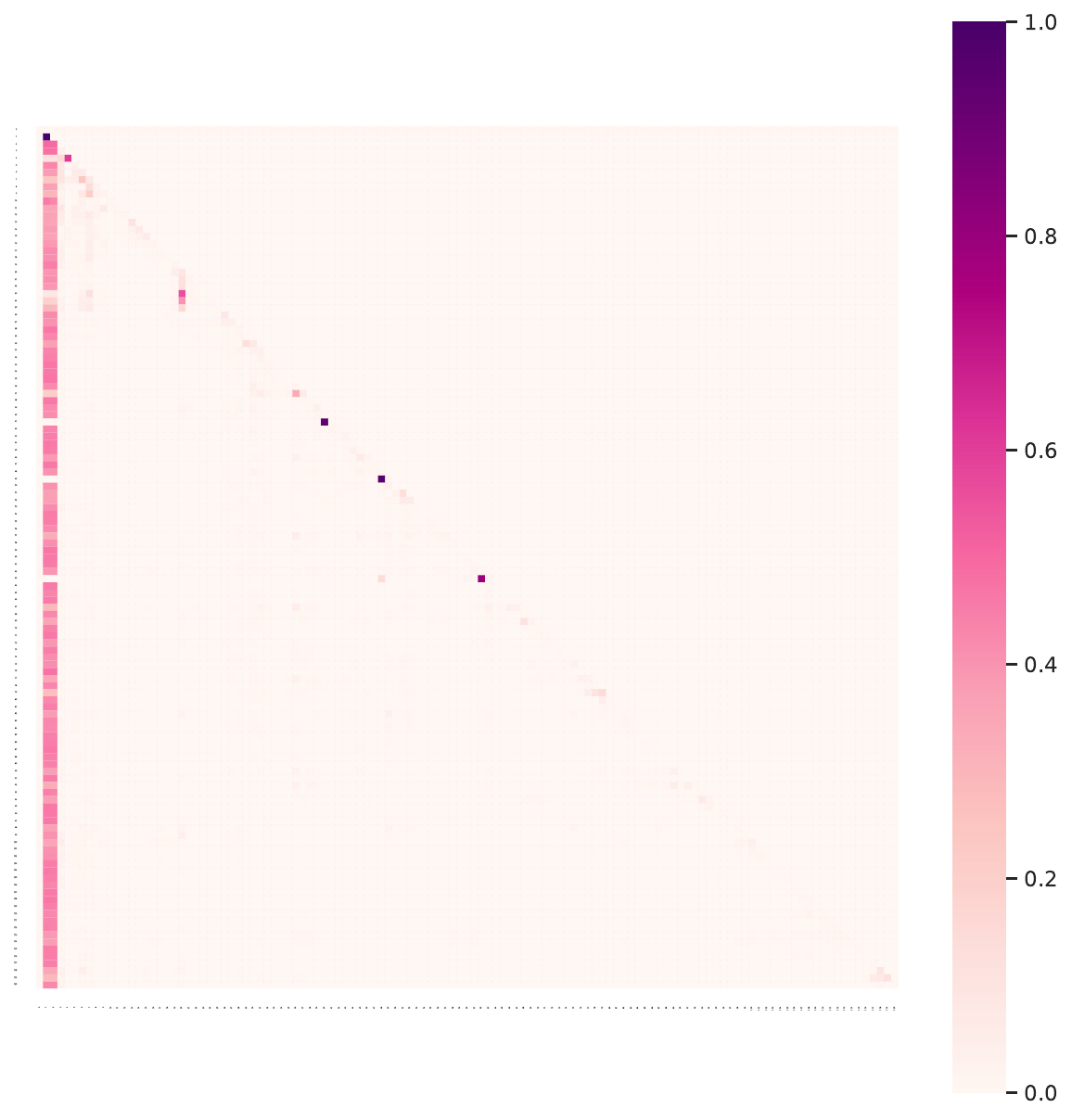}
        \caption*{Head 1}
    \end{minipage}
    \begin{minipage}{0.23\textwidth}
        \includegraphics[width=\linewidth]{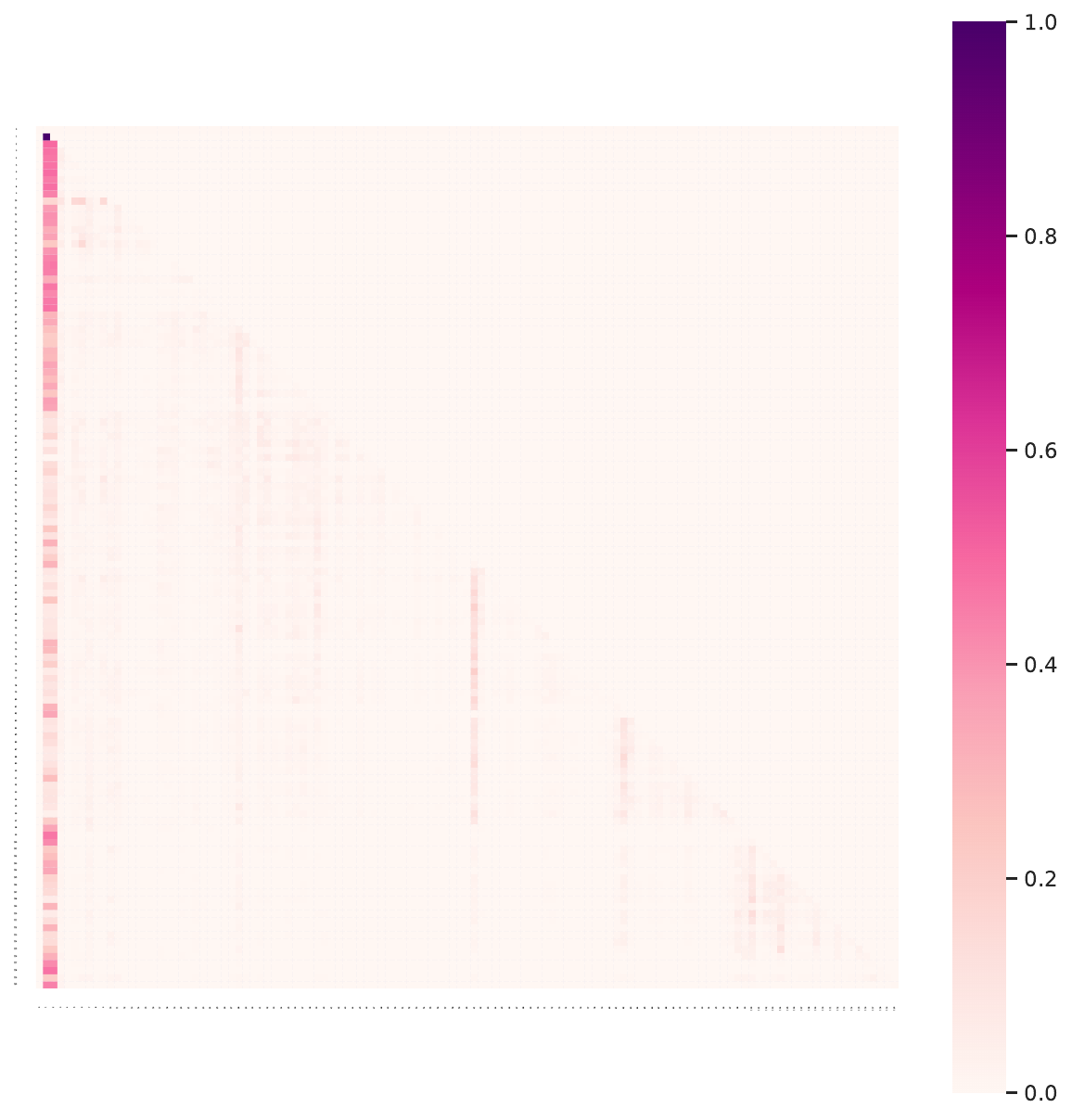}
        \caption*{Head 2}
    \end{minipage}
    \begin{minipage}{0.23\textwidth}
        \includegraphics[width=\linewidth]{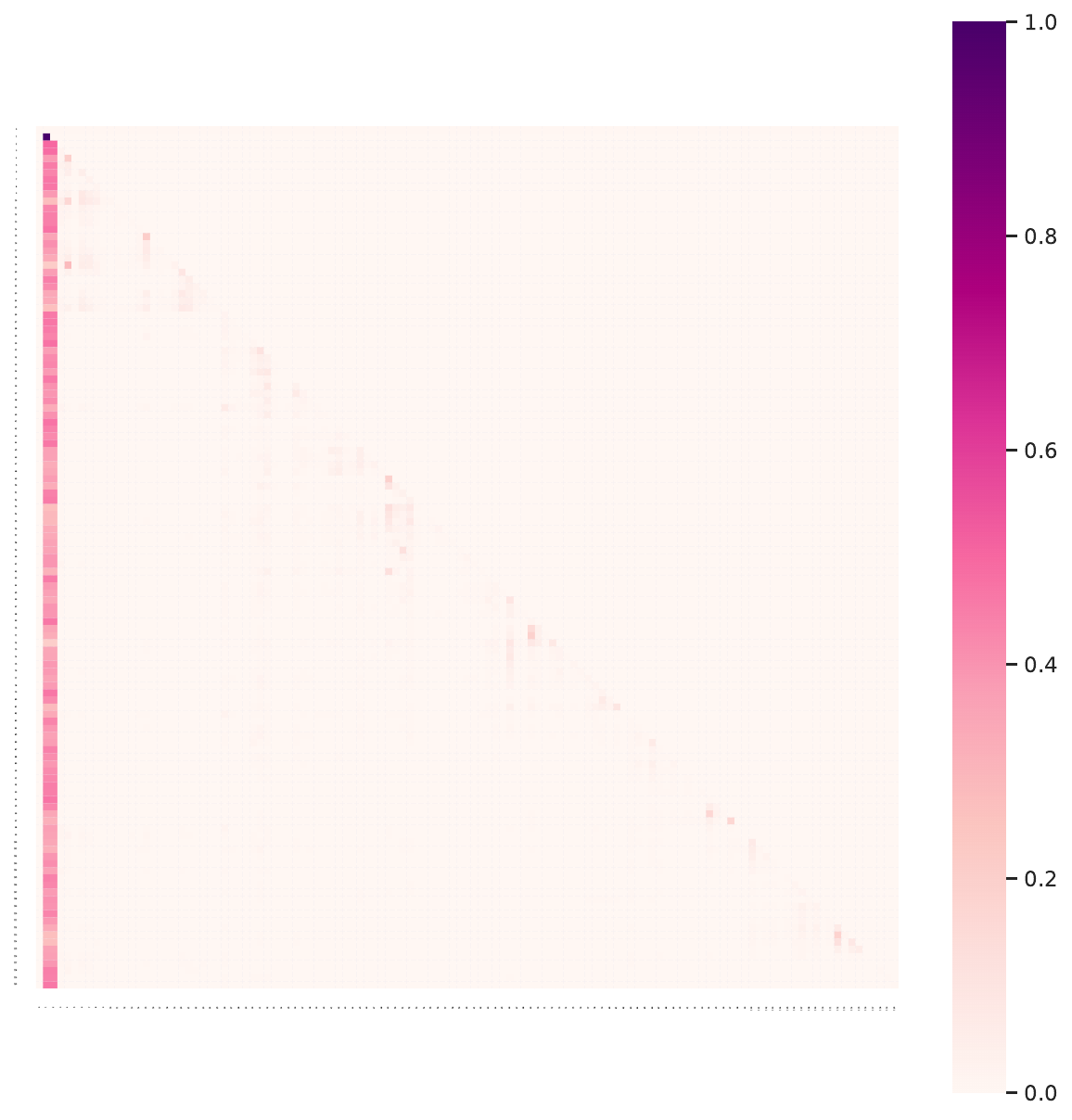}
        \caption*{Head 3}
    \end{minipage}
\vspace{1em}
\begin{minipage}{0.23\textwidth}
        \includegraphics[width=\linewidth]{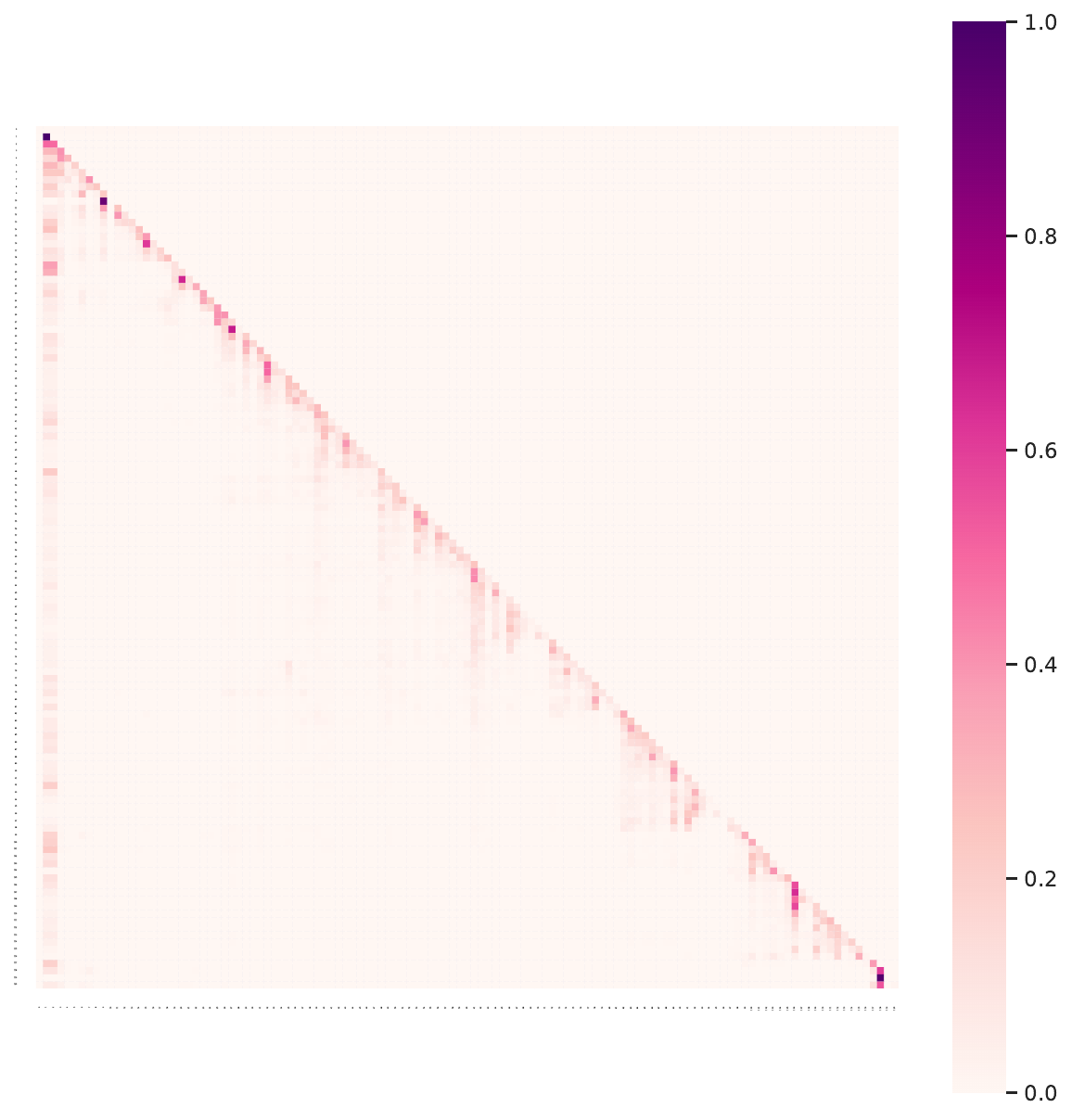}
        \caption*{Head 4}
    \end{minipage}
    \begin{minipage}{0.23\textwidth}
        \includegraphics[width=\linewidth]{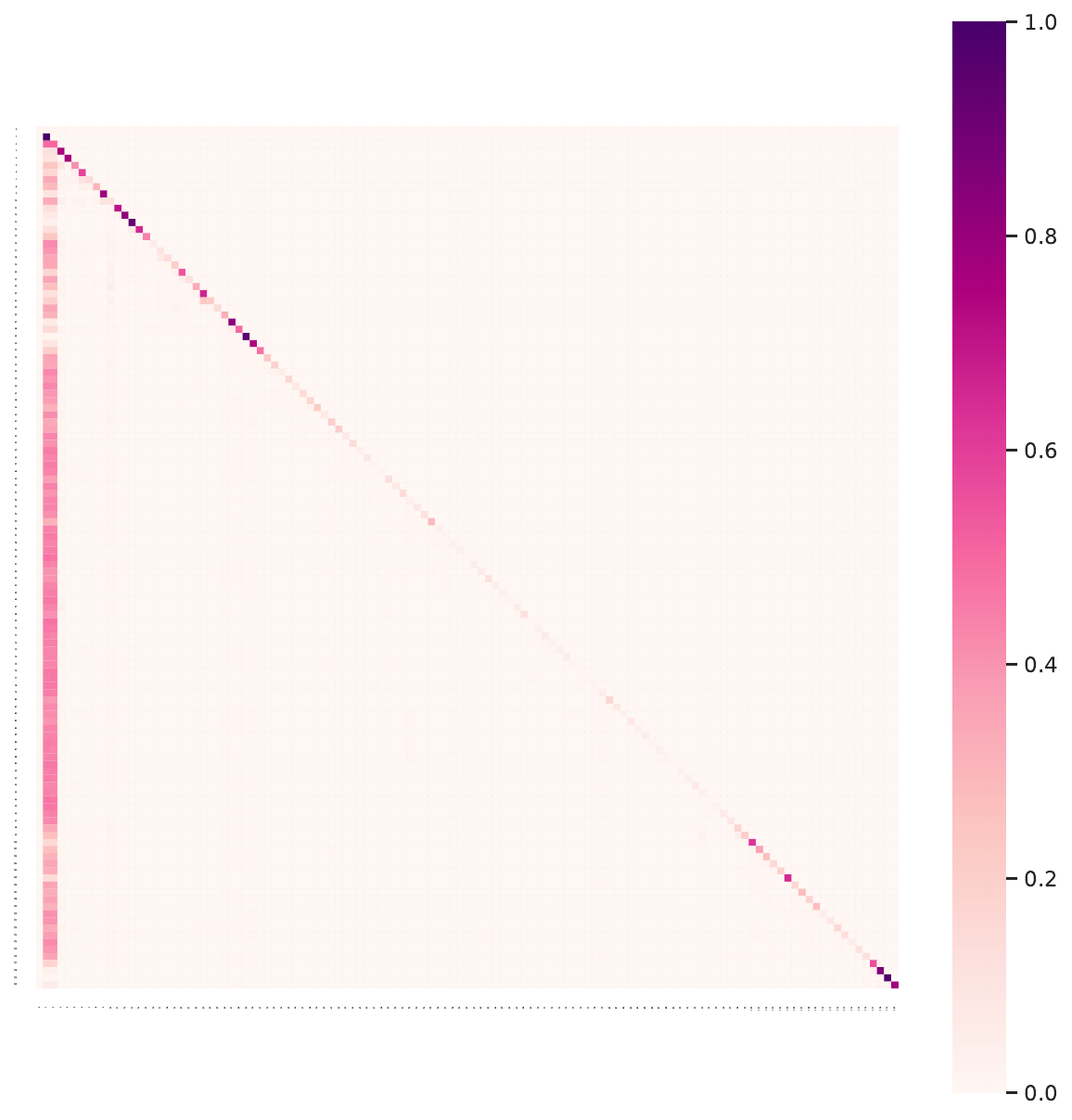}
        \caption*{Head 5}
    \end{minipage}
    \begin{minipage}{0.23\textwidth}
        \includegraphics[width=\linewidth]{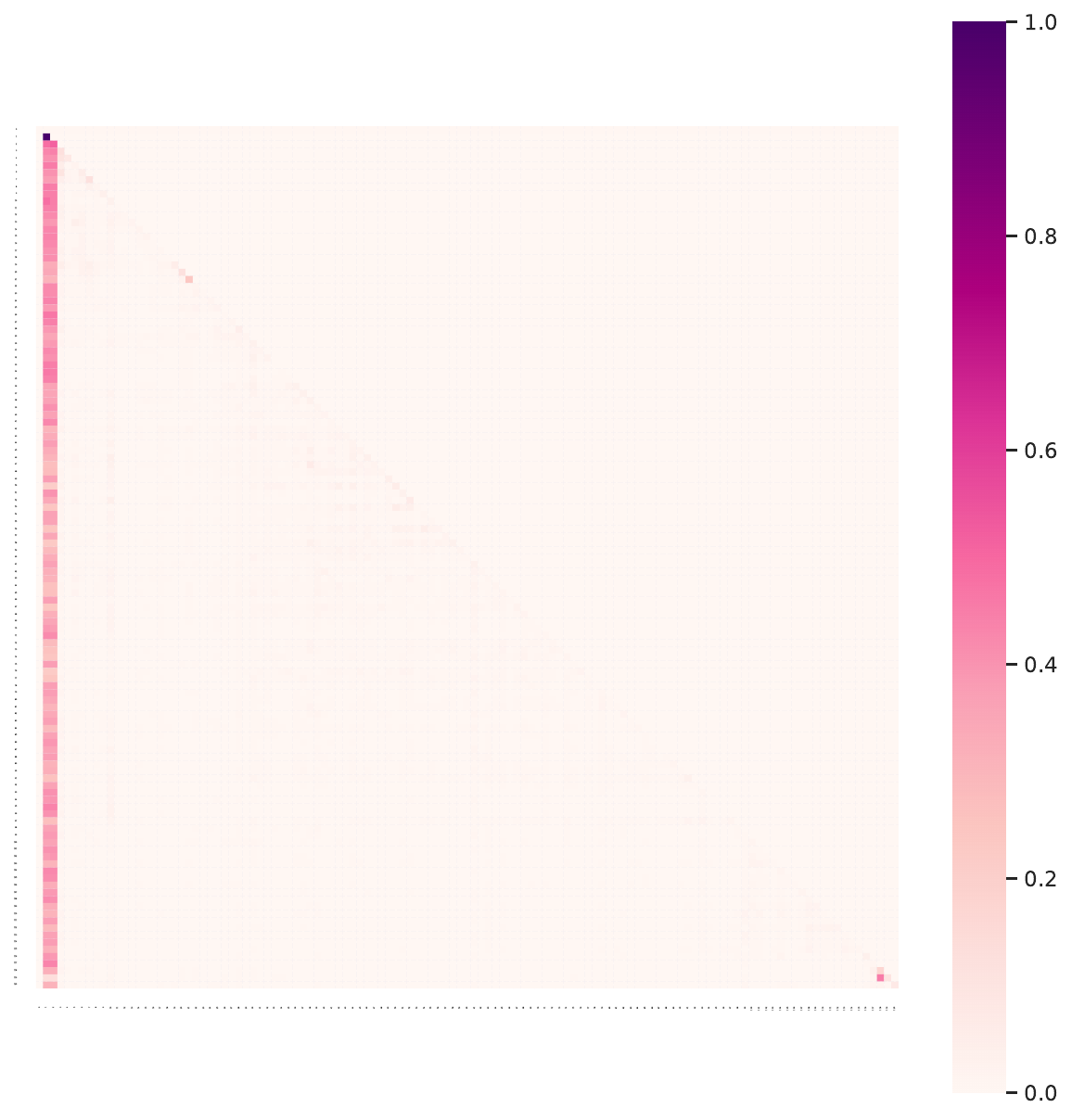}
        \caption*{Head 6}
    \end{minipage}
    \begin{minipage}{0.23\textwidth}
        \includegraphics[width=\linewidth]{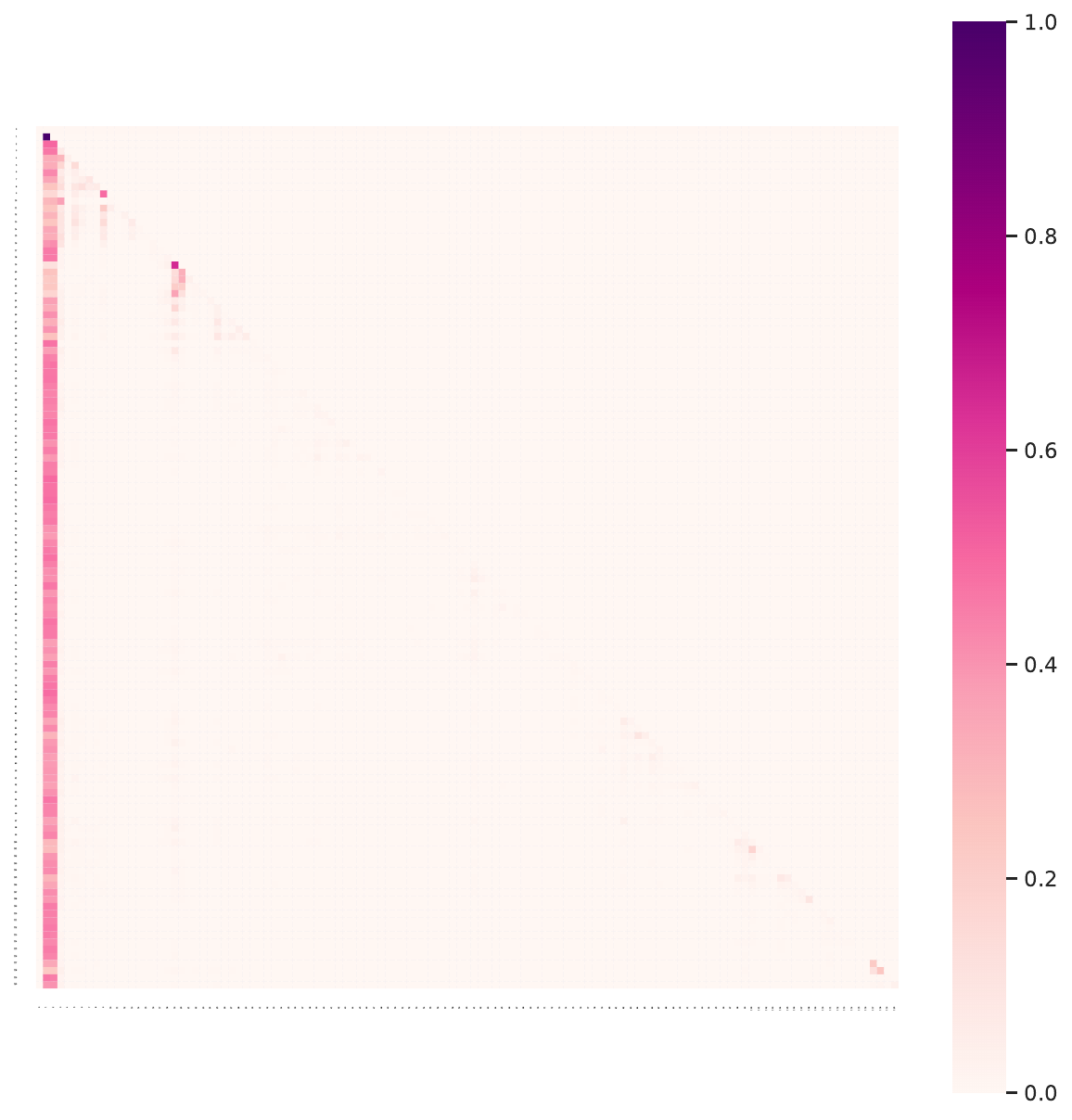}
        \caption*{Head 7}
    \end{minipage}
\vspace{1em}
\begin{minipage}{0.23\textwidth}
        \includegraphics[width=\linewidth]{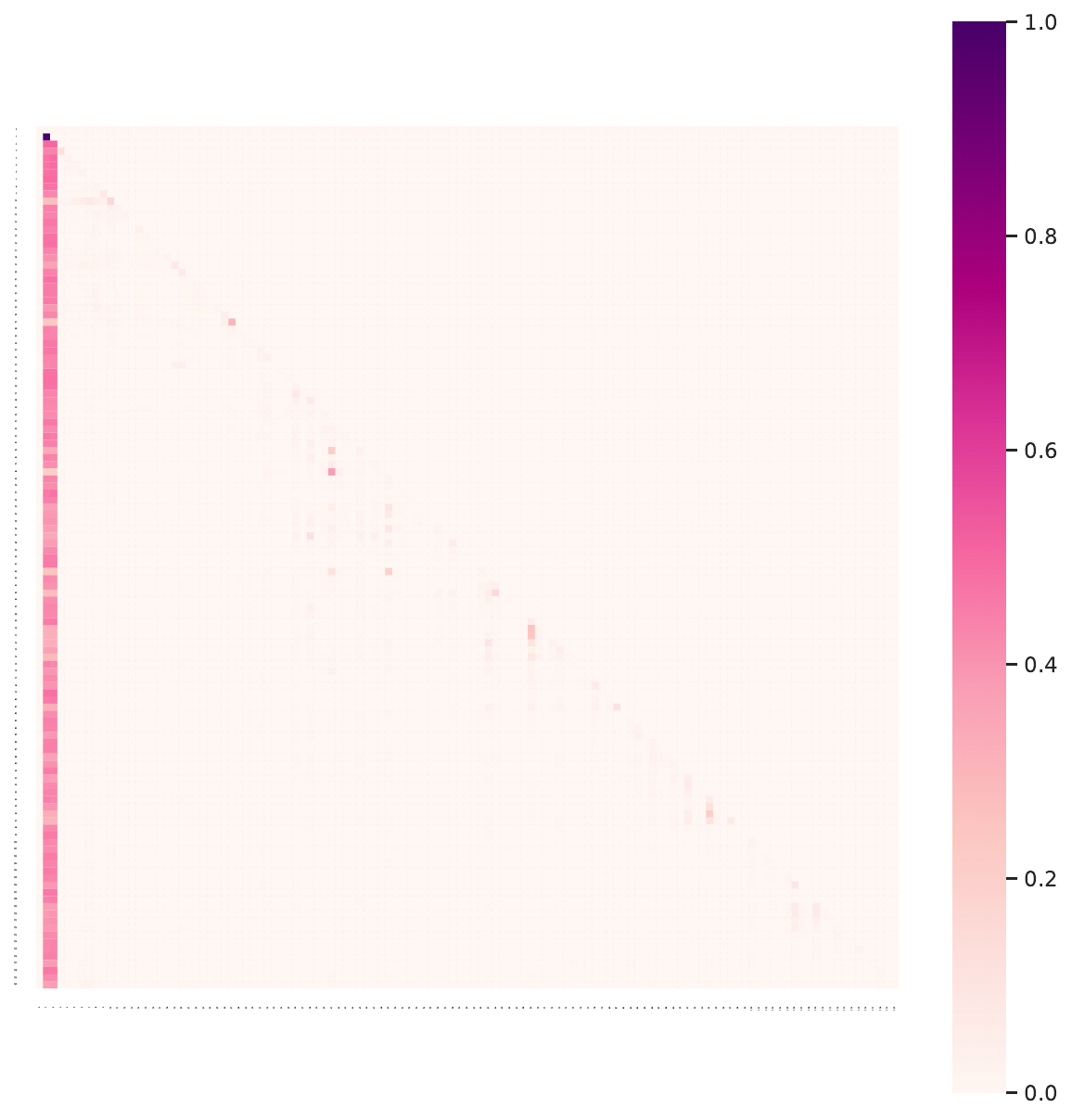}
        \caption*{Head 8}
    \end{minipage}
    \begin{minipage}{0.23\textwidth}
        \includegraphics[width=\linewidth]{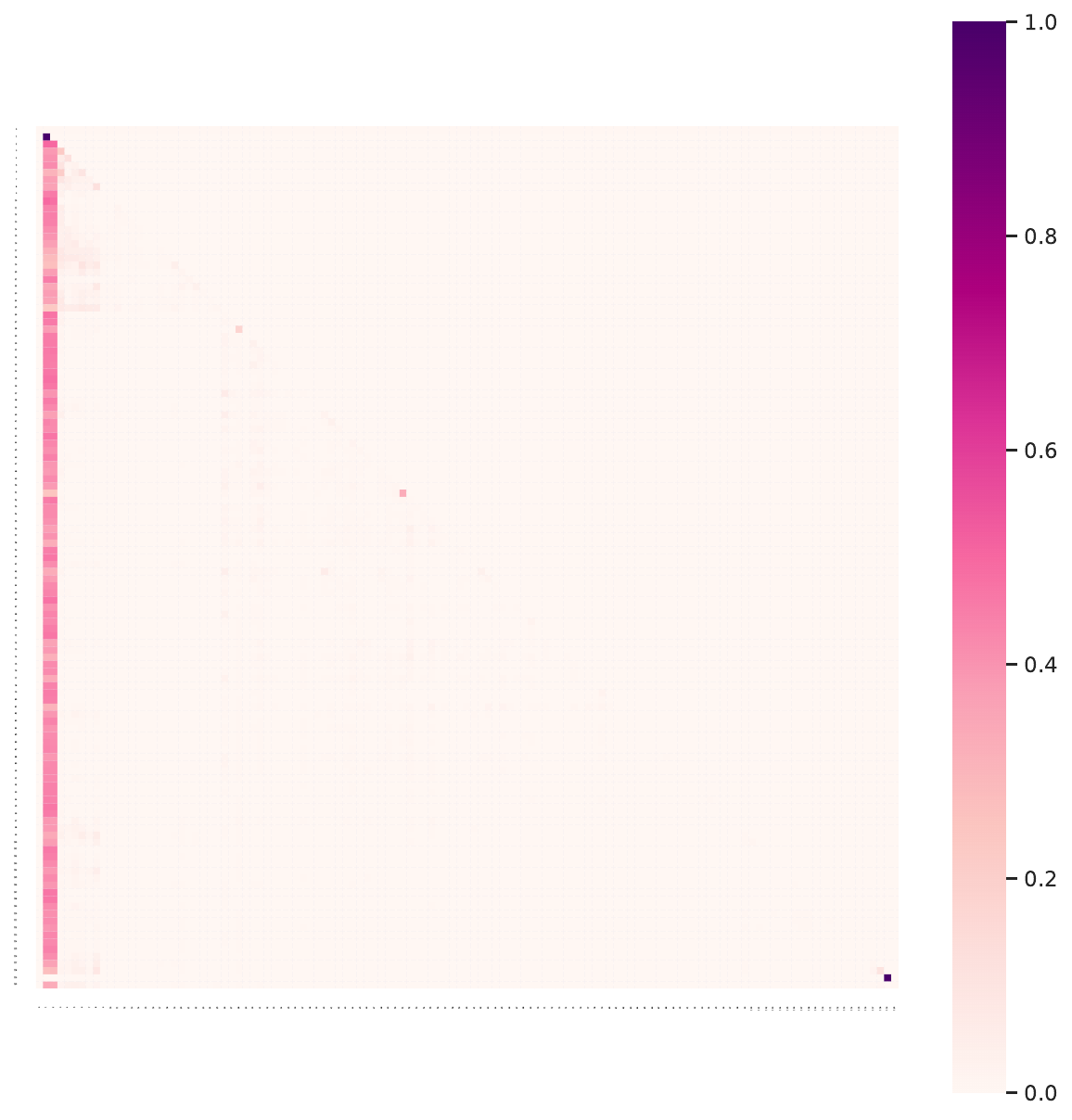}
        \caption*{Head 9}
    \end{minipage}
    \begin{minipage}{0.23\textwidth}
        \includegraphics[width=\linewidth]{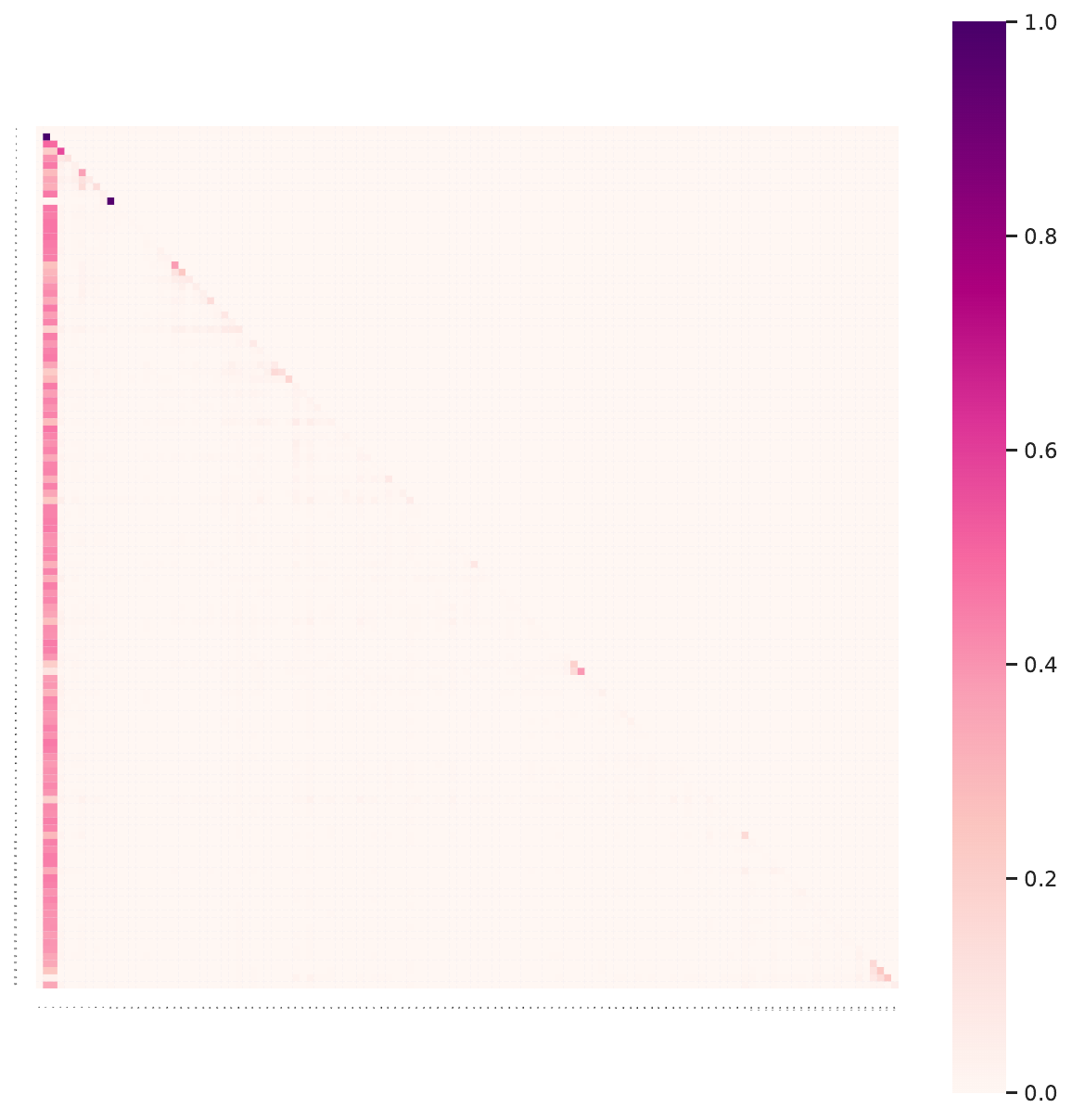}
        \caption*{Head 10}
    \end{minipage}
    \begin{minipage}{0.23\textwidth}
        \includegraphics[width=\linewidth]{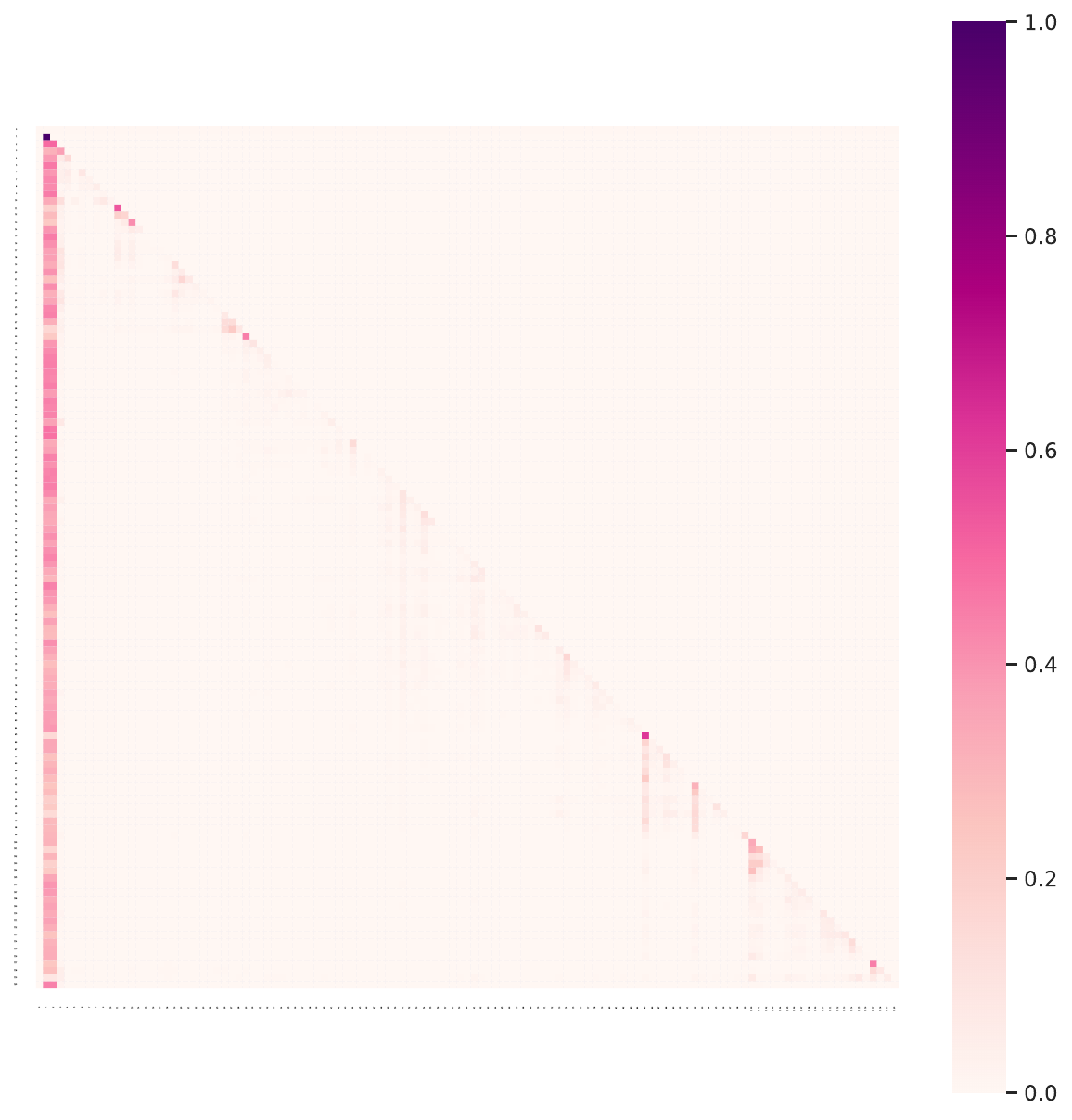}
        \caption*{Head 11}
    \end{minipage}
    \vspace{1em}
\begin{minipage}{0.23\textwidth}
        \includegraphics[width=\linewidth]{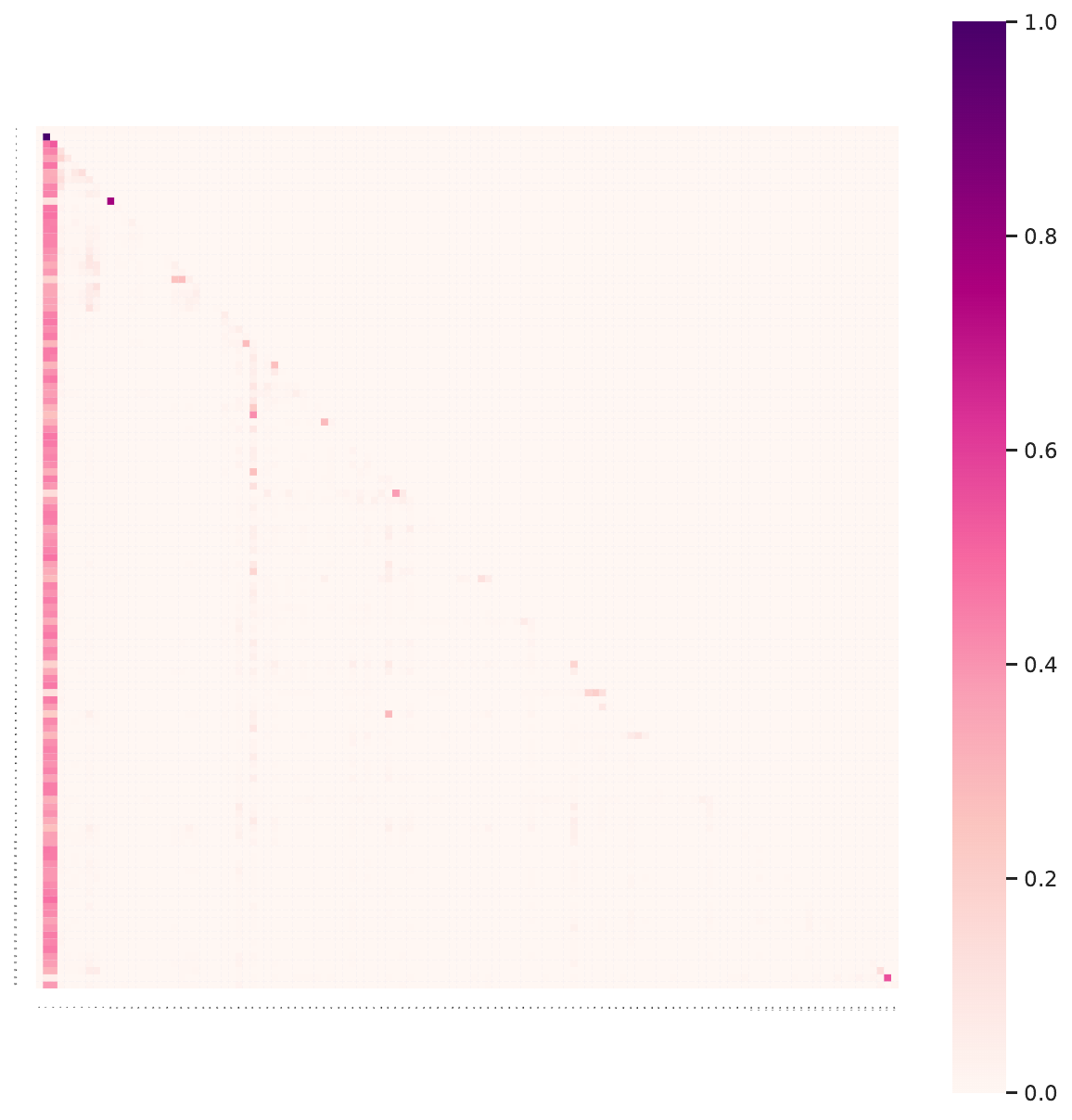}
        \caption*{Head 12}
    \end{minipage}
    \begin{minipage}{0.23\textwidth}
        \includegraphics[width=\linewidth]{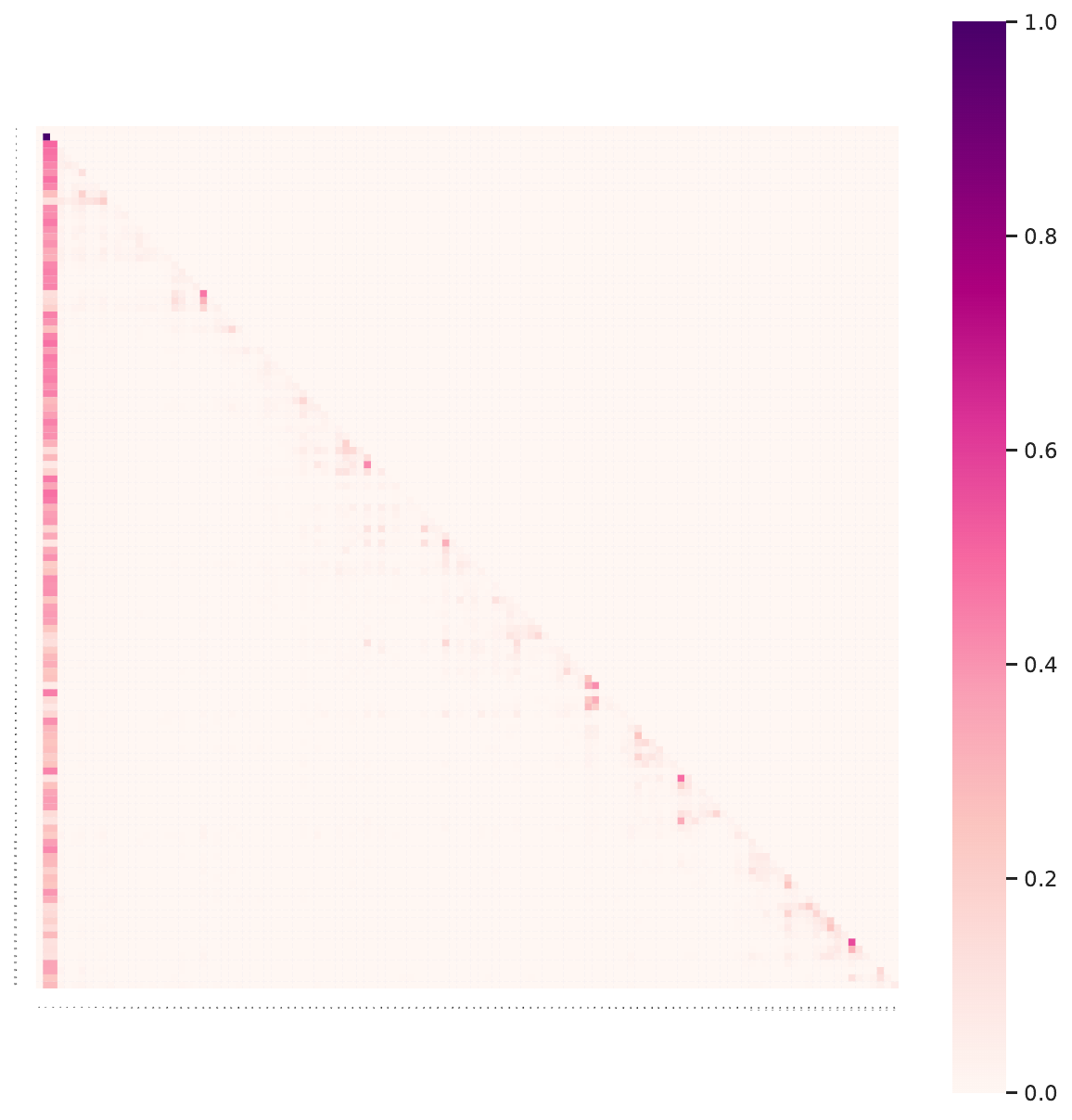}
        \caption*{Head 13}
    \end{minipage}
    \begin{minipage}{0.23\textwidth}
        \includegraphics[width=\linewidth]{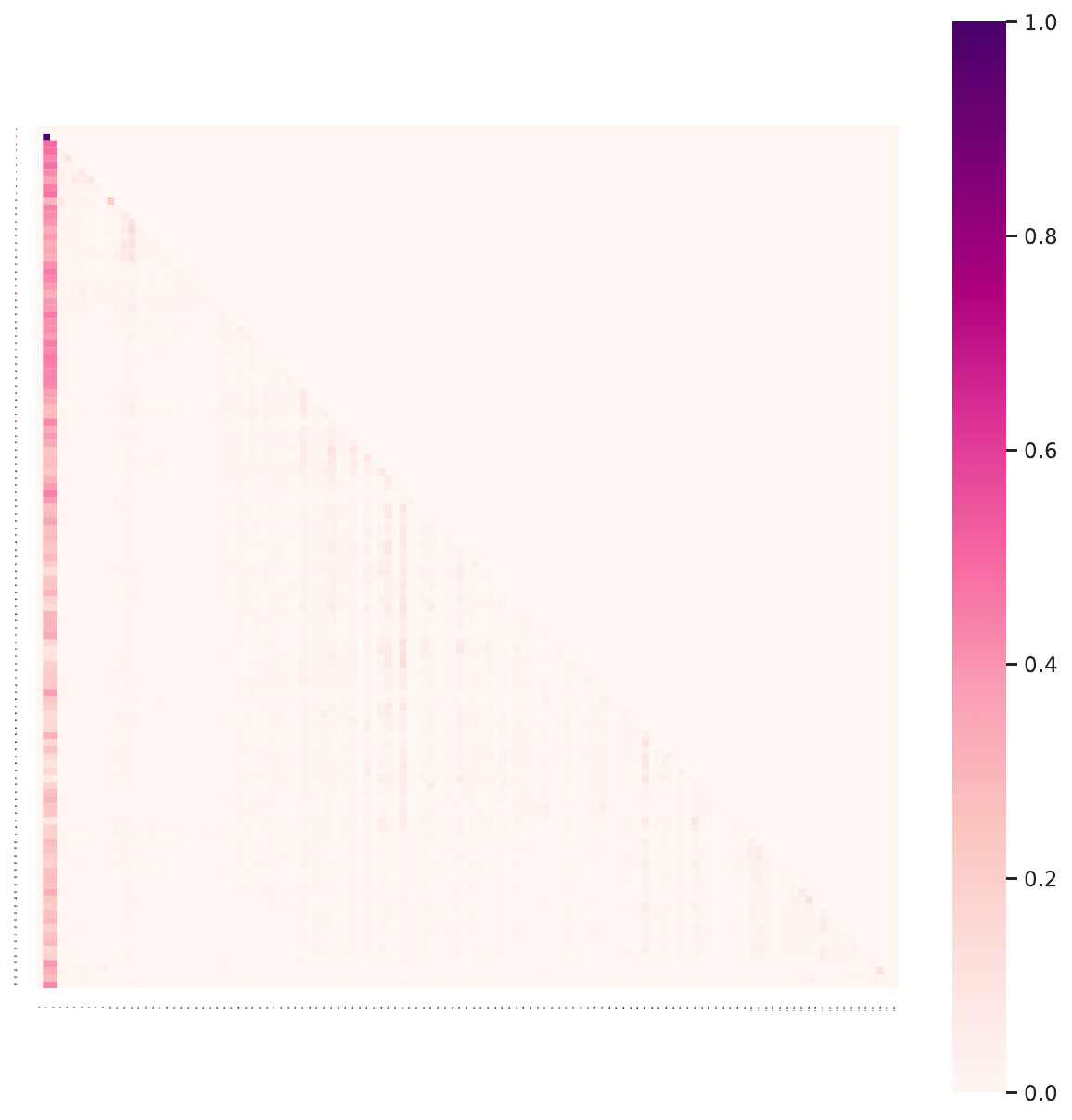}
        \caption*{Head 14}
    \end{minipage}
    \begin{minipage}{0.23\textwidth}
        \includegraphics[width=\linewidth]{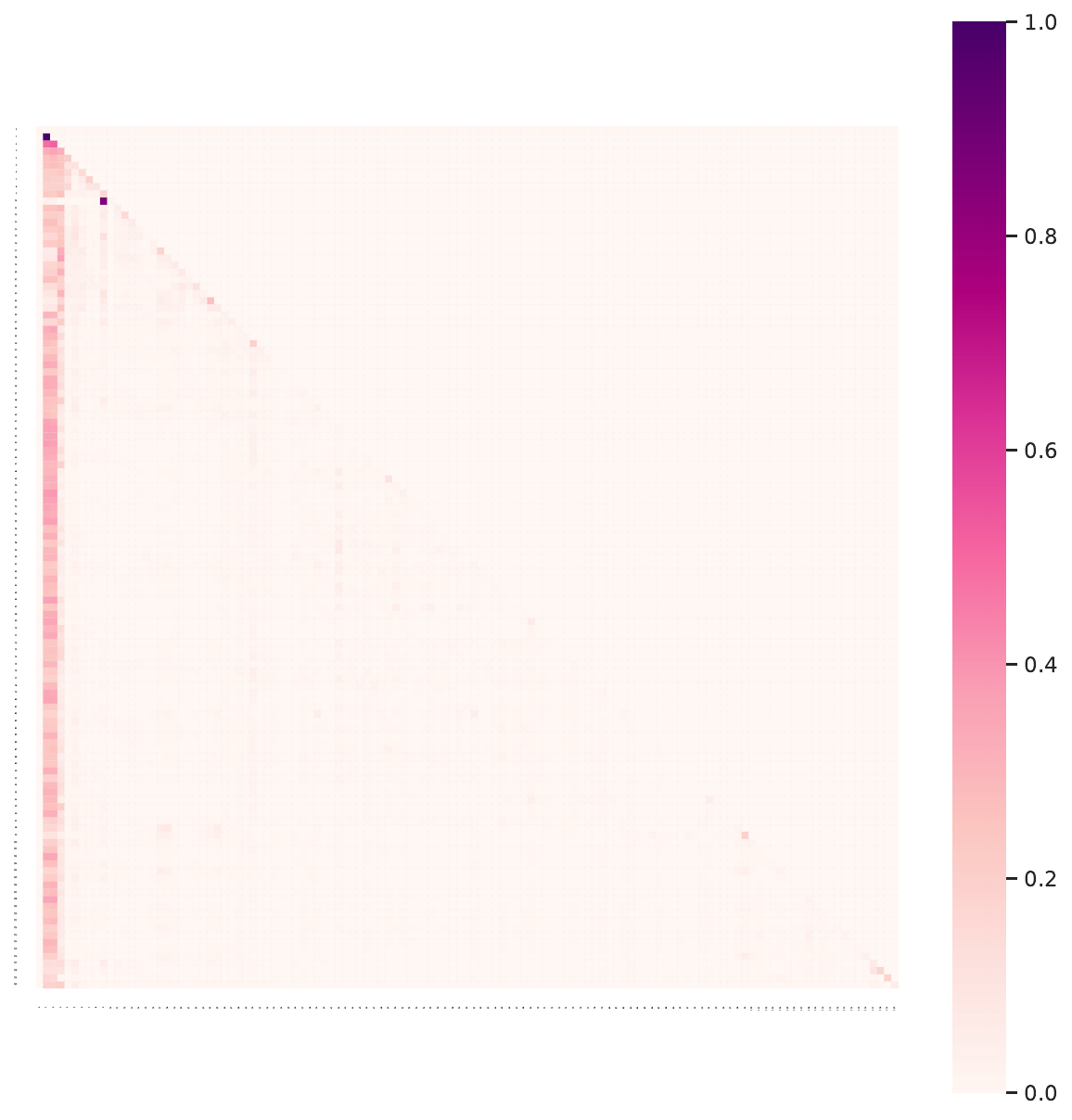}
        \caption*{Head 15}
    \end{minipage}
\caption{\textbf{The attention score of LLaMA-2-7B in layer 31.} (part 1 of 2)}
\label{Fig: vislayerori31_1}
\end{figure*}

\begin{figure*}[tbp]
\centering
\begin{minipage}{0.23\textwidth}
        \includegraphics[width=\linewidth]{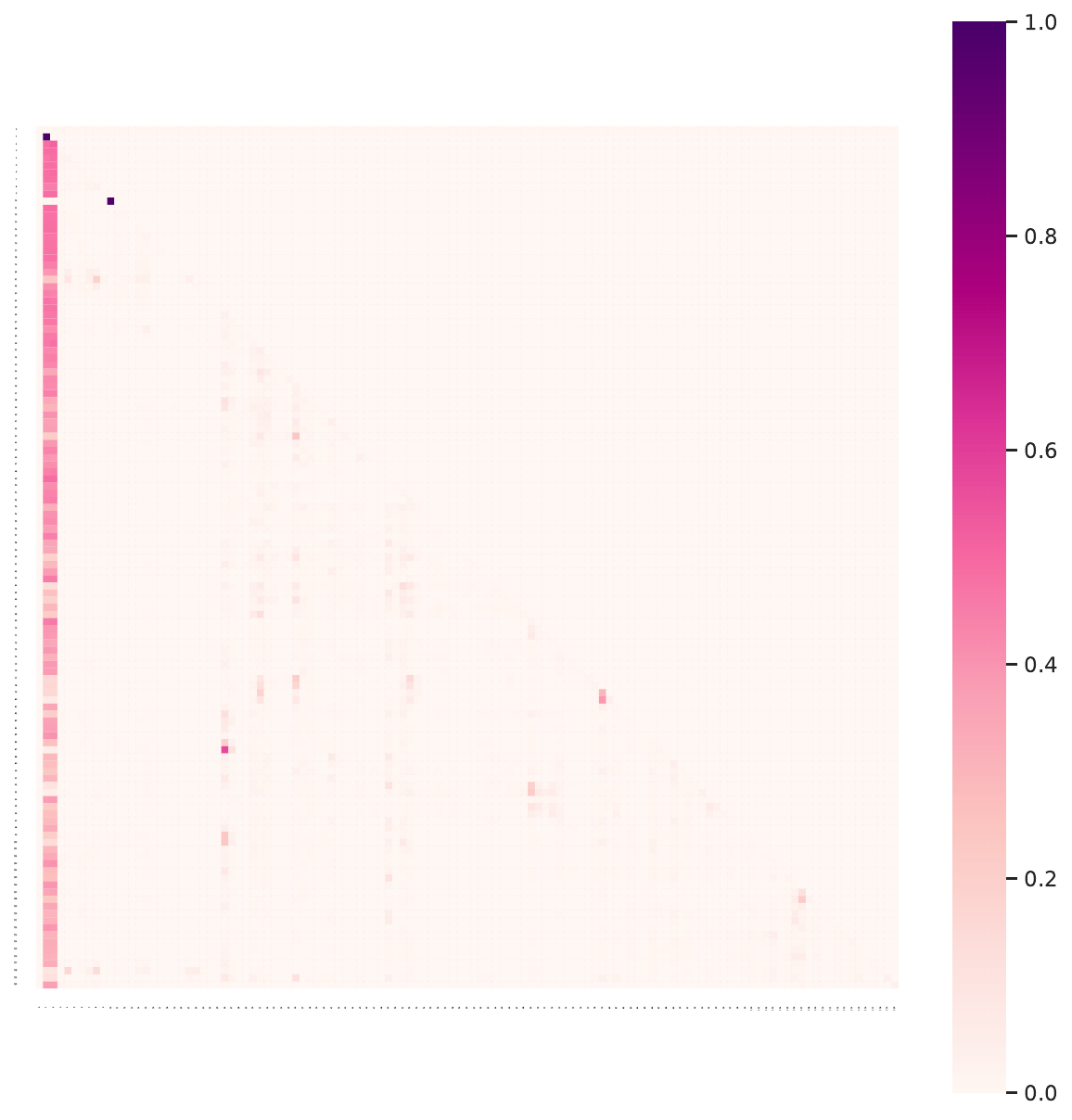}
        \caption*{Head 16}
    \end{minipage}
    \begin{minipage}{0.23\textwidth}
        \includegraphics[width=\linewidth]{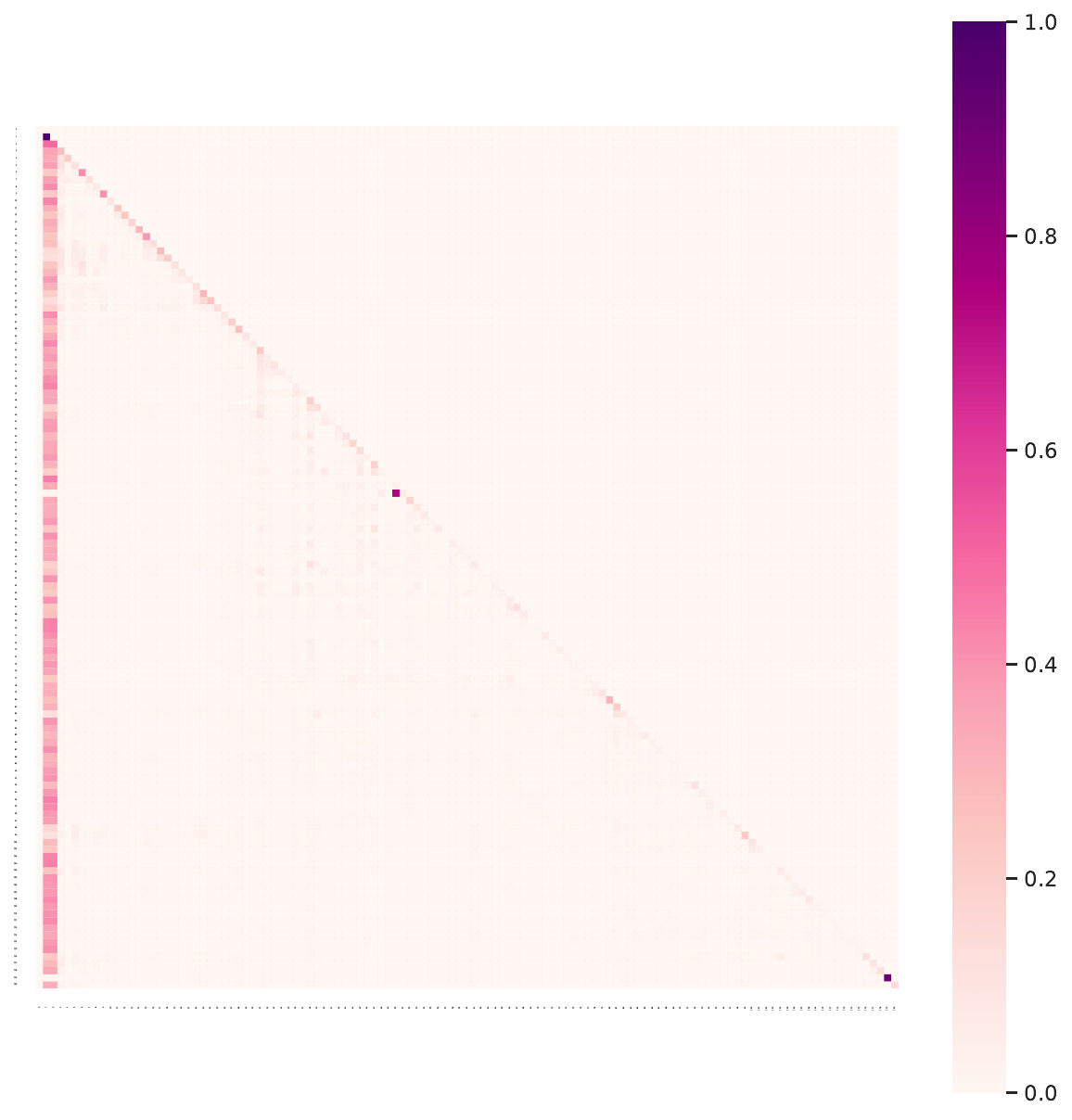}
        \caption*{Head 17}
    \end{minipage}
    \begin{minipage}{0.23\textwidth}
        \includegraphics[width=\linewidth]{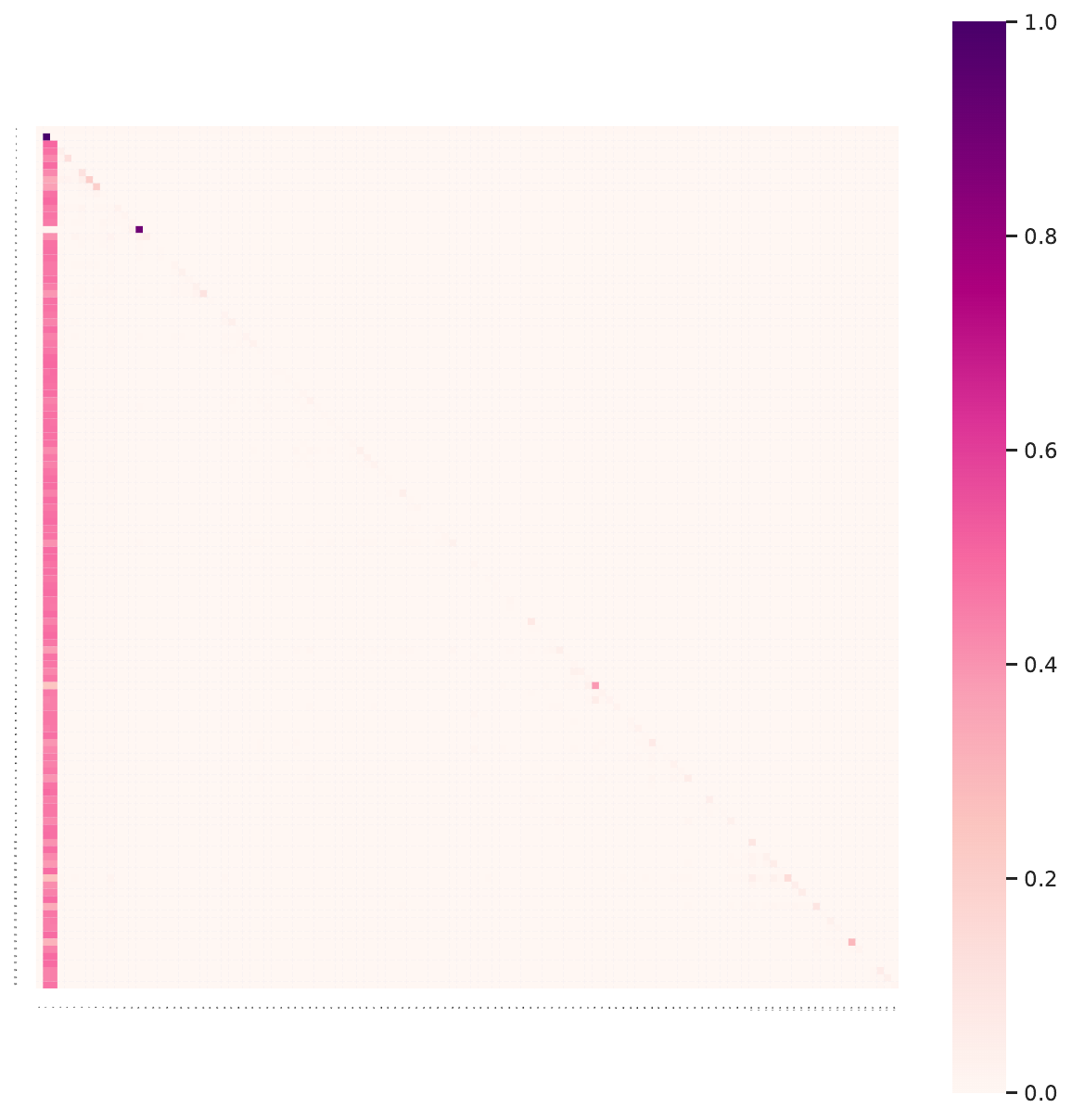}
        \caption*{Head 18}
    \end{minipage}
    \begin{minipage}{0.23\textwidth}
        \includegraphics[width=\linewidth]{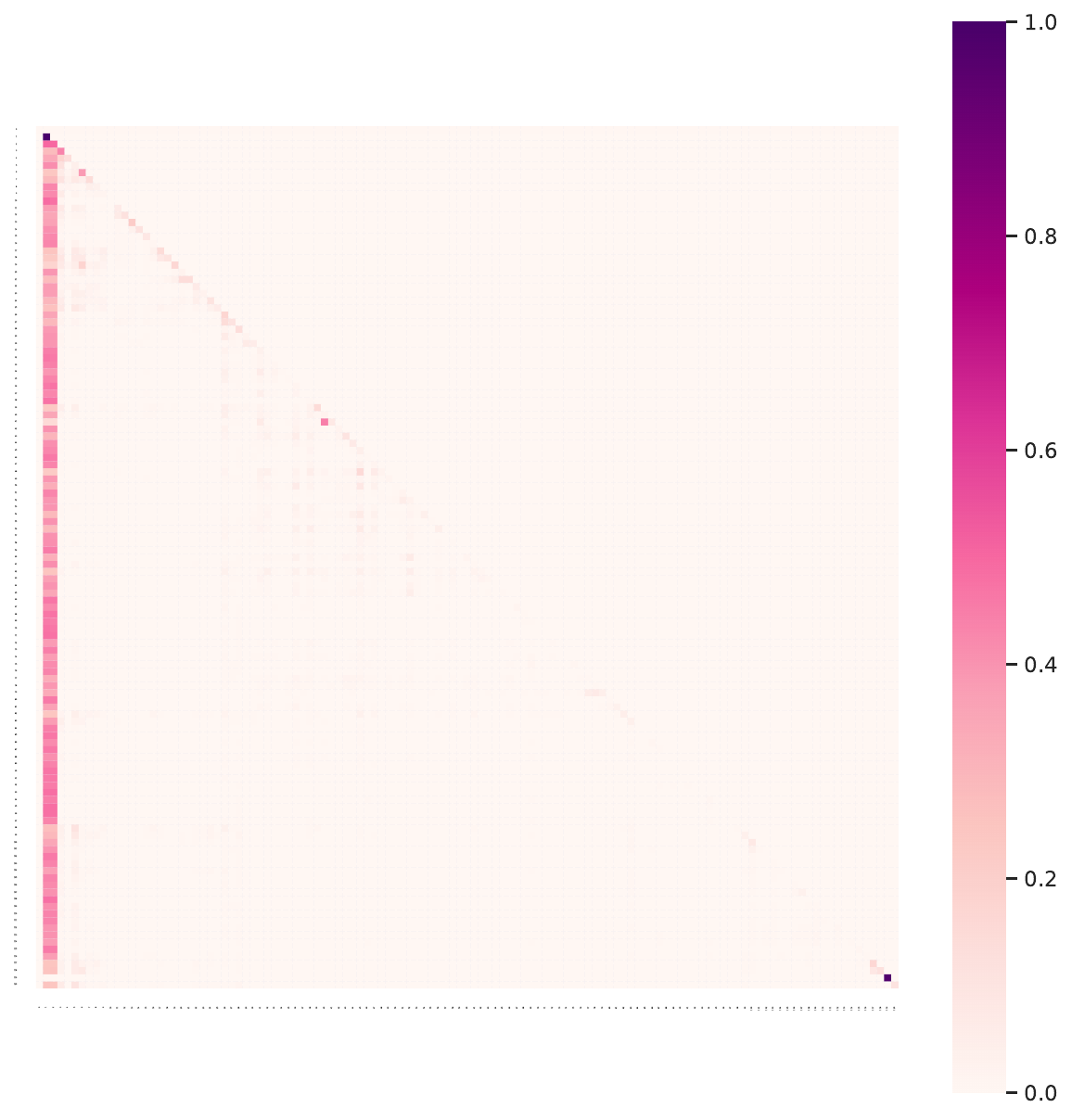}
        \caption*{Head 19}
\end{minipage}
\vspace{1em}
\begin{minipage}{0.23\textwidth}
        \includegraphics[width=\linewidth]{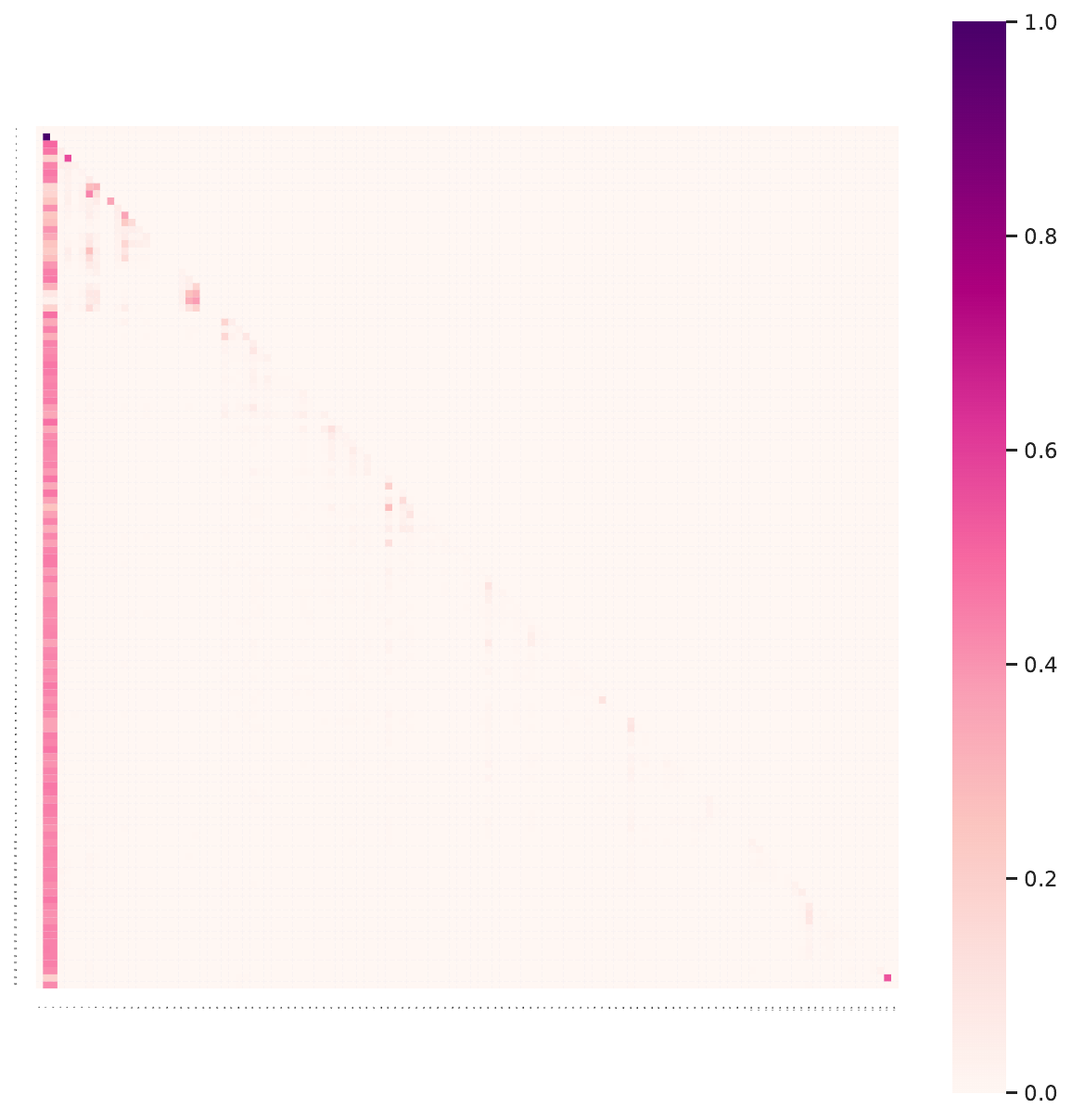}
        \caption*{Head 20}
    \end{minipage}
    \begin{minipage}{0.23\textwidth}
        \includegraphics[width=\linewidth]{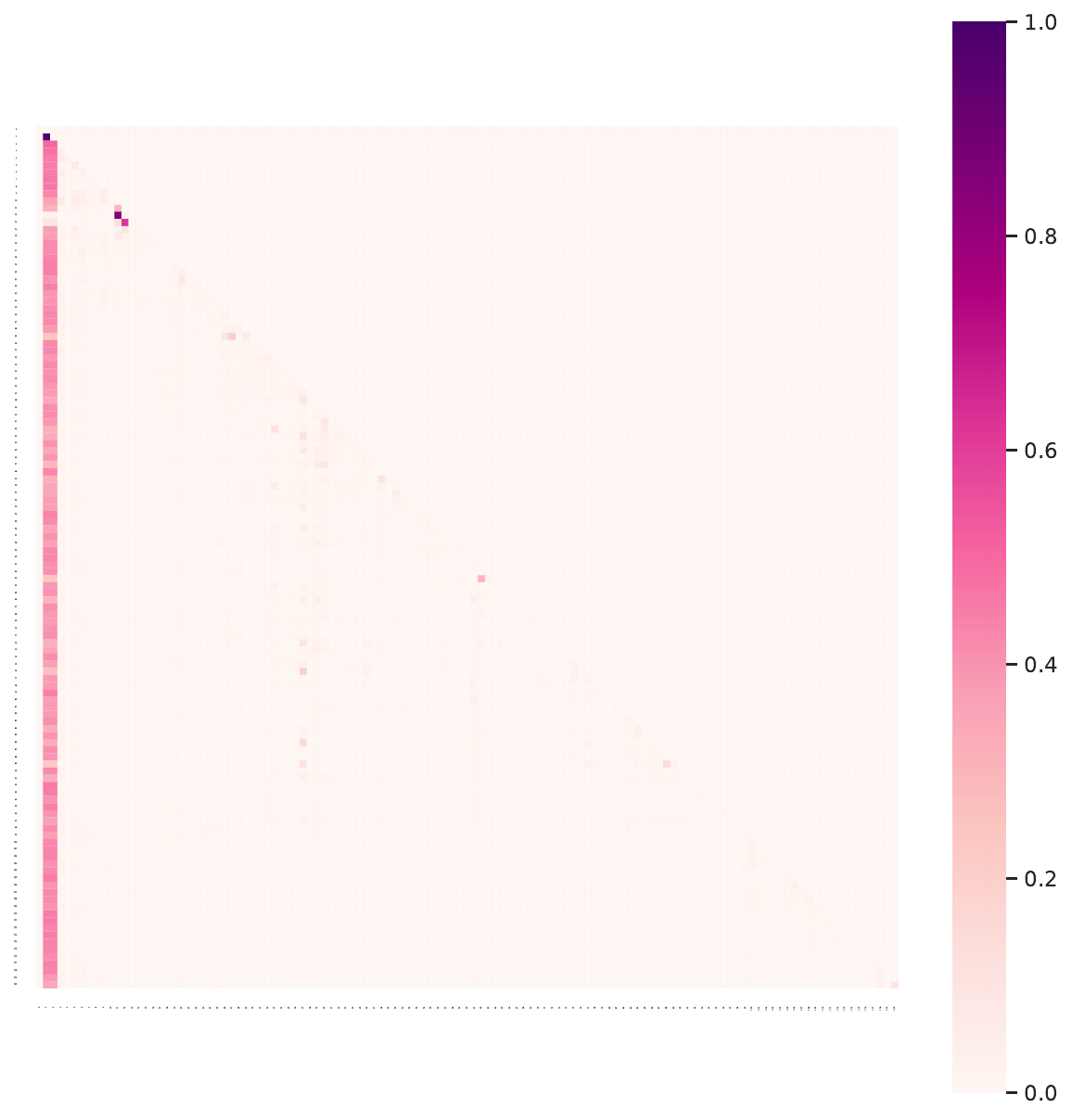}
        \caption*{Head 21}
    \end{minipage}
    \begin{minipage}{0.23\textwidth}
        \includegraphics[width=\linewidth]{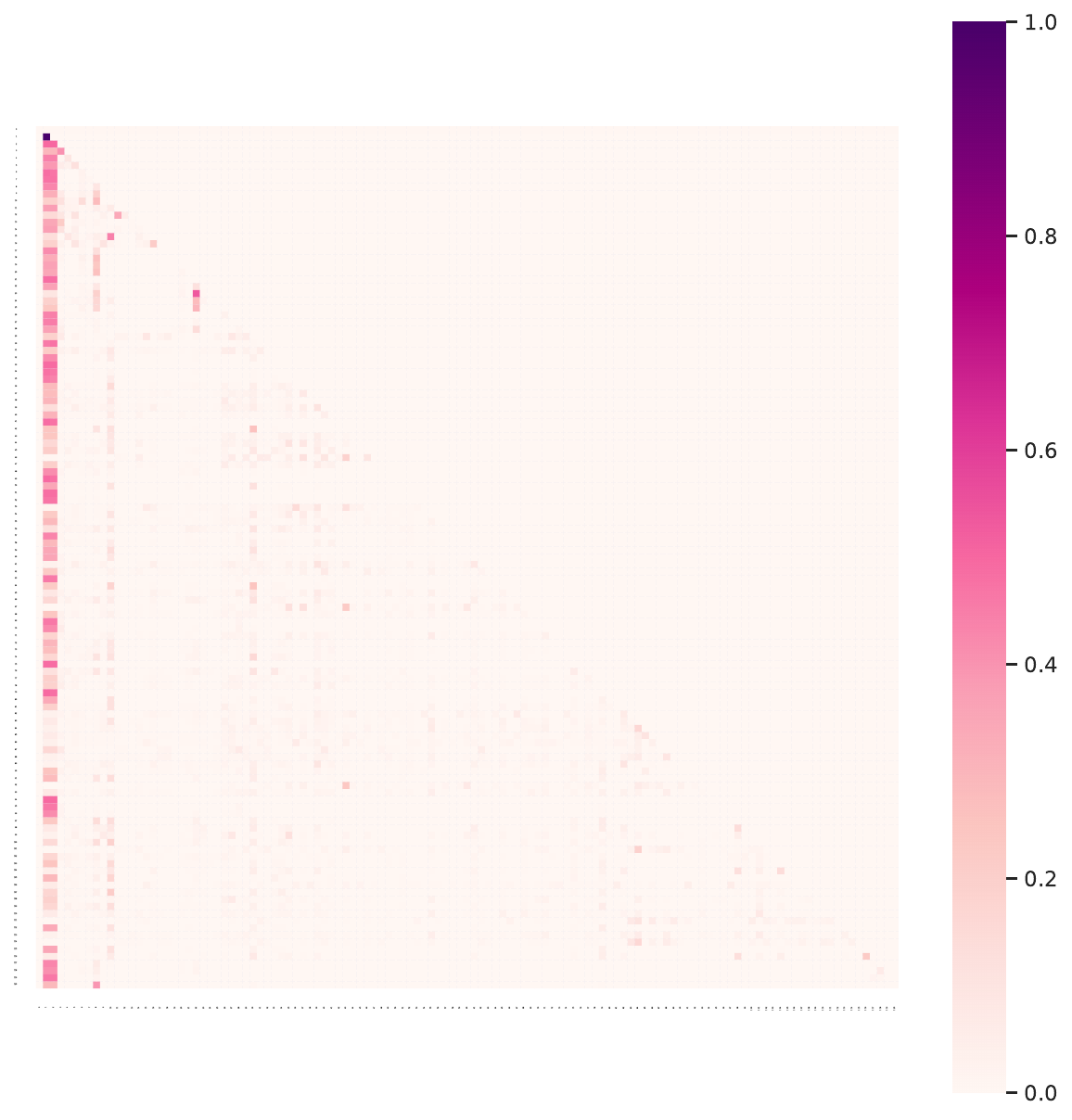}
        \caption*{Head 22}
    \end{minipage}
    \begin{minipage}{0.23\textwidth}
        \includegraphics[width=\linewidth]{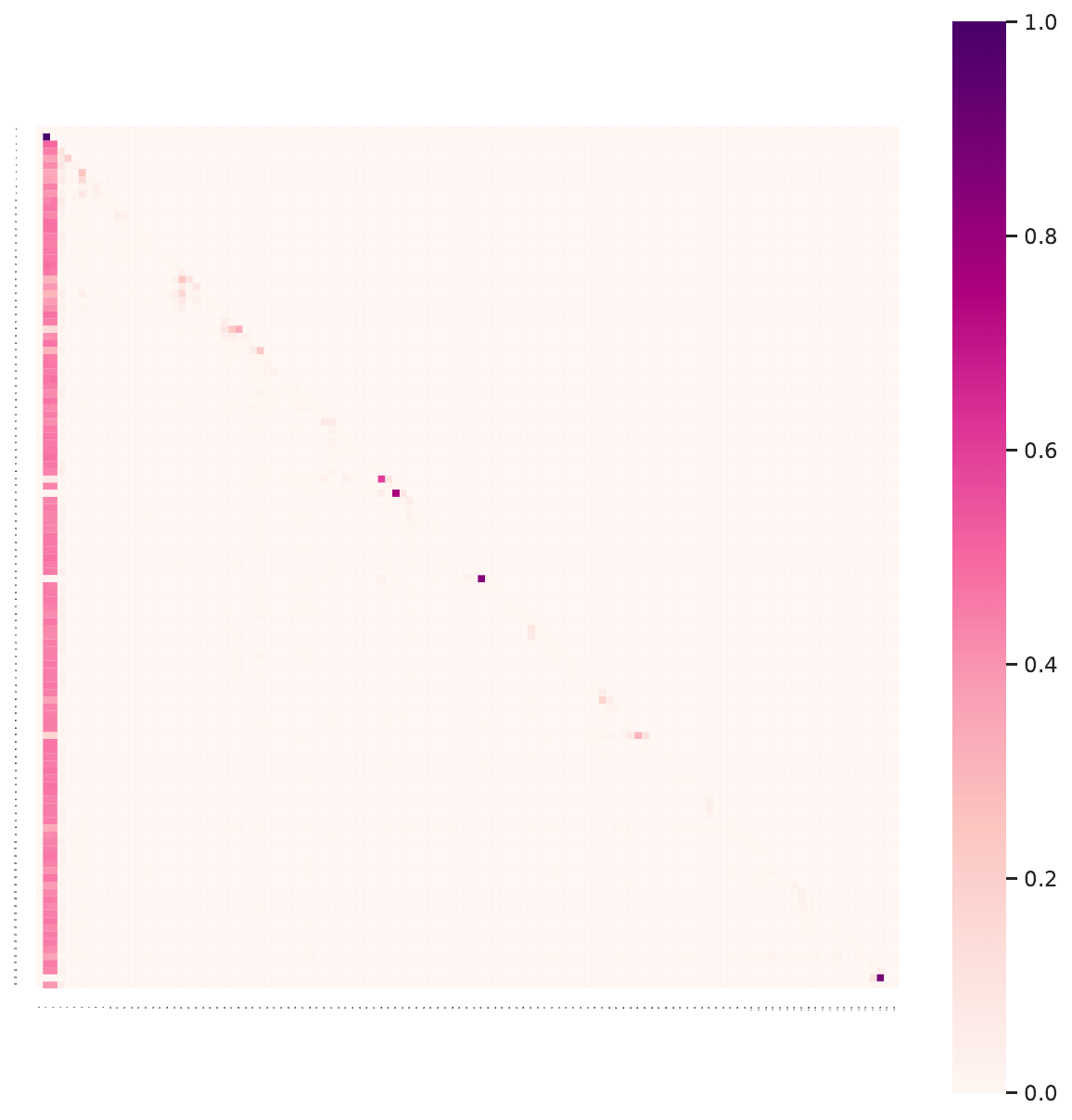}
        \caption*{Head 23}
    \end{minipage}
    \vspace{1em}
\begin{minipage}{0.23\textwidth}
        \includegraphics[width=\linewidth]{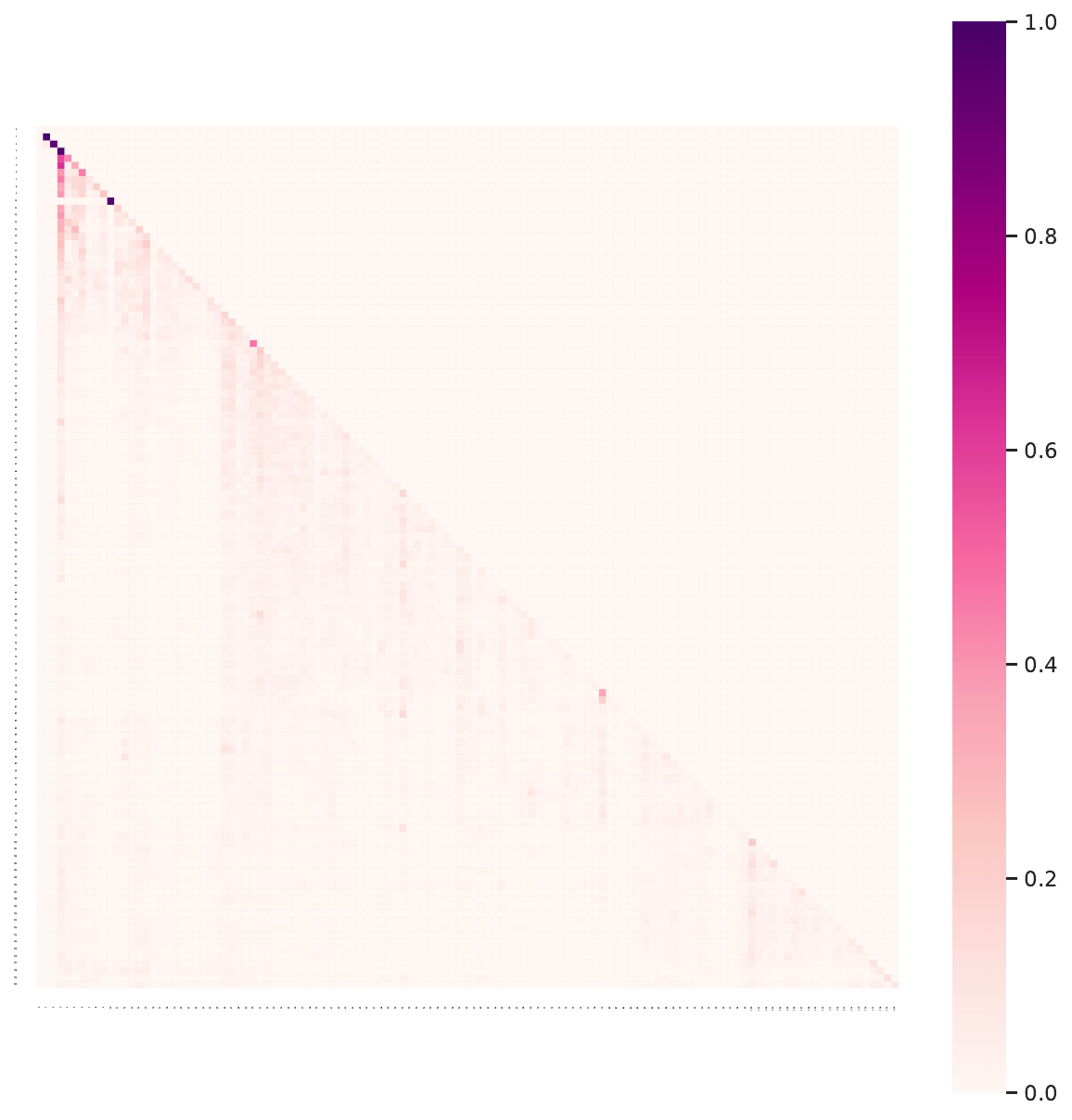}
        \caption*{Head 24}
    \end{minipage}
    \begin{minipage}{0.23\textwidth}
        \includegraphics[width=\linewidth]{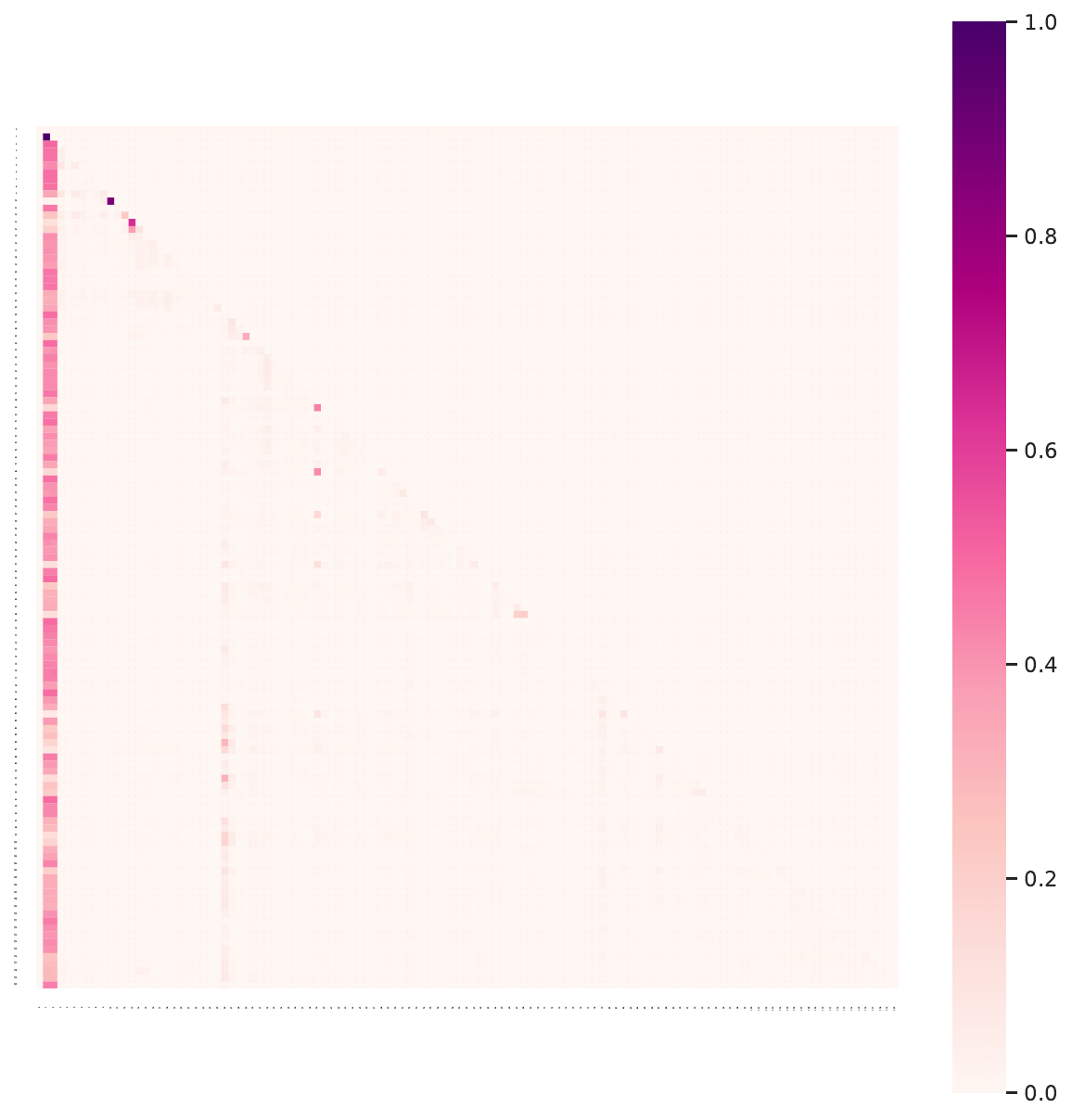}
        \caption*{Head 25}
    \end{minipage}
    \begin{minipage}{0.23\textwidth}
        \includegraphics[width=\linewidth]{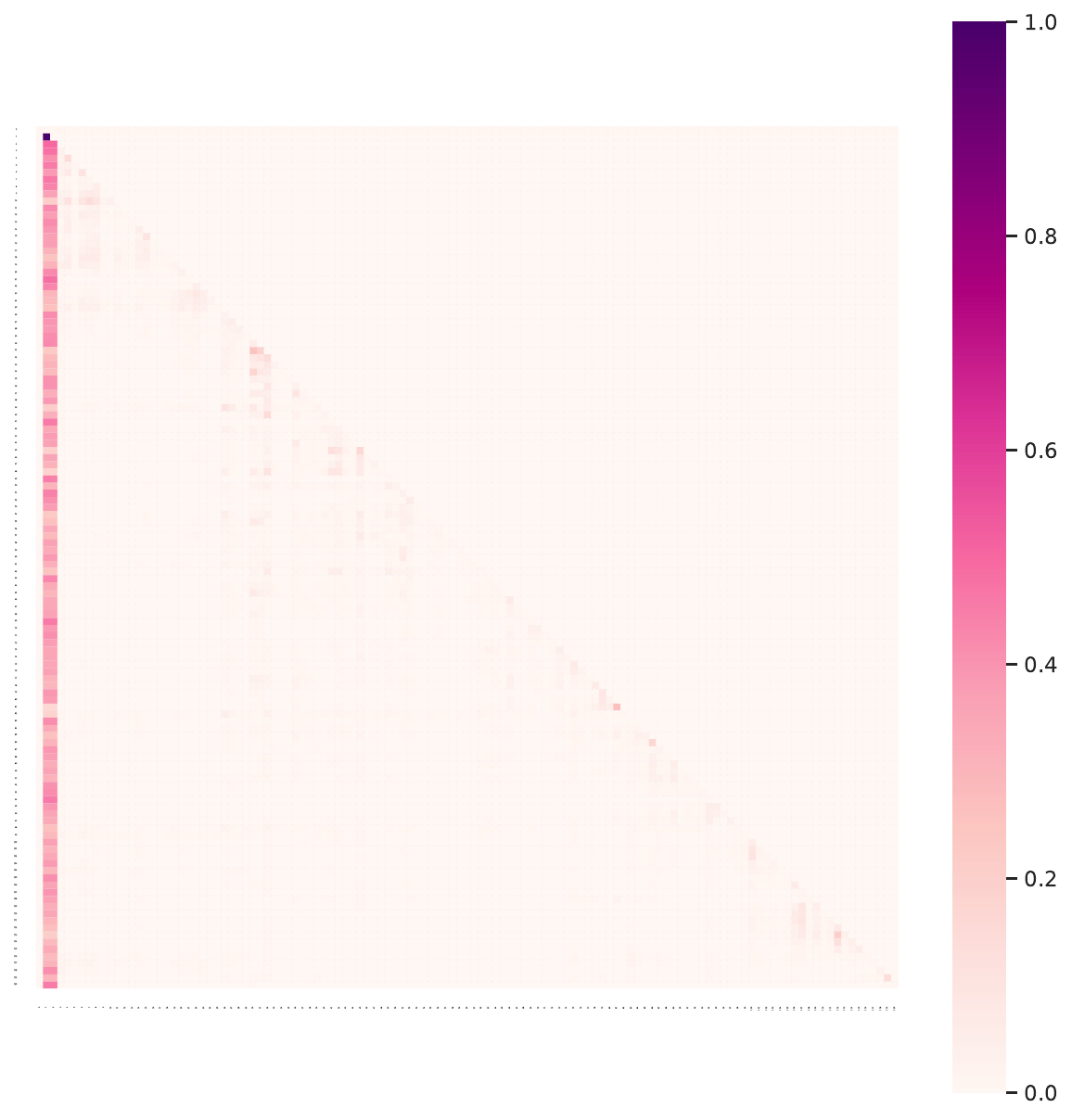}
        \caption*{Head 26}
    \end{minipage}
    \begin{minipage}{0.23\textwidth}
        \includegraphics[width=\linewidth]{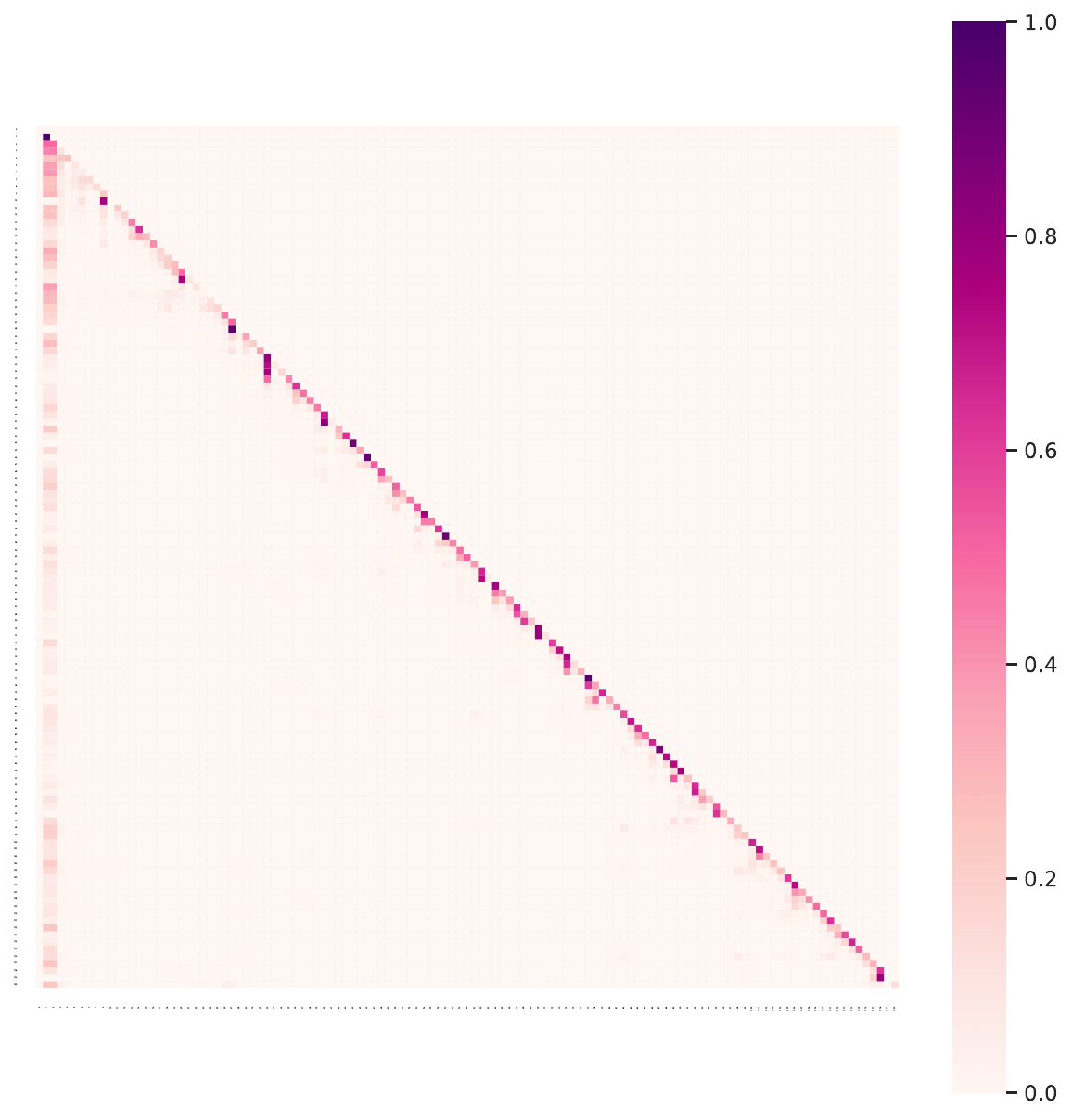}
        \caption*{Head 27}
    \end{minipage}
    \vspace{1em}
\begin{minipage}{0.23\textwidth}
        \includegraphics[width=\linewidth]{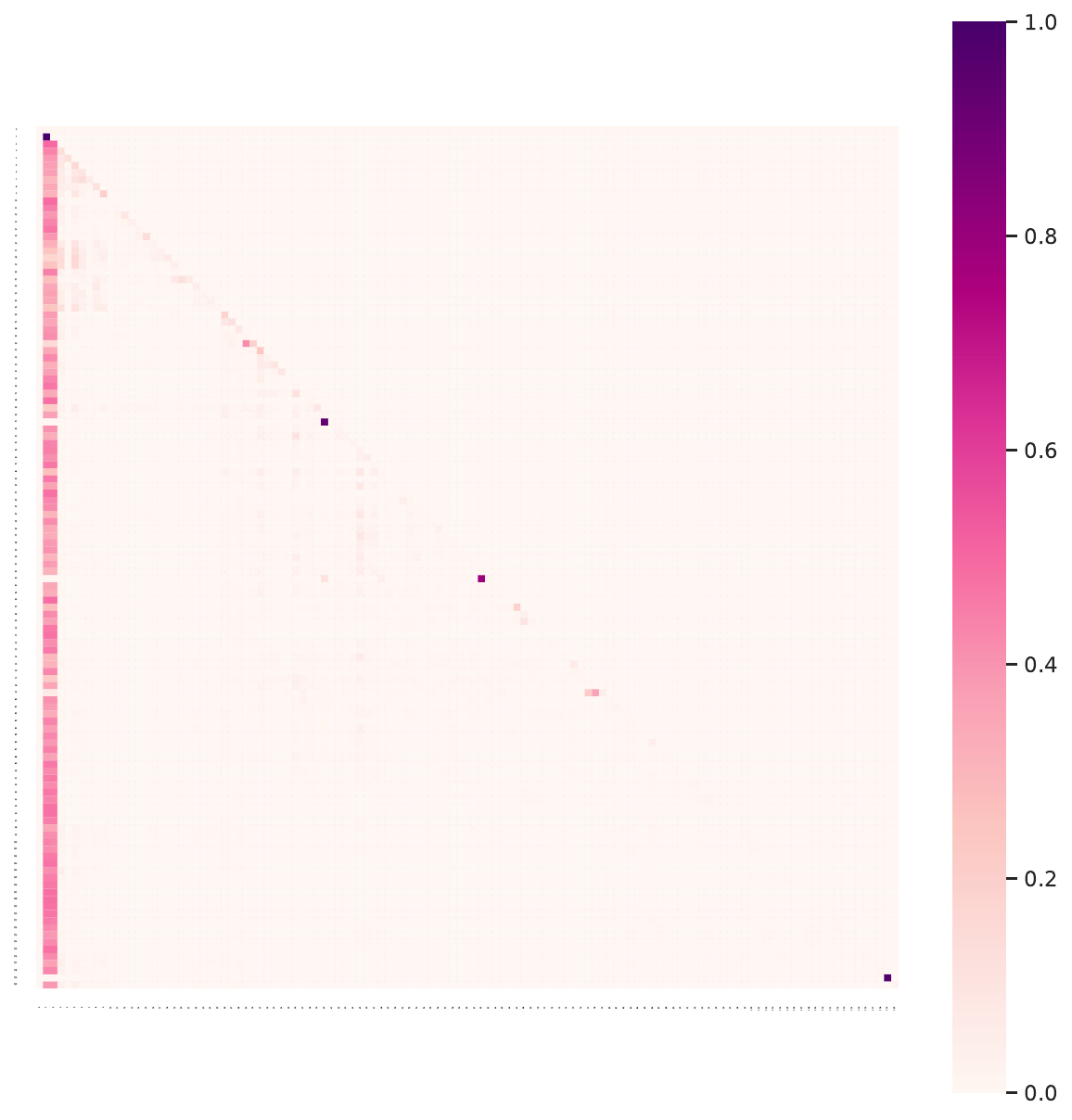}
        \caption*{Head 28}
    \end{minipage}
    \begin{minipage}{0.23\textwidth}
        \includegraphics[width=\linewidth]{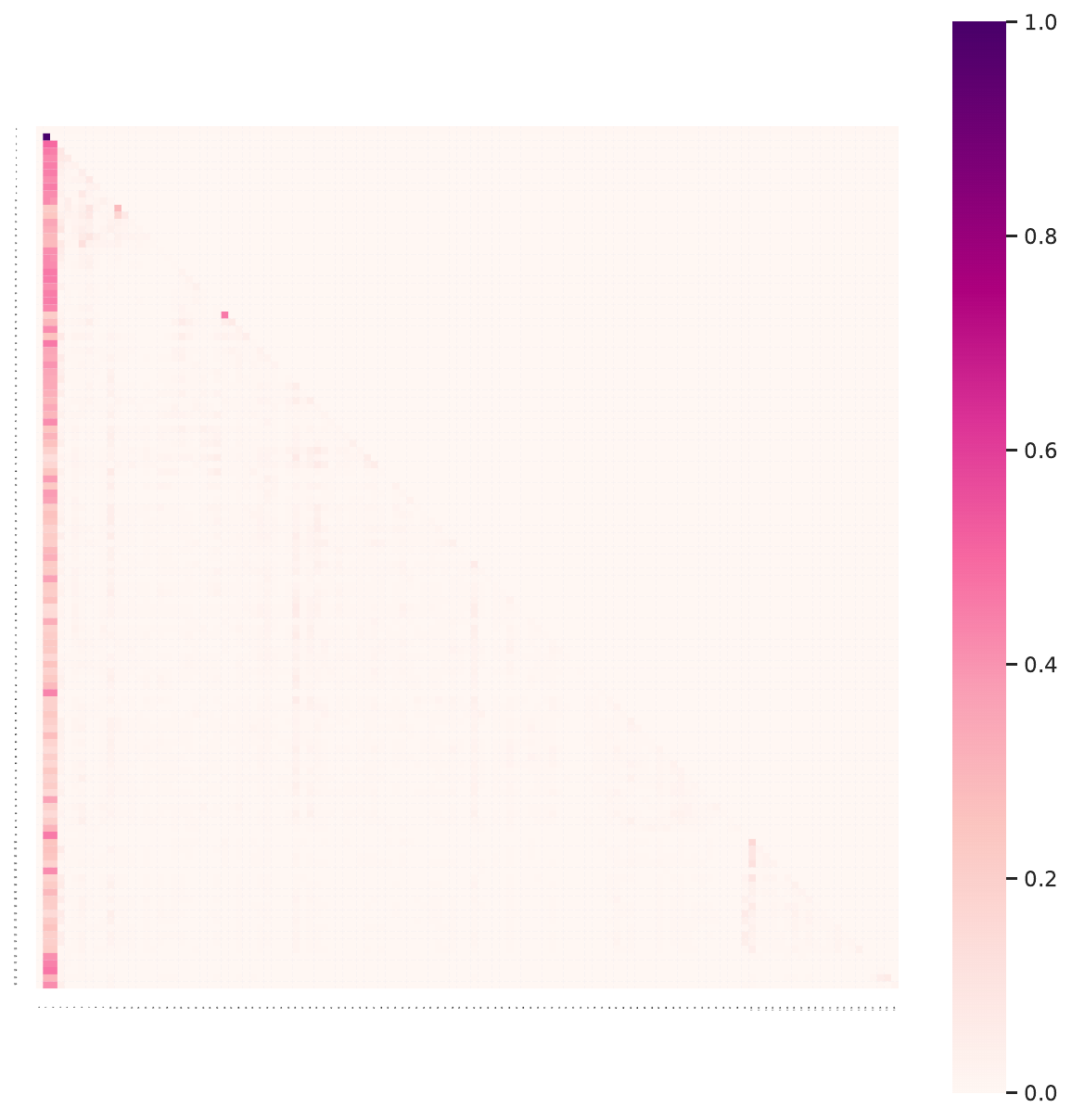}
        \caption*{Head 29}
    \end{minipage}
    \begin{minipage}{0.23\textwidth}
        \includegraphics[width=\linewidth]{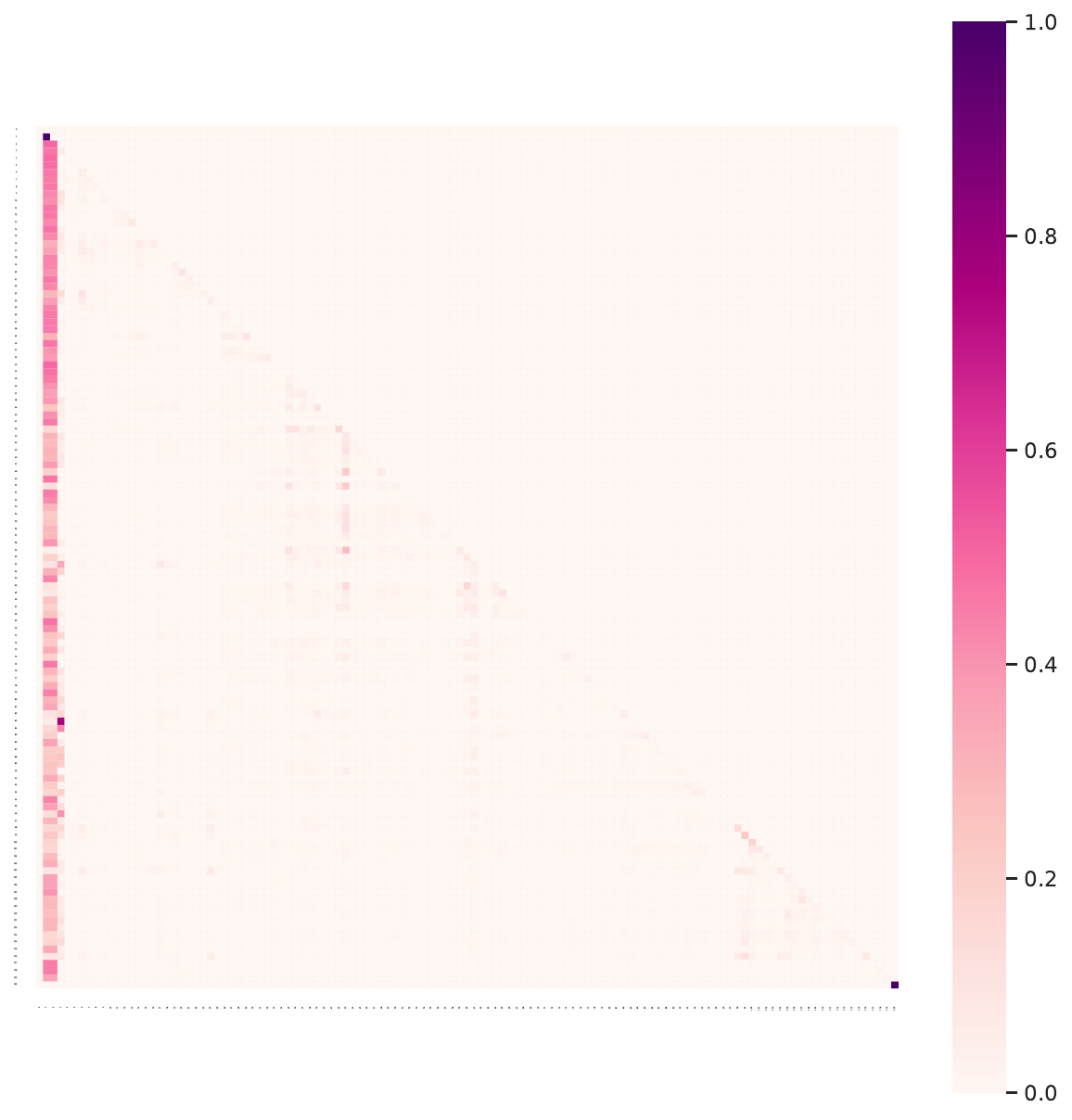}
        \caption*{Head 30}
    \end{minipage}
    \begin{minipage}{0.23\textwidth}
        \includegraphics[width=\linewidth]{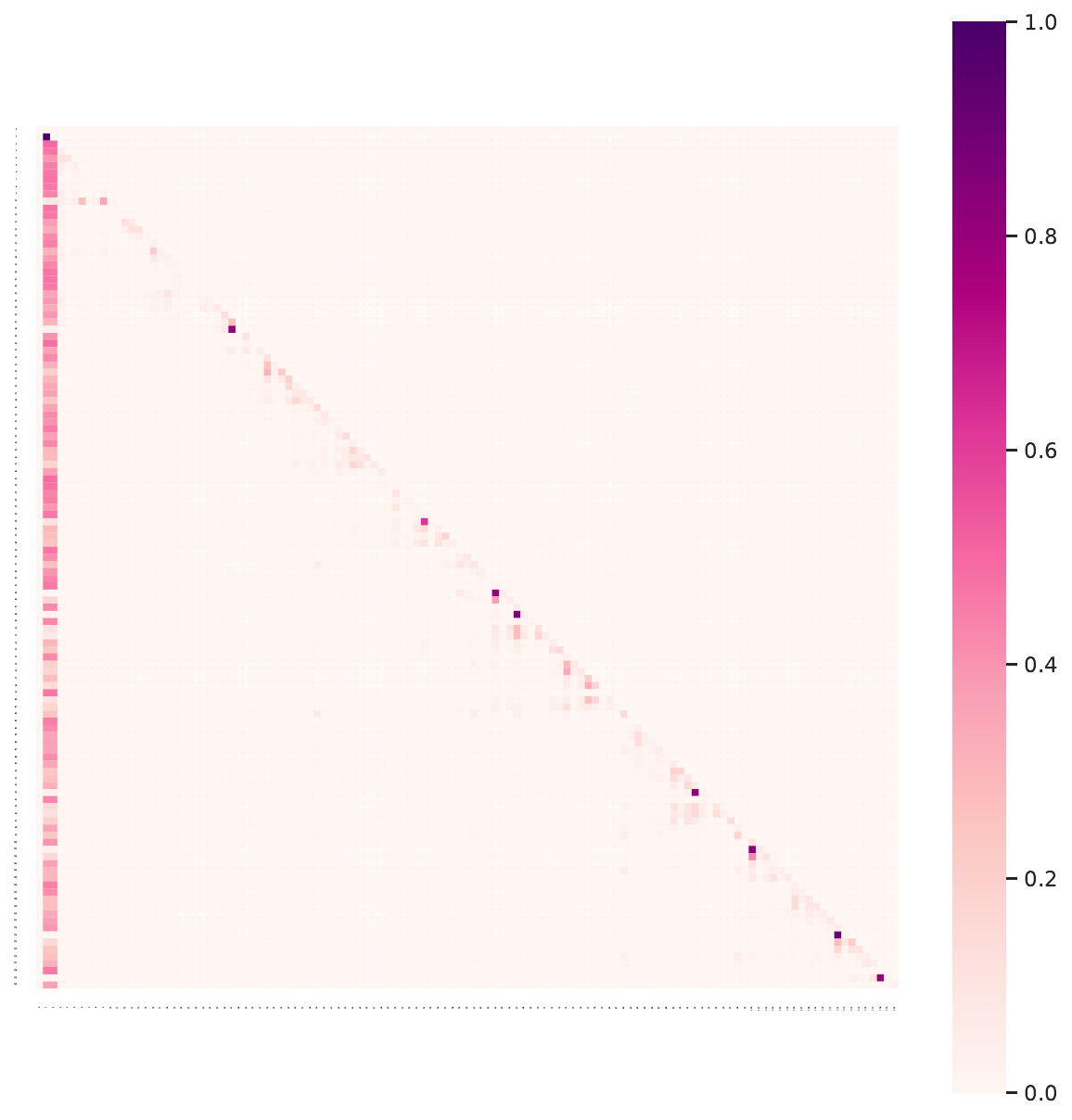}
        \caption*{Head 31}
    \end{minipage}
\caption{\textbf{The attention score of LLaMA-2-7B in layer 31.} (part 2 of 2)}
\label{Fig: vislayerori31_2}
\end{figure*}

\begin{figure*}[tbp]
\centering
    \begin{minipage}{0.23\textwidth}
        \includegraphics[width=\linewidth]{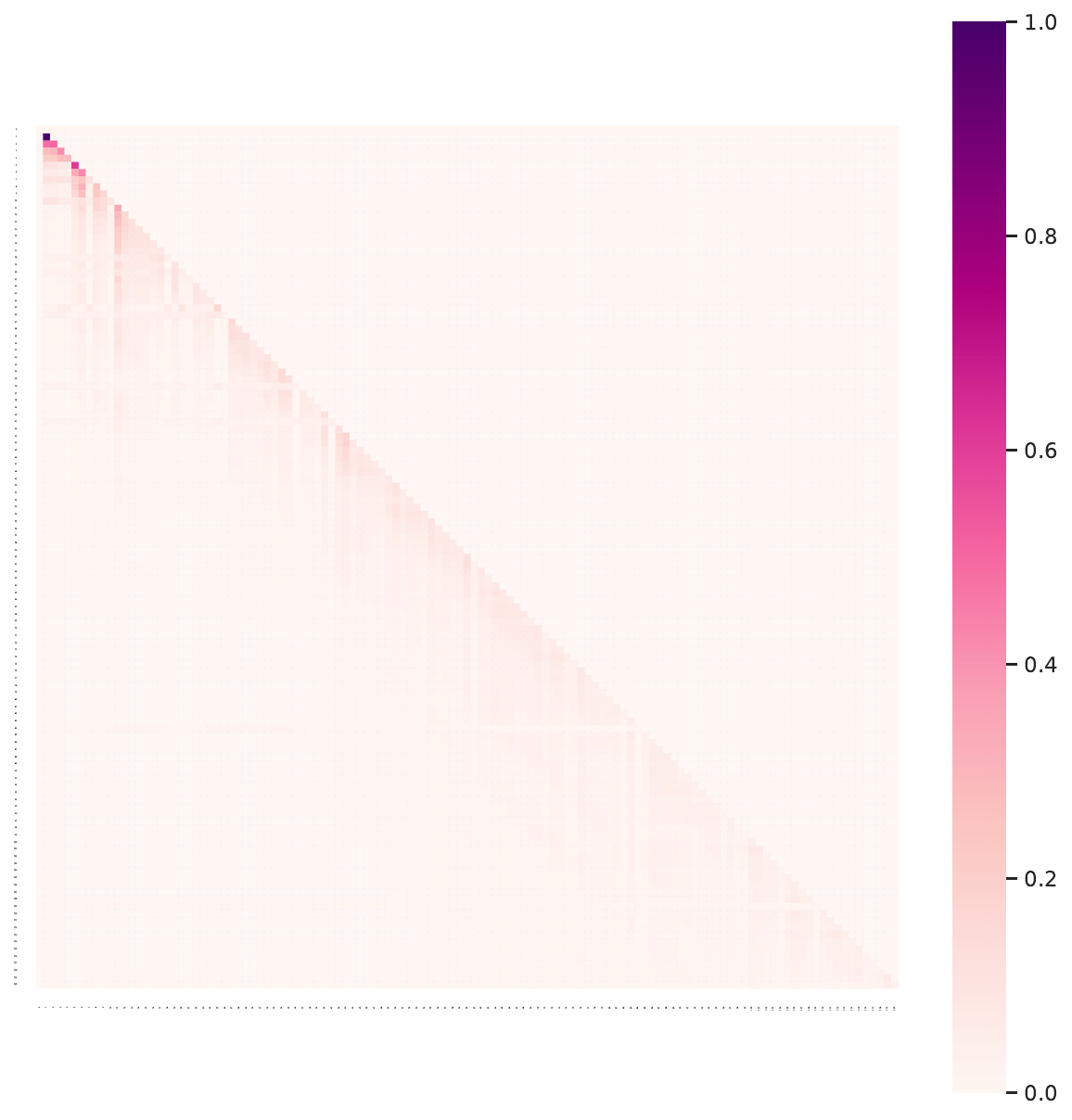}
        \caption*{Head 0}
    \end{minipage}
    \begin{minipage}{0.23\textwidth}
        \includegraphics[width=\linewidth]{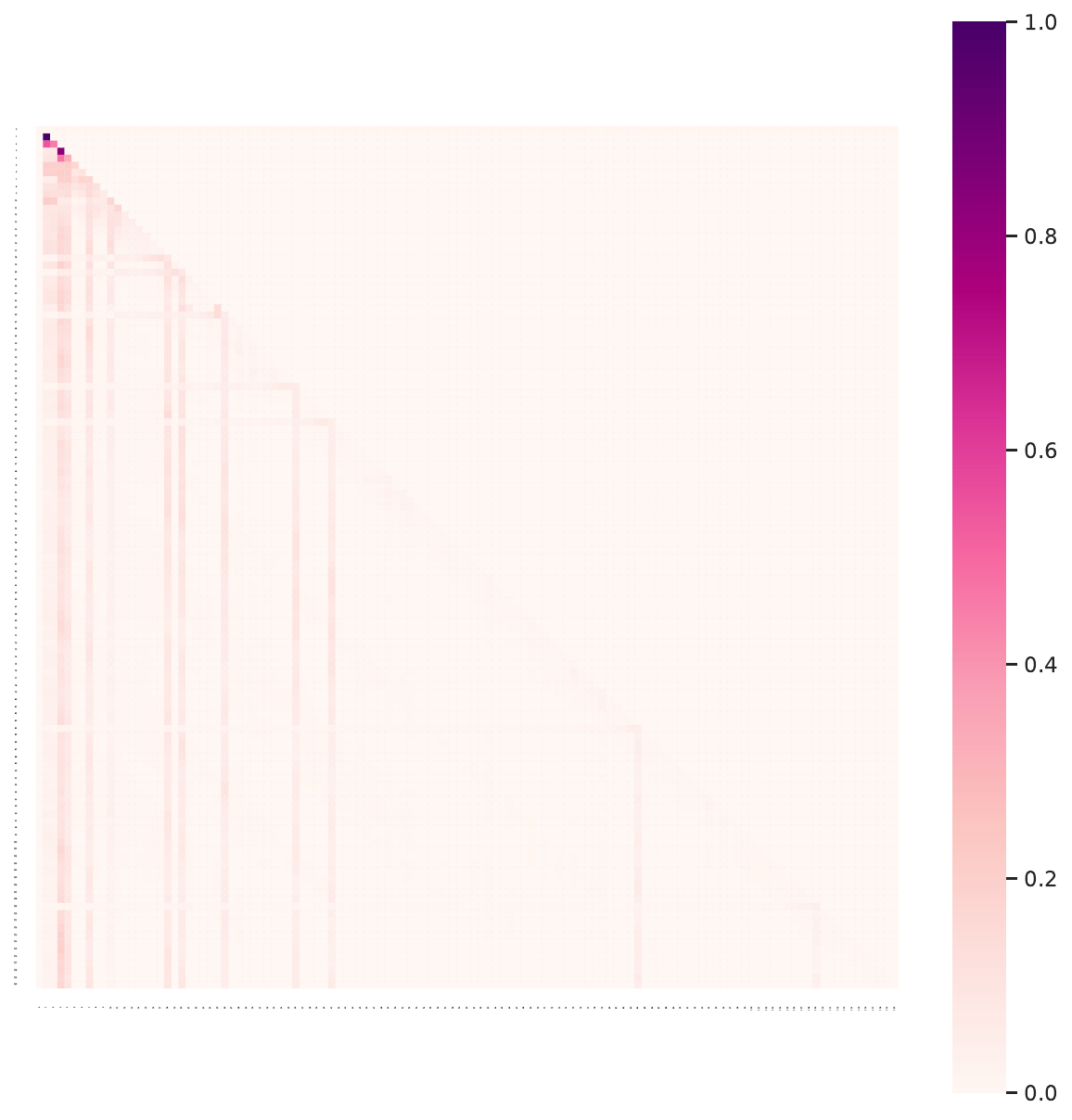}
        \caption*{Head 1}
    \end{minipage}
    \begin{minipage}{0.23\textwidth}
        \includegraphics[width=\linewidth]{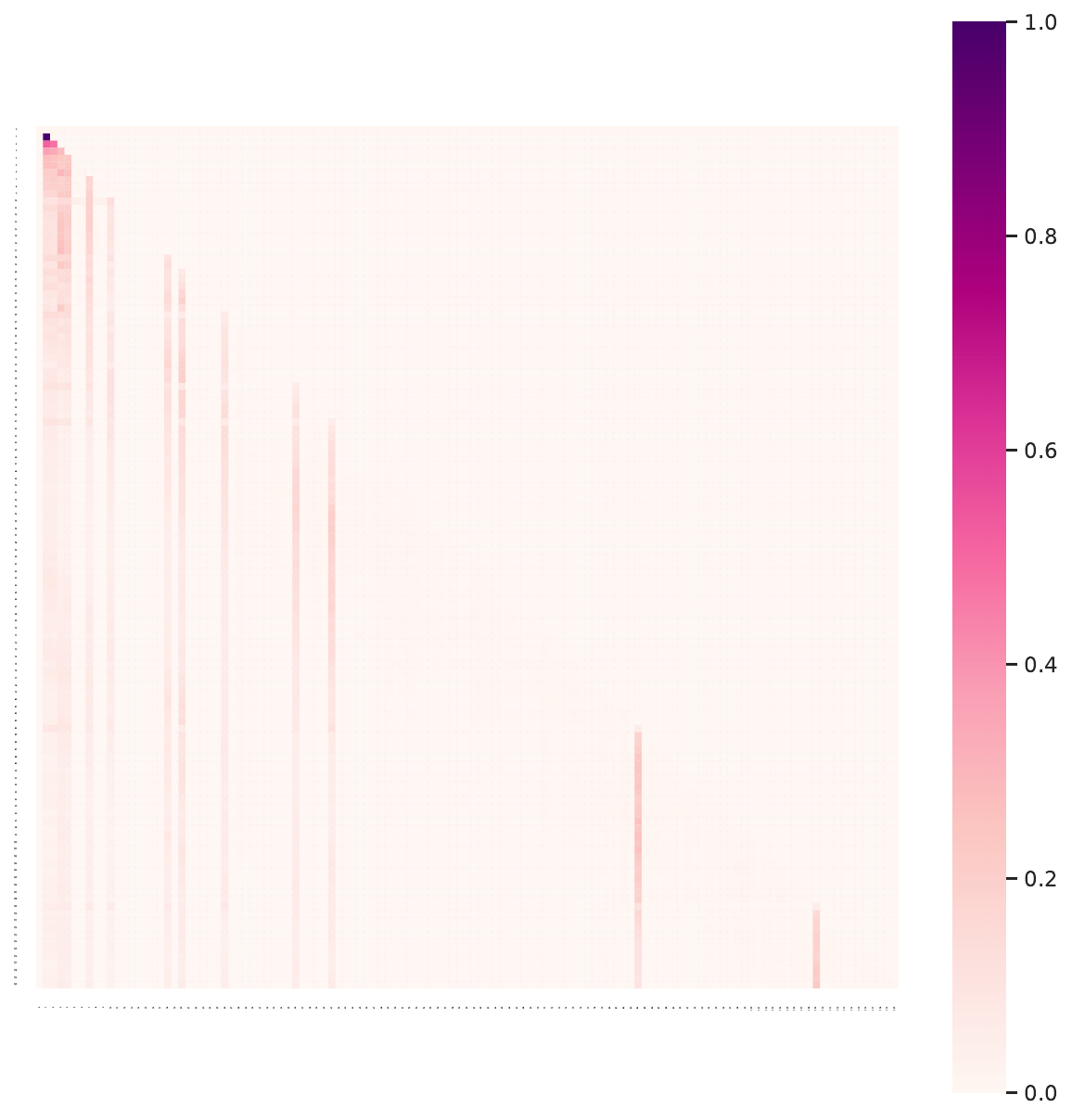}
        \caption*{Head 2}
    \end{minipage}
    \begin{minipage}{0.23\textwidth}
        \includegraphics[width=\linewidth]{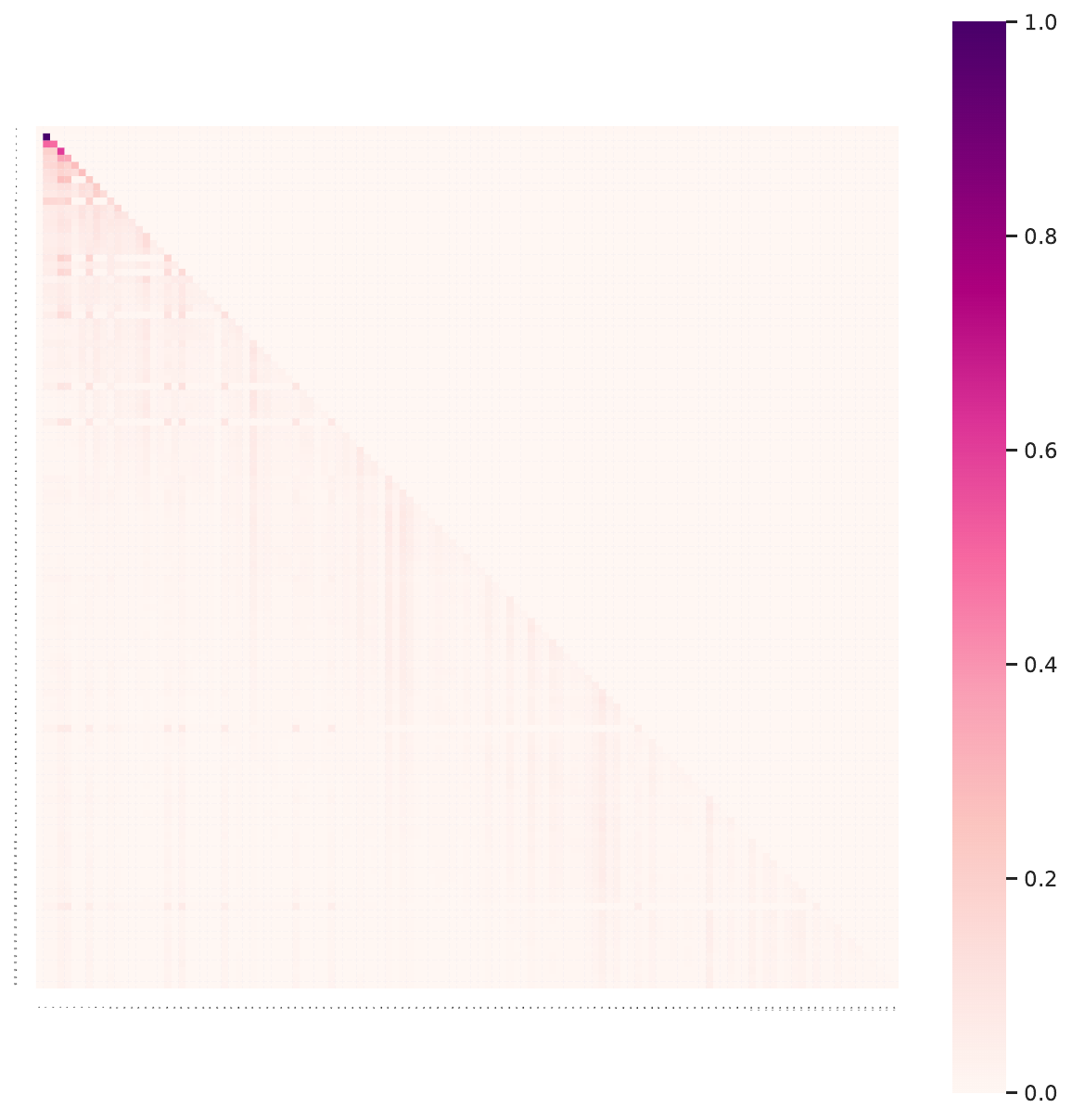}
        \caption*{Head 3}
    \end{minipage}
\vspace{1em}
\begin{minipage}{0.23\textwidth}
        \includegraphics[width=\linewidth]{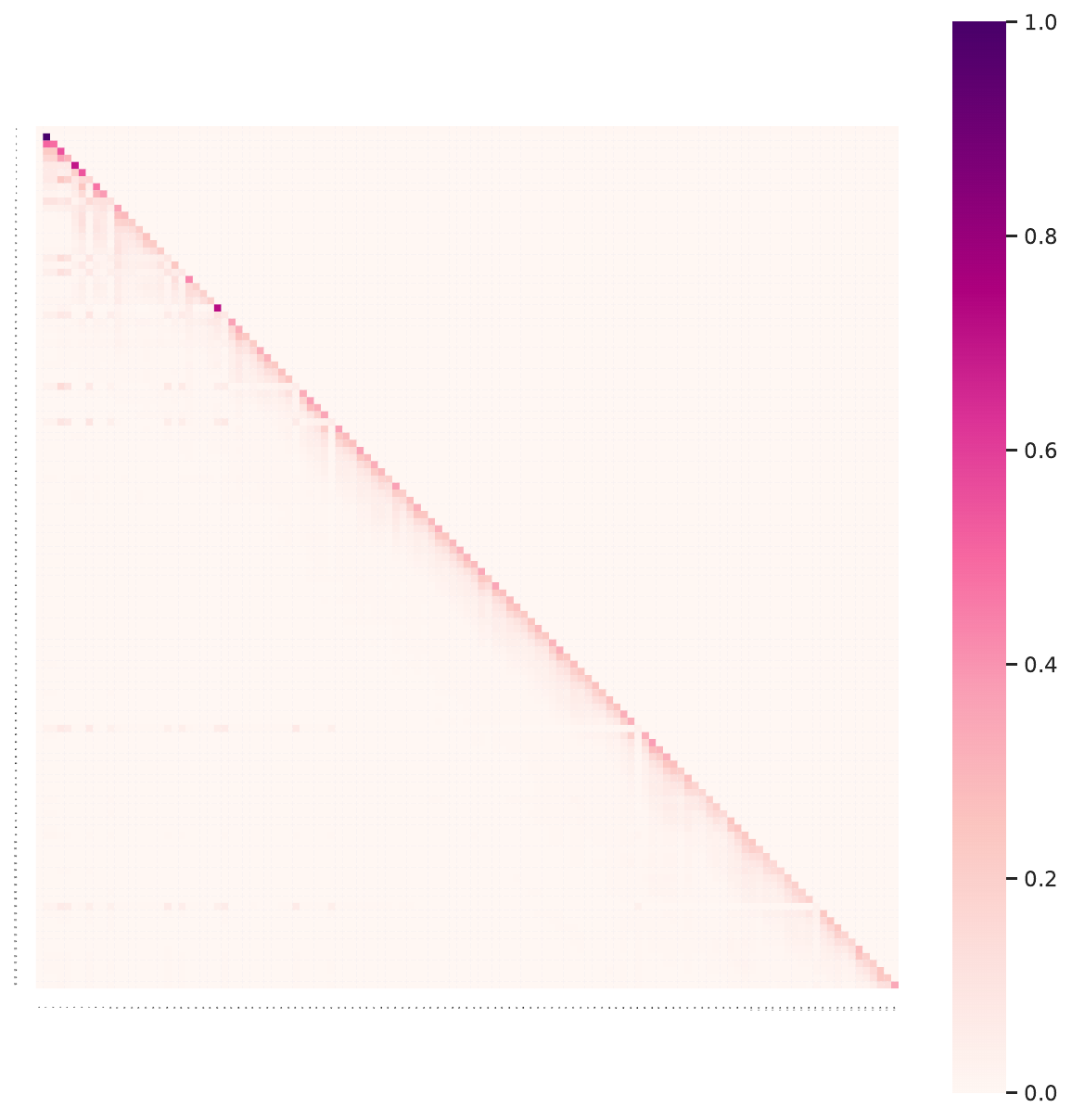}
        \caption*{Head 4}
    \end{minipage}
    \begin{minipage}{0.23\textwidth}
        \includegraphics[width=\linewidth]{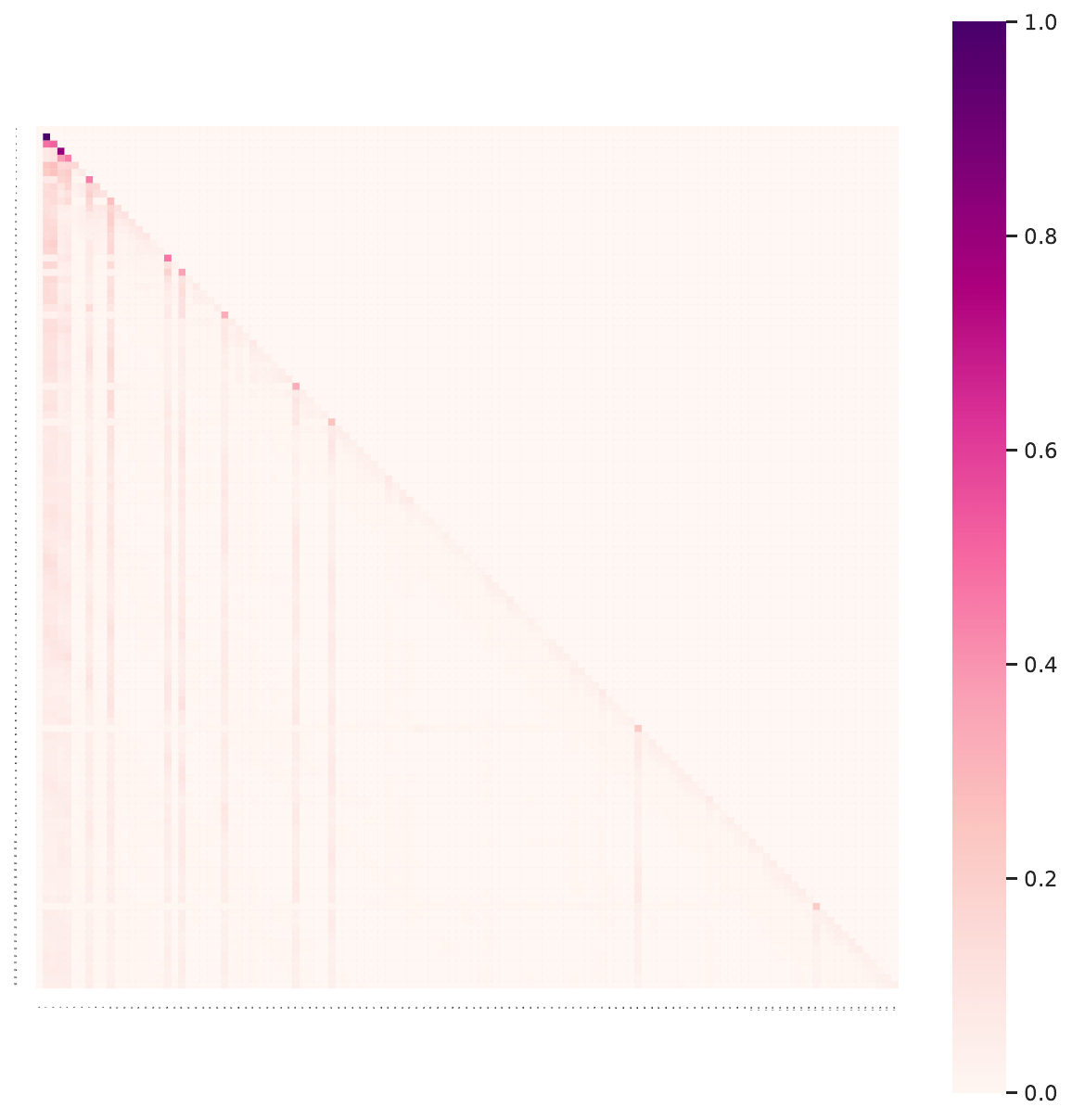}
        \caption*{Head 5}
    \end{minipage}
    \begin{minipage}{0.23\textwidth}
        \includegraphics[width=\linewidth]{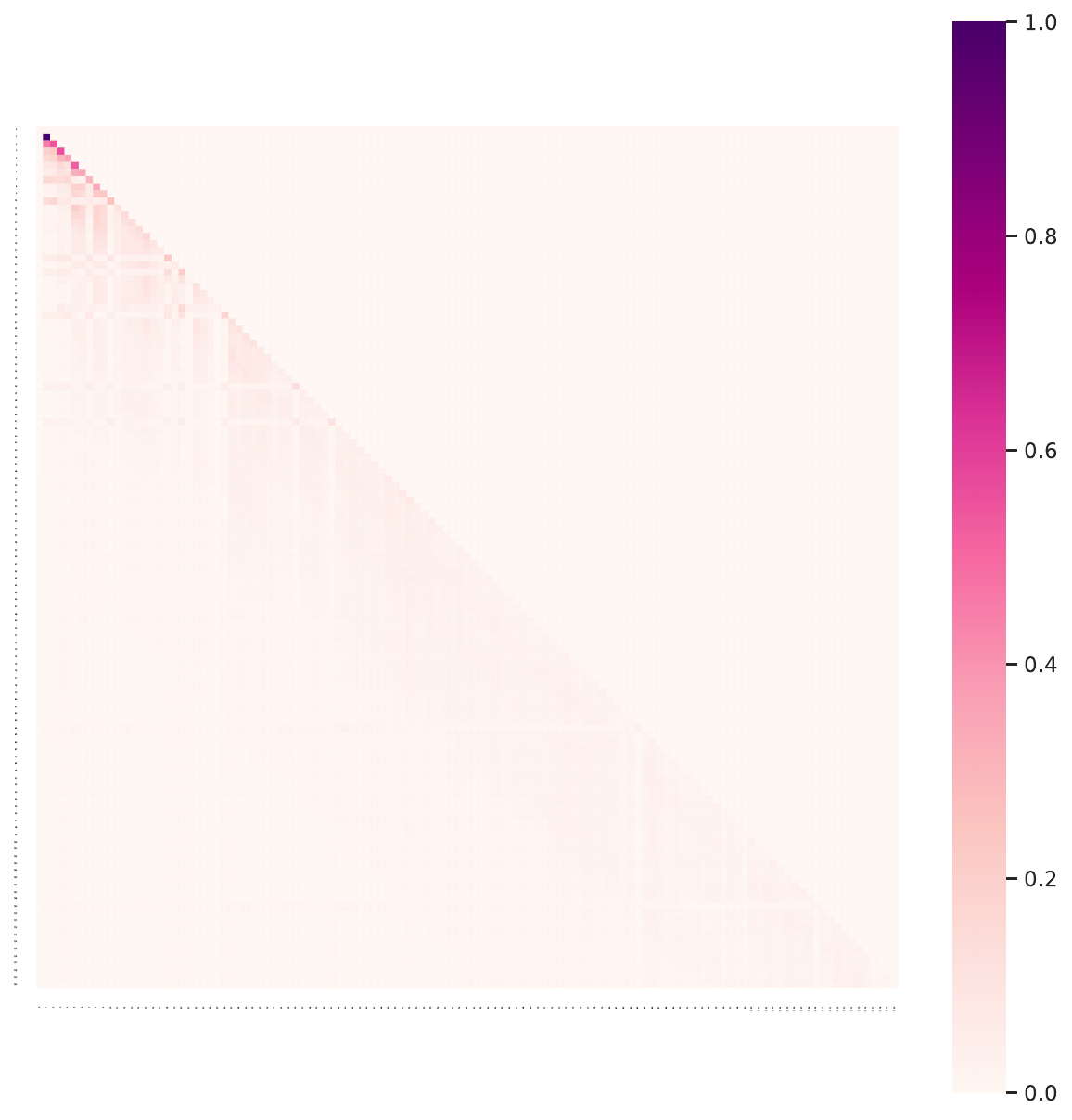}
        \caption*{Head 6}
    \end{minipage}
    \begin{minipage}{0.23\textwidth}
        \includegraphics[width=\linewidth]{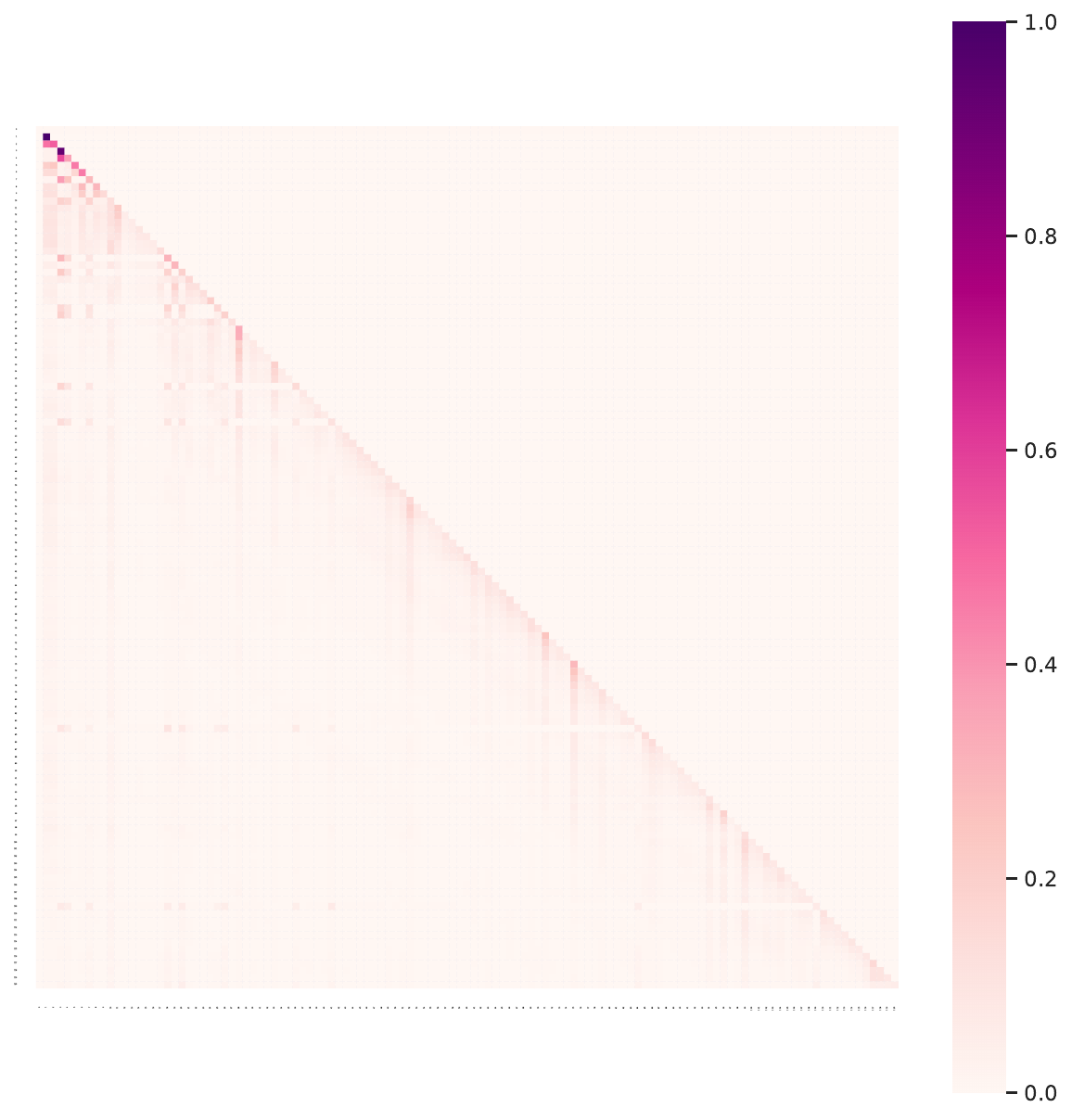}
        \caption*{Head 7}
    \end{minipage}
\vspace{1em}
\begin{minipage}{0.23\textwidth}
        \includegraphics[width=\linewidth]{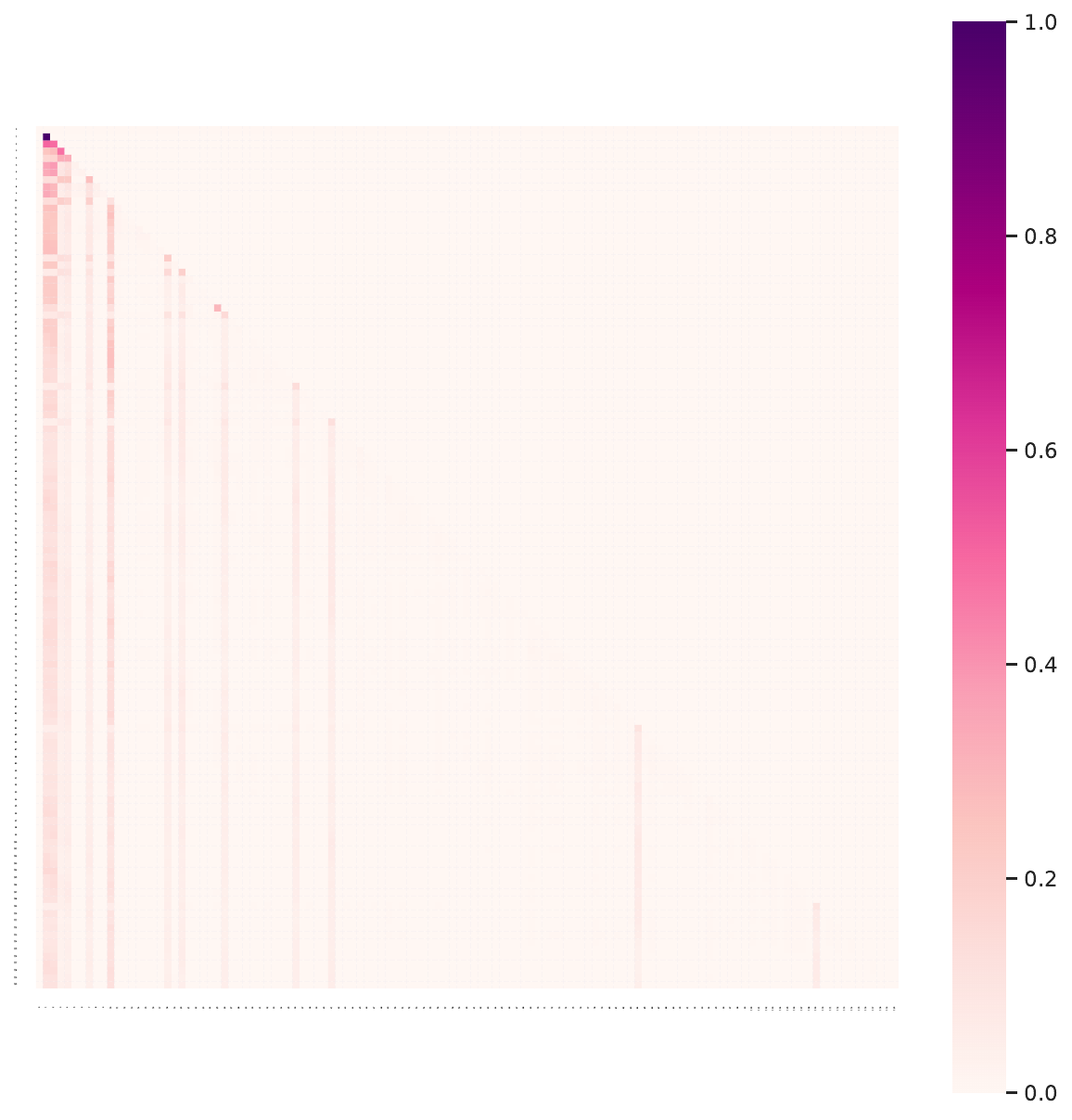}
        \caption*{Head 8}
    \end{minipage}
    \begin{minipage}{0.23\textwidth}
        \includegraphics[width=\linewidth]{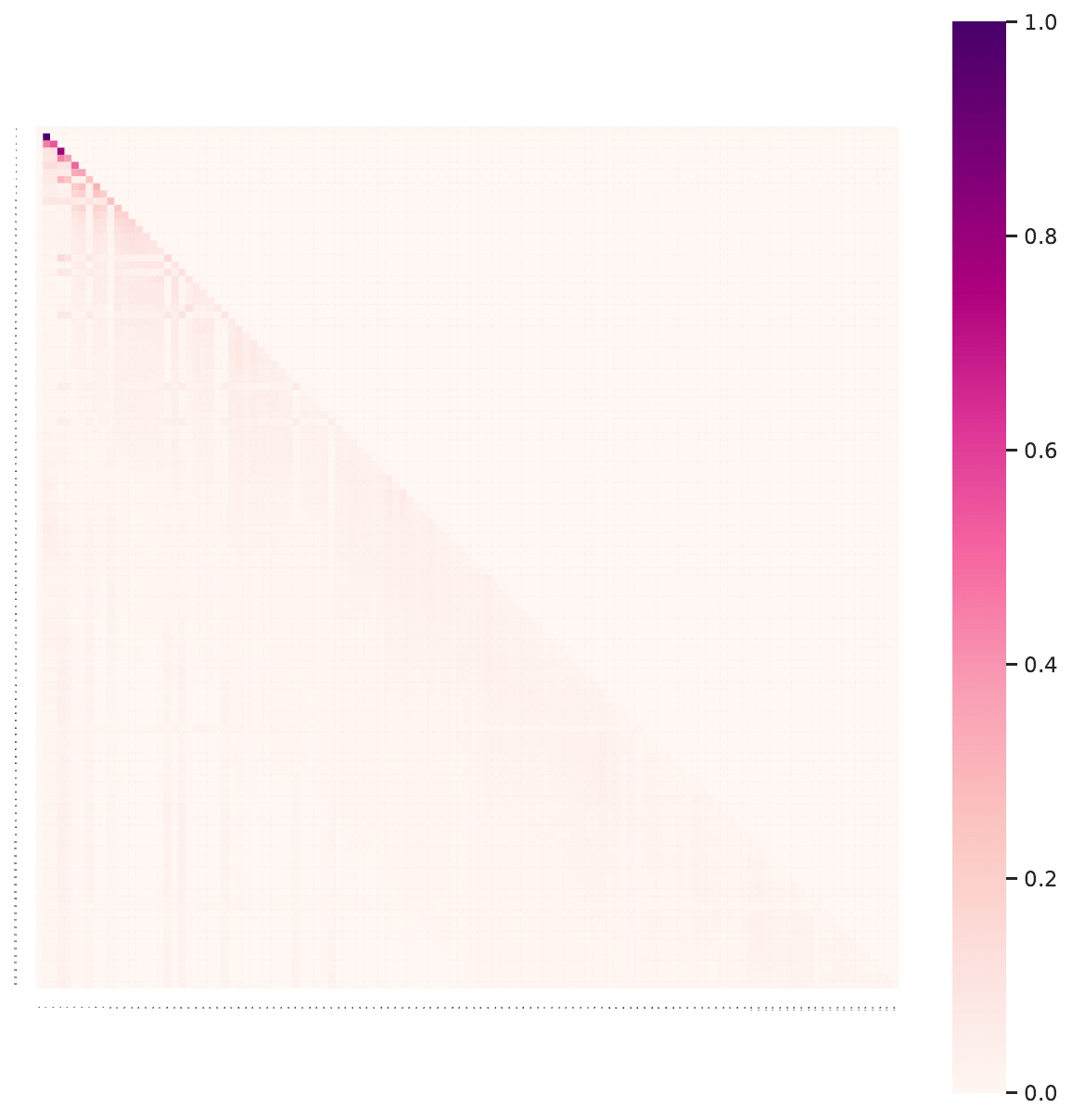}
        \caption*{Head 9}
    \end{minipage}
    \begin{minipage}{0.23\textwidth}
        \includegraphics[width=\linewidth]{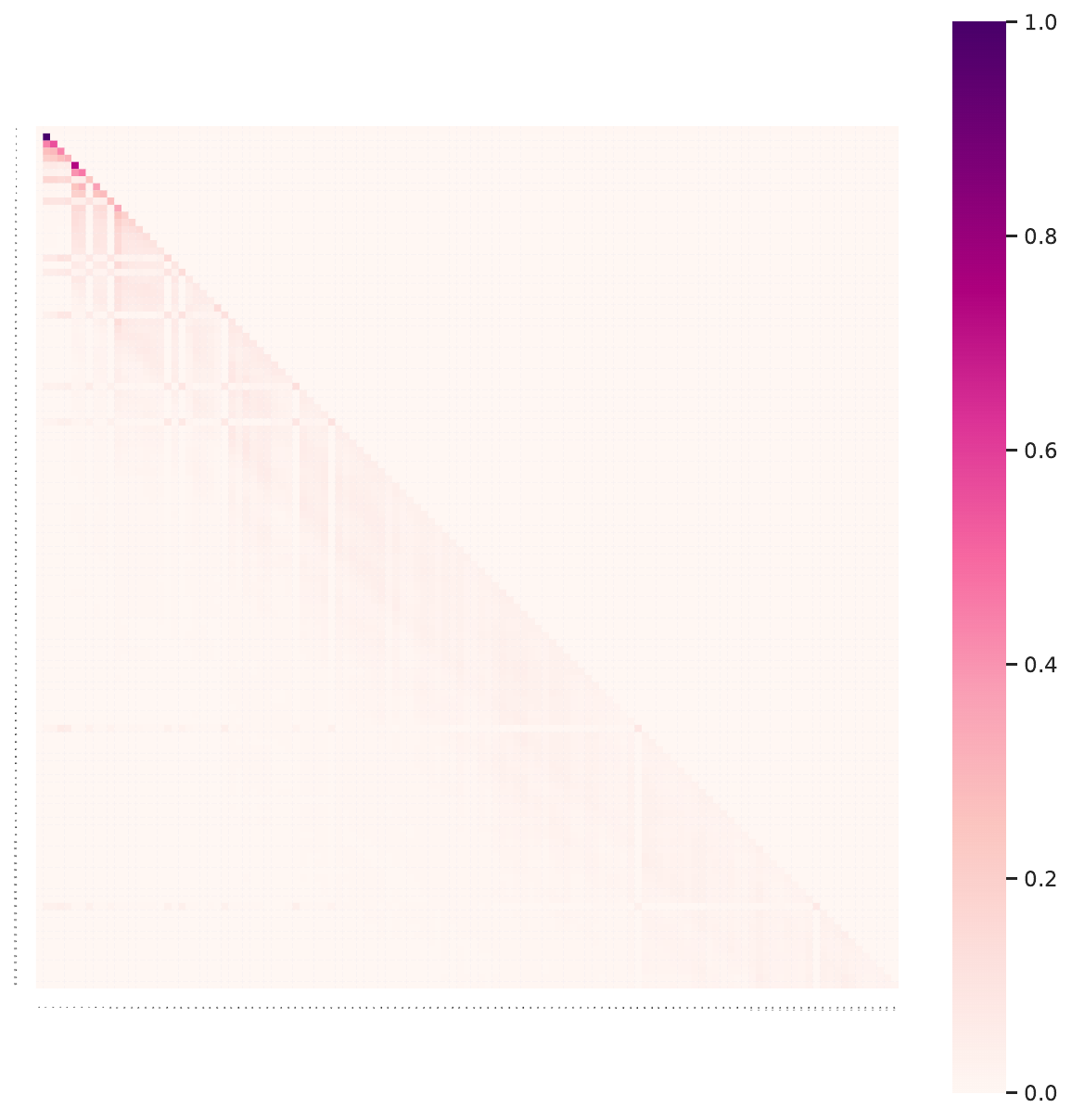}
        \caption*{Head 10}
    \end{minipage}
    \begin{minipage}{0.23\textwidth}
        \includegraphics[width=\linewidth]{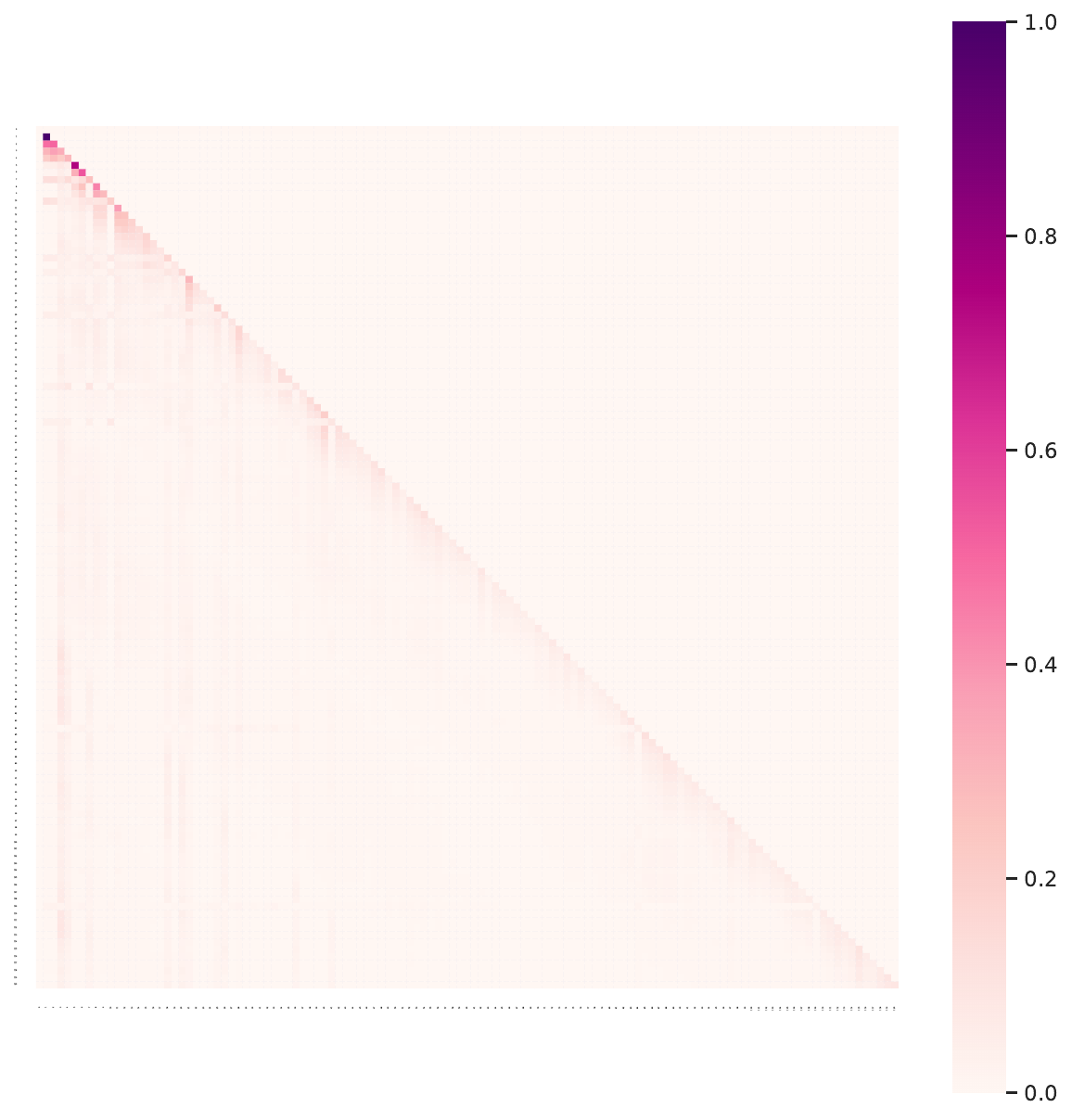}
        \caption*{Head 11}
    \end{minipage}
    \vspace{1em}
\begin{minipage}{0.23\textwidth}
        \includegraphics[width=\linewidth]{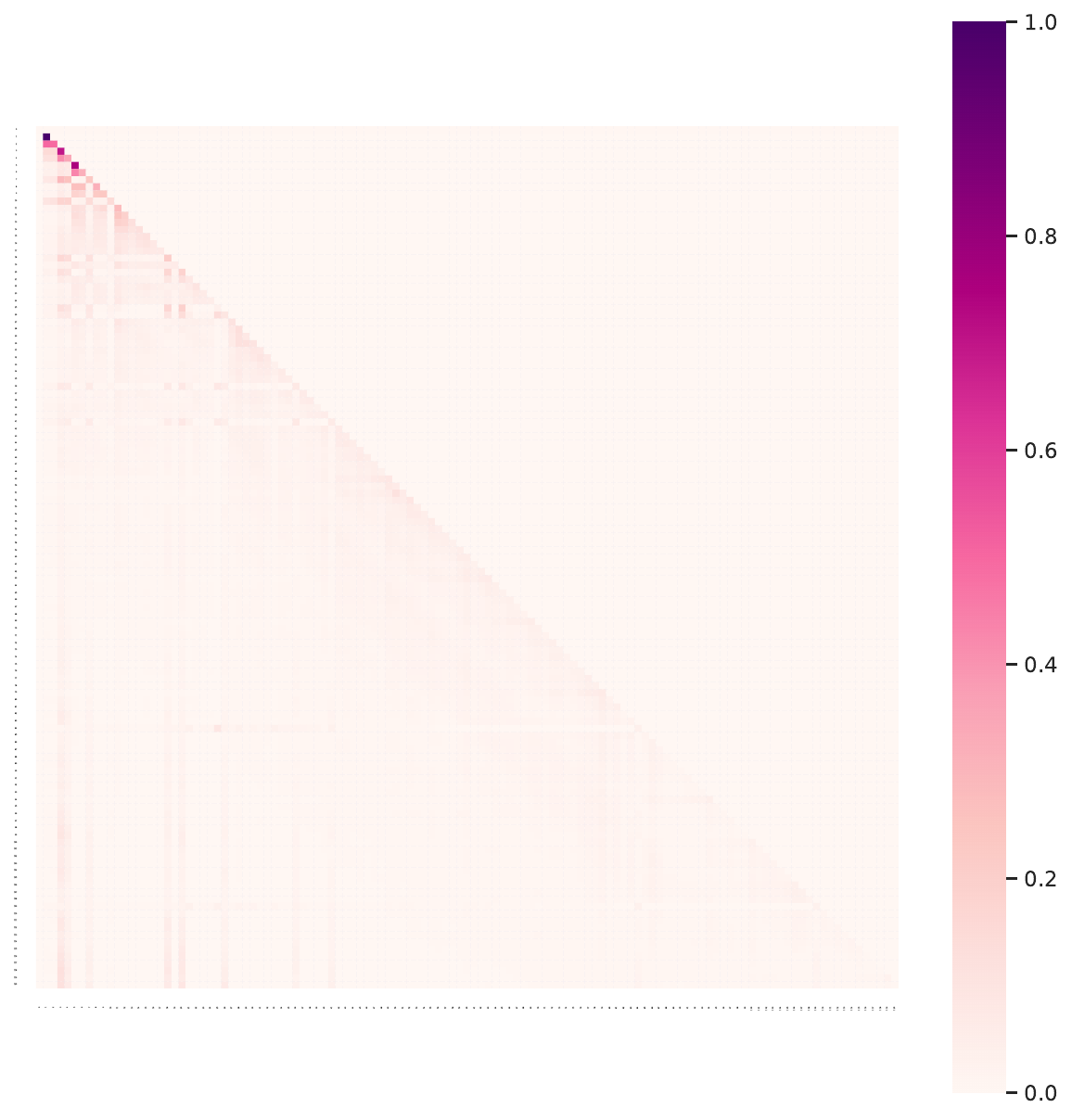}
        \caption*{Head 12}
    \end{minipage}
    \begin{minipage}{0.23\textwidth}
        \includegraphics[width=\linewidth]{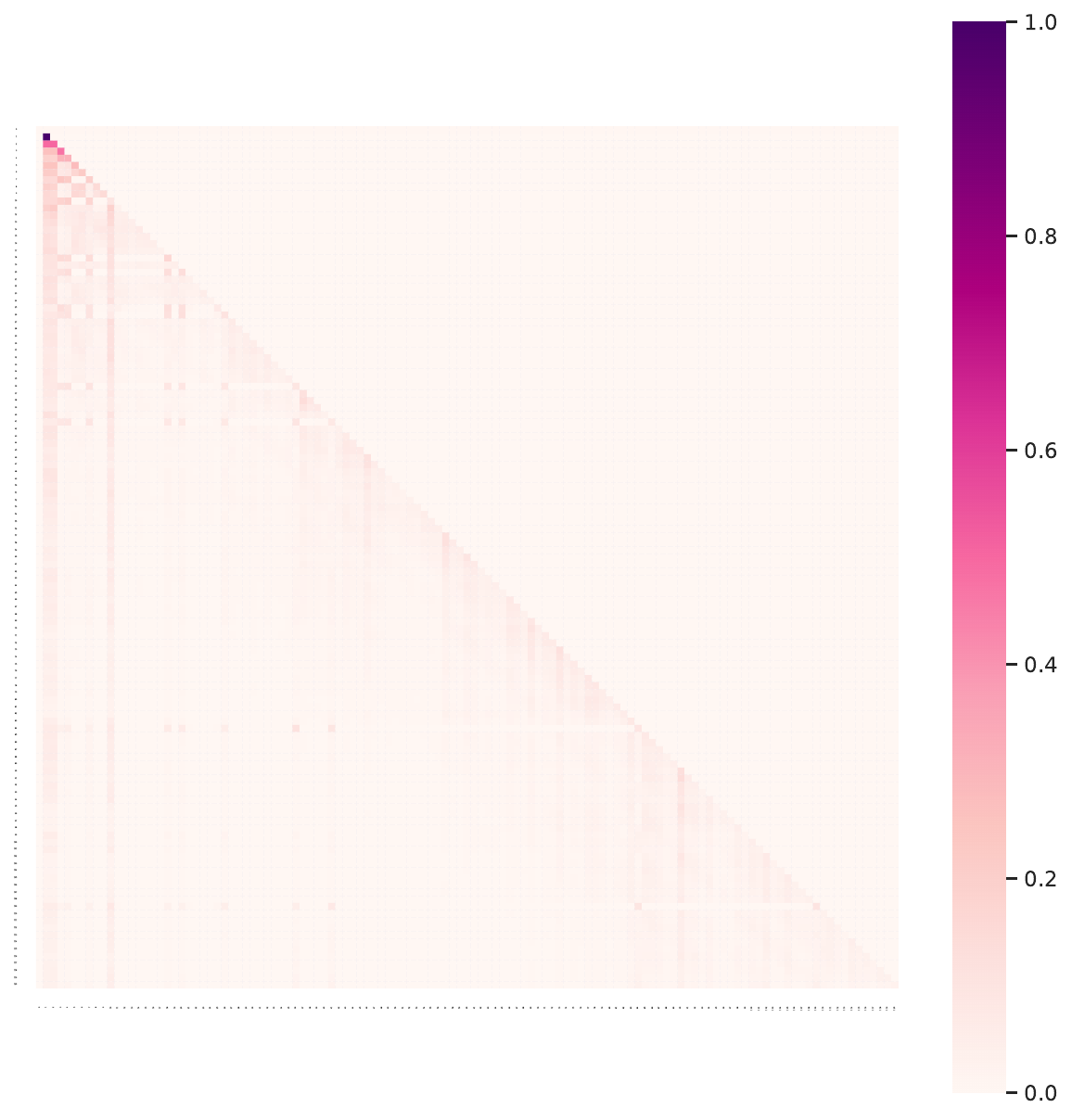}
        \caption*{Head 13}
    \end{minipage}
    \begin{minipage}{0.23\textwidth}
        \includegraphics[width=\linewidth]{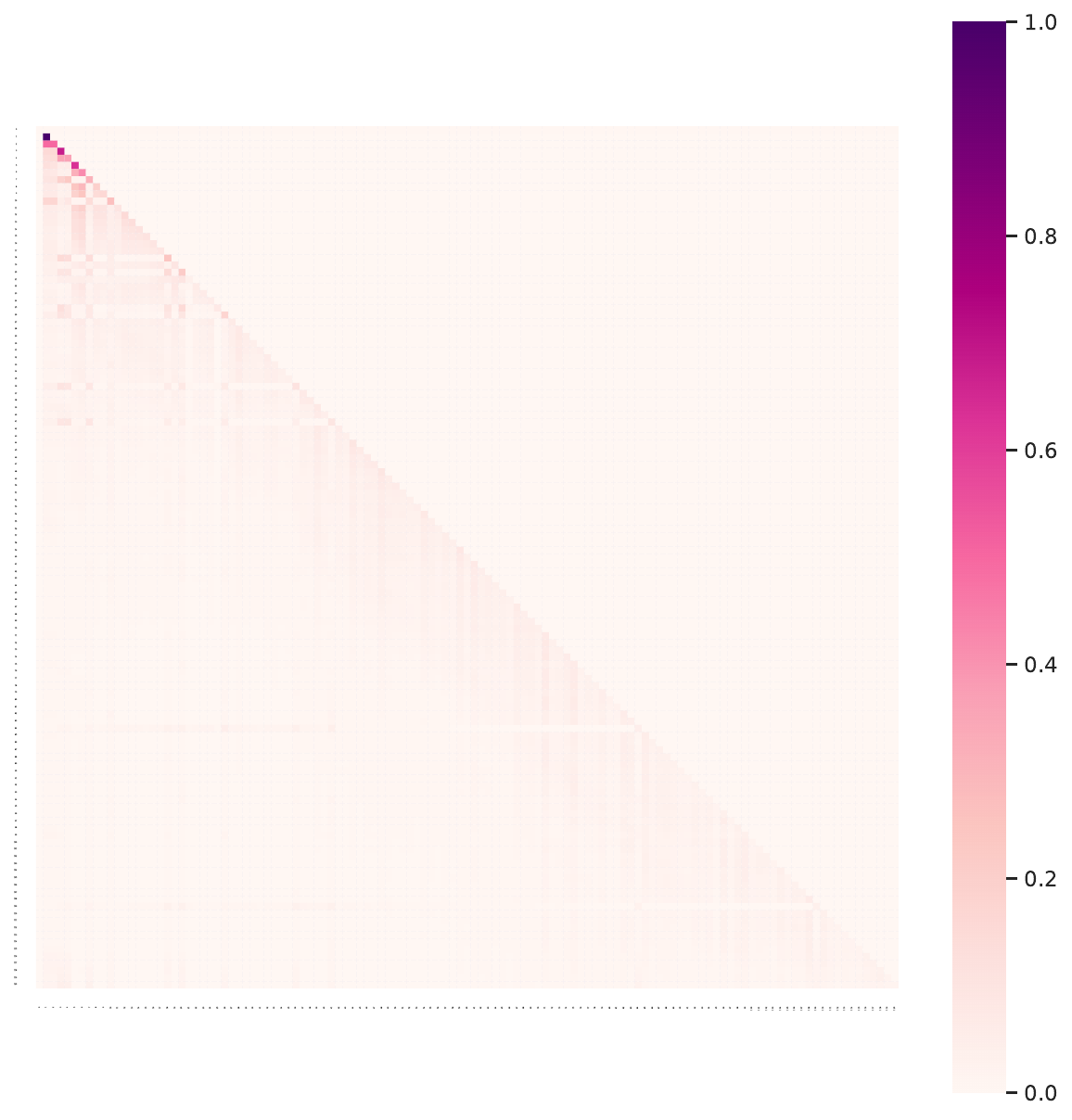}
        \caption*{Head 14}
    \end{minipage}
    \begin{minipage}{0.23\textwidth}
        \includegraphics[width=\linewidth]{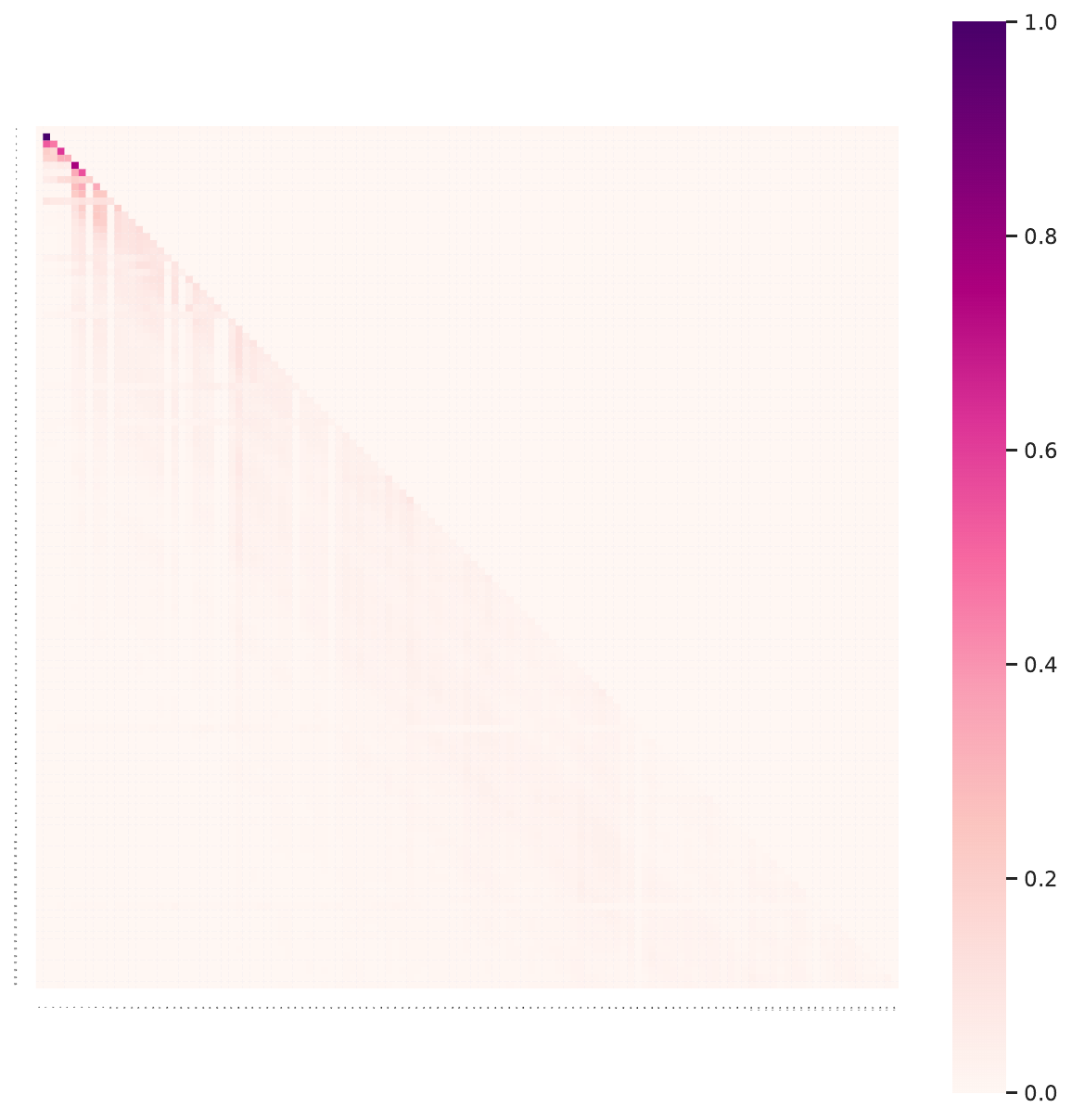}
        \caption*{Head 15}
    \end{minipage}
\caption{\textbf{The attention score of our CRFT in layer 31.} (part 1 of 2)}
\label{Fig: vislayerdde31_1}
\end{figure*}

\begin{figure*}[tbp]
\centering
\begin{minipage}{0.23\textwidth}
        \includegraphics[width=\linewidth]{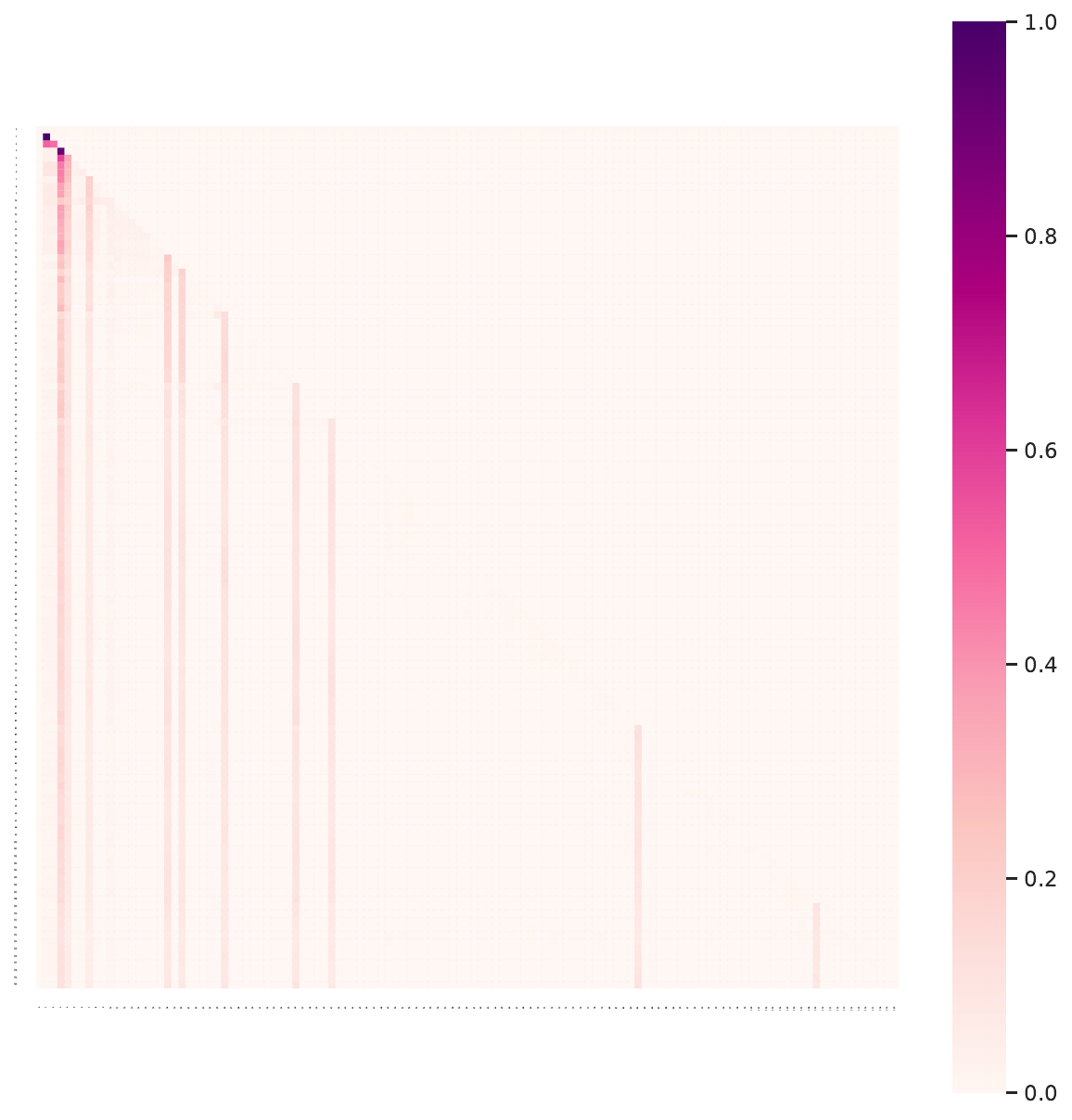}
        \caption*{Head 16}
    \end{minipage}
    \begin{minipage}{0.23\textwidth}
        \includegraphics[width=\linewidth]{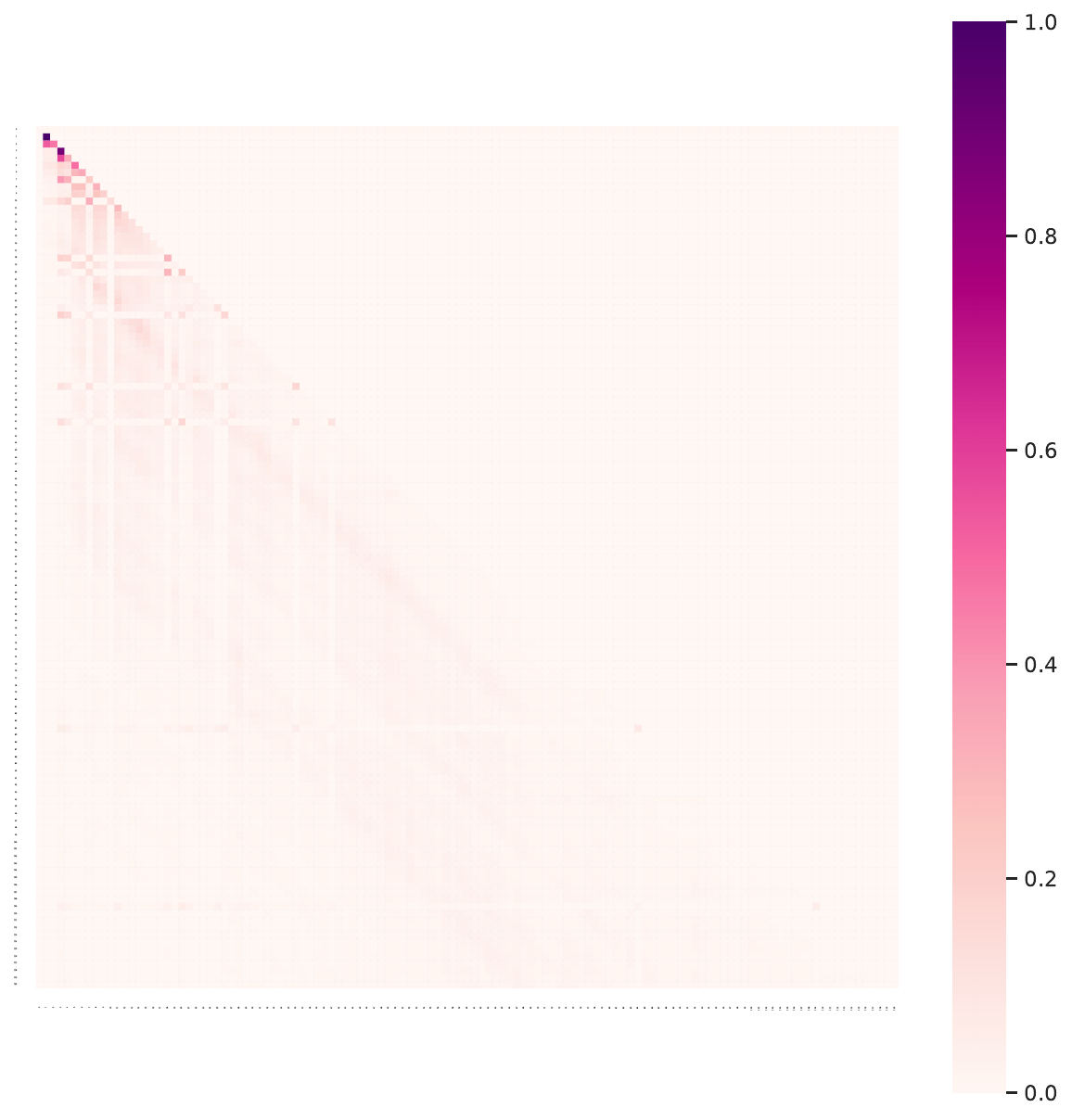}
        \caption*{Head 17}
    \end{minipage}
    \begin{minipage}{0.23\textwidth}
        \includegraphics[width=\linewidth]{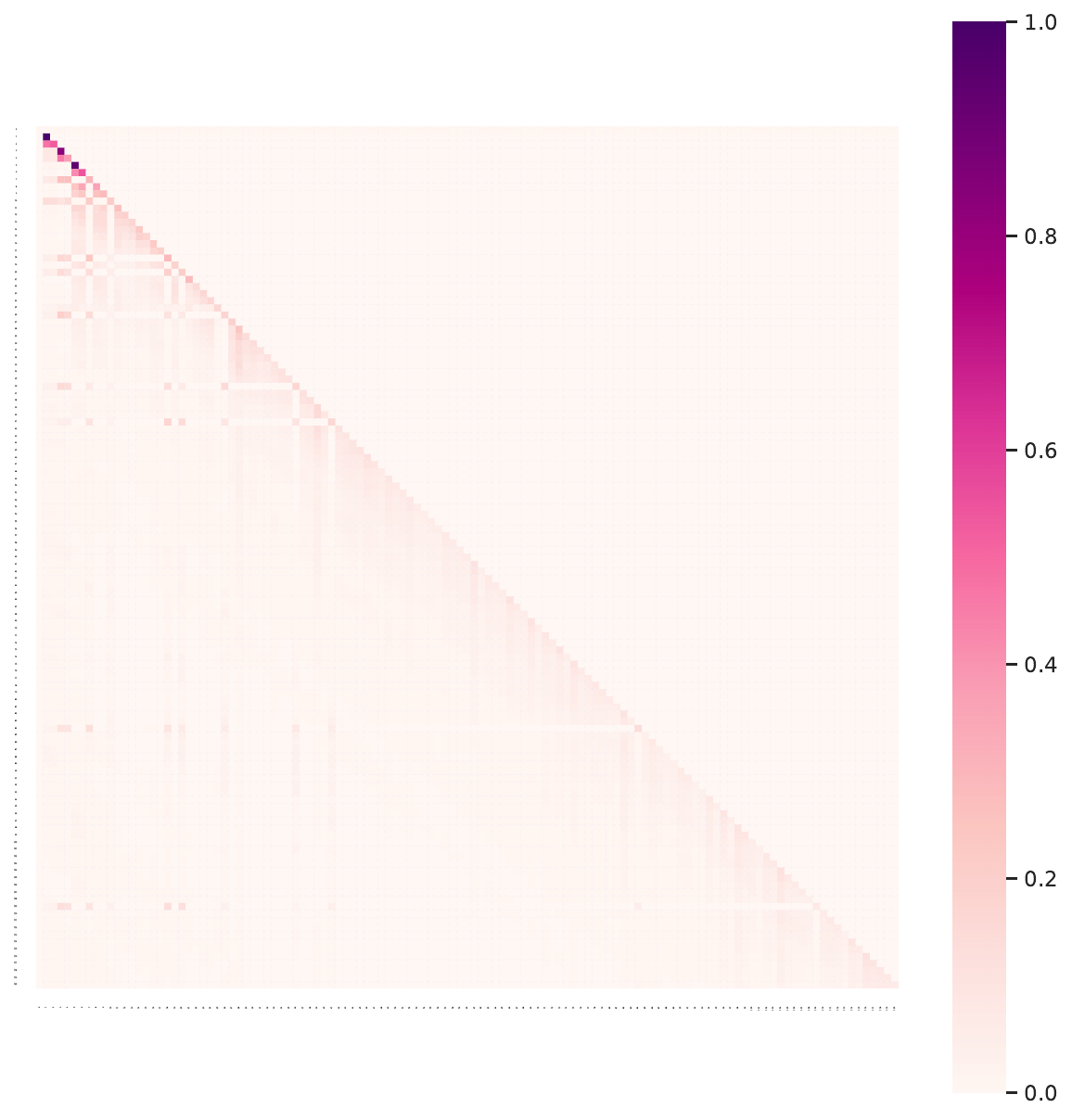}
        \caption*{Head 18}
    \end{minipage}
    \begin{minipage}{0.23\textwidth}
        \includegraphics[width=\linewidth]{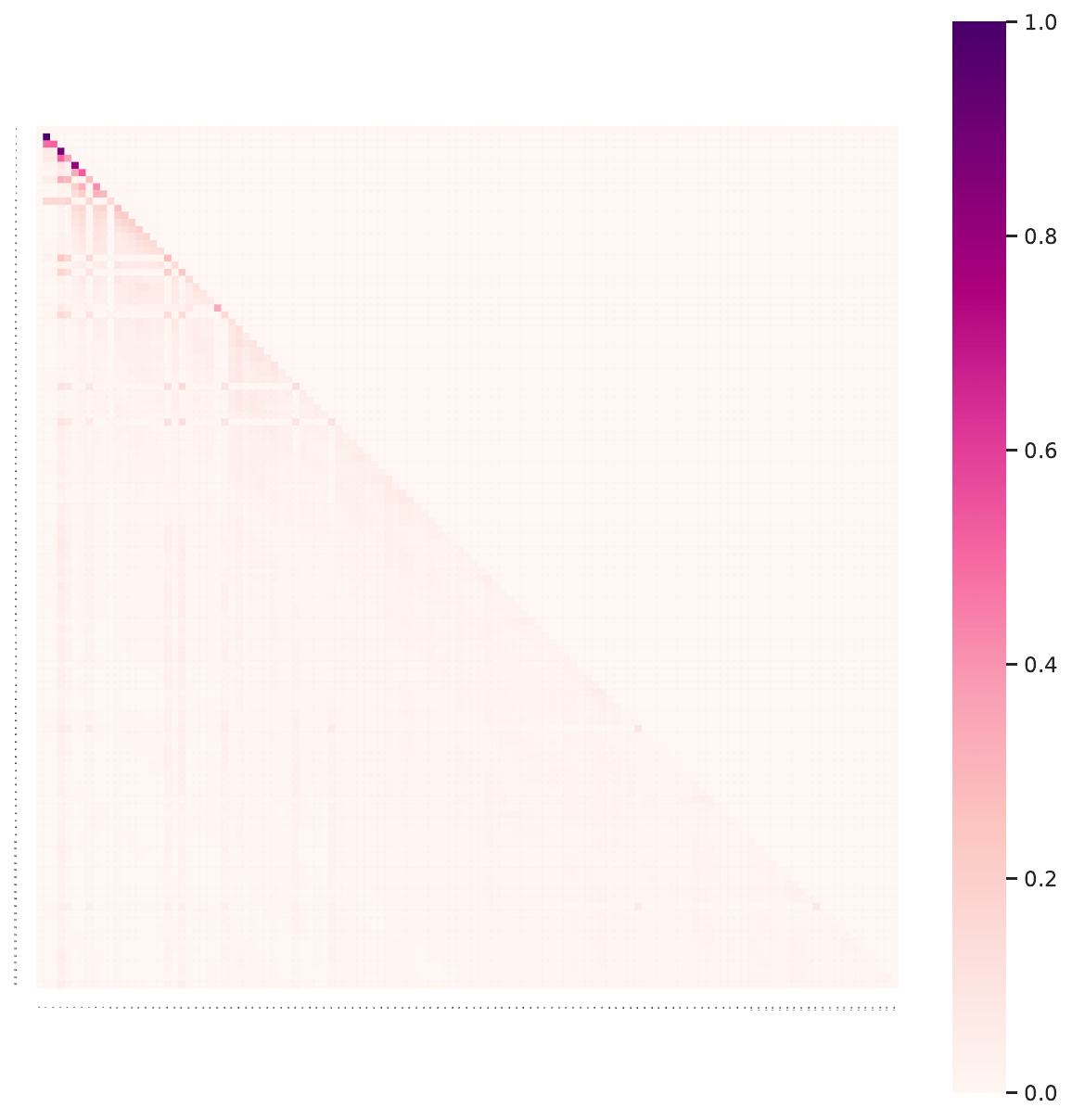}
        \caption*{Head 19}
    \end{minipage}
\vspace{1em}
\begin{minipage}{0.23\textwidth}
        \includegraphics[width=\linewidth]{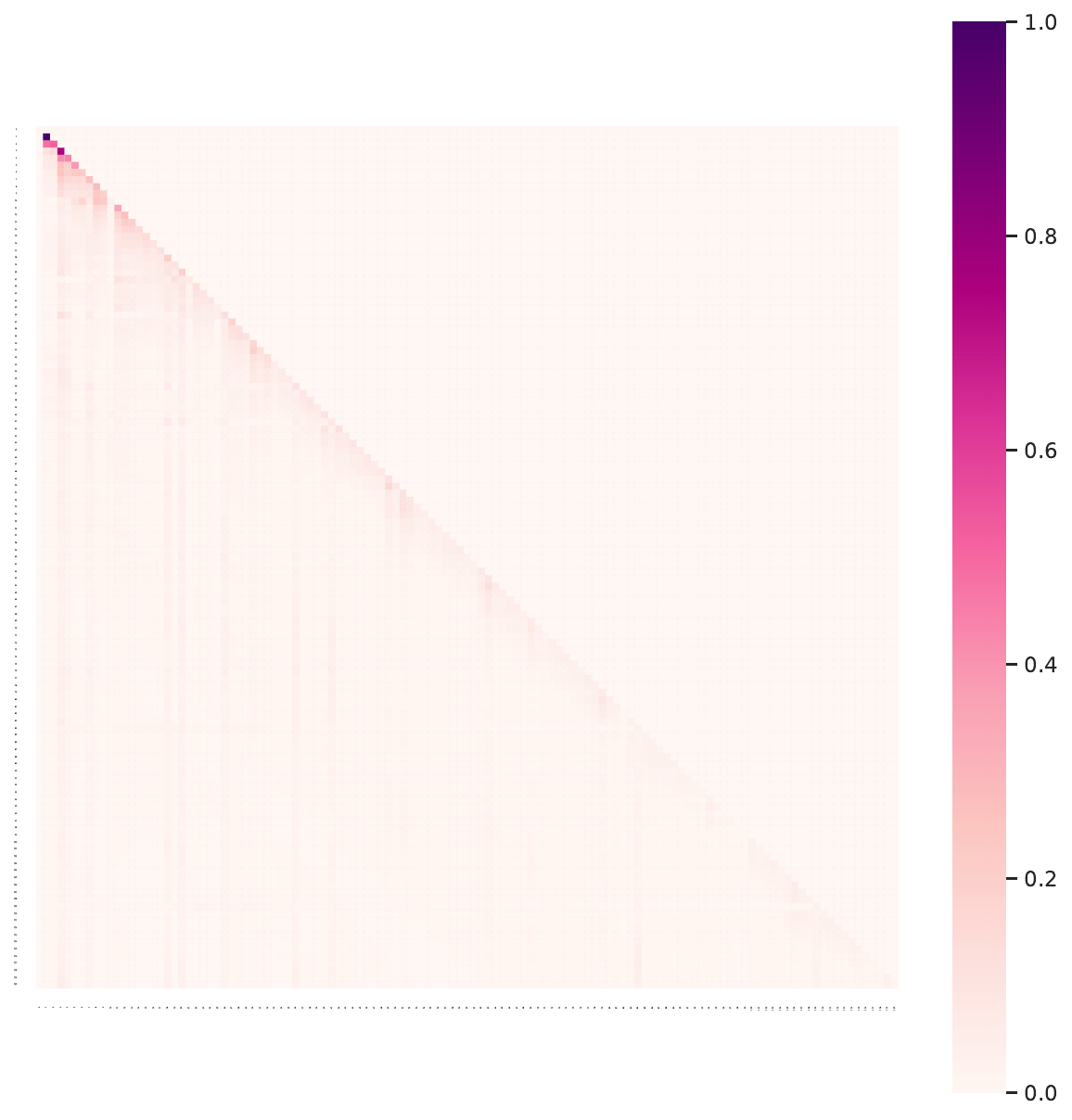}
        \caption*{Head 20}
    \end{minipage}
    \begin{minipage}{0.23\textwidth}
        \includegraphics[width=\linewidth]{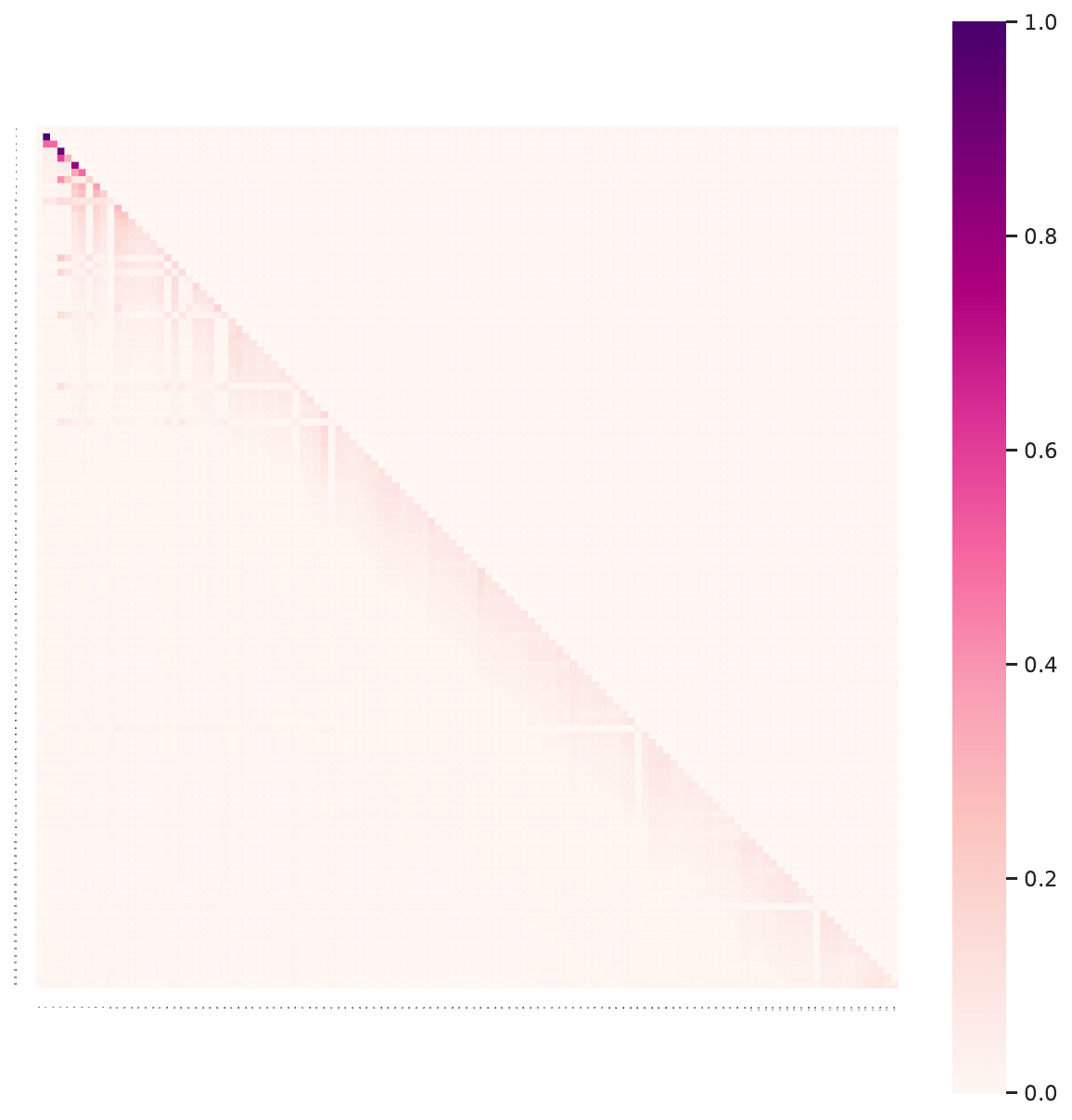}
        \caption*{Head 21}
    \end{minipage}
    \begin{minipage}{0.23\textwidth}
        \includegraphics[width=\linewidth]{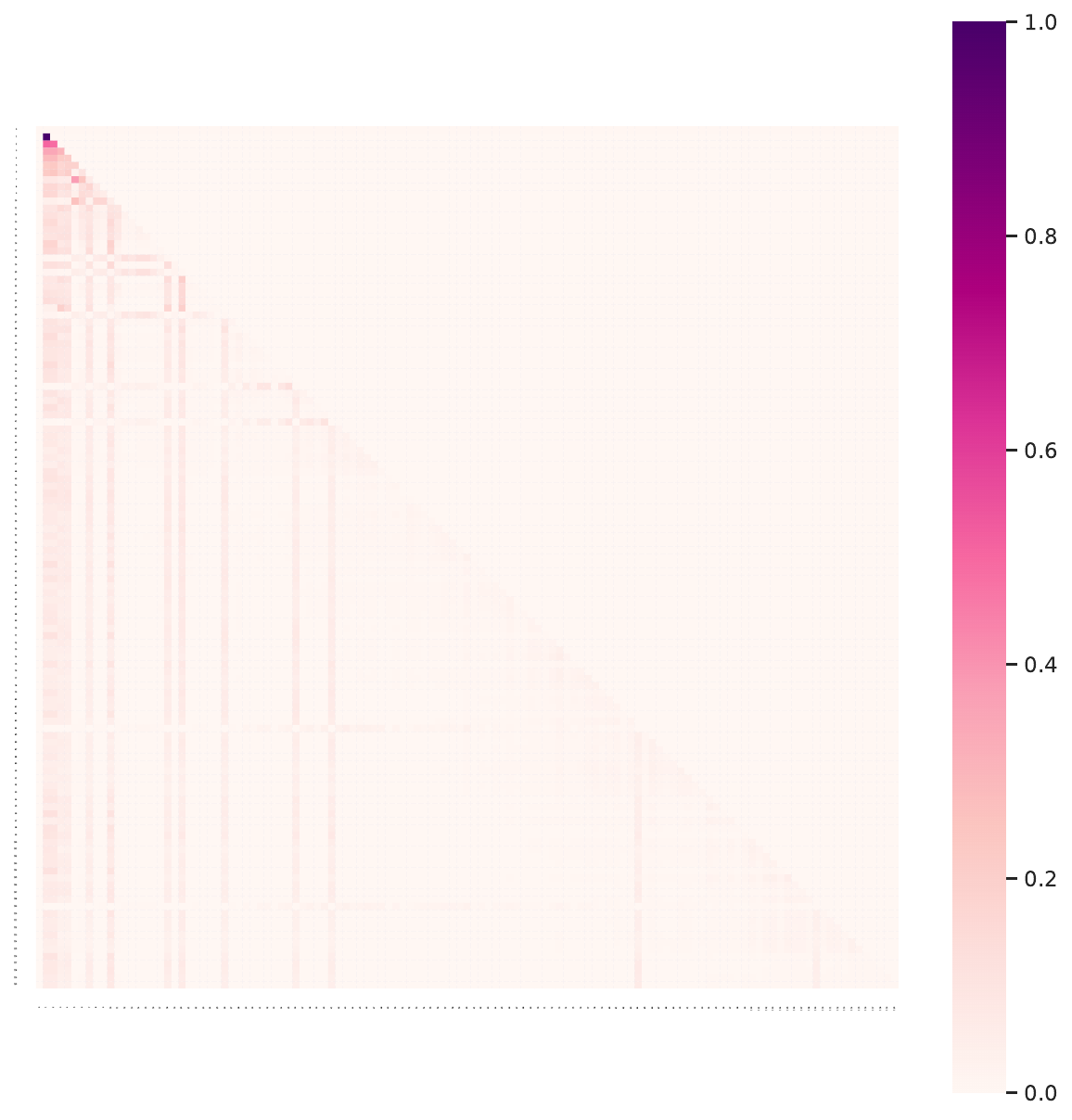}
        \caption*{Head 22}
    \end{minipage}
    \begin{minipage}{0.23\textwidth}
        \includegraphics[width=\linewidth]{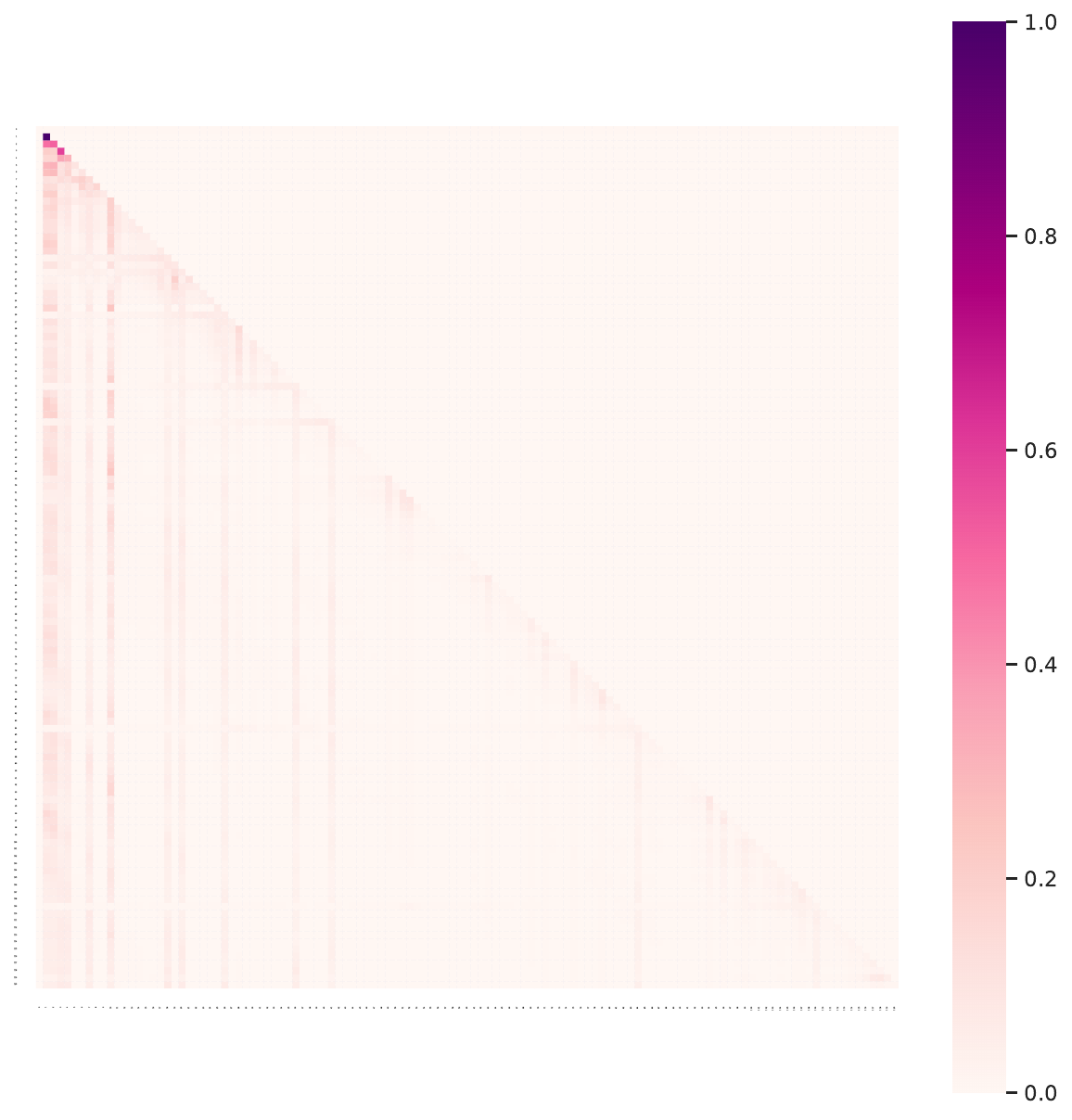}
        \caption*{Head 23}
    \end{minipage}
    \vspace{1em}
\begin{minipage}{0.23\textwidth}
        \includegraphics[width=\linewidth]{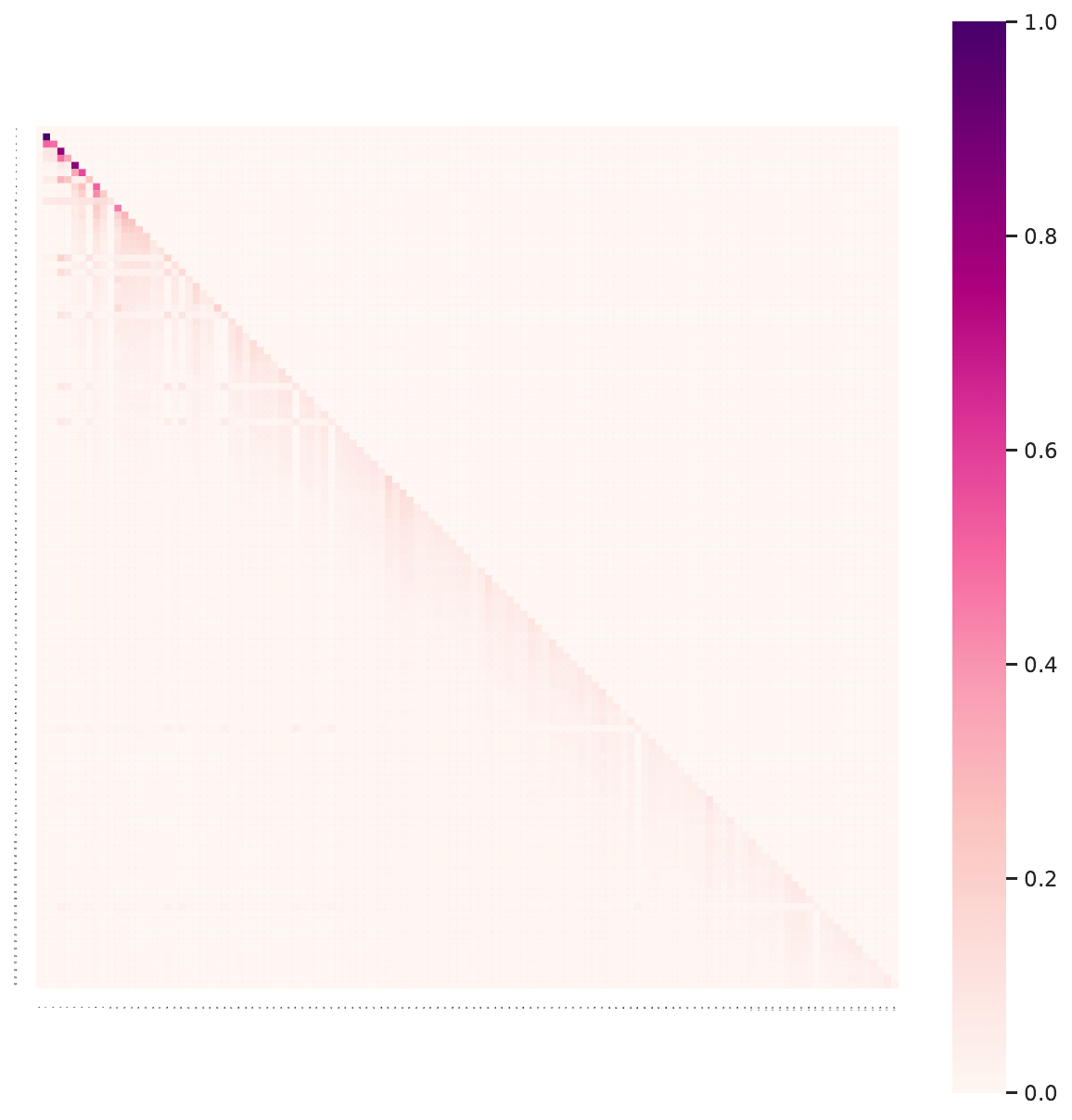}
        \caption*{Head 24}
    \end{minipage}
    \begin{minipage}{0.23\textwidth}
        \includegraphics[width=\linewidth]{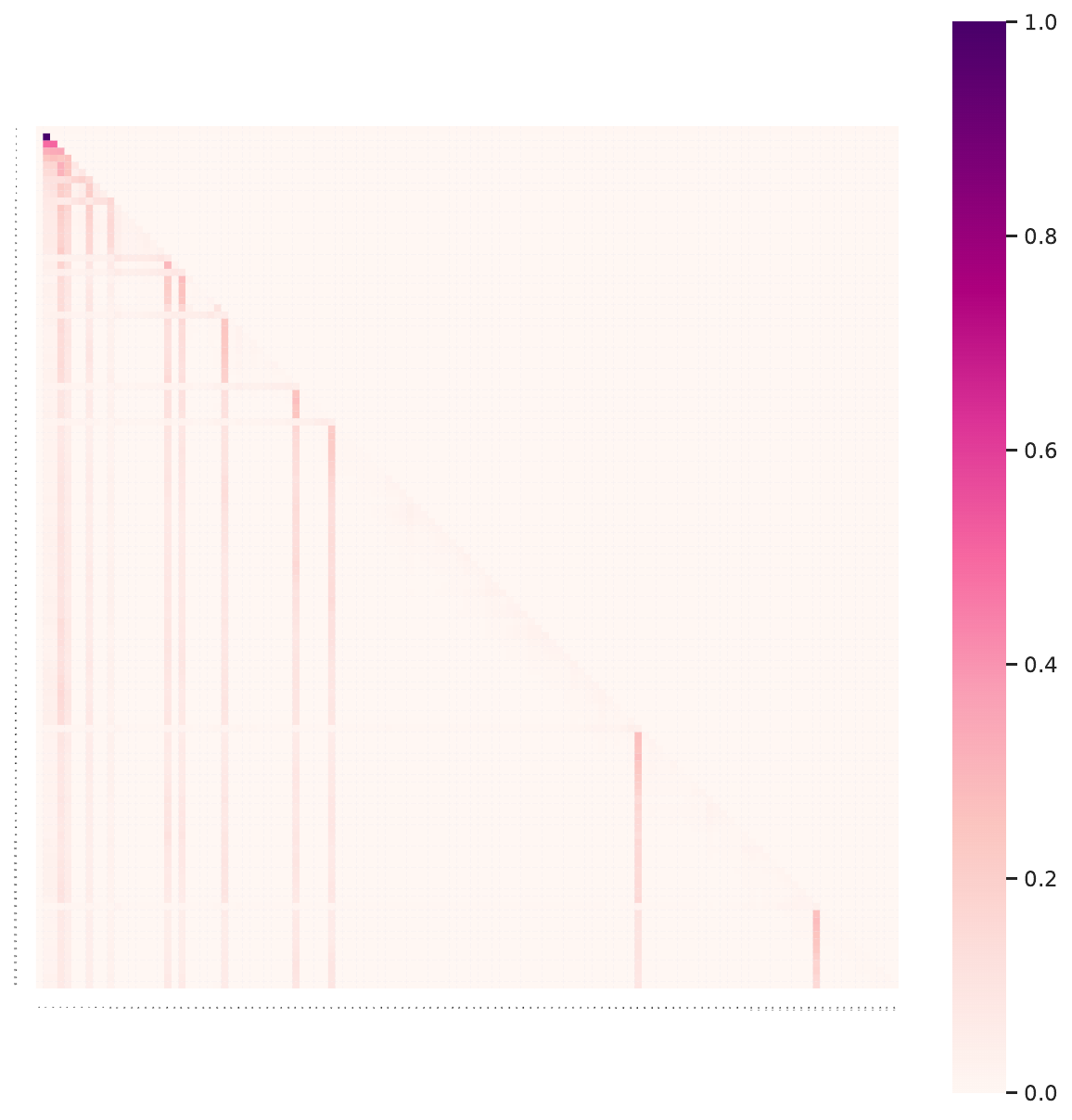}
        \caption*{Head 25}
    \end{minipage}
    \begin{minipage}{0.23\textwidth}
        \includegraphics[width=\linewidth]{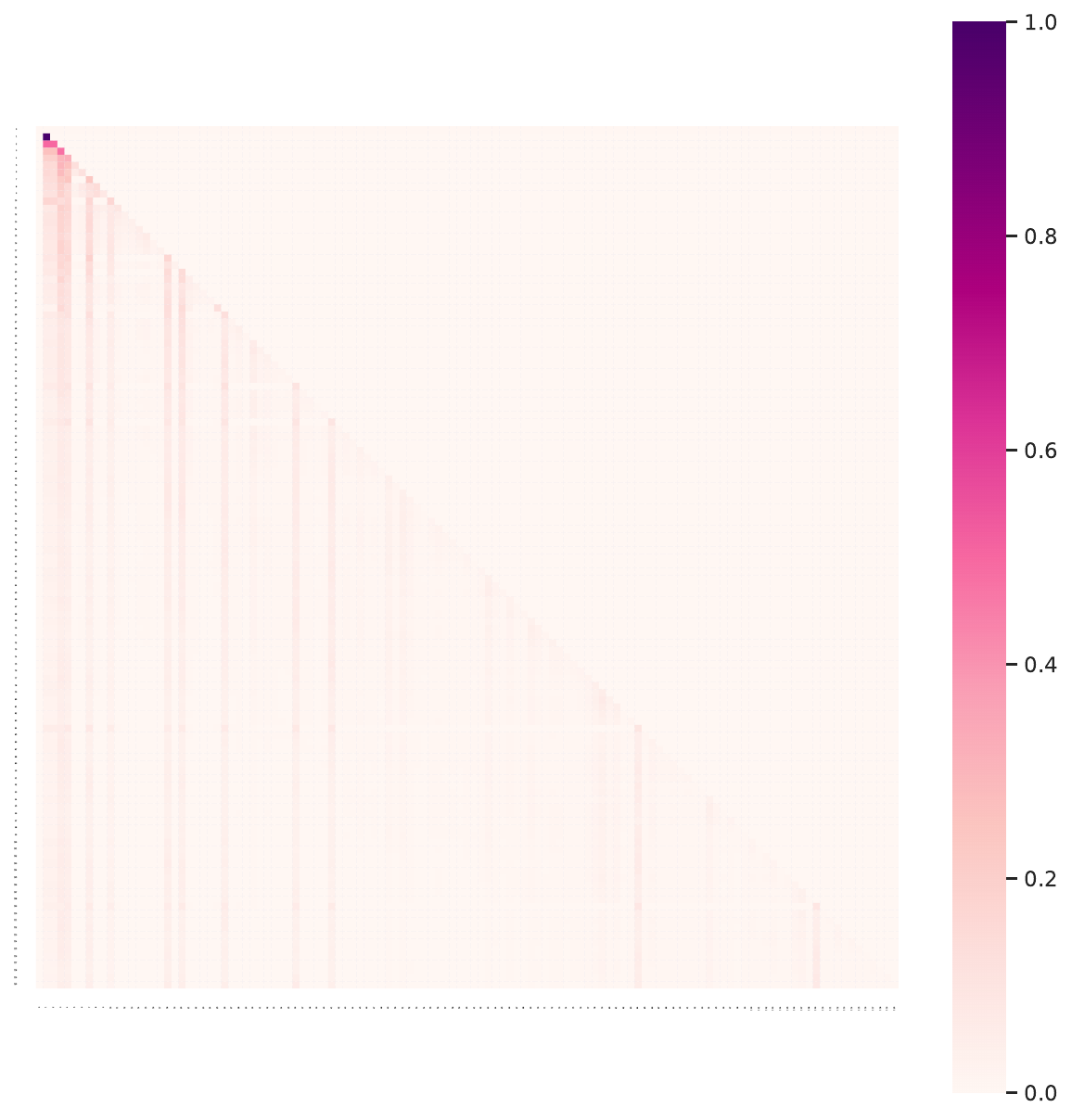}
        \caption*{Head 26}
    \end{minipage}
    \begin{minipage}{0.23\textwidth}
        \includegraphics[width=\linewidth]{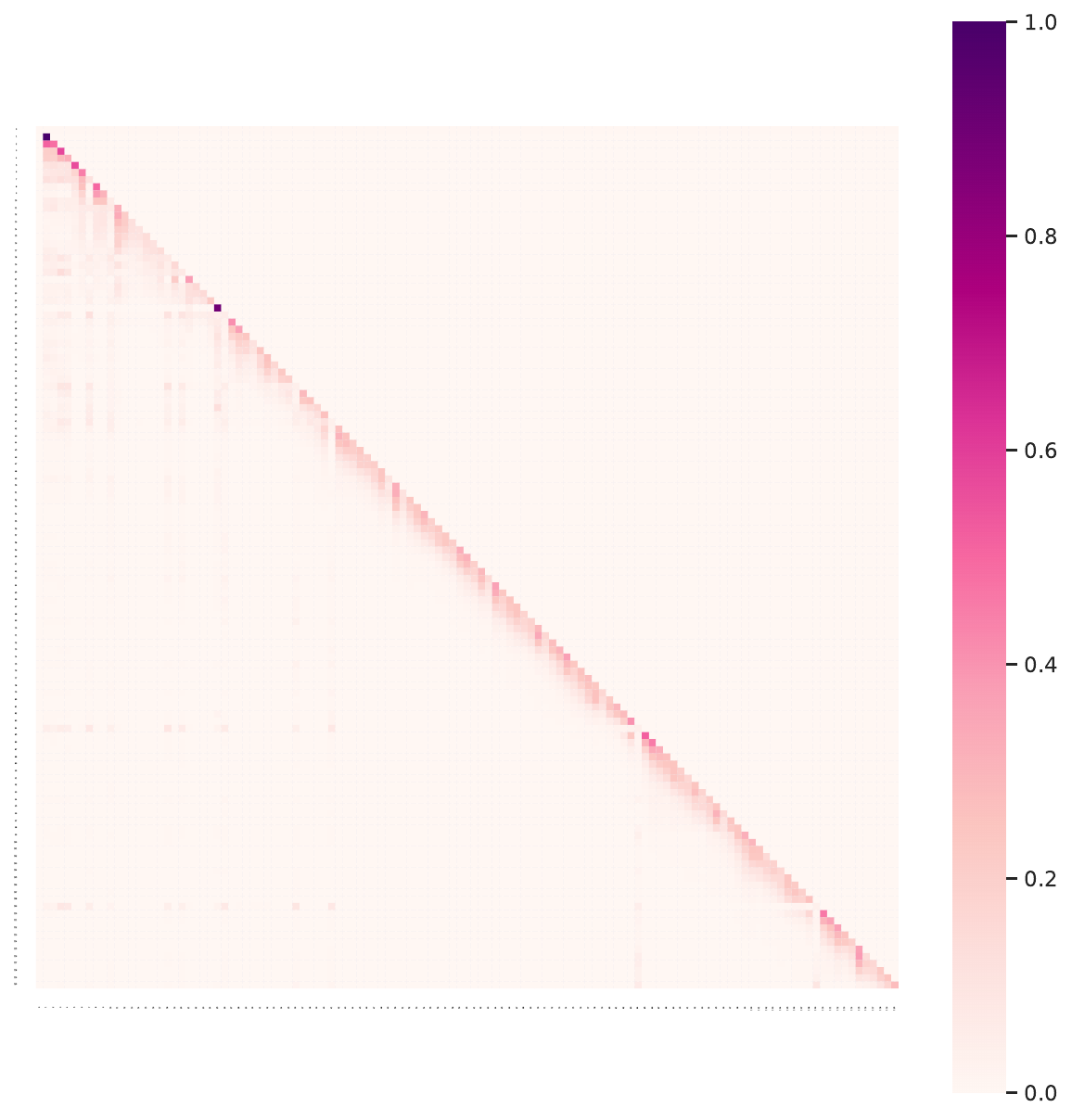}
        \caption*{Head 27}
    \end{minipage}
    \vspace{1em}
\begin{minipage}{0.23\textwidth}
        \includegraphics[width=\linewidth]{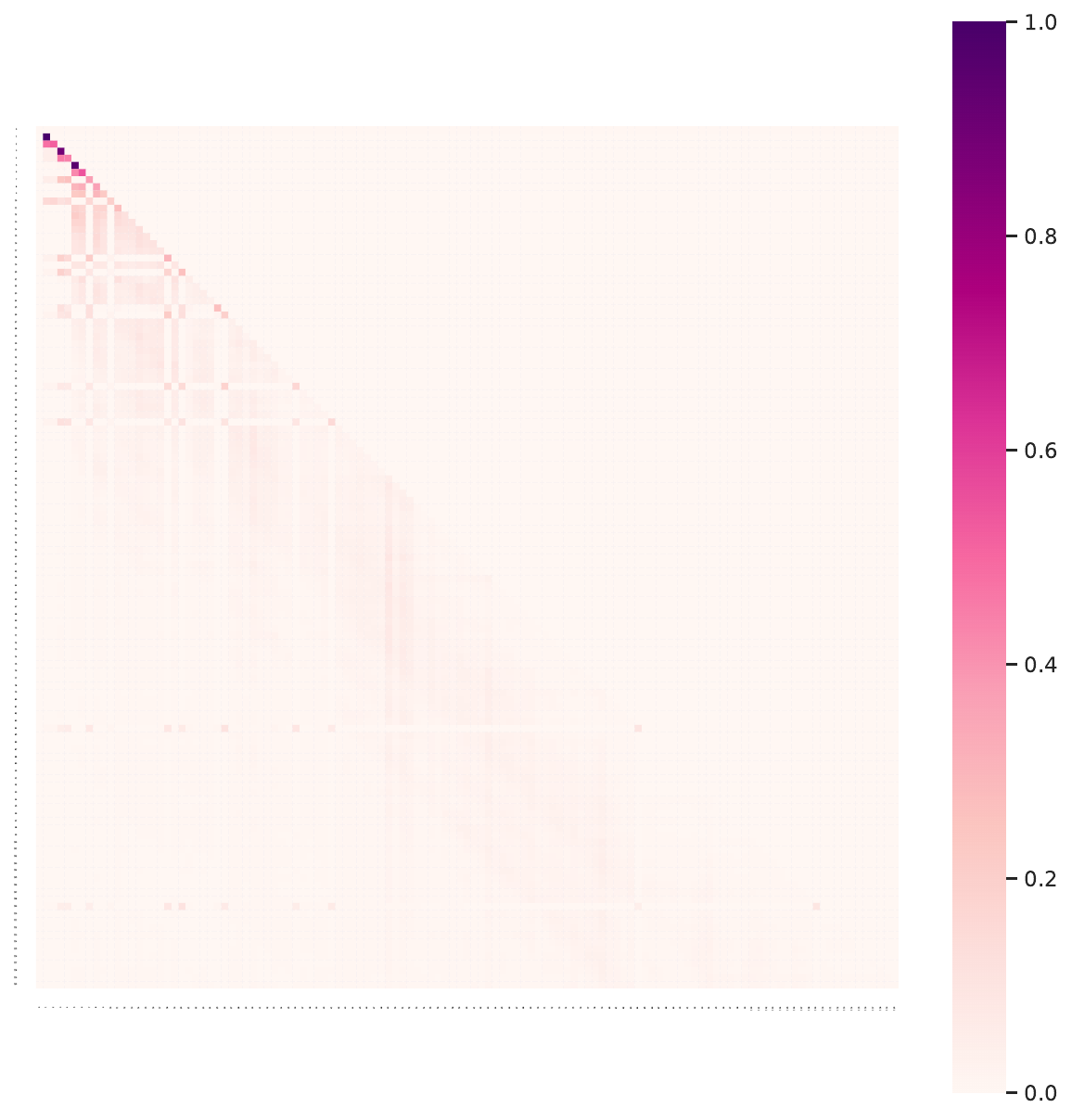}
        \caption*{Head 28}
    \end{minipage}
    \begin{minipage}{0.23\textwidth}
        \includegraphics[width=\linewidth]{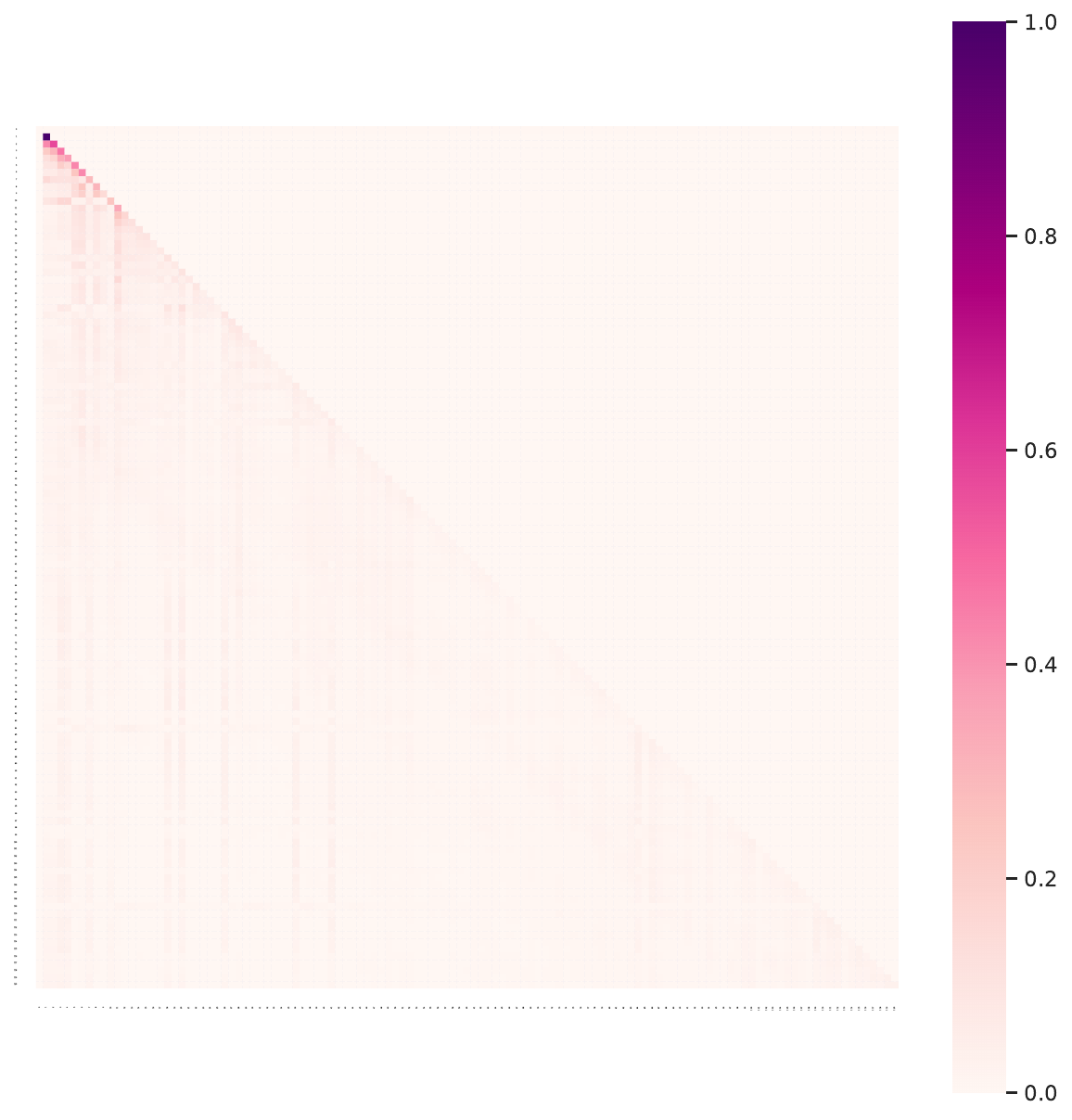}
        \caption*{Head 29}
    \end{minipage}
    \begin{minipage}{0.23\textwidth}
        \includegraphics[width=\linewidth]{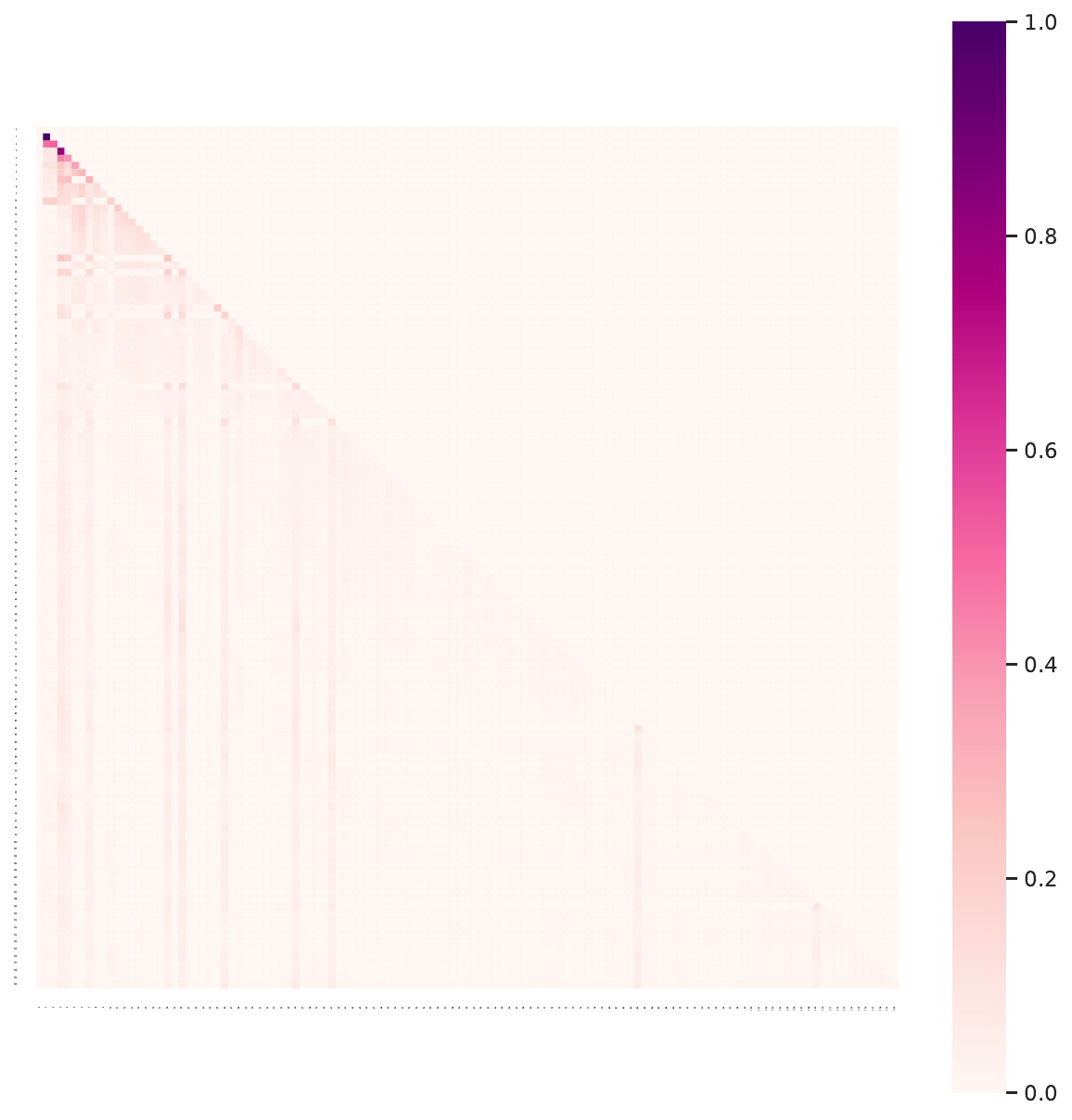}
        \caption*{Head 30}
    \end{minipage}
    \begin{minipage}{0.23\textwidth}
        \includegraphics[width=\linewidth]{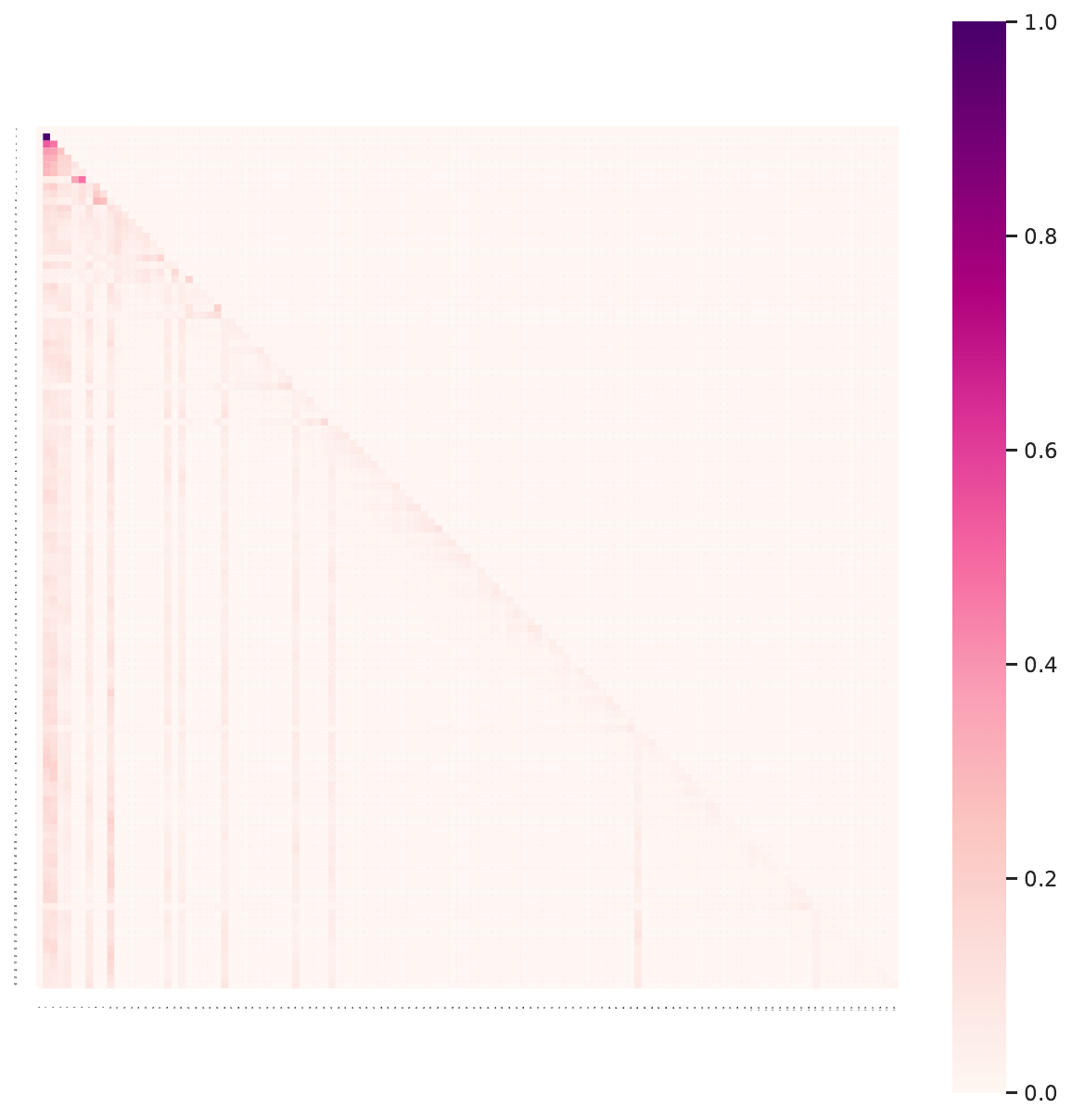}
        \caption*{Head 31}
    \end{minipage}
\caption{\textbf{The attention score of our CRFT in layer 31.} (part 2 of 2)}
\label{Fig: vislayerdde31_2}
\end{figure*}

\begin{figure*}[tbp]
\centering
    \begin{minipage}{0.23\textwidth}
        \includegraphics[width=\linewidth]{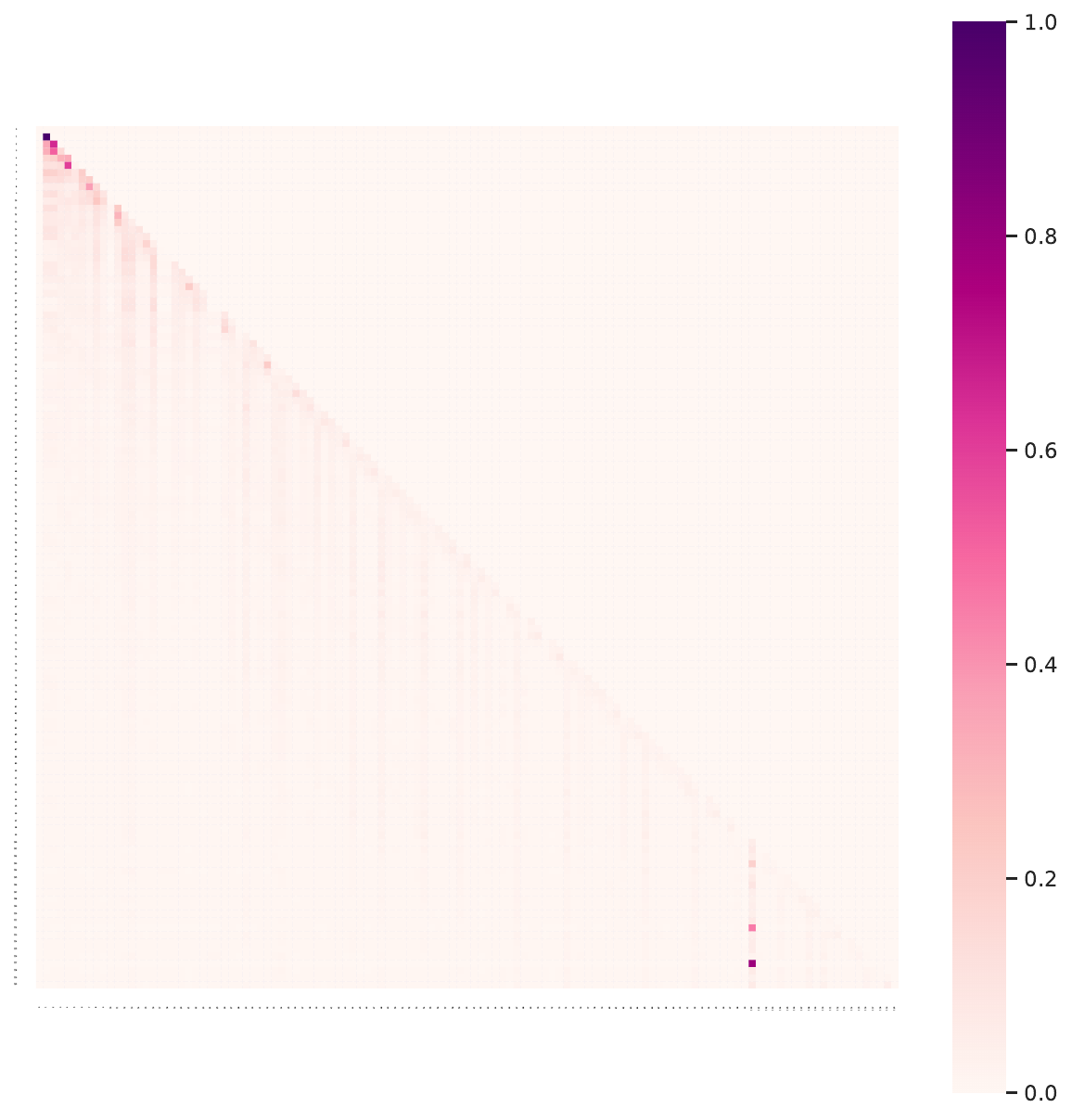}
        \caption*{Layer 0}
    \end{minipage}
    \begin{minipage}{0.23\textwidth}
        \includegraphics[width=\linewidth]{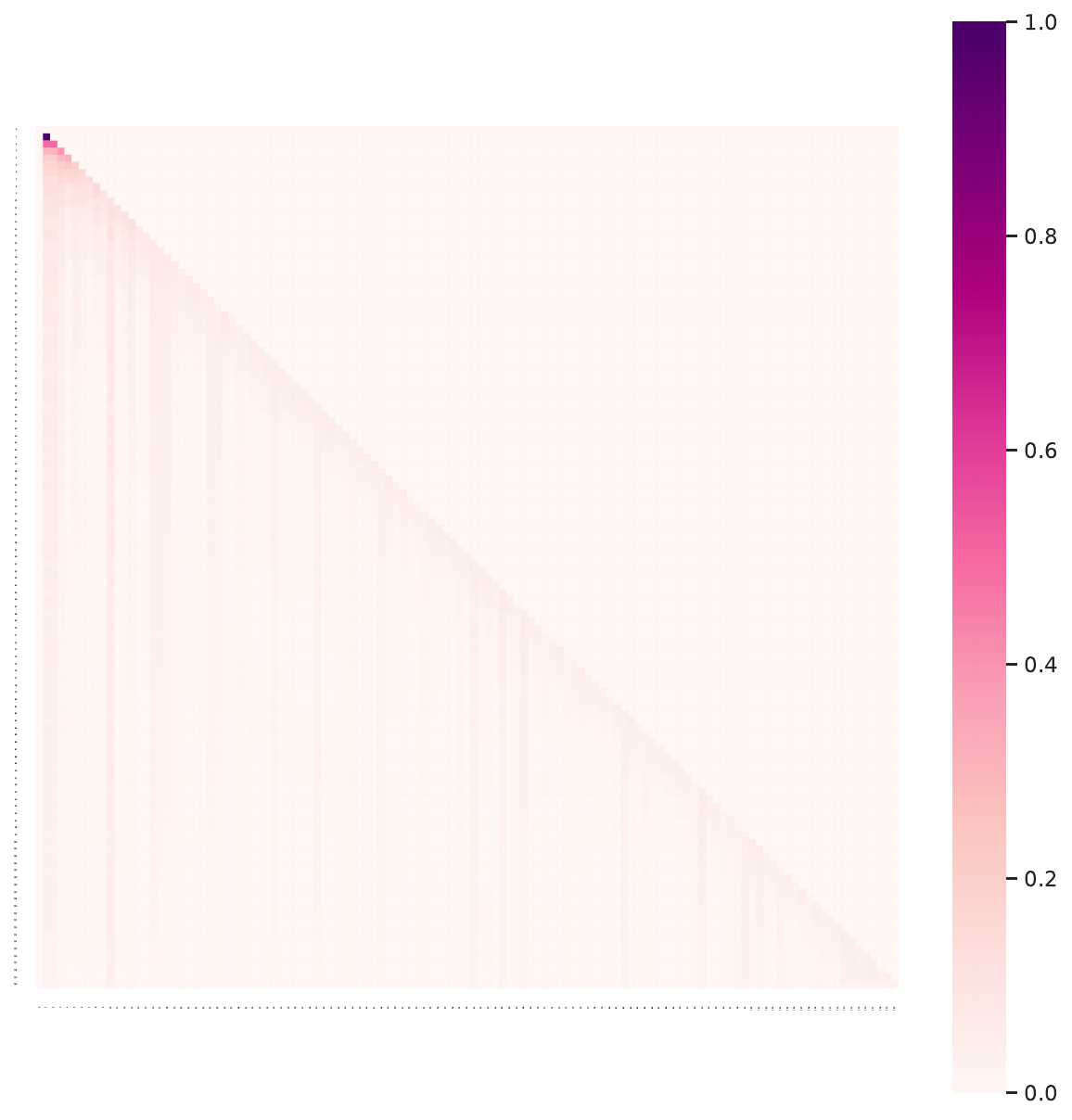}
        \caption*{Layer 1}
    \end{minipage}
    \begin{minipage}{0.23\textwidth}
        \includegraphics[width=\linewidth]{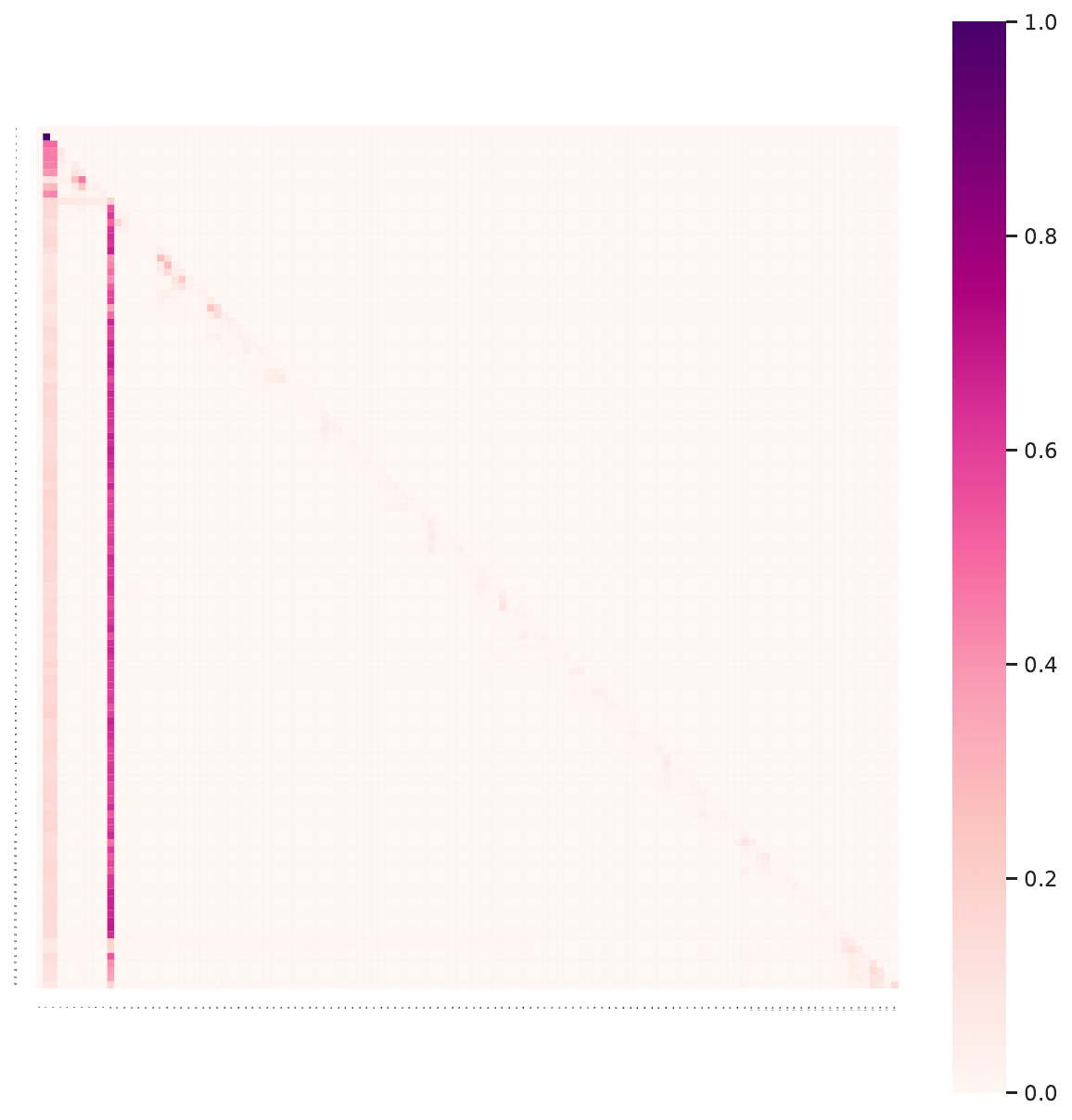}
        \caption*{Layer 2}
    \end{minipage}
    \begin{minipage}{0.23\textwidth}
        \includegraphics[width=\linewidth]{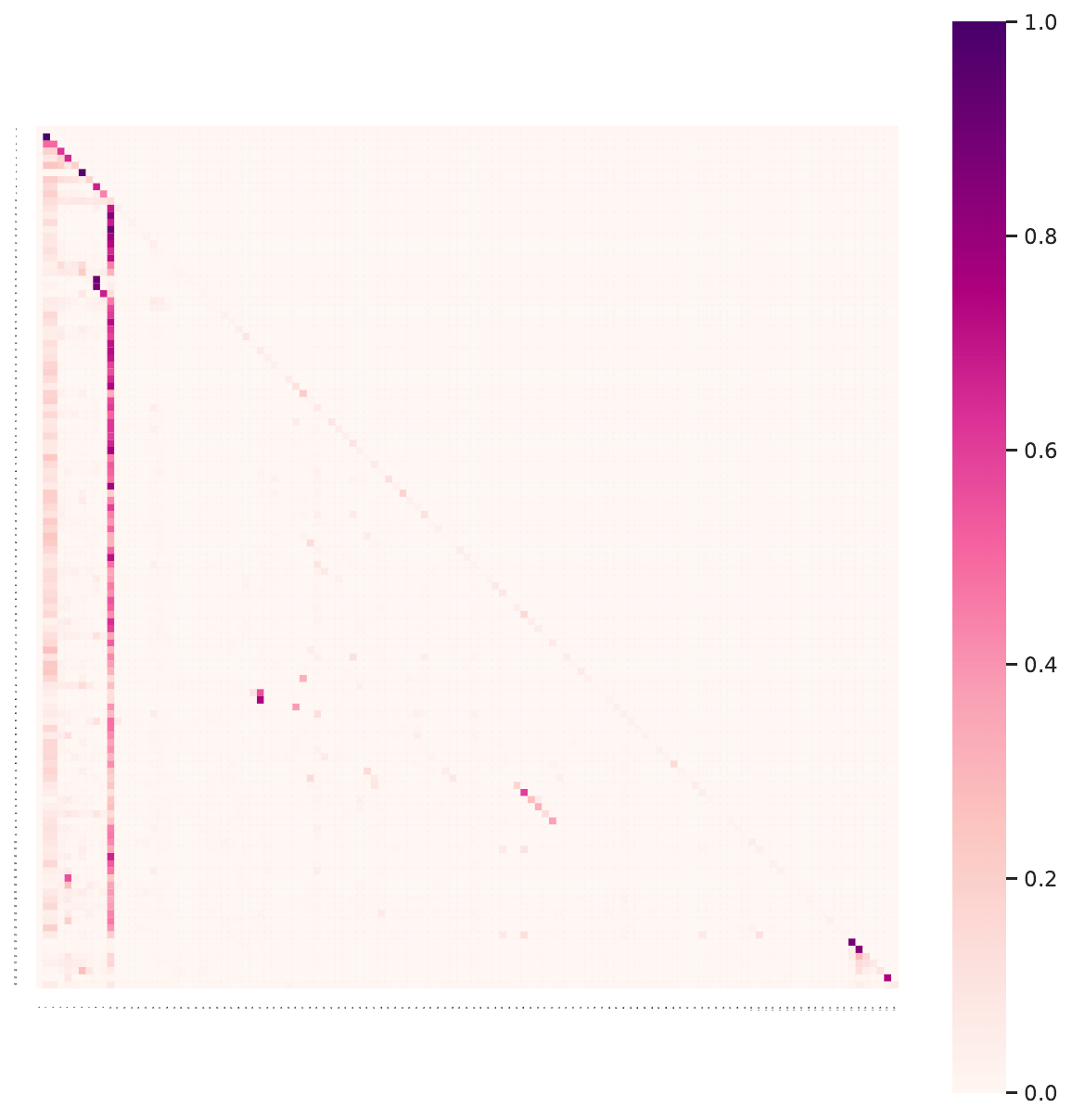}
        \caption*{Layer 3}
    \end{minipage}
\vspace{1em}
\begin{minipage}{0.23\textwidth}
        \includegraphics[width=\linewidth]{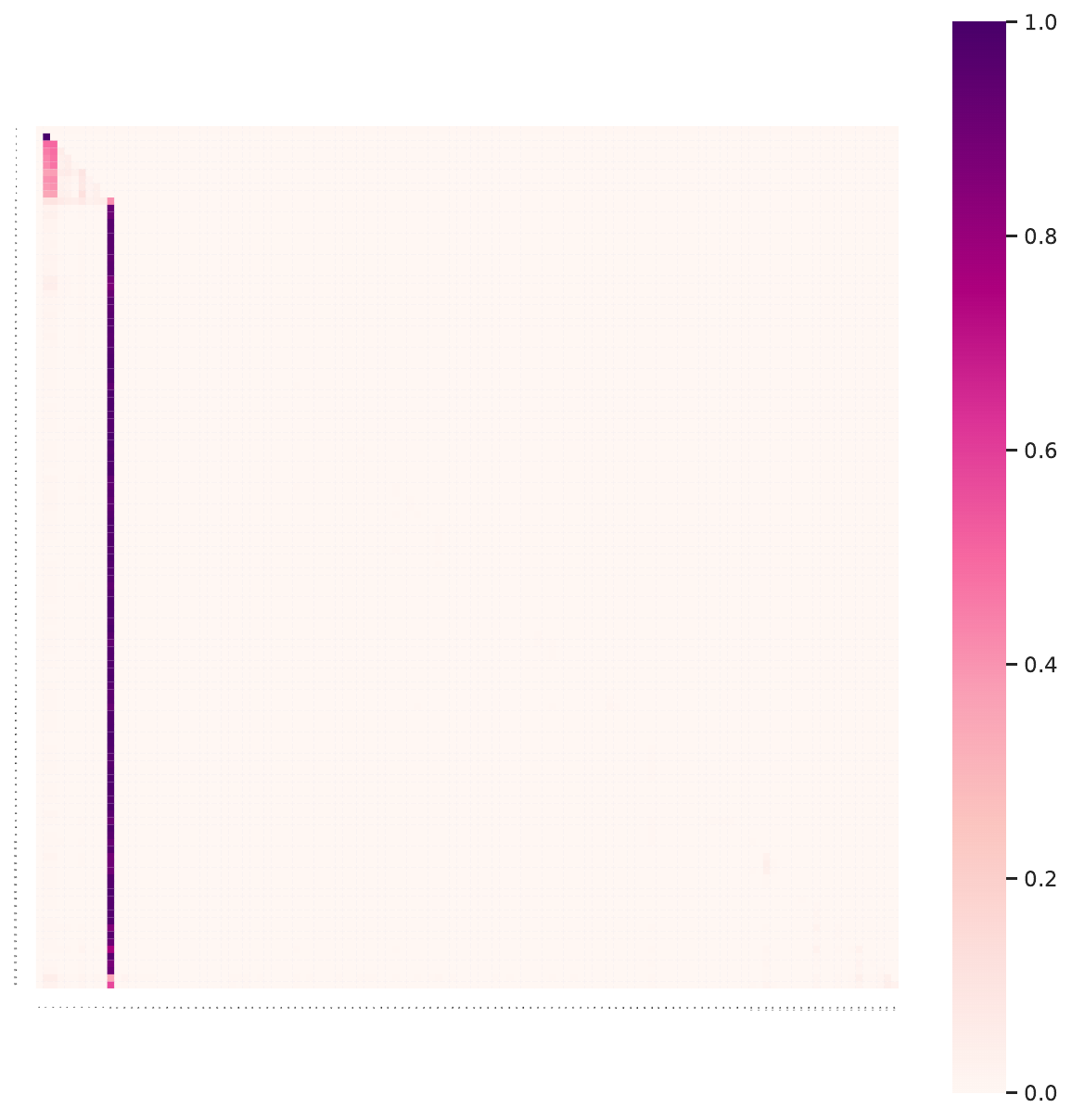}
        \caption*{Layer 4}
    \end{minipage}
    \begin{minipage}{0.23\textwidth}
        \includegraphics[width=\linewidth]{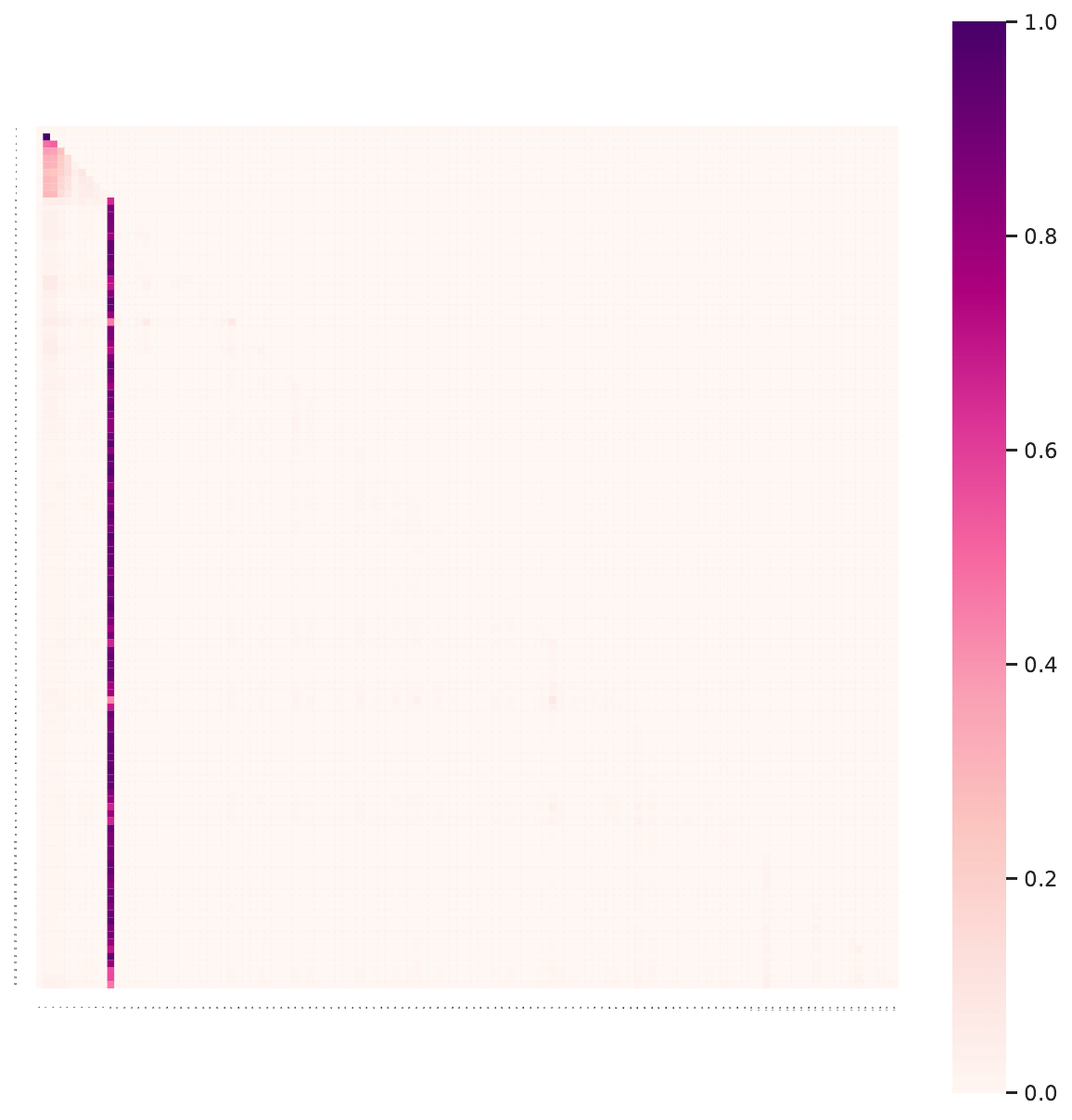}
        \caption*{Layer 5}
    \end{minipage}
    \begin{minipage}{0.23\textwidth}
        \includegraphics[width=\linewidth]{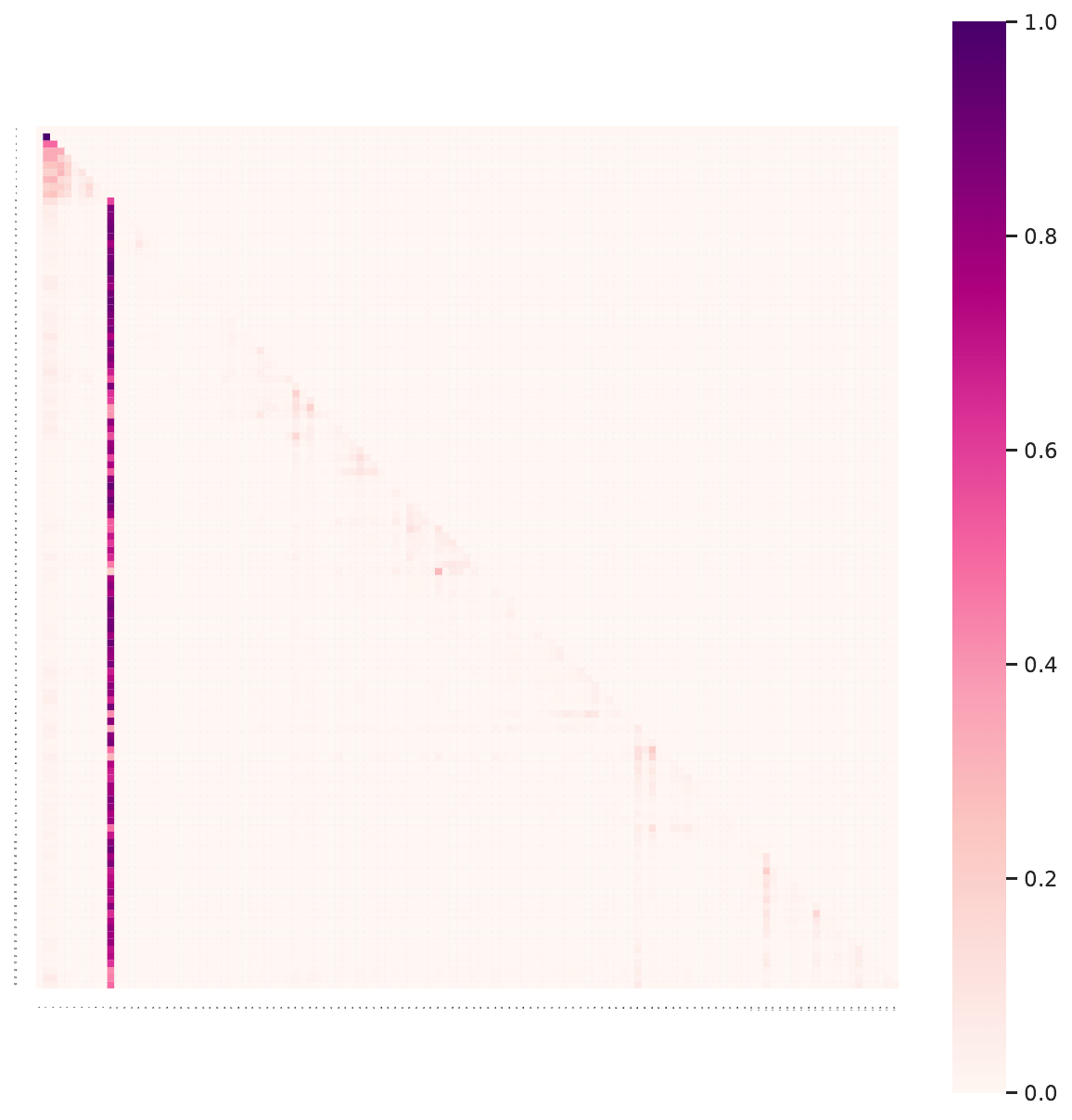}
        \caption*{Layer 6}
    \end{minipage}
    \begin{minipage}{0.23\textwidth}
        \includegraphics[width=\linewidth]{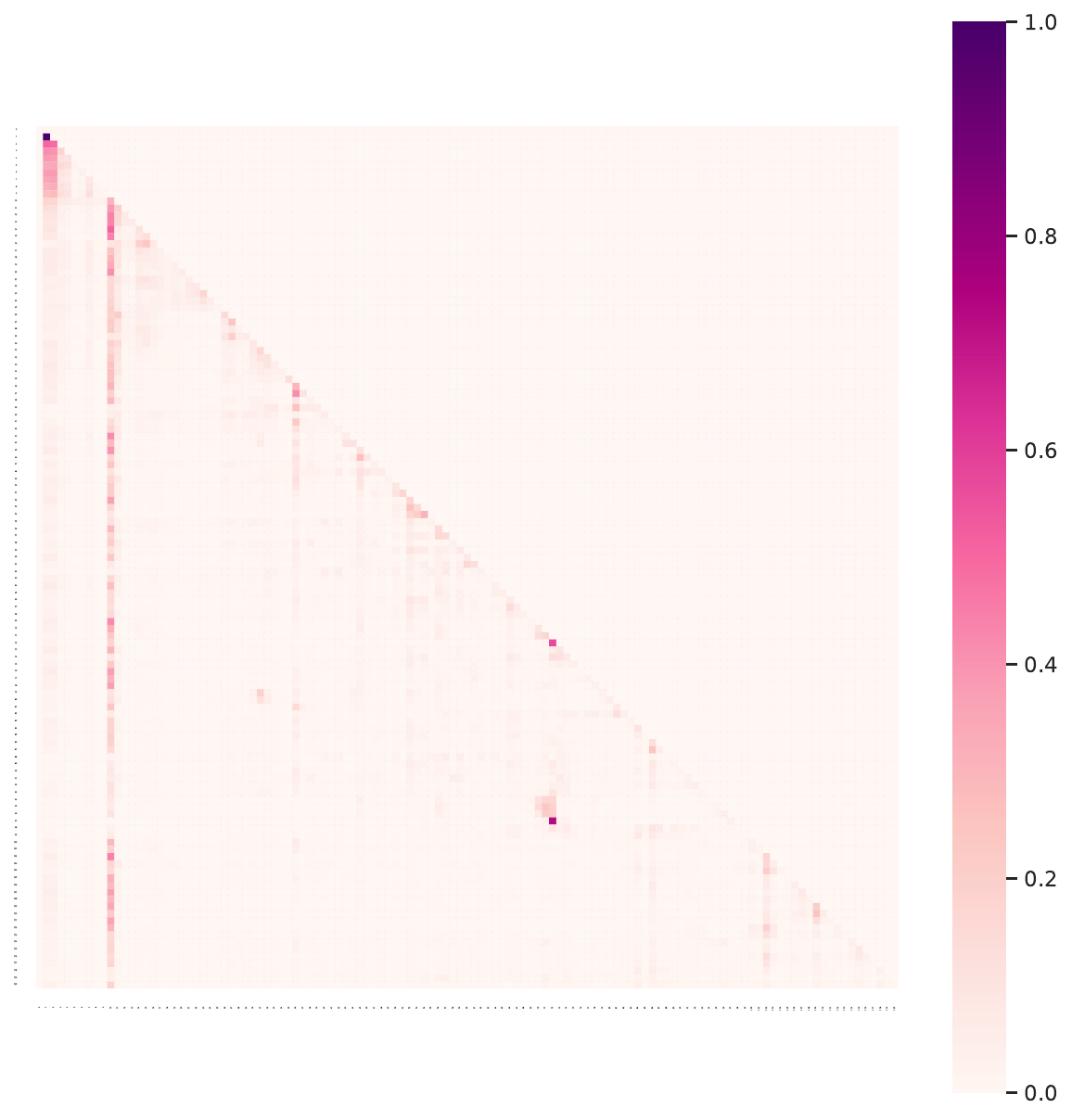}
        \caption*{Layer 7}
    \end{minipage}
\vspace{1em}
\begin{minipage}{0.23\textwidth}
        \includegraphics[width=\linewidth]{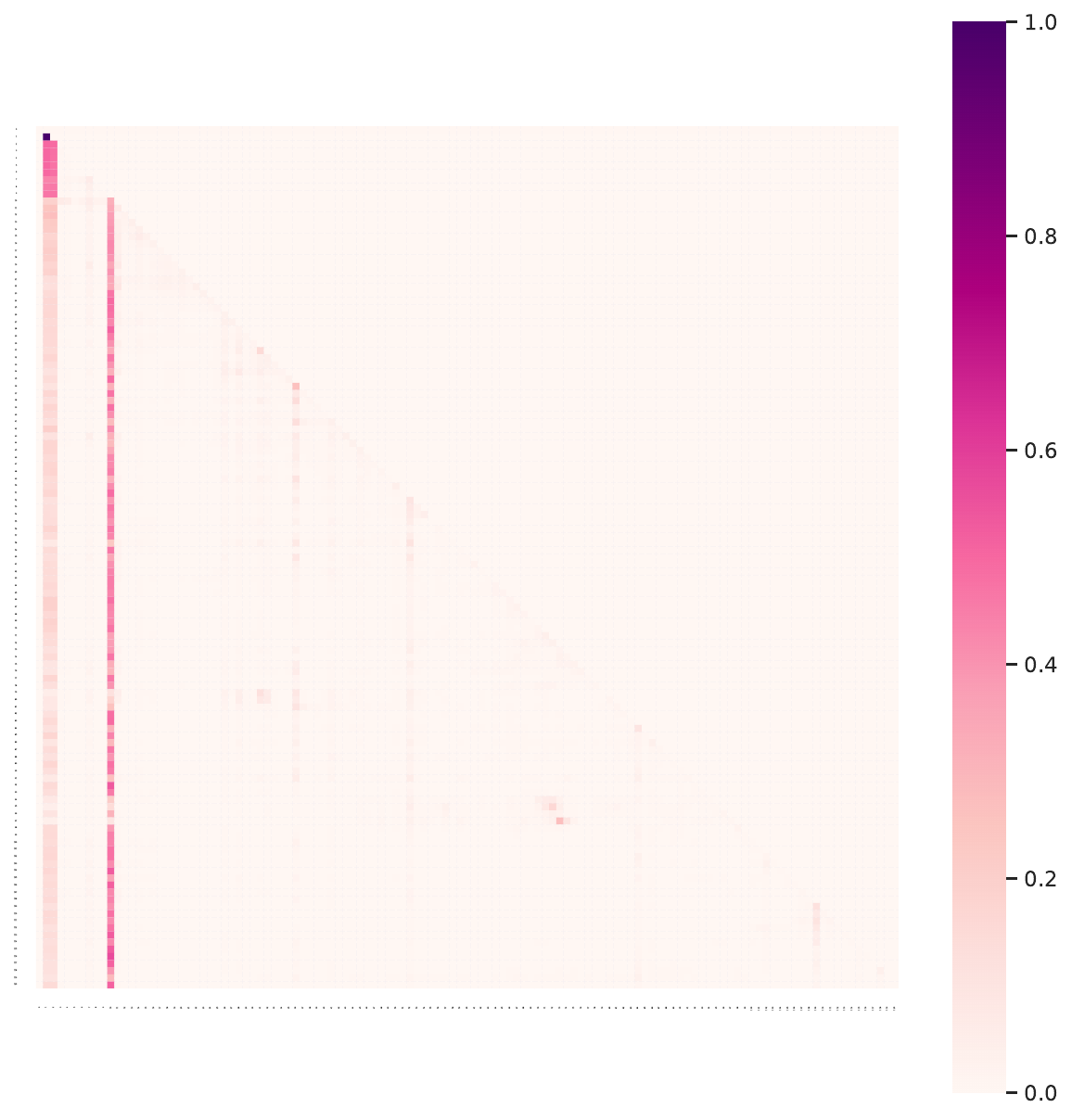}
        \caption*{Layer 8}
    \end{minipage}
    \begin{minipage}{0.23\textwidth}
        \includegraphics[width=\linewidth]{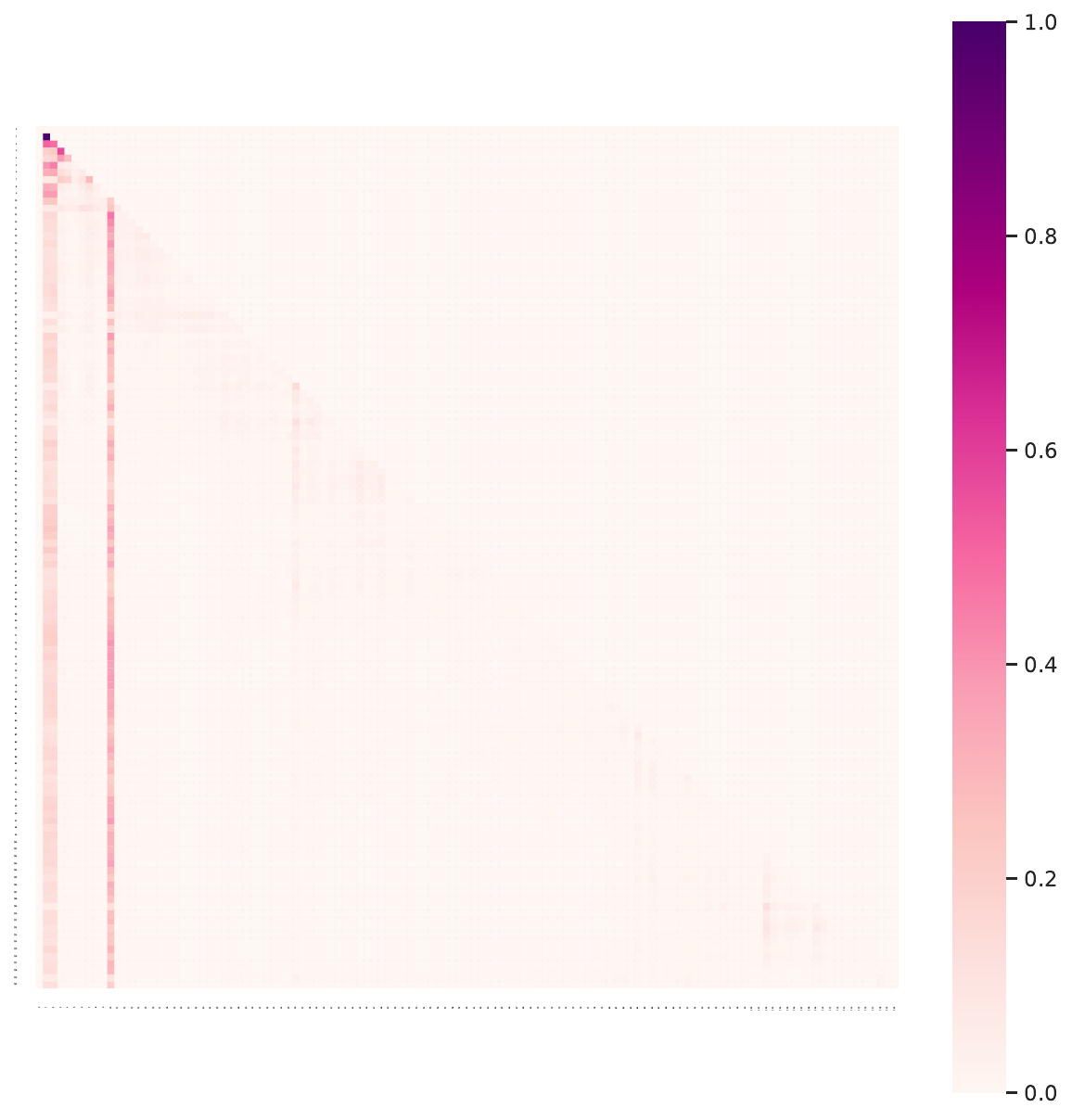}
        \caption*{Layer 9}
    \end{minipage}
    \begin{minipage}{0.23\textwidth}
        \includegraphics[width=\linewidth]{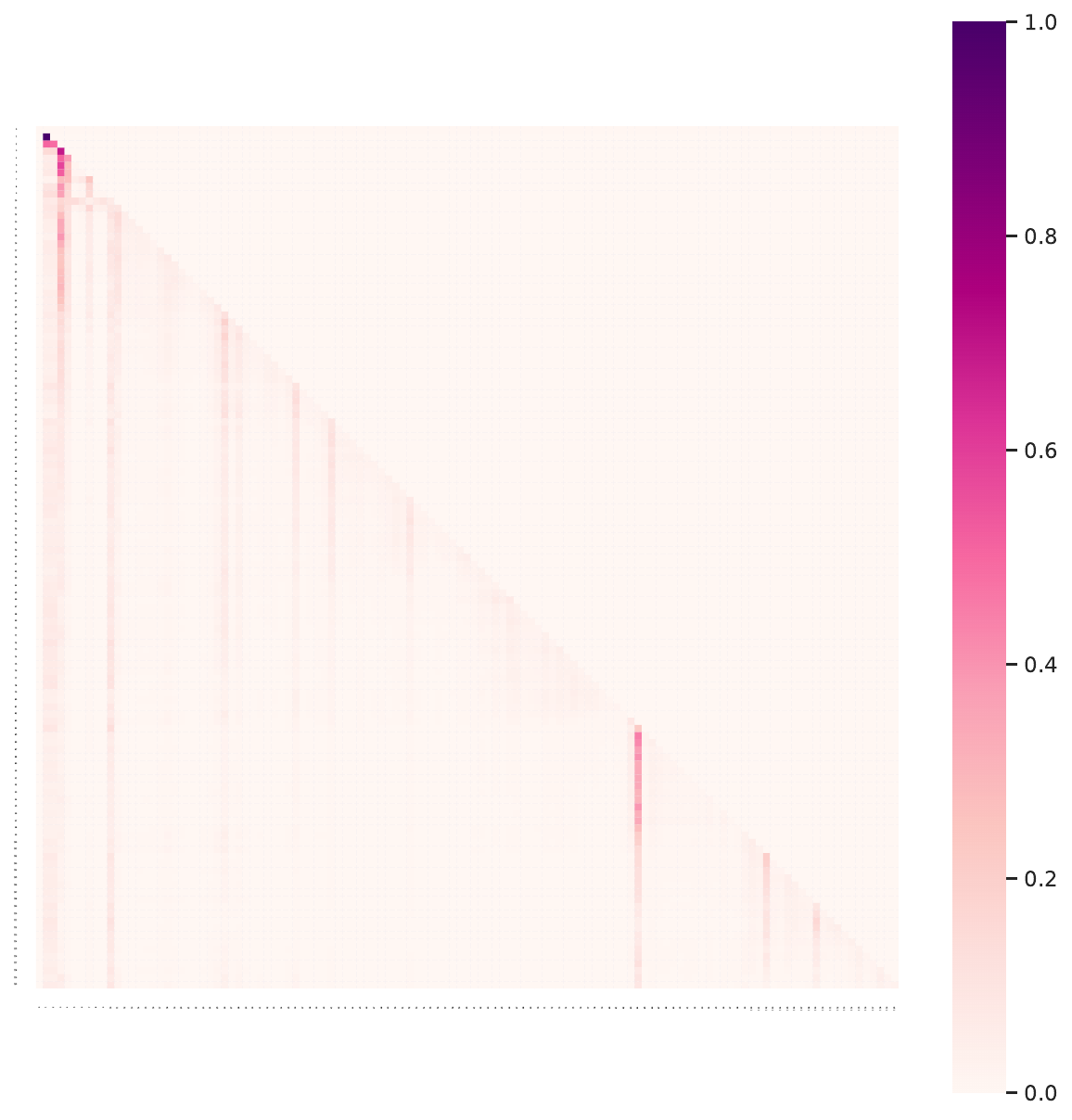}
        \caption*{Layer 10}
    \end{minipage}
    \begin{minipage}{0.23\textwidth}
        \includegraphics[width=\linewidth]{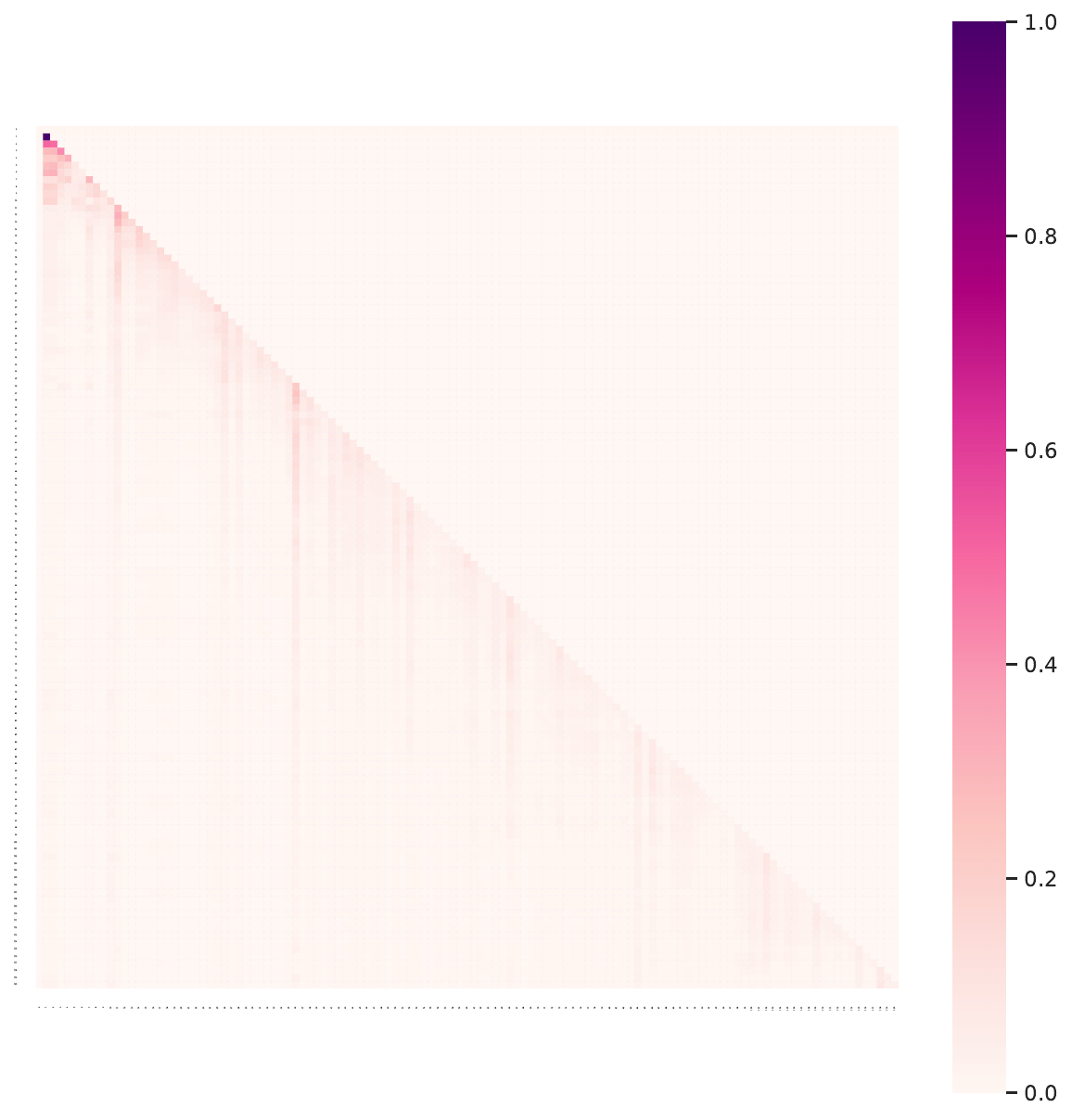}
        \caption*{Layer 11}
    \end{minipage}
    \vspace{1em}
\begin{minipage}{0.23\textwidth}
        \includegraphics[width=\linewidth]{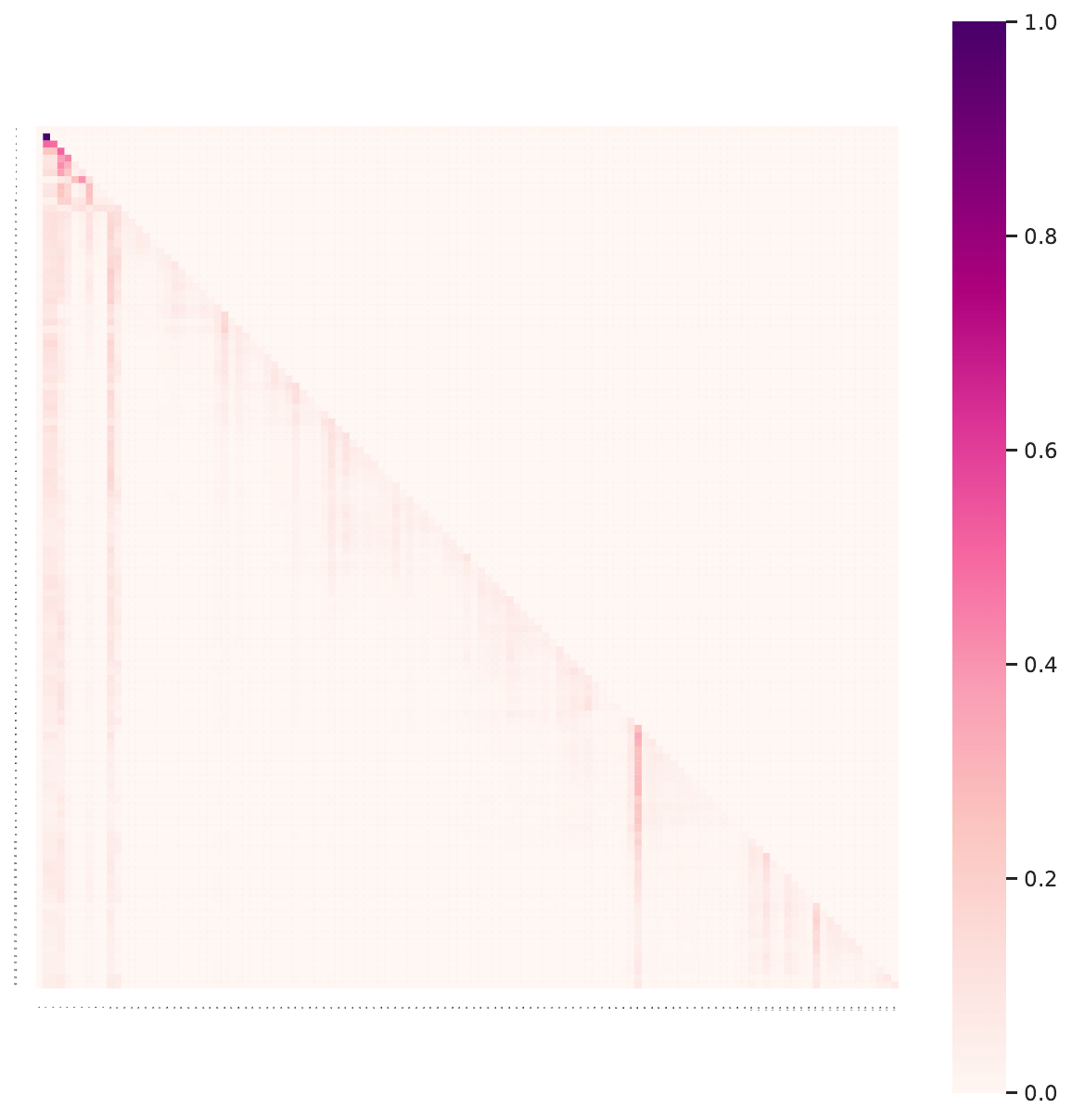}
        \caption*{Layer 12}
    \end{minipage}
    \begin{minipage}{0.23\textwidth}
        \includegraphics[width=\linewidth]{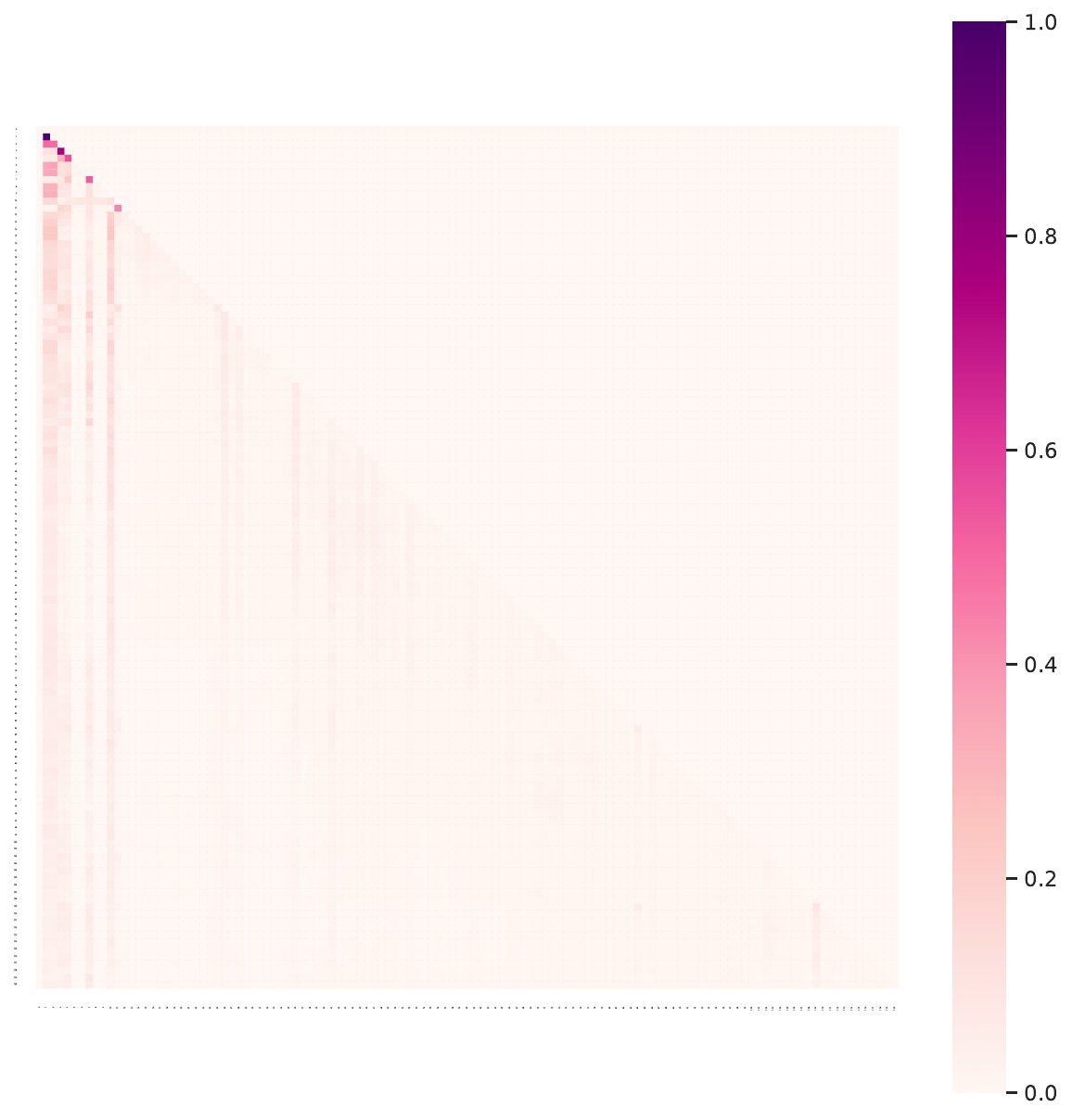}
        \caption*{Layer 13}
    \end{minipage}
    \begin{minipage}{0.23\textwidth}
        \includegraphics[width=\linewidth]{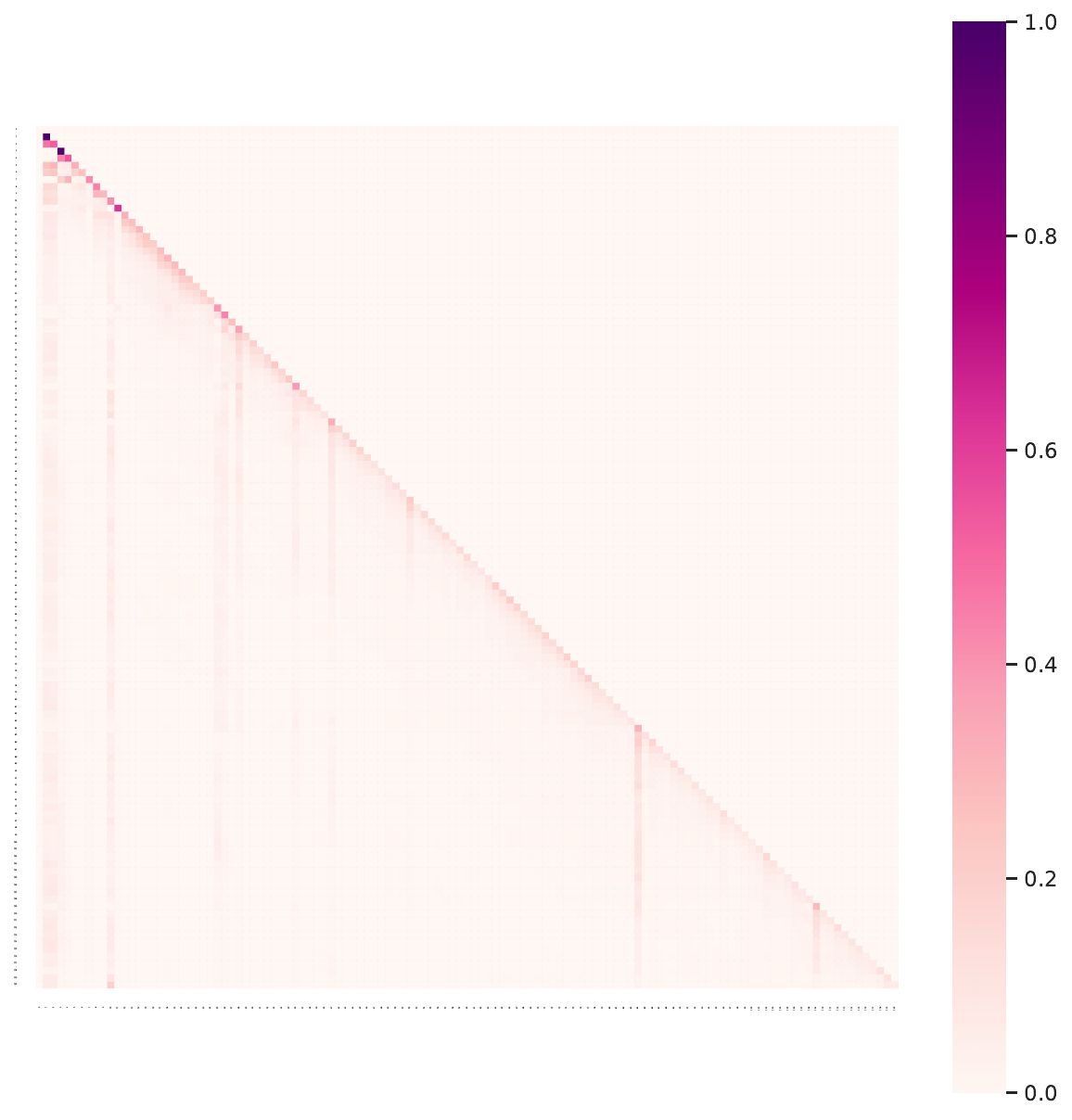}
        \caption*{Layer 14}
    \end{minipage}
    \begin{minipage}{0.23\textwidth}
        \includegraphics[width=\linewidth]{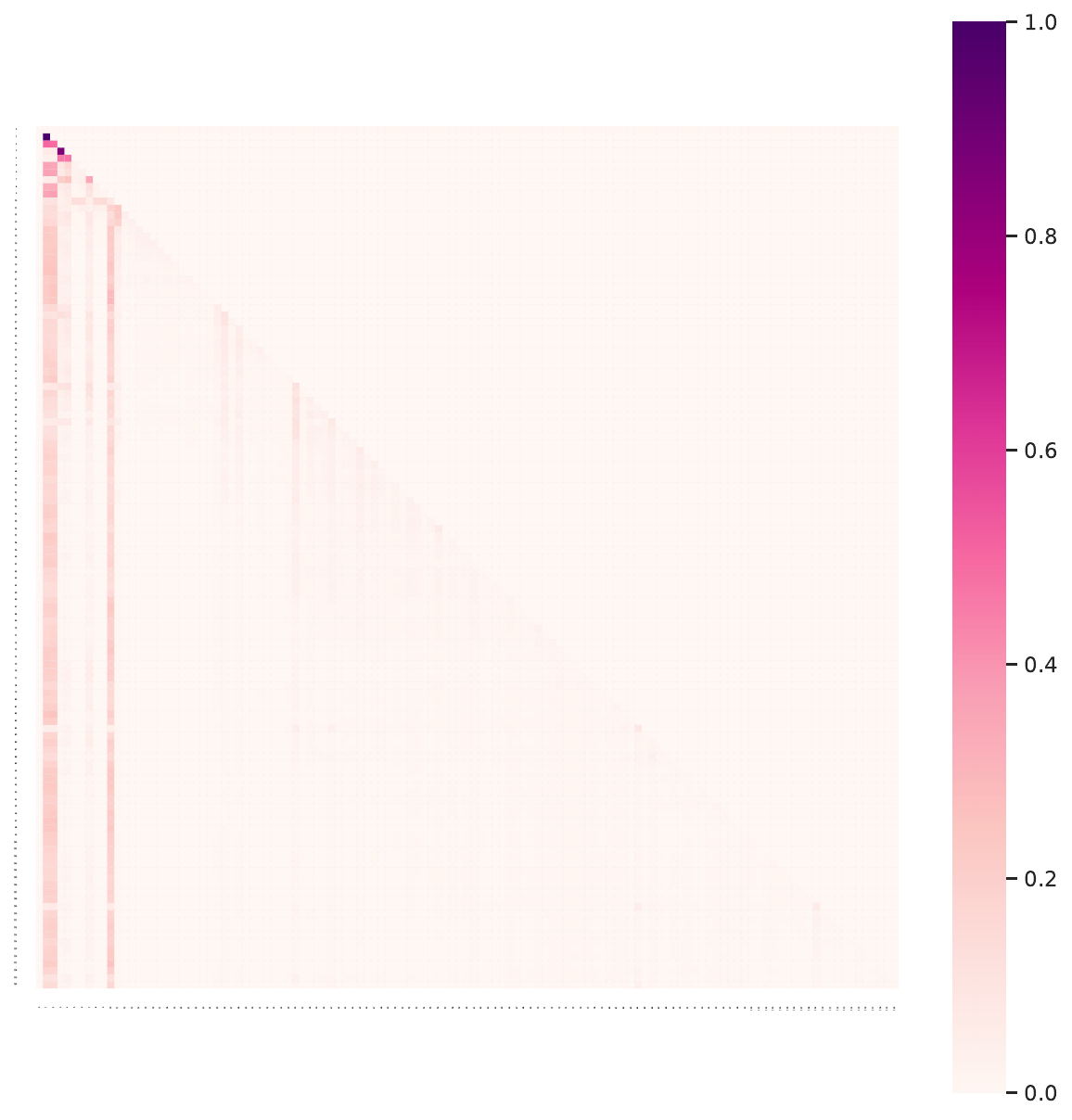}
        \caption*{Layer 15}
    \end{minipage}
\caption{\textbf{The attention score of our CRFT on head 31 in all layers.} (part 1 of 2)}
\label{Fig: vislayerh31dde_1}
\end{figure*}

\begin{figure*}[tbp]
\centering
\begin{minipage}{0.23\textwidth}
        \includegraphics[width=\linewidth]{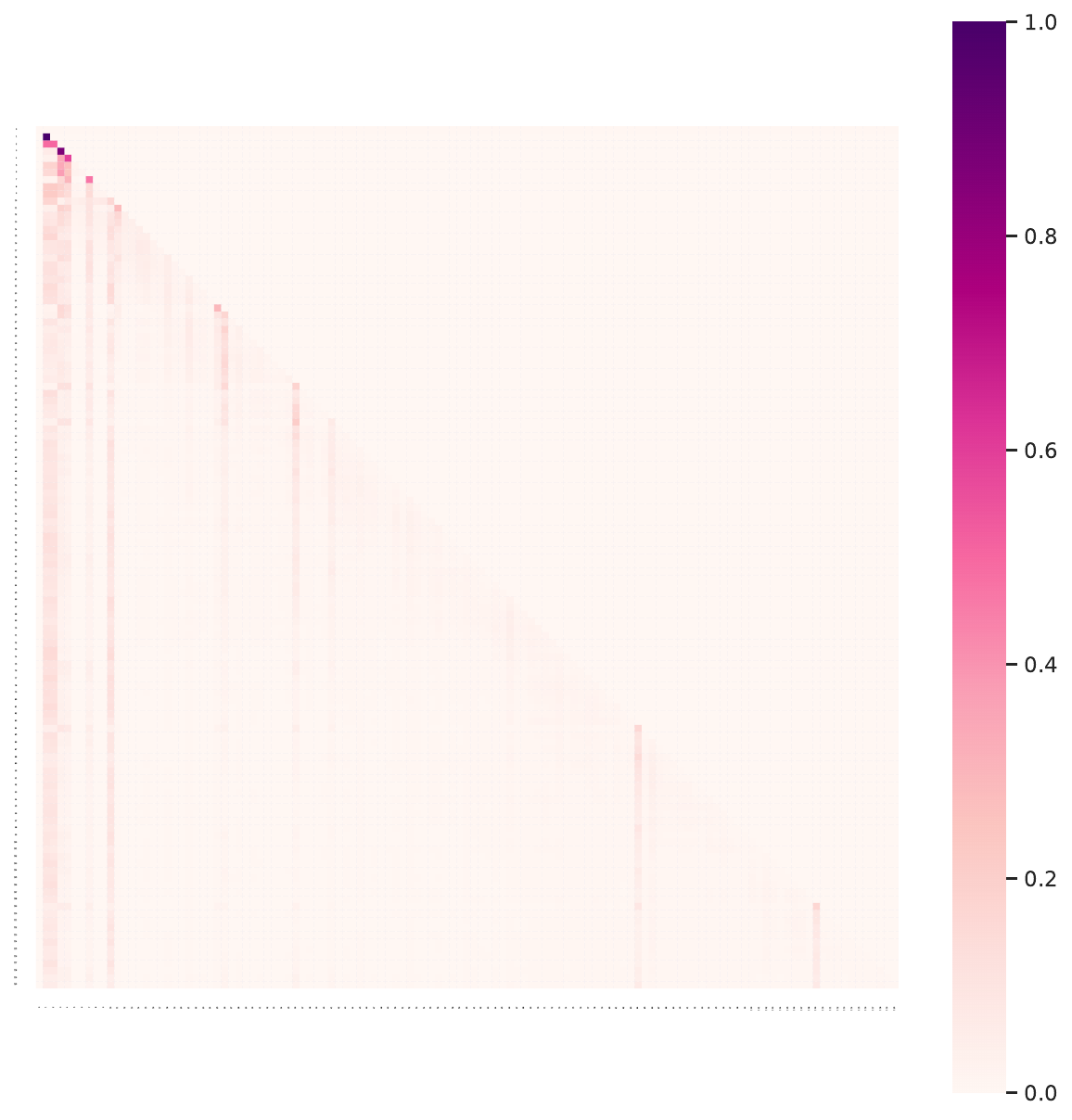}
        \caption*{Layer 16}
    \end{minipage}
    \begin{minipage}{0.23\textwidth}
        \includegraphics[width=\linewidth]{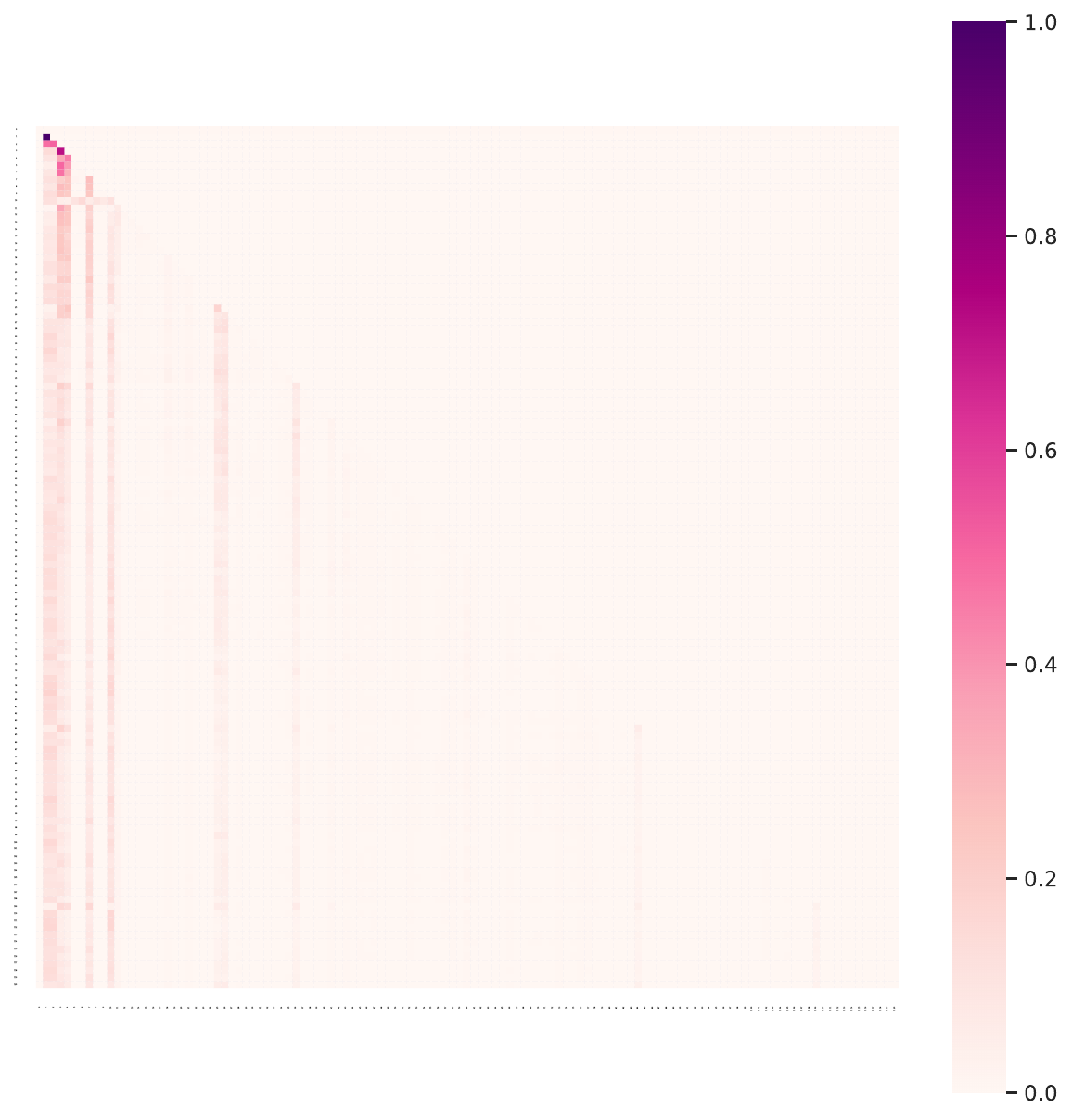}
        \caption*{Layer 17}
    \end{minipage}
    \begin{minipage}{0.23\textwidth}
        \includegraphics[width=\linewidth]{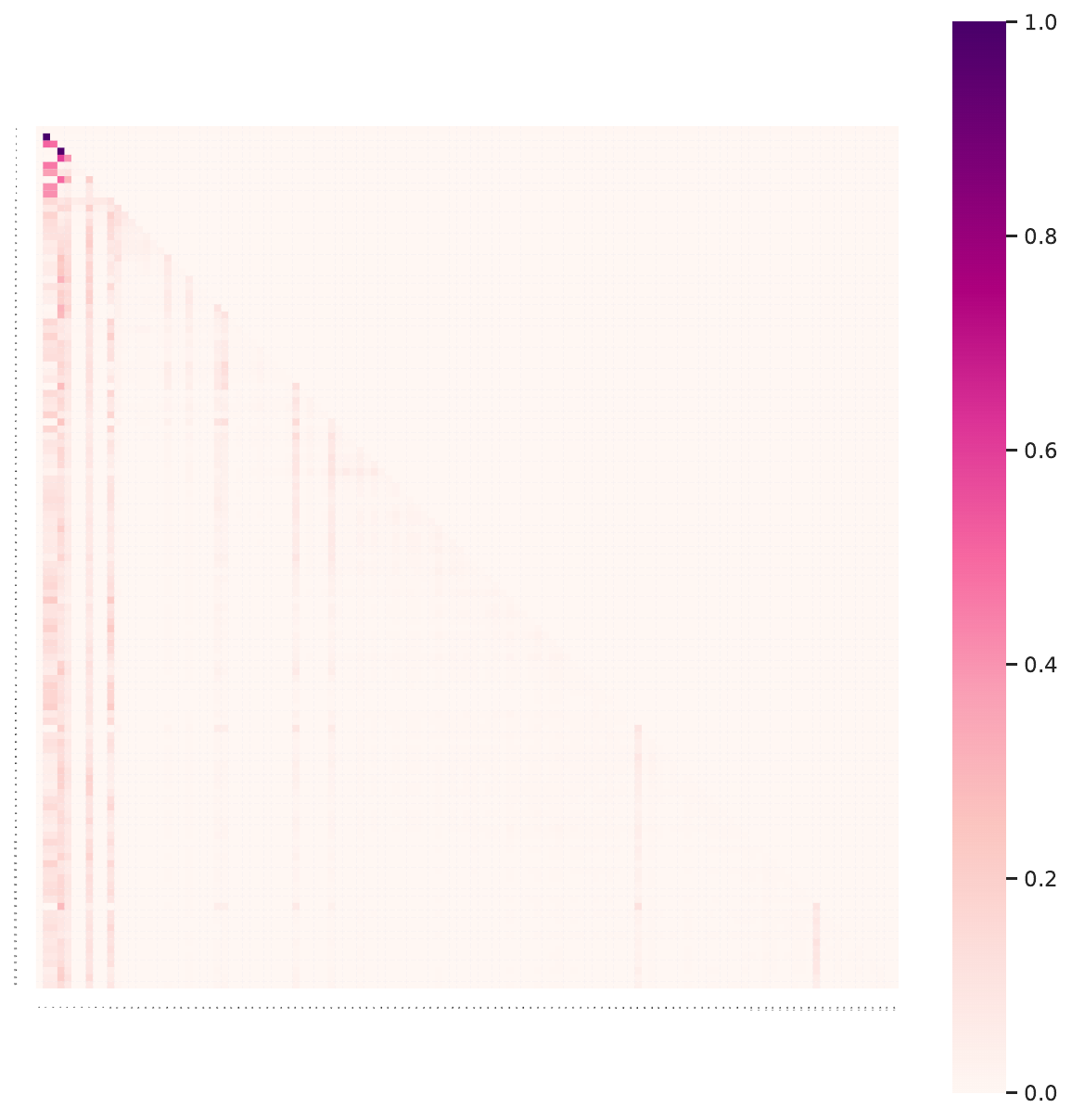}
        \caption*{Layer 18}
    \end{minipage}
    \begin{minipage}{0.23\textwidth}
        \includegraphics[width=\linewidth]{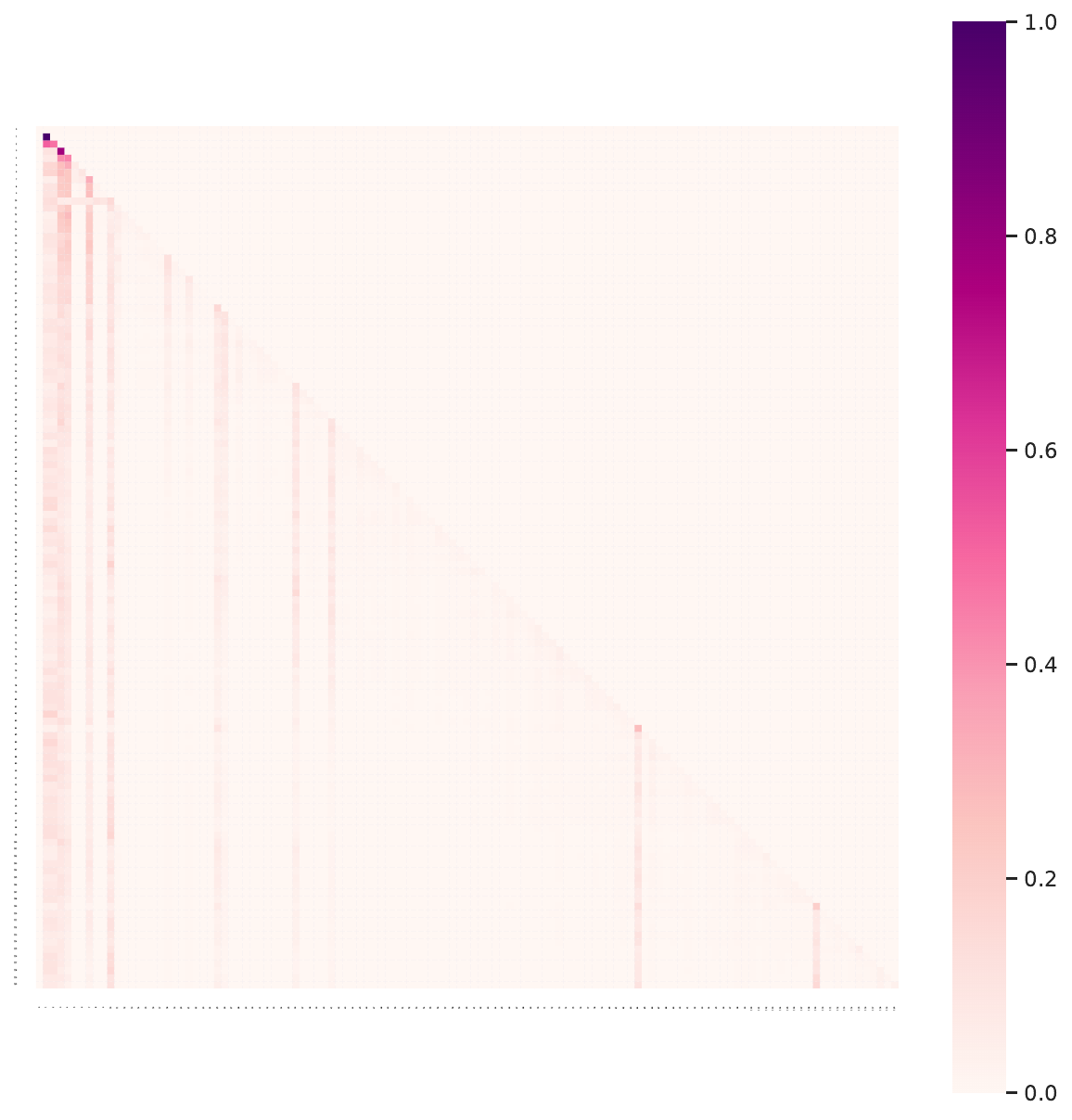}
        \caption*{Layer 19}
    \end{minipage}
\vspace{1em}
\begin{minipage}{0.23\textwidth}
        \includegraphics[width=\linewidth]{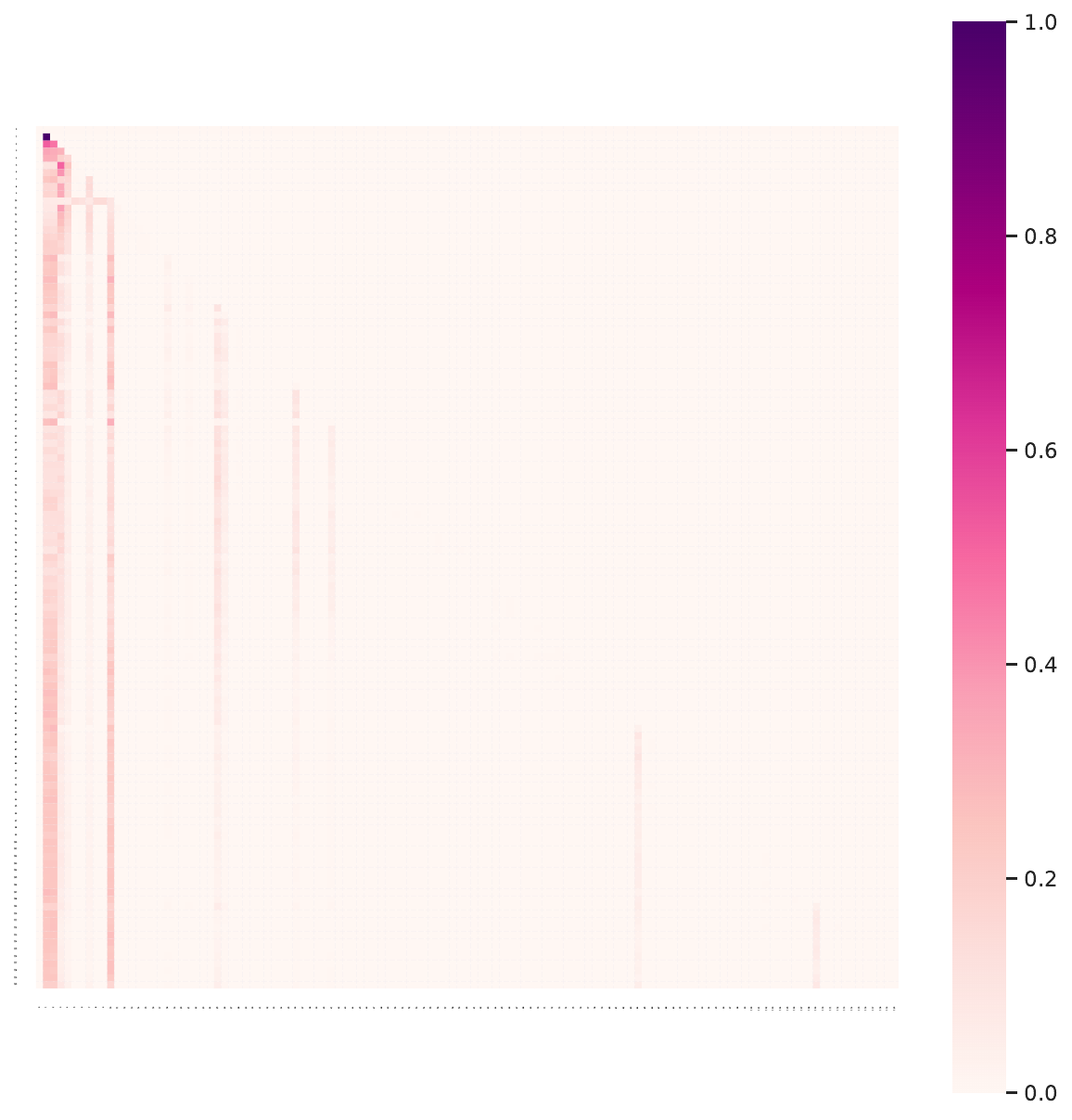}
        \caption*{Layer 20}
    \end{minipage}
    \begin{minipage}{0.23\textwidth}
        \includegraphics[width=\linewidth]{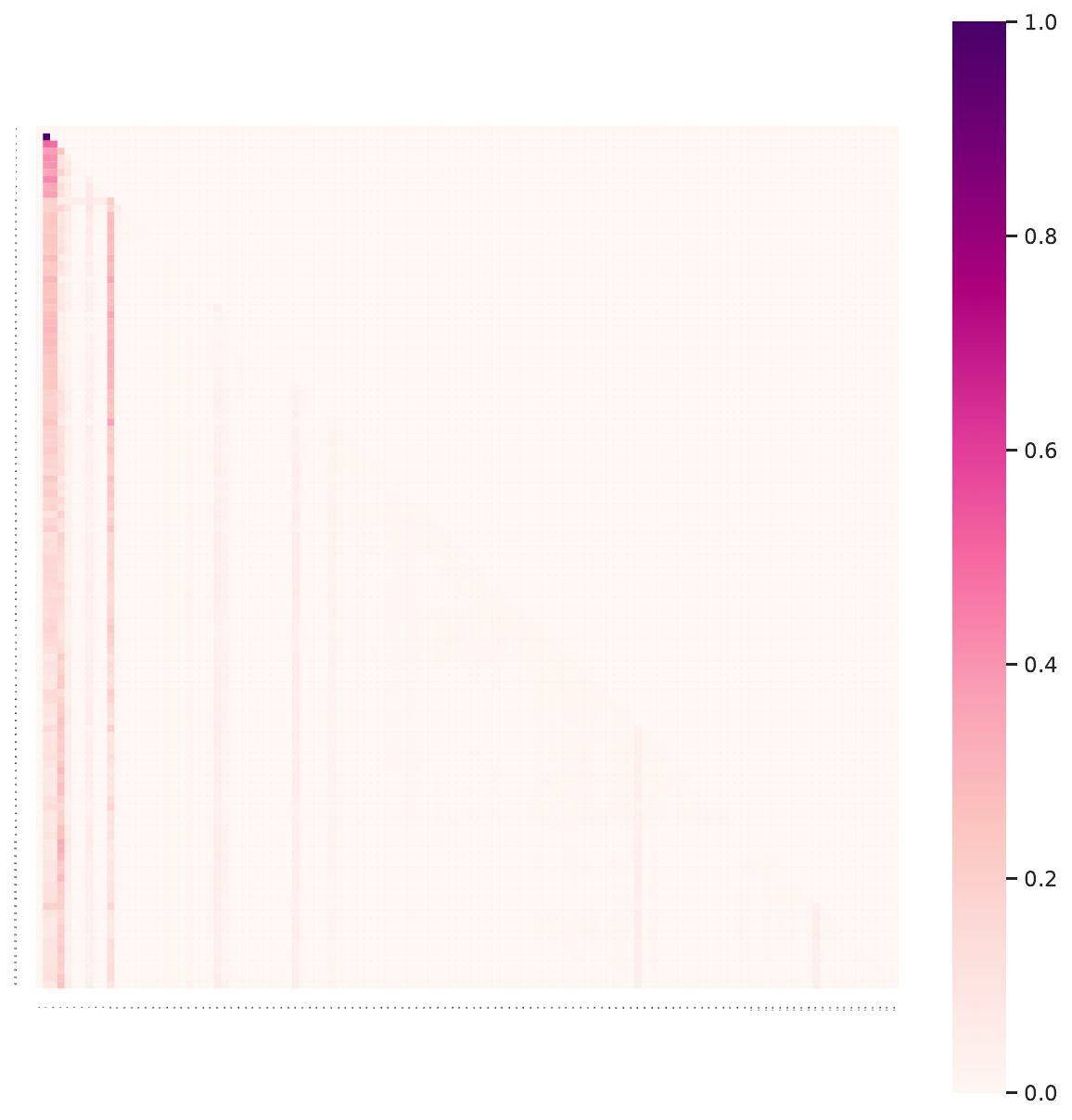}
        \caption*{Layer 21}
    \end{minipage}
    \begin{minipage}{0.23\textwidth}
        \includegraphics[width=\linewidth]{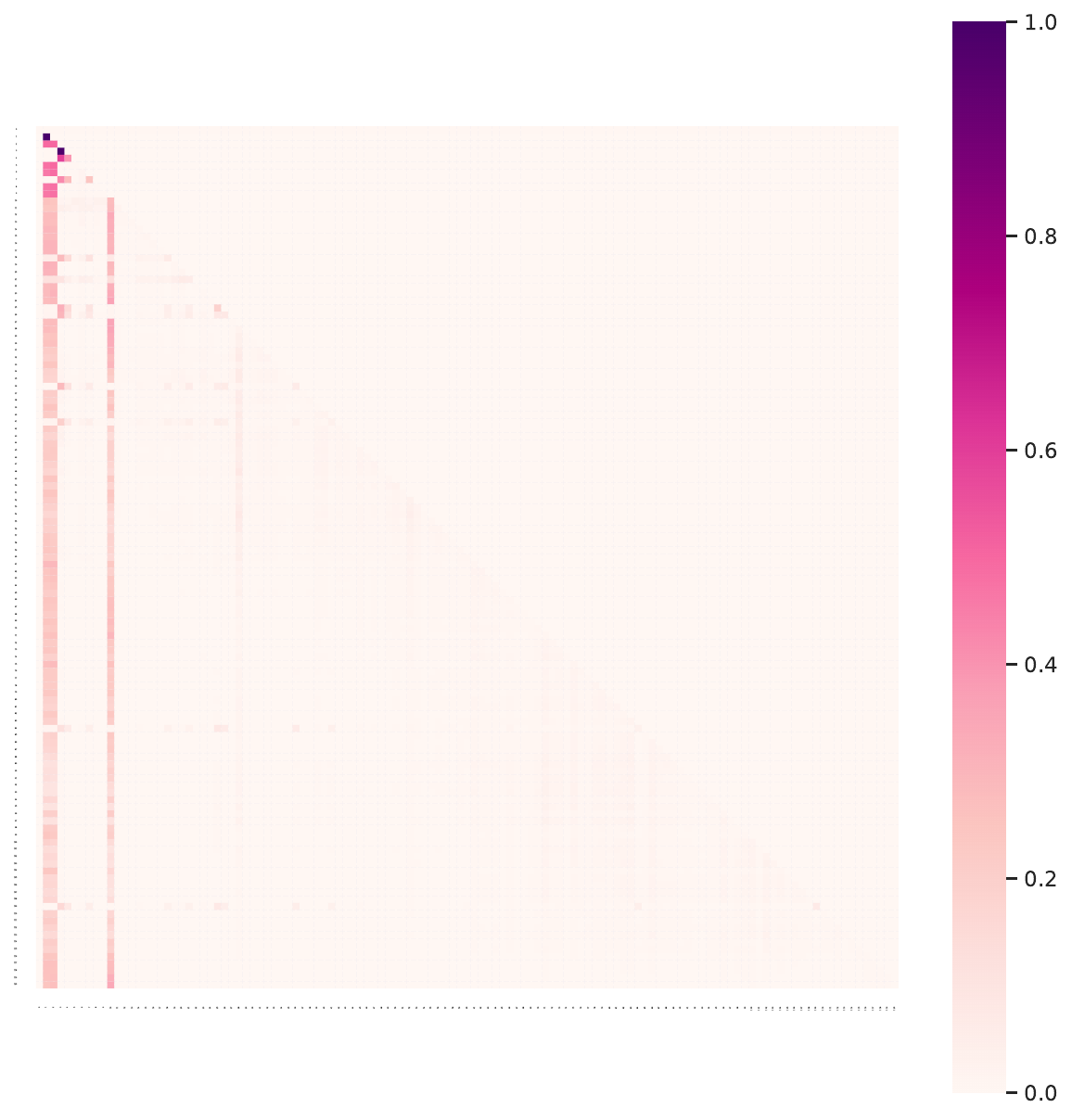}
        \caption*{Layer 22}
    \end{minipage}
    \begin{minipage}{0.23\textwidth}
        \includegraphics[width=\linewidth]{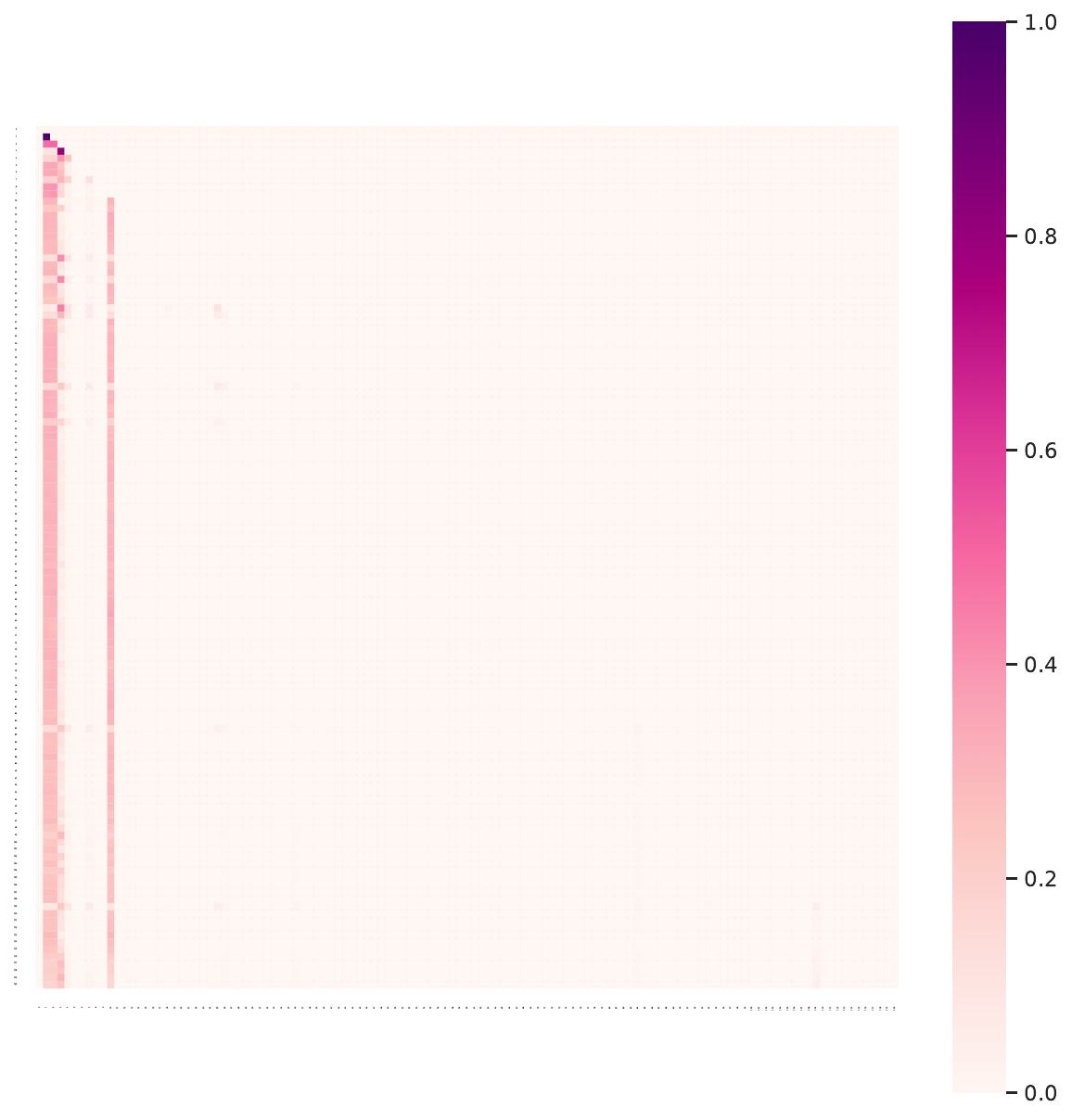}
        \caption*{Layer 23}
    \end{minipage}
    \vspace{1em}
\begin{minipage}{0.23\textwidth}
        \includegraphics[width=\linewidth]{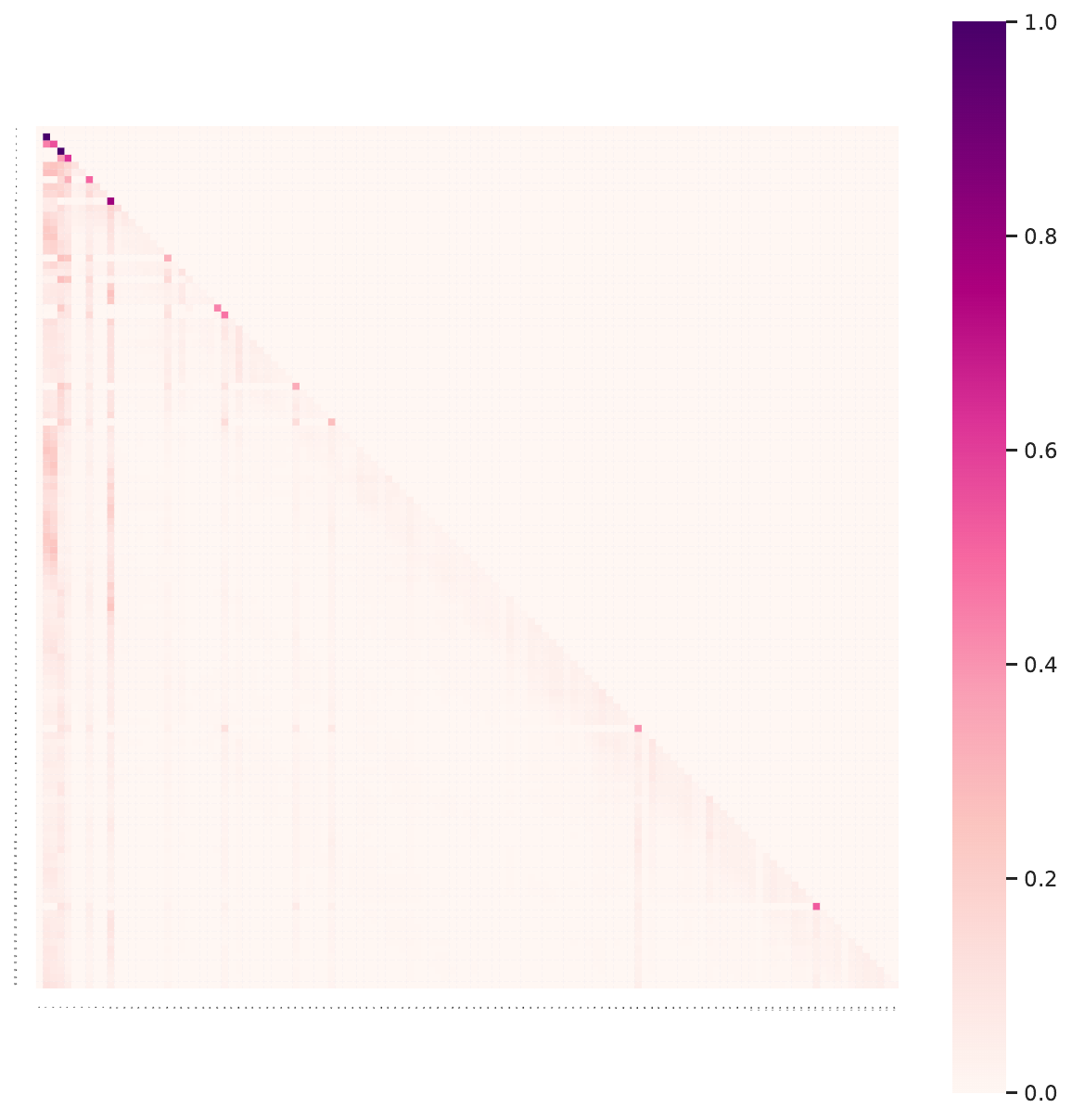}
        \caption*{Layer 24}
    \end{minipage}
    \begin{minipage}{0.23\textwidth}
        \includegraphics[width=\linewidth]{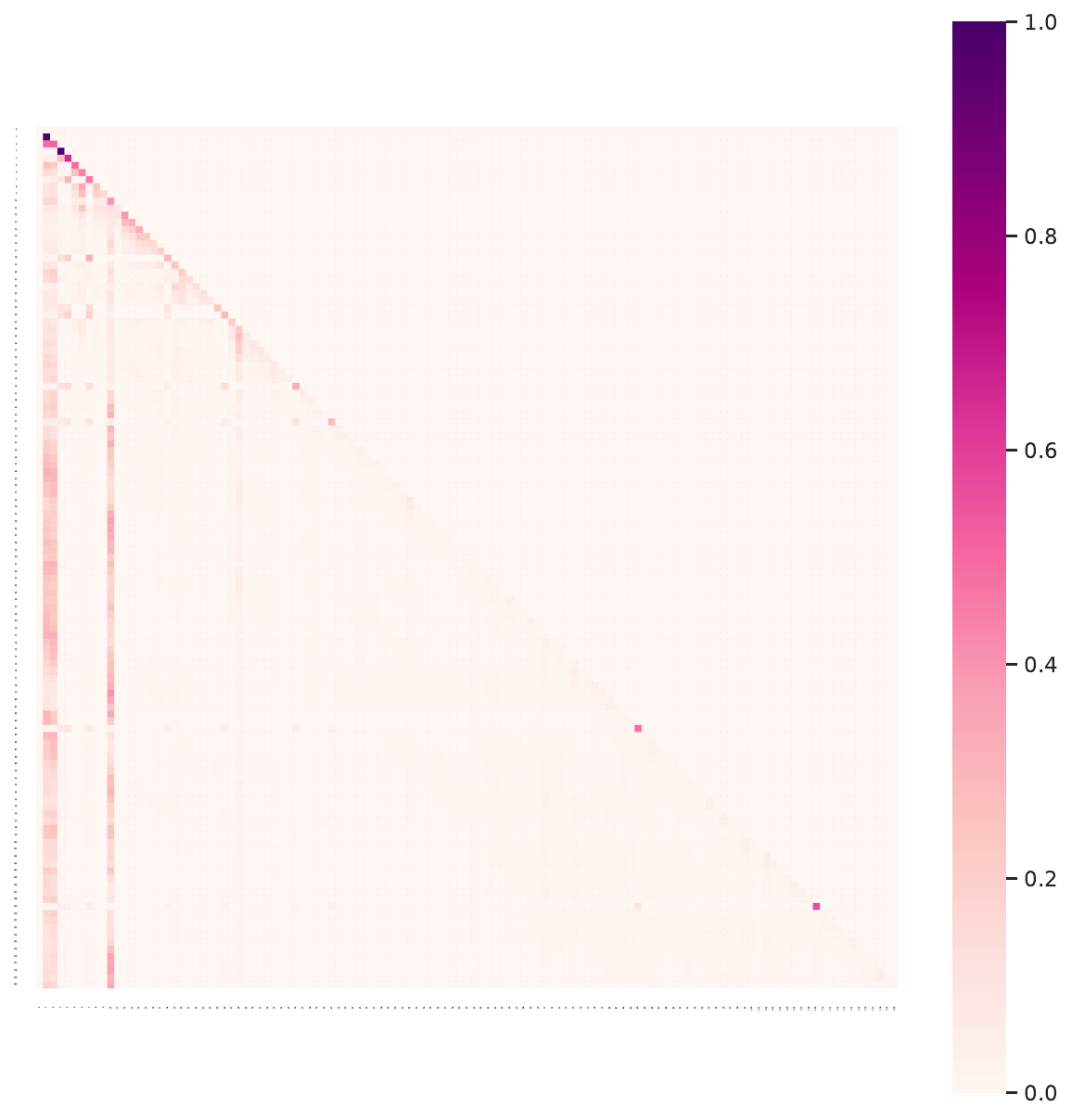}
        \caption*{Layer 25}
    \end{minipage}
    \begin{minipage}{0.23\textwidth}
        \includegraphics[width=\linewidth]{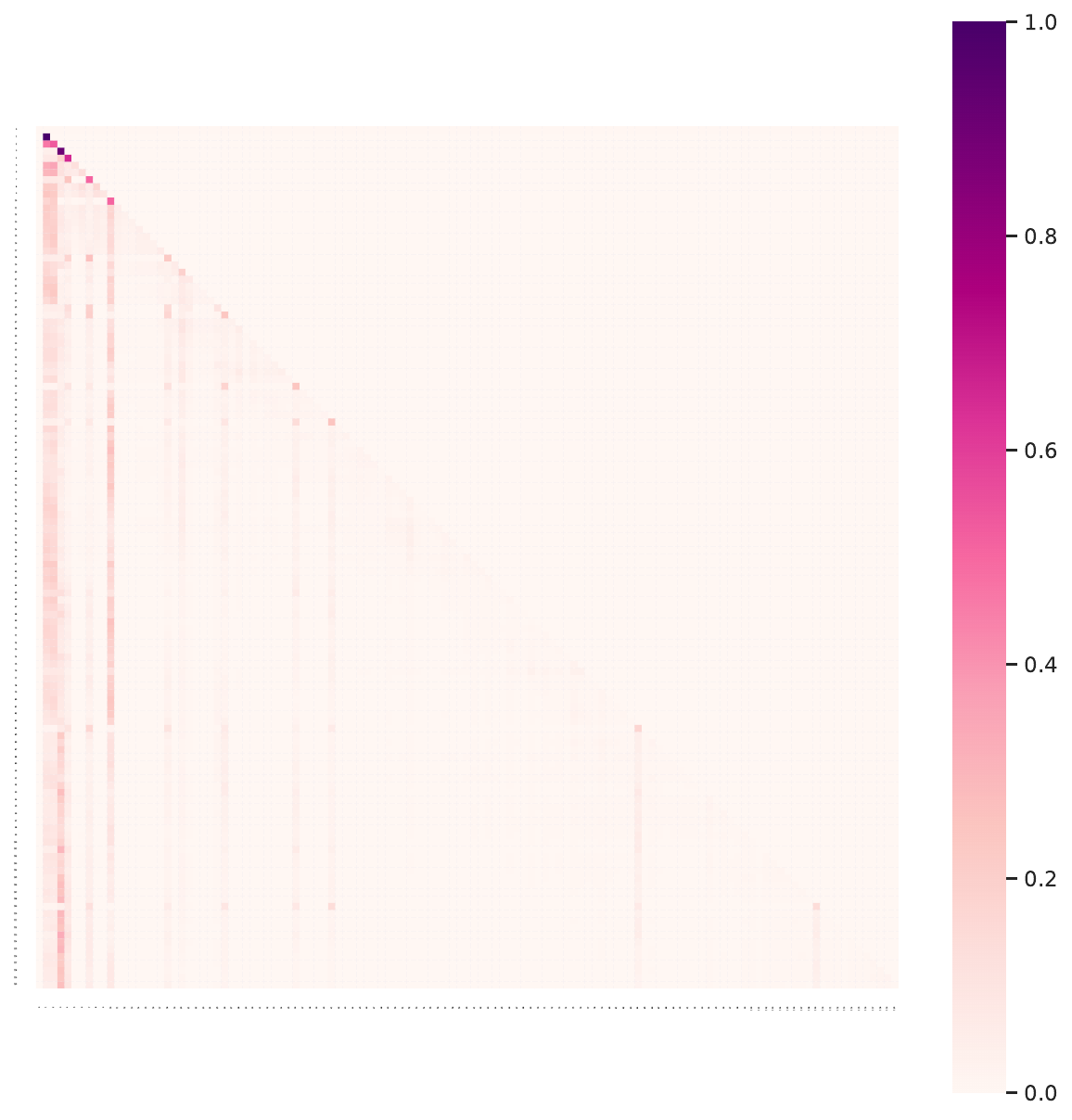}
        \caption*{Layer 26}
    \end{minipage}
    \begin{minipage}{0.23\textwidth}
        \includegraphics[width=\linewidth]{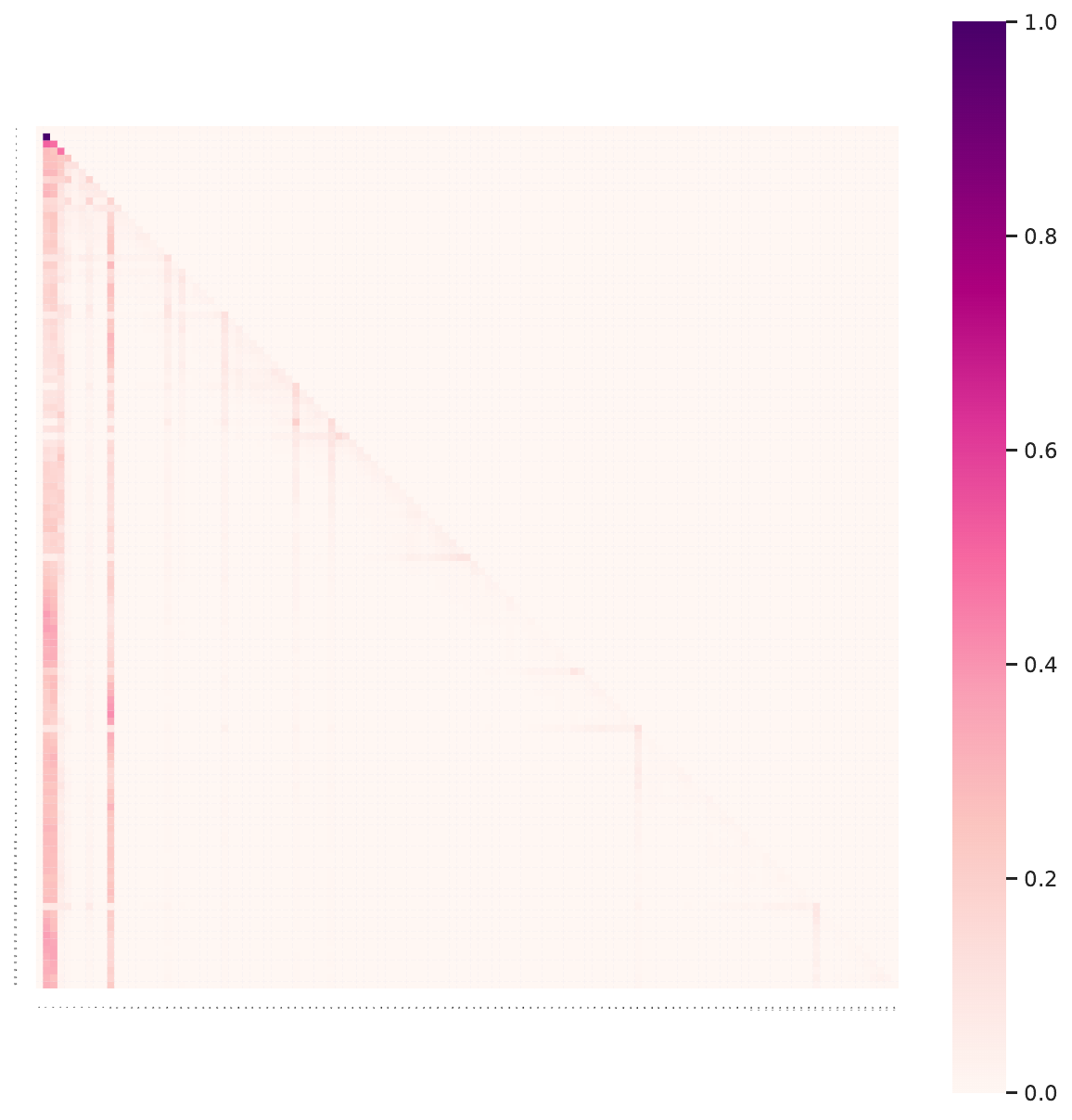}
        \caption*{Layer 27}
    \end{minipage}
    \vspace{1em}
\begin{minipage}{0.23\textwidth}
        \includegraphics[width=\linewidth]{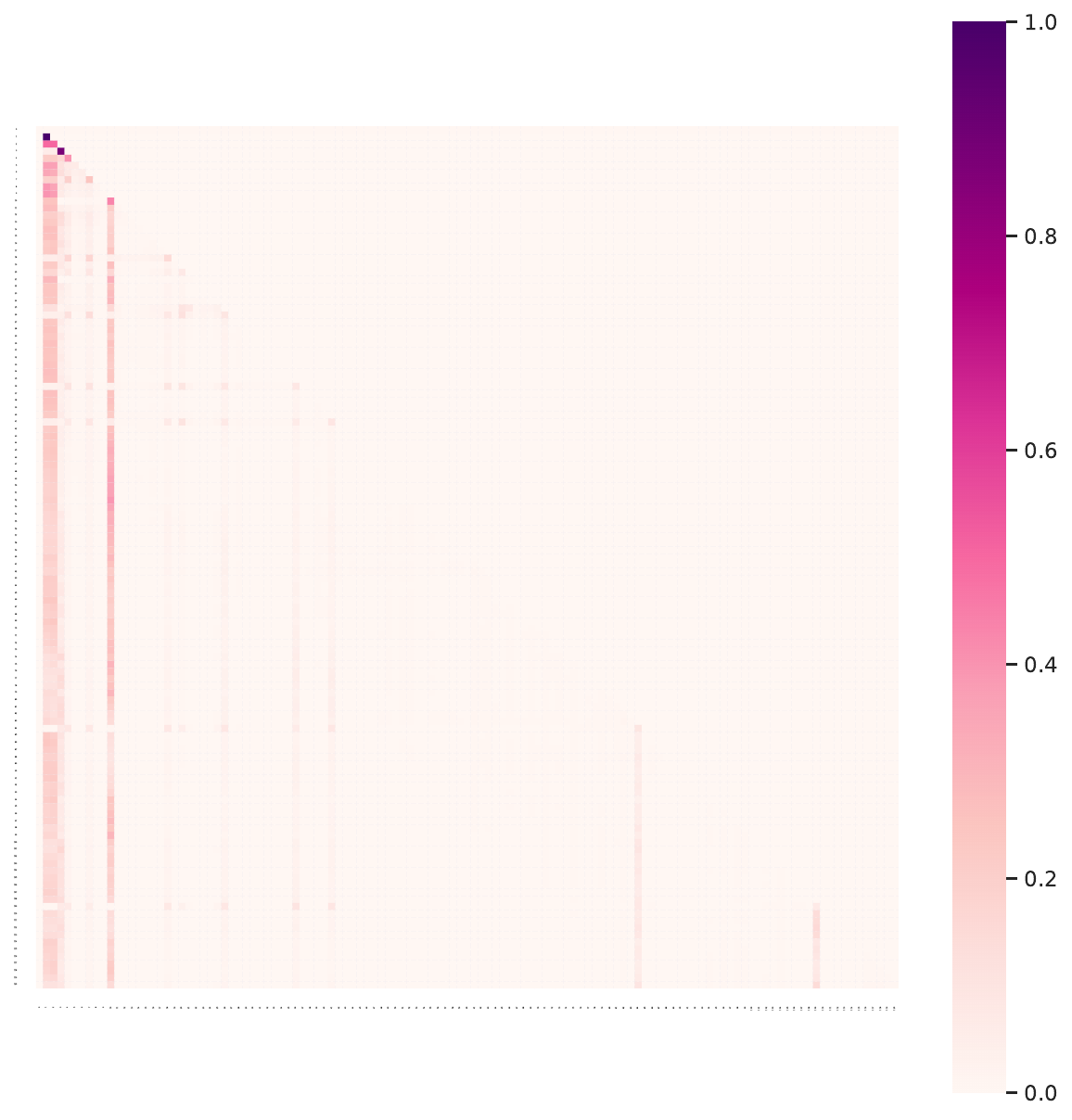}
        \caption*{Layer 28}
    \end{minipage}
    \begin{minipage}{0.23\textwidth}
        \includegraphics[width=\linewidth]{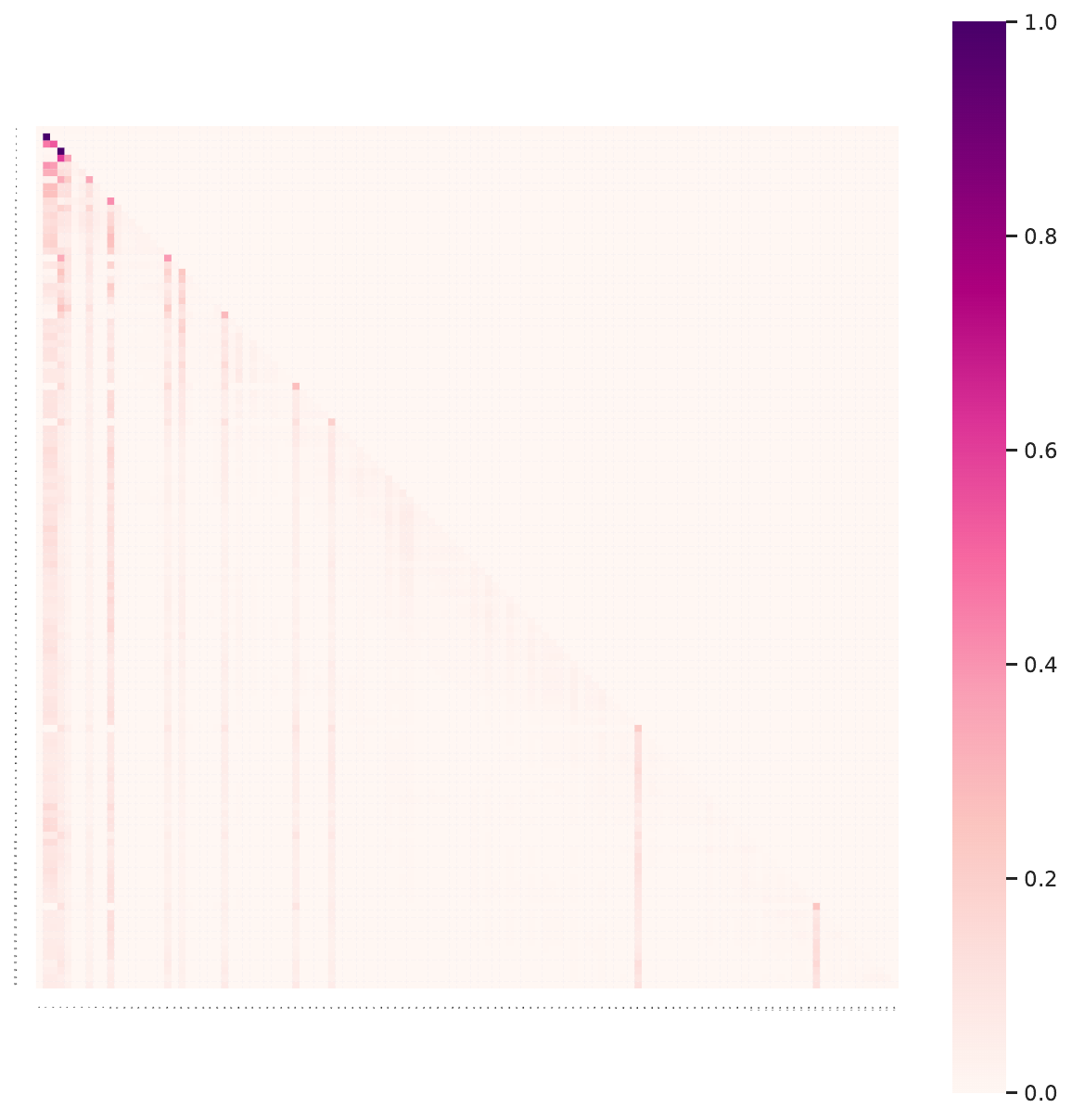}
        \caption*{Layer 29}
    \end{minipage}
    \begin{minipage}{0.23\textwidth}
        \includegraphics[width=\linewidth]{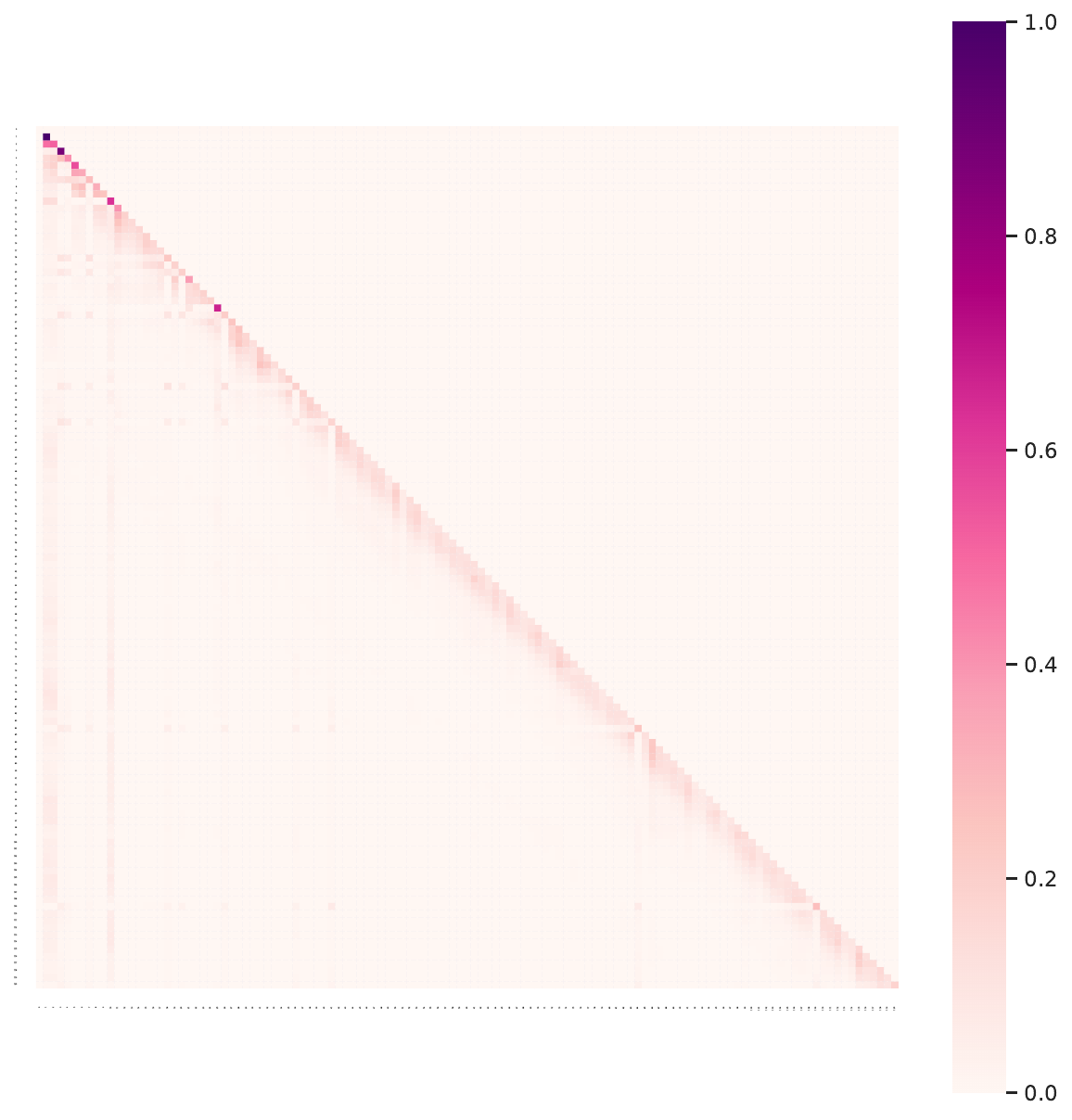}
        \caption*{Layer 30}
    \end{minipage}
    \begin{minipage}{0.23\textwidth}
        \includegraphics[width=\linewidth]{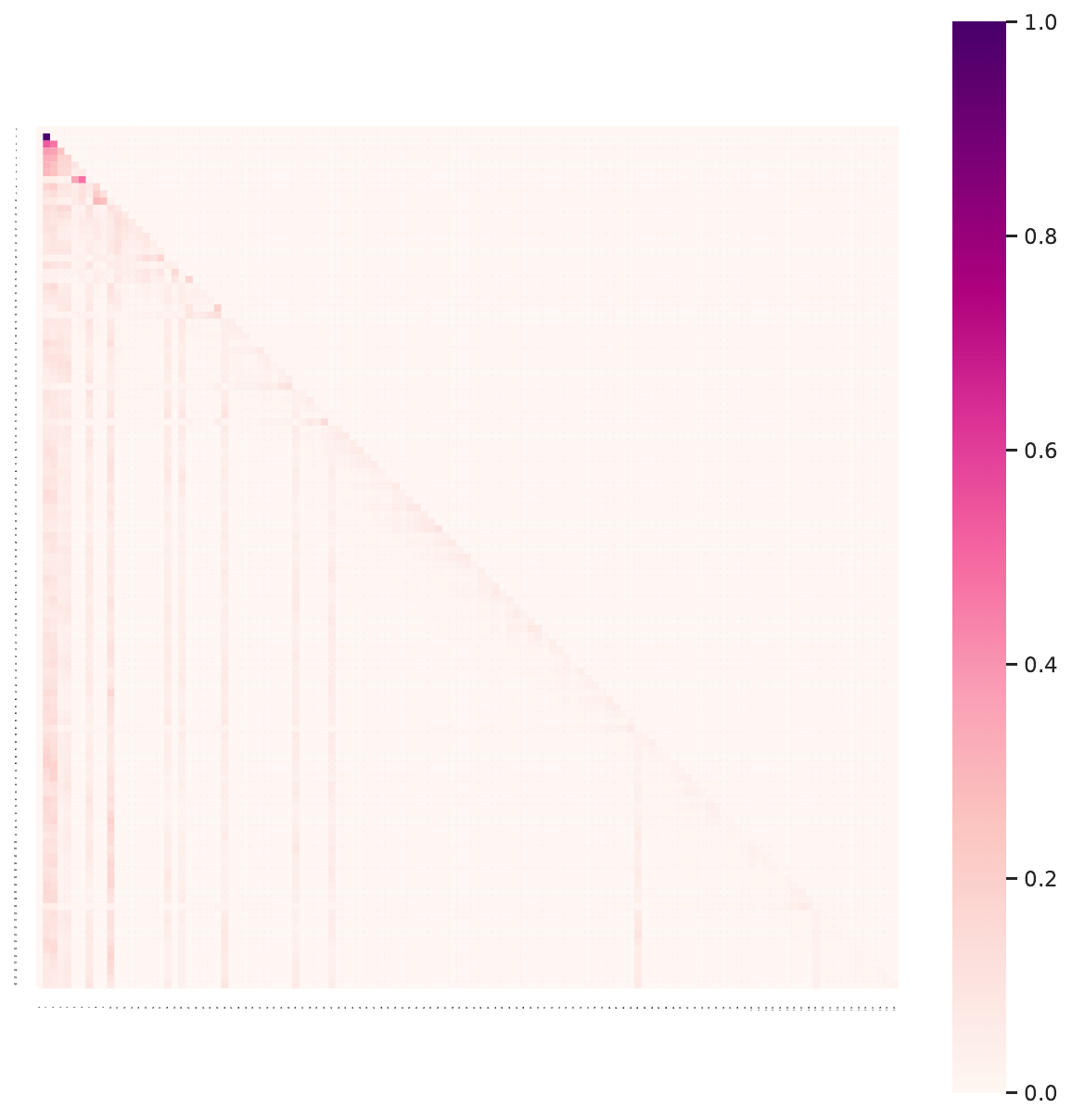}
        \caption*{Layer 31}
    \end{minipage}
\caption{\textbf{The attention score of our CRFT on head 31 in all layers.} (part 2 of 2)}
\label{Fig: vislayerh31dde_2}
\end{figure*}

\end{document}